%% file: Literature_Review.tex
\documentclass[12pt,letterpaper]{report}
\usepackage[letterpaper,width=150mm,top=25mm,bottom=25mm,bindingoffset=6mm]{geometry} 

\usepackage{titlesec}
\usepackage[nottoc]{tocbibind}
	\settocbibname{References}
\usepackage[titletoc,title]{appendix}

\usepackage{fancyhdr}
\usepackage{multirow}
\usepackage{tabularx,tabulary}
	
\usepackage[newcommands]{ragged2e} 
\usepackage{booktabs} 
\usepackage{subcaption}
\usepackage{graphicx}
\usepackage{amsmath, amssymb,amsthm}
	\DeclareMathOperator{\Tr}{Tr}
	
\usepackage{xfrac} 
\graphicspath{{images/}} 
\usepackage[hidelinks]{hyperref} 
\usepackage{pdflscape} 
\usepackage{afterpage}
\usepackage{color} 
\usepackage{mdwlist} 
\usepackage{diagbox} 
\usepackage{titlesec} 
\usepackage{xcolor}

\title{Review of \\ Action Recognition and Detection \\ Methods}
\author{ \Large{Soo Min Kang and Richard P. Wildes} \\[12pt]
	Department of Electrical Engineering and Computer Science\\
	York University \\
	Toronto, Ontario \\
	Canada}
\date{}

\begin{document}
\maketitle

\chapter*{Abstract}
\input{chapters/abstract}

\tableofcontents

\chapter{Introduction \label{ch:intro}}
\input{chapters/introduction}

\chapter{Benchmark Datasets \label{ch:datasets}}
\input{chapters/chapter2_datasets}

\chapter[Image Representation]{Image Representation:\\ Features and their Encodings \label{ch:image_representation}}
\input{chapters/chapter3_representation}

\chapter{Classification \label{ch:classification}}
\input{chapters/chapter4_classification}

\chapter{Current Status \label{ch:current_status}}
\input{chapters/chapter5_currentstatus}

\begin{appendices}
\chapter{Related Fields \label{ch:appendix}}
	\input{chapters/appendix_related_fields}
\end{appendices}

\bibliographystyle{plain}
\bibliography{references}

\end{document}

%% file: chapters/abstract.tex
In computer vision, \textit{action recognition} refers to the act of classifying an action that is present in a given video and \textit{action detection} involves locating actions of interest in space and/or time. Videos, which contain photometric information (e.g. RGB, intensity values) in a lattice structure, contain information that can assist in identifying the action that has been imaged. The process of action recognition and detection often begins with extracting useful features and encoding them to ensure that the features are specific to serve the task of action recognition and detection. Encoded features are then processed through a classifier to identify the action class and their spatial and/or temporal locations. In this report, a thorough review of various action recognition and detection algorithms in computer vision is provided by analyzing the two-step process of a typical action recognition and detection algorithm: (i) extraction and encoding of features, and (ii) classifying features into action classes. In efforts to ensure that computer vision-based algorithms reach the capabilities that humans have of identifying actions irrespective of various nuisance variables that may be present within the field of view, the state-of-the-art methods are reviewed and some remaining problems are addressed in the final chapter.

%% file: chapters/introduction.tex
Videos have become a vital component of our lives as it contains important information about the world. Its information has served humans in various domains: from security to robotics to entertainment and many more. 
The practicality of videos have led to immense advancements for video recording, viewing, and distribution. One major drawback of such availability, however, is the overwhelming amount of videos that are produced for viewing and analysis by humans. An alternative to this tedious task is to use machines to automatically extract useful information in a video. Consequently, detecting and localizing human actions has been a topic of high interest in computer vision for many years. \\

Various terms (e.g. action recognition, action spotting, event recognition, etc.) have been coined to describe similar tasks. Thus, it is important that we define the terms precisely to avoid any misunderstandings. First, we must distinguish the difference between an \textit{action} and an \textit{event}. An action refers to motion created by the human body, which may or may not be cyclic. An event is composed of multiple primitive actions and can involve more than a single individual. While `run' and `jump' are some examples of cyclic and non-cyclic actions, respectively, `hurdle' would be an example of an event since it can be broken down into two primitive actions: `run' and `jump'. Second, we must identify the similarities and differences between the following terms: \textit{recognition}, \textit{classification}, \textit{detection}, \textit{localization}, and \textit{spotting}. Action \textit{recognition} and \textit{classification} are terms that are used interchangeably to describe the act of categorizing an action in a clip to one of the pre-defined set of actions. Action \textit{detection}, \textit{localization}, and \textit{spotting} are also synonymous terms, which aim to determine the action and its location (in space and/or time). In this survey, we focus on actions rather than events, and both recognition and detection algorithms will be studied. \\

With the emergence of wearable cameras (e.g. GoPro and Google Glass), first-person action recognition has also been of interest to many in the computer vision community. First- and third-person action recognition algorithms are two very closely related tasks. However, there is a significant difference between the two. \textit{First-person action recognition} involves determining the action executed by the person wearing the camera from an egocentric viewpoint. \textit{Third-person action recognition}, on the other hand, involves determining the action executed by a person as captured by someone other than the actor. This difference results in contrasting datasets, actions of interest, and viewpoints. Thus, we emphasize here that this paper primarily reviews third-person action recognition and detection algorithms. First-person action recognition algorithms along with a select few other related fields of action recognition and detection are briefed in Appendix \ref{ch:appendix}.\\

\begin{figure}[htbp]
	\begin{center}
		\includegraphics[width=0.9\textwidth]{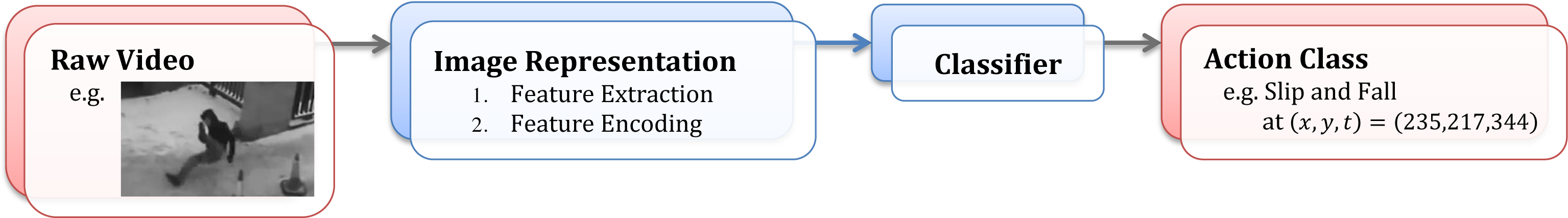}
	\end{center}
	\caption{General stages of a typical action recognition and detection algorithm. A video containing an action (e.g. slipping and falling) is inserted into the system. Features are extracted and encoded to represent the input video. The encoded features are processed by a classifier to output the class of the action (e.g. `slip and fall') for action recognition and its spatiotemporal coordinates (e.g. $(x,y,t)=(235,217,344)$) for action detection algorithms. The input (raw video) and the output (class label + spatiotemporal coordinates) of the system are marked in red while the intermediate processes are marked in blue. \label{fig:ar_overview}}
\end{figure}

To identify the action class of a given video, features must be extracted from a video and encoded to enter a classifier (see Figure \ref{fig:ar_overview}). 
In this report, benchmark datasets that appear in the field of action recognition and detection will be surveyed in Chapter \ref{ch:datasets}. A variety of ways to encode discriminative features in videos followed by various classification methods that have appeared in the action recognition and detection literature will be studied in Chapters \ref{ch:image_representation} and \ref{ch:classification}, respectively. Finally, some recent state-of-the-art algorithms in action recognition and detection as well as some outstanding challenges that remain in the field will be addressed in Chapter \ref{ch:current_status} to conclude the report.

%% file: chapters/chapter2_datasets.tex
With the growing popularity of various action recognition and detection algorithms, it is important to understand the comparative and absolute strengths and weaknesses of each approach. One of the most just ways to draw comparisons is to quantitatively evaluate each approach on the same database with the same protocol. Thus, it is important to survey the commonly used datasets and their key features to understand the capabilities and limitations of each tested approach \cite{Aha11, Cha13, Liu11}. In this chapter, some common testing protocols will be reviewed, benchmark datasets used for evaluation in subsequent chapters will be studied, then a quantitative summary of the datasets will follow. The datasets have been categorized by some common features that they share and a thorough analysis was conducted for each dataset by surveying their key characteristics, quantitative summary including the number of actors, actions, and conditions, video specifications (e.g. spatial resolution, video duration, frame rate), test protocols, and its intended use (recognition and/or detection).

\begin{section}{Testing Protocol}\label{sec:test_protocols}
To make a fair comparison between algorithms, it is very important to test them under the same protocol. First, the training, validation, and test data that are used to evaluate these algorithms must be consistent. As its name suggests, the purpose of a \textit{training set} is to train the classifier (i.e. to optimize the parameters of the classifier (e.g. weights in neural networks)). The \textit{validation set}, which is optional, is comprised of data distinct from those in the training set. It is used to make adjustments on the selected model such that the algorithm can perform well on both the training and the validation set. A validation set often is used to find the most optimal hyperparameters (e.g. number of hidden units, length of training, training rate in neural networks) for the model. The model that performs the best on the training and validation sets is finally assessed using the \textit{test set} to measure the performance of the overall system \cite{Dou13}. Separating a dataset into three disjoint sets (training, validation, and testing) allows researchers to tune their system and estimate the error simultaneously. \\

Second, the method of splitting a dataset into training, validation, and test must be uniform. There are three general ways to divide a set \cite{Dou13}: (i) using a pre-defined split, (ii) through $n$-fold cross-validation, and (iii) through leave-one-out cross-validation. 
	The \textit{pre-defined split} separates the dataset into two (or three) uneven components: training and testing (and validation), which is specified by the authors of the dataset. 
	The \textit{$n$-fold cross-validation} divides the dataset into $n$ mutually exclusive equal-sized folds. Videos in $n-1$ folds, which is approximately $\frac{n-1}{n}$ videos of the entire set, are used for training, and the remaining fold, approximately $\frac{1}{n}$ videos, is used for testing. This process is repeated $n$ times such that all clips are used once for testing. The average error rate of each fold is the estimated error rate of the classifier. 
	The \textit{leave-one-out cross-validation} is a special instance of cross-validation, where each removed sequence is compared to the remaining sequences. Leave-one-out is computationally expensive, but it determines the most accurate estimate of a classifier's error rate. \\

	Third, a single quantitative measure should be used for comparison. To evaluate how an action recognition algorithm performs with respect to each action class, an interpolated \textit{average precision} (AP) can be used. AP is defined as: 
\begin{equation}\label{eq:AP}
	AP(c) = \frac{\sum_{k=1}^{n}{\left( P(k) \times rel(k) \right)}}{\sum_{k=1}^{n}{rel(k)}}
\end{equation}
for test class $c$, where $n$ is the total number of videos, $P(k)$ is the precision at cutoff $k$ of the list, and $rel(k)$ is an indicator function which equals 1 if the video ranked $k$ is a true positive and 0 otherwise. The denominator in \eqref{eq:AP} represents the total number of true positives in the list.
The overall performance of the system can be evaluated using the \textit{mean average precision} (mAP) measure, which is defined as:
\begin{equation}\label{eq:mAP}
	mAP = \frac{1}{C} \sum_{c=1}^{C}{AP(c)} \text{,}
\end{equation}
where $C$ is the total number of test classes (i.e. $C = 101$ for UCF101). 
To determine whether the prediction should be considered a true or false positive for a detection algorithm, a threshold value can be associated with the \textit{intersection-over-union} (IoU) to accept or reject a detected result. That is, if $o$ denotes IoU between the predicted location, $L_p$, and the ground truth location, $L_{gt}$, then $o$ can be written mathematically as: 
\begin{equation}\label{eq:eval_overlap}
	o = \frac{L_p \cap L_{gt}}{L_p \cup L_{gt}} \text{,}
\end{equation}
and $L_p$ is considered correct if $o \geq \kappa$ for some constant $\kappa$.
\end{section}

\begin{section}[Static Background]{Static Camera with Clean Background}
One of the earliest goals in action recognition was to classify the action of a single individual in a video given a set of actions. Thus, a benchmark dataset containing a heterogeneous set of actions with systematic variations of parameters was in great demand. The KTH and Weizmann datasets met these requirements and became two of the earliest standard datasets of which to test action recognition algorithms. These datasets share a common characteristic of actors performing the actions in front of a simple background recorded with a static camera. Here, KTH, Weizmann, and the more recent MPII Cooking Activities datasets will be surveyed. 
	\begin{subsection}{The KTH Dataset}
		The efforts to create a non-trivial and publicly available dataset for action recognition was initiated at the KTH Royal Institute of Technology in 2004. The \textit{KTH dataset} \cite{Sch04} is one of the most standard datasets, which contains six actions: walk, jog, run, box, hand-wave, and hand clap (see Figure \ref{fig:kth}). To account for performance nuance, each action is performed by 25 different individuals, and the setting is systematically altered for each action per actor. Setting variations include: outdoor (s1), outdoor with scale variation (s2), outdoor with different clothes (s3), and indoor (s4). These variations test the ability of each algorithm to identify actions independent of the background, appearance of the actors, and the scale of the actors. \\
		
		The KTH dataset contains 6 actions performed by 25 individuals in 4 different settings (6 actions $\times$ 25 actors $\times$ 4 settings) resulting in a total of 600 clips\footnotemark. Each clip contains multiple instances of a single action and is recorded on a static camera with a frame rate of 25 frames per second (fps). The videos were down-sampled to have a spatial resolution of 160$\times$120 pixels and each clip ranges from 8 seconds (204 frames) to 59 seconds (1492 frames) averaging 18.9 seconds. The test protocol of the KTH dataset divides the videos into training, validation, and test sets, which contains 8, 8, and 9 actors, respectively. The dataset is useful for the task of recognition and temporal detection, as the ground truth indicates when specific actions occur but not where (the location).
		
		\footnotetext{A clip of person 13 performing \textit{hand clap} in the \textit{outdoor with different clothes} (s3) setting is missing in the KTH dataset resulting in a total of 599 clips instead of 600.}
		
		\begin{figure}[htbp]
			\begin{center}
			\includegraphics[width=0.95\textwidth]{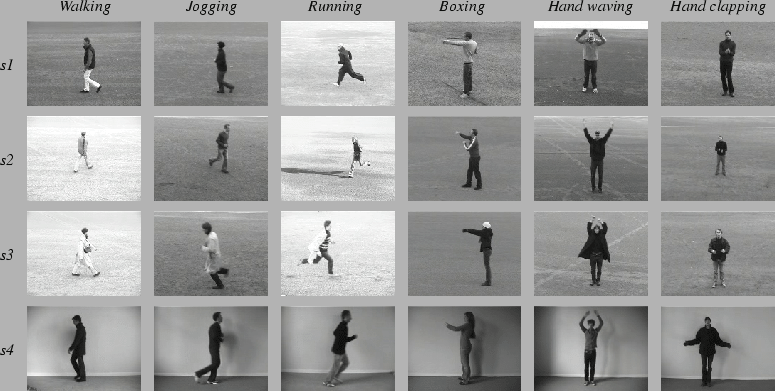}
			\caption{The KTH Dataset. The KTH dataset contains six different actions (left-to-right): \textit{walk}, \textit{jog}, \textit{run}, \textit{box}, \textit{hand-wave}, and \textit{hand clap}; taken at four different settings (top-to-bottom): \textit{outdoor} (s1), \textit{outdoor with scale variation} (s2), \textit{outdoor with different clothes} (s3), and \textit{indoor} (s4). Redrawn from \cite{Sch04}.}
			\label{fig:kth}
			\end{center}
		\end{figure}
	\end{subsection}

	\begin{subsection}{The Weizmann Dataset}
		The following year after the KTH dataset was released, the Weizmann Actions as Space-Time Shapes dataset (or the \textit{Weizmann dataset} \cite{Bla05}) at the Weizmann Institute of Science in the Department of Computer Science and Applied Mathematics in Israel also became available in the field of action recognition. The Weizmann dataset contains more actions than the KTH (bend, wave one hand, wave two hands, jumping jack, jump in place on two legs, jump forward on two legs, walk, run, skip, and gallop sideways (see Figure \ref{fig:weizmann})), but each action is performed by fewer individuals. Nevertheless, performance by nine individuals is enough to take into consideration the nuance between individuals. The actors repeat most actions, namely skip, jump, run, gallop sideways, walk, in opposite directions to account for the asymmetry of these actions. Like the KTH dataset, the videos in this dataset are recorded using a static camera on a uniform background. The actors move horizontally across the frame, maintaining the consistency in the size of the actor as they perform each action. \\
		
		The Weizmann dataset contains 10 actions performed by 9 individuals (10 actions $\times$ 9 actors) resulting in a total of 90 clips\footnotemark. Each clip contains multiple instances of a single action. Each clip was recorded on a static camera with 50 fps, but has been deinterlaced to 25 fps. The videos have a spatial resolution of 180$\times$144 pixels and each clip ranges from 1 second (36 frames) to 5 seconds (125 frames) averaging 3.66 seconds. The recommended testing protocol for using the Weizmann dataset is to perform a leave-one-out procedure. 
		Although the intended use of the dataset is for action recognition, it is also useful for the task of detection, as the ground truth are silhouette masks, which can be applied to extract both spatial and temporal information of the action.
		
		\footnotetext{Select actions (run, skip, and walk) by one of the individuals, Lena, are split into two clips resulting in 10 clips per action instead of 9. Thus, there are a total of 93 clips instead of 90.}
		
		\begin{figure}[h]
			\begin{center}
			\includegraphics[width=0.95\textwidth]{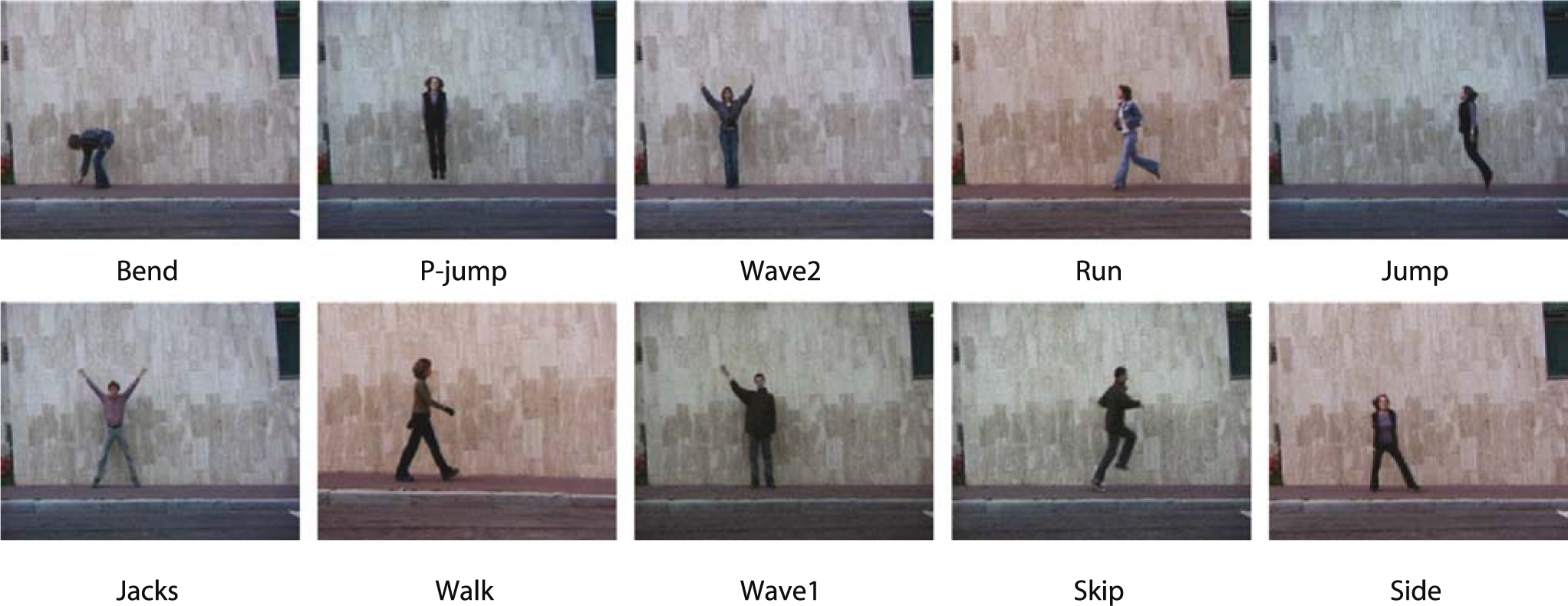}
			\caption{The Weizmann Dataset. The Weizmann dataset contains ten actions (left-to-right, top-to-bottom): \textit{bend}, \textit{jump in place on two legs} (P-jump), \textit{wave two hands} (wave2), \textit{run}, \textit{jump forward on two legs} (jump), \textit{jumping jack} (jacks), \textit{walk}, \textit{wave one hand} (wave1), \textit{skip}, and \textit{gallop sideways} (side). Redrawn from \cite{Bla05}.}
			\label{fig:weizmann}
			\end{center}
		\end{figure}
	\end{subsection}

	\begin{subsection}{MPII Cooking Activities Dataset}
		A group from the Max Planck Institute for Informatics (MPII) compiled the \textit{MPII Cooking Activities} \cite{Roh12} and its extension \textit{MPII Cooking 2} \cite{Roh15} datasets, which consist of actions related to cooking. The goal of these datasets is to distinguish between fine-actions, which is a very challenging task since there is high intra-class variation (e.g. peeling a carrot vs. peeling a pineapple) and low inter-class variation (e.g. mixing vs. stirring or dicing vs. slicing).
Participants, whose cooking skills range from beginner to amateur chefs, were instructed to cook one to six of pre-defined dishes (e.g. fruit salad) for the MPII Cooking dataset. The individuals were not given a specific recipe to follow. As a result, each individual used different ingredients to prepare each dish and very dissimilar videos were obtained. For each cooking video, actions (e.g. cut, peel) were annotated. A list of the 14 (and 59 additional) pre-defined dishes and the annotated 65 (and 67) actions for the MPII Cooking Activities (and MPII Cooking 2) dataset are listed in Table \ref{tab:mpii_cooking} (and \ref{tab:mpii_cooking2}).\\
	
	The MPII Cooking Activity dataset contains 12 subjects, where 7 of the subjects are used to perform leave-one-out cross-validation. That is, one of the subjects are removed from training, and the other 11 are used and this process is repeated 7 times. The MPII Cooking 2 dataset contains 30 subjects in 273 videos. The dataset is split into 201 training, 17 validation, and 42 testing with no overlap between the subjects. The training, validation, and test splits do not sum to the full dataset because for all composite actions in the testing set, the authors ensured that there were at least 3 training and validation videos from the same actor. Since some subjects had less than 3 training or validation videos, some test subjects were not used. Each video was recorded on a mounted camera attached to the ceiling, recording the actor working at the counter from the frontal view. The videos in both datasets have a spatial resolution of $1624 \times 1224$ with a frame rate of 29.4 fps, and the duration of the videos in the MPII Cooking 2 dataset ranges from 2 minutes and 44 seconds to 24 minutes and 34 seconds for a total of 8 hours and 19 minutes. 
Both datasets are useful for the task of action recognition as well as detection. Average precision (AP) is computed to compare per class results and mean average precision is used to report the overall performance of the algorithm on the datasets. The mid-point criterion is used to decide the correctness of the detection. That is, if the mid-point of the detection is within the ground truth, then it is considered correct.

\begin{table}[htbp]
	\begin{center}
	\begin{tabular}{| l | p{0.82\textwidth}|}
		\hline
		\small{Dishes} & \small{sandwich, salad, fried potatoes, potato pancake, omelet, soup, pizza, casserole, mashed potato, snack plate, cake, fruit salad, cold drink, and hot drink} \\
		\hline
		\small{Actions} & \small{background activity, change temperature, cut apart, cut dice, cut in, cut off ends, cut out inside, cut slices, cut stripes, dry, fill water from tap, grate, put on lid, remove lid, mix, move from X to Y, open egg, open tin, open/close cupboard, open/close drawer, open/close fridge, open/close oven, package X, peel, plug in/out, pour, pull out, puree, put in bowl, put in pan/pot, put on bread/dough, put on cutting-board, put on plate, read, remove from package, rip open, scratch off, screw close, screw open, shake, smell, spice, spread, squeeze, stamp, stir, strew, take and put in cupboard, take and put in drawer, take and put in fridge, take and put in oven, take and put in spice holder, take ingredient apart, take out from cupboard, take out from drawer, take out from fridge, take out from oven, take out from spice holder, taste, throw in garbage, unroll dough, wash hands, wash objects, whisk, and wipe clean} \\
		\hline
	\end{tabular}
	\caption{MPII Cooking Dataset \cite{Roh12}. 14 pre-defined dishes and 65 annotated actions are listed. \label{tab:mpii_cooking}}
	\end{center}
	\begin{center}
	\begin{tabular}{| l | p{0.82\textwidth}|}
		\hline
		\small{Dishes} & \small{cooking pasta, juicing \{lime, orange\}, making \{coffee, hot dog, tea\}, pouring beer, preparing \{asparagus, avocado, borad beans, broccoli and cauliflower, broccoli, carrot and potatoes, carrots, cauliflower, chilli, cucumber, figs, garlic, ginger, herbs, kiwi, leeks, mango, onion, orange, peach, peas, pepper, pineapple, plum, pomegranate, potatoes, scrambled eggs, spinach, spinach and leeks\}, separating egg, sharpening knives, slicing loaf of bread, using \{microplane grater, pestle and mortar, speed peeler, toaster, tongs\}, zesting lemon} \\
		\hline
		\small{Actions} & \small{add, arrange, change temperature, chop, clean, close, cut apart, cut dice, cut off ends, cut out inside, cut stripes, cut, dry, enter, fill, gather, grate, hang, mix, move, open close, open egg, open tin, open, package, peel, plug, pour, pull apart, pull up, pull, puree, purge, push down, put in, put lid, put on, read, remove from package, rip open, scratch off, screw close, screw open, shake, shape, slice, smell, spice, spread, squeeze, stamp, stir, strew, take apart, take lid, take out, tap, taste, test temperature, throw in garbage, turn off, turn on, turn over, unplug, wash, whip, wring out} \\
		\hline
	\end{tabular}
\caption{MPII Cooking 2 Dataset \cite{Roh15}. Additional 41 dishes that were added to the MPII Cooking 2 dataset and 67 annotated actions are listed. The dishes that were added are slightly shorter and simpler than the dishes in the MPII Cooking dataset. \label{tab:mpii_cooking2}}
	\end{center}
\end{table}
	\end{subsection}
	
	\begin{subsection}{Discussion}
	The KTH and Weizmann datasets set a good stepping stone for the field of action recognition through their heterogeneous selection of actions and systematic variations in its parameters. The controlled settings, such as absence of occlusion and clutter, limited variations in illumination and camera motion, allow these datasets to be ideal for standard testing. Unfortunately, good performance on the KTH and Weizmann datasets does not suffice to determine the algorithm's proficiency in real-world videos due to the richness and complexity of the videos in the real-world. In fact, while state-of-the-art action recognition algorithms routinely achieve greater than 90\% recognition accuracy on these datasets, they perform far less well on the more naturalistic datasets that are to be introduced in the remainder of this chapter. For this reason, strong performance on the KTH and Weizmann datasets is no longer of much interest in the field.\\

	The MPII Cooking 2 dataset shifts the focus of recognizing full-body movements (e.g. run, jump) to classifying actions with small motions. This fine-grained categorization can assist in differentiating visually similar activities that frequently occur in daily living (e.g. hug vs. hold someone and throw in garbage vs. put in drawer). The MPII Cooking 2 dataset also provides data for the often neglected but more challenging and realistic temporal detection task.
	\end{subsection}
\end{section}

\begin{section}[Dynamic Background]{Still Camera with Background Motion}
	To accommodate the lack of naturalistic settings in the KTH and Weizmann datasets, in particular the clean nature of the background, the next step was to test algorithms on videos with a dynamic background. In this section, the CMU Crowded Videos dataset and the MSR Action Dataset I, II, which contain videos with background motion and clutter will be examined. Dynamic background was obtained by recording videos in environments with moving cars and people. 

	\begin{subsection}{The CMU Crowded Videos Dataset}
		A group from Carnegie Mellon University (CMU) was one of the first to assemble a dataset, called the \textit{CMU Crowded Videos Dataset} \cite{Ke07}, for the action recognition and detection tasks that contain background motion. The CMU Crowded Videos Dataset focuses on five actions: pick-up, one-hand wave, push button, jumping jack, and two-hand wave. As many of the actions in the CMU Crowded Video dataset overlap those in the KTH and Weizmann, it was also one of the first cross-datasets that appeared in the field. That is, one of the training videos that is supplied in this dataset is the exact same video as the two-hand wave in the KTH dataset. \\
		
		The CMU Crowded Videos dataset contains 5 training videos for each action and 48 test videos. Each training video is performed by a single individual on a static background. The test videos contain three to six individuals different from those in the training set, and contains one to six instances of any three actions in no particular order (see Figure \ref{fig:cmu_clutter}). All videos, training and testing, have been scaled such that the spatial resolution of each video is $120 \times 160$. All videos have a frame rate of 30 fps, except the two handed wave, which has a frame rate of 25 fps. The test videos range from 5 to 37 seconds (166 to 1115 frames). The authors provide spatial and temporal coordinates (x, y, height, width, start, and end frames) for specified actions as ground truth, giving researchers the option to evaluate the ability of an algorithm to recognize and detect actions of interest. The detected action is considered a true positive if there is greater than 50\% overlap (in space and time) with the labelled action.
		
		\begin{figure}[!h]
			\begin{center}
			\begin{subfigure}{0.22\textwidth}
				\includegraphics[width=0.95\textwidth]{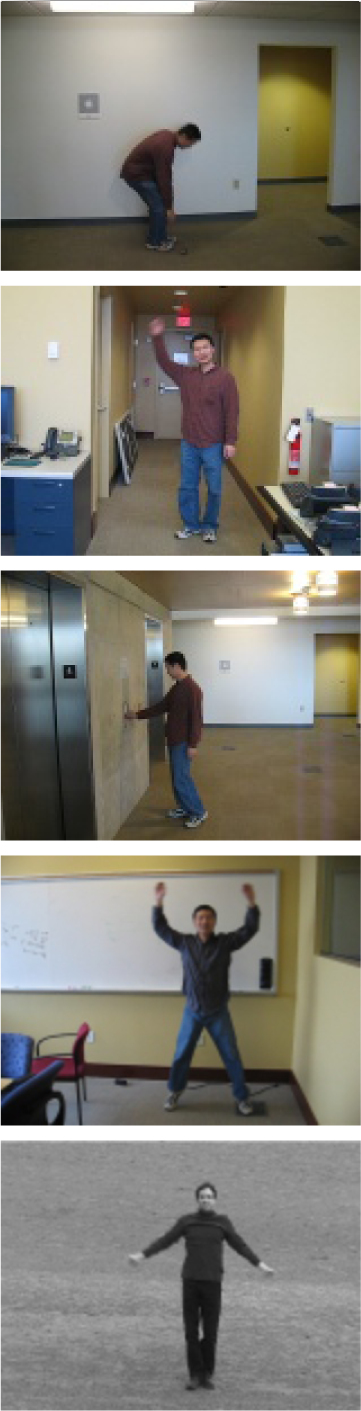}
				\caption{Templates}
				\label{fig:cmu_template}
			\end{subfigure}
			~
			\begin{subfigure}{0.72\textwidth}
				\includegraphics[width=0.95\textwidth]{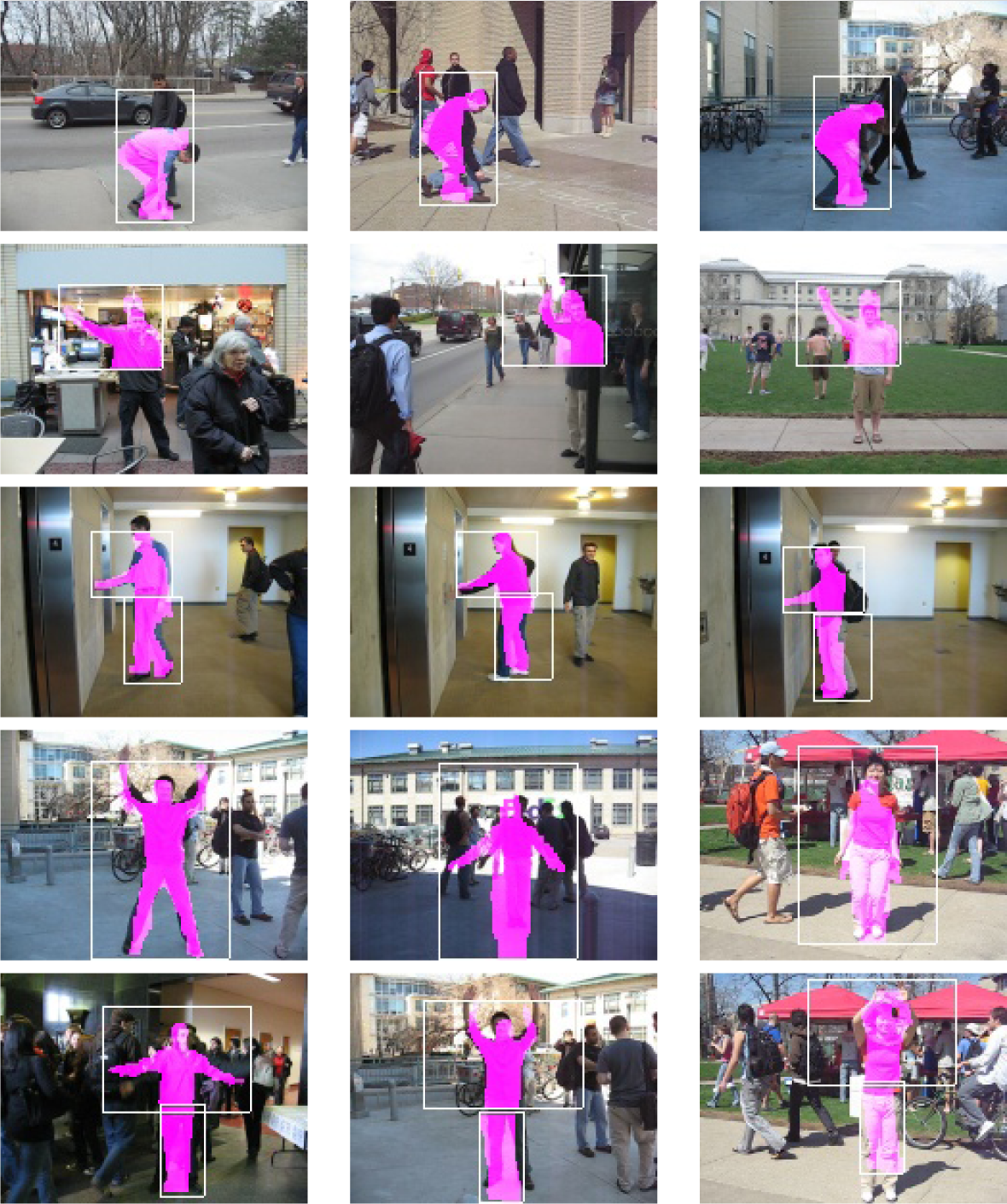}
				\caption{Test Videos}
				\label{fig:cmu_search}
			\end{subfigure}
			\caption{The CMU Clutter Dataset. The CMU Clutter dataset contains five actions (top-to-bottom): \textit{pick-up}, \textit{one-hand wave}, \textit{push button}, \textit{jumping jack}, and \textit{two-hand wave}. Select frames of the (a) templates and (b) test/search set are shown. The pink silhouettes overlaid on the test sequences are the best matches obtained from the template action, and the white bounding boxes indicate the match location of the upper and lower body parts. Redrawn from \cite{Ke07}.}
			\label{fig:cmu_clutter}
			\end{center}
		\end{figure}
	\end{subsection}
	
	\begin{subsection}{The MSR Action Dataset I, II}
		The Microsoft Research Group (MSR) also created action recognition datasets, referred to as the \textit{MSR Action dataset I} \cite{Yua09} and \textit{MSR Action dataset II} \cite{Cao10}, where II is a direct extension of I. These were made available in 2009 and 2010, respectively. Similar to the CMU Crowded dataset, the purpose of the MSR Action dataset construction was to obtain videos that contain cluttered and/or dynamic backgrounds \cite{Cao10,Yua09}. The datasets were assembled to detect 3 actions: clap, (two-)hand wave, and boxing. The MSR Action datasets are instances of a full cross-dataset\footnotemark. That is, to use the test videos in the MSR datasets, the actions must be trained using the videos in the KTH dataset. Each test sequence contains multiple actions, varies in the number of participants performing the action, the number of individuals in the video, and the number of actions that occur simultaneously. Some sequences contain actions performed by a single individual, some performed by different individuals at a time, and some performed by two individuals simultaneously. \\
\footnotetext{Cross-datasets allow researchers to develop general algorithms deviating from action- or dataset-specific recognition algorithms.}
		
		The MSR Action dataset I contains 24 instances of box, 24 instances of a two-hand wave, and 14 instances of clap, tallying 62 instances in total for 16 video sequences. The MSR Action dataset II, on the other hand, contains 81, 71, and 51 instances of box, wave, and clap, respectively, to sum up to a total of 203 instances of the three actions in a set of 54 videos. All videos in the MSR Action dataset I have a frame rate of 15 fps, and ranges from 32 to 76 seconds (480 to 1149 frames). Videos in the MSR Action dataset II, on the other hand, have varying frame rates ranging from 14 to 15 fps, and are 21 to 85 seconds (321 to 1284 frames) long. All videos in both the MSR Action dataset I and II have a spatial resolution of $240 \times 320$, and are filmed using a static camera. As mentioned before, the videos from the KTH dataset that correspond to the three actions: box, wave, and clap are used for training, and the videos provided by MSR are used for testing. Both the spatial and temporal coordinates of each action instance are provided for ground truth allowing the dataset to be used for action detection, as well as recognition. Although the original documentation of the MSR datasets do not specify the evaluation criterion, many papers that have used the MSR dataset for spatiotemporal action detection \cite{van15} consider the localized result a true positive if the IoU \eqref{eq:eval_overlap} between the ground truth data and the detected result is greater than or equal to some constant $\kappa$, where $\kappa = 0.2$ \cite{Tian13} and $\kappa = 0.5$ \cite{van15}.

		\begin{figure}[h]
			\begin{center}
				\begin{subfigure}{0.95\textwidth}
					\includegraphics[width=0.3\textwidth]{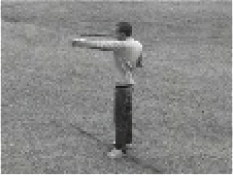}
					~
					\includegraphics[width=0.3\textwidth]{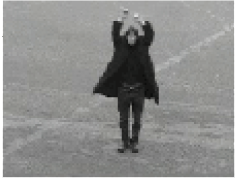}
					~
					\includegraphics[width=0.3\textwidth]{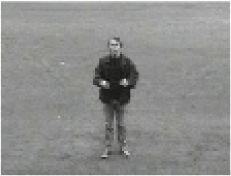}
				\end{subfigure}
			\\
				\begin{subfigure}{0.95\textwidth}
					\includegraphics[width=0.3\textwidth]{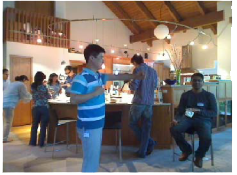}
					~
					\includegraphics[width=0.3\textwidth]{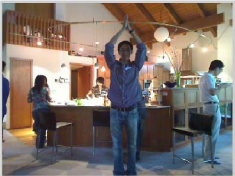}
					~
					\includegraphics[width=0.3\textwidth]{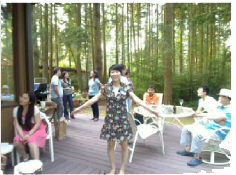}
				\end{subfigure}
			\caption{KTH vs. MSR. Comparison between the KTH dataset (top row) and the MSR dataset (bottom row) for actions boxing, two-hand wave, and clap (left-to-right). Redrawn from \cite{Cao10}.}
			\label{fig:msr}
			\end{center}
		\end{figure}
	\end{subsection} 
\end{section}

\begin{section}[Activities]{Action Recognition in Activity Videos}
	Along with many other videos, there are also plentiful sports and performance videos online that require categorization for accessible browsing and organization. A group from UC Berkeley collected videos from various sources to gather clips that frequently appear in ballet, tennis, and soccer \cite{Efr03}. This marked the beginning stages of collecting videos from multiple angles and moving cameras. In the following section, four activity-related action recognition/detection datasets will be introduced: the UC Berkeley Sports Dataset, the UCF Sports dataset, the Olympic Dataset, and Sports-1M. 
	\begin{subsection}{The UC Berkeley Dataset}
		The \textit{UC Berkeley dataset} consists of videos from three types of activities: ballet, tennis, and soccer.
		The ballet videos were collected from instructional videos, which contain four professional ballet dancers (two ballerinas and two ballerinos) performing mostly standard ballet moves. 16 ballet actions (standard moves) were chosen for the task of action detection: second position plies, first position plies, releve, down from releve, point toe and step right, point toe and step left, arms first position to second position, rotate arms in second position, degage, arms first position forward and out to second position, arms circle, arms second to high fifth, arms high fifth to first, port de dras, right arm from high fifth to right, and port de bra flowy arms (refer to Figure \ref{fig:ucb_ballet} to view select frames of each action). Each action was choreographed and all videos were filmed with a stationary camera. \\
		
		Two amateur tennis players playing tennis outdoors were recorded to gather videos for the tennis portion of the dataset. Videos were filmed on different days at different courts with slightly different camera positions to test variation in setting and perspective. Six actions were selected to complete the task of action recognition in tennis videos, which are: swing, move left, move right, move left and swing, move right and swing, and stand (refer to Figure \ref{fig:ucb_tennis} to see select frames from the tennis set). \\
		
		The videos for the soccer component were gathered from footages of the World Cup games. Among many angles that were available, only wide-angle shots of the playing field were collected. This angle forces each human figure to span $30\times30$ pixels on average, which is coarse for a video with a resolution of $640 \times 480$. Unlike the ballet and tennis videos, there is camera motion in the videos, a new challenge in the field of action recognition that has yet to have been introduced. The task is to differentiate between running and walking motions in specific directions. There are a total of eight categories for the soccer component: run left 45$^{\circ}$, run left, walk left, walk in/out, run in/out, walk right, run right, and run right 45$^{\circ}$. \\
		
		Unfortunately, the UC Berkeley dataset is no longer available for use and cannot be accessed anywhere. Therefore, a quantitative summary of this dataset is omitted. 
		\begin{figure}[!h]
		\begin{center}
			\begin{subfigure}{0.95\textwidth}
				\includegraphics[width=\textwidth]{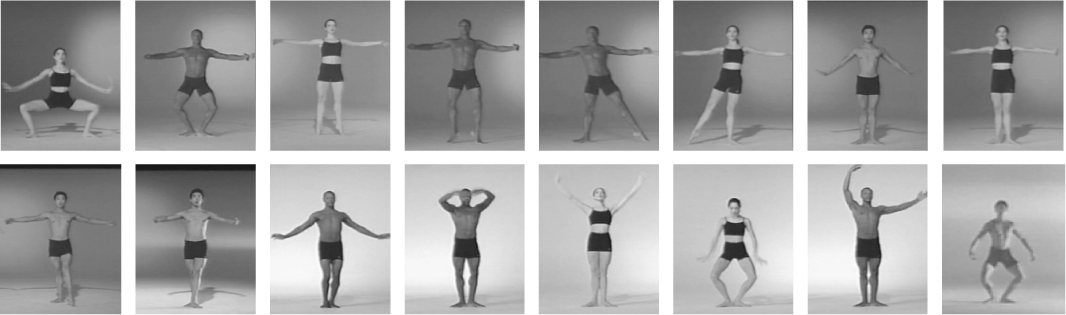}
			\caption{The UC Berkeley Ballet Dataset. Select frames that represent the 16 ballet actions are shown (left to right): (i) \textit{second position plies}, (ii) \textit{first position plies}, (iii) \textit{releve}, (iv) \textit{down from releve}, (v) \textit{point toe and step right}, (vi) \textit{point toe and step left}, (vii) \textit{arms first position to second position}, (viii) \textit{rotate arms in second position}, (ix) \textit{degage}, (x) \textit{arms first position forward and out to second position}, (xi) \textit{arms circle}, (xii) \textit{arms second to high fifth}, (xiii) \textit{arms high fifth to first}, (xiv) \textit{port de dras}, (xv) \textit{right arm from high fifth to right}, and (xvi) \textit{port de bra flowy arms}.}
			\label{fig:ucb_ballet}
			\end{subfigure}
			
			\begin{subfigure}{0.95\textwidth}
				\begin{center}
				\includegraphics[width=0.325\textwidth]{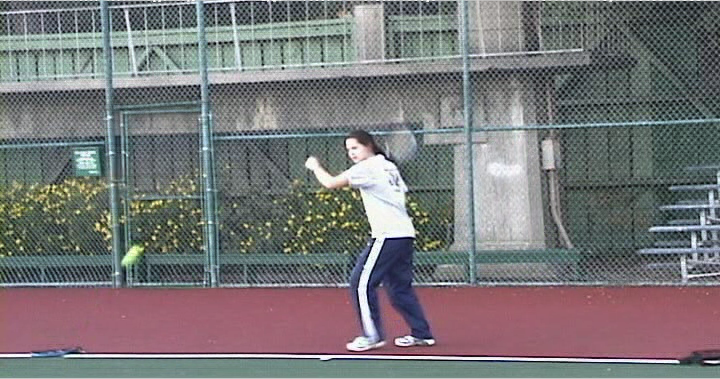}
				\includegraphics[width=0.325\textwidth]{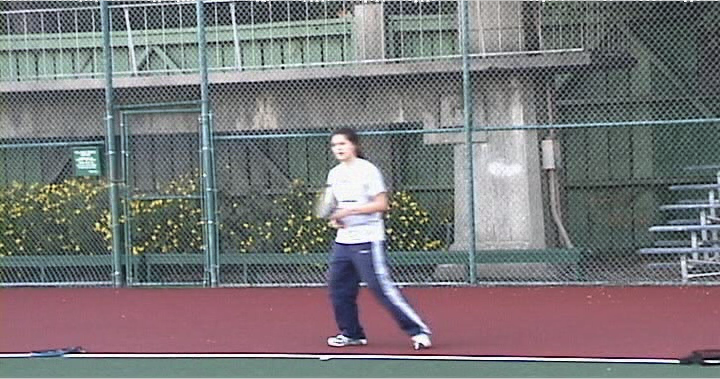}
				\includegraphics[width=0.325\textwidth]{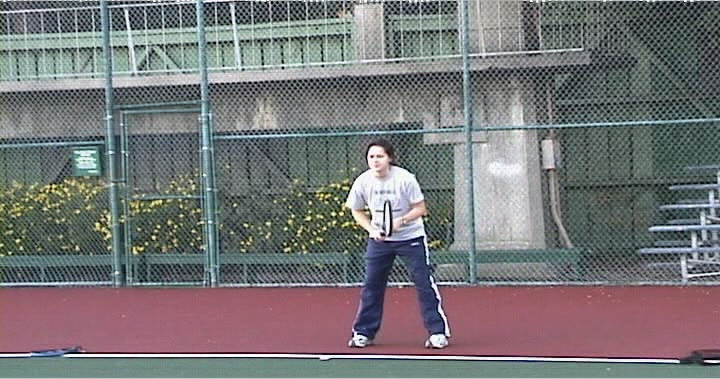}
				\caption{The UC Berkeley Tennis Dataset. Select frames of tennis player \textit{swing}, \textit{move left} and \textit{stand} are illustrated amongst the 6 tennis actions: swing, move left, move right, move left and swing, move right and swing, stand in the UC Berkeley Tennis Dataset.}
				\label{fig:ucb_tennis}
				\end{center}
			\end{subfigure}		
			
			\begin{subfigure}{0.95\textwidth}
				\includegraphics[width=0.34\textwidth]{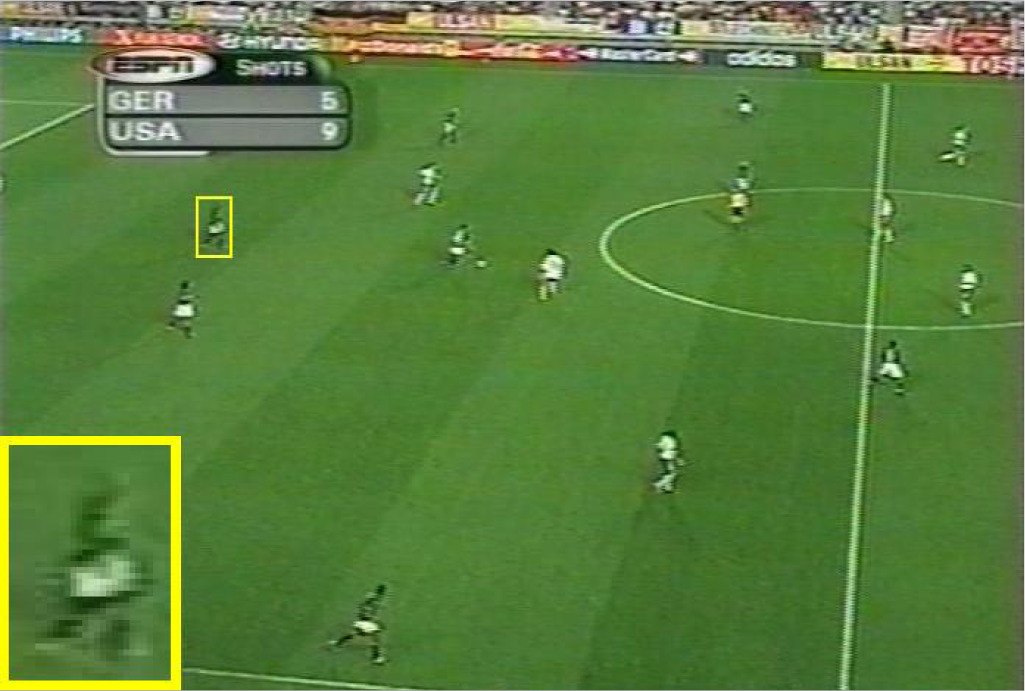}
				\includegraphics[width=0.31\textwidth]{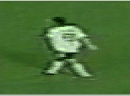}
				\includegraphics[width=0.31\textwidth]{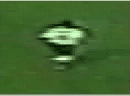}
				\caption{The UC Berkeley Soccer Dataset. A frame from a wide-angle shot of the playing field (left). Illustration of a player \textit{walking to the left} (centre) and \textit{running 45$^{\circ}$ to the right} (right). }
				\label{fig:ucb_soccer}
			\end{subfigure}

			\caption{The UC Berkeley Dataset. The UC Berkeley dataset contains actions in ballet, tennis, and soccer. Redrawn from \cite{Efr03}.}
			\label{fig:ucb_sports}
		\end{center}
		\end{figure} 
	\end{subsection}
	
	\begin{subsection}{UCF Sports Dataset}
		The actions in the \textit{UCF Sports} \cite{Rod08,Soo14} dataset were selected based on those that are typically featured in broadcast television channels, such as BBC and ESPN. The initial release of the dataset \cite{Rod08} consisted of nine actions: diving, golf swing, kicking, lifting, horseback riding, running, skateboarding, swinging a baseball bat, and pole vaulting (see Figure \ref{fig:ucf_sports1}). However, in the next release of the dataset \cite{Soo14}, swinging a baseball bat and pole vaulting, had been removed and swinging on a pommel horse and floor, swinging on parallel bars, and walking have been added to the second (and final) release of the UCF Sports dataset (see Figure \ref{fig:ucf_sports2}). Similar to the soccer videos of the UC Berkeley Dataset, the videos in the UCF Sports dataset contain camera motion and complex backgrounds. \\
		
		The UCF Sports dataset contains 150 clips ranging from 6 to 22 clips for the ten actions. Each clip has a frame rate of 10 fps. The spatial resolution of the videos range from $480 \times 360$ to $720 \times 576$ and are 2.20 to 14.40 seconds in duration, averaging 6.39 seconds. Two experimental setups for the task of action recognition (leave-one-out and five-fold cross-validation) and one for action detection (pre-defined split) are used with this dataset. 
		The authors provide temporal, as well as spatial coordinates for each action for the ground truth allowing this dataset to be used for both action recognition and spatiotemporal detection tasks\footnotemark.				
		\footnotetext{Although there are 150 clips in the UCF Sports dataset, only 140 clips contain ground truth data.}
		
		\begin{figure}[htbp]
			\begin{center}
			\begin{subfigure}{0.95\textwidth}
				\includegraphics[width=\textwidth]{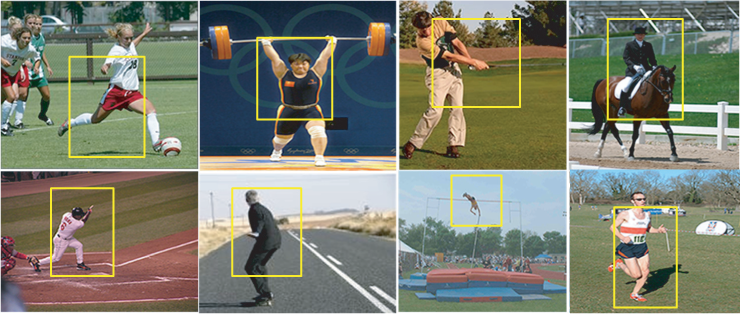}
				\caption{UCF Sports I. Select frames for eight of nine actions (left-to-right, then top-to-bottom): \textit{kicking}, \textit{lifting}, \textit{golf swing}, \textit{horseback riding}, \textit{baseball swing}, \textit{skateboarding}, \textit{pole vaulting}, and \textit{running}  from the first version of the UCF Sports Dataset are displayed. Redrawn from \cite{Rod08}.}
				\label{fig:ucf_sports1}
			\end{subfigure}
			
			\begin{subfigure}{0.95\textwidth}
				\includegraphics[width=\textwidth]{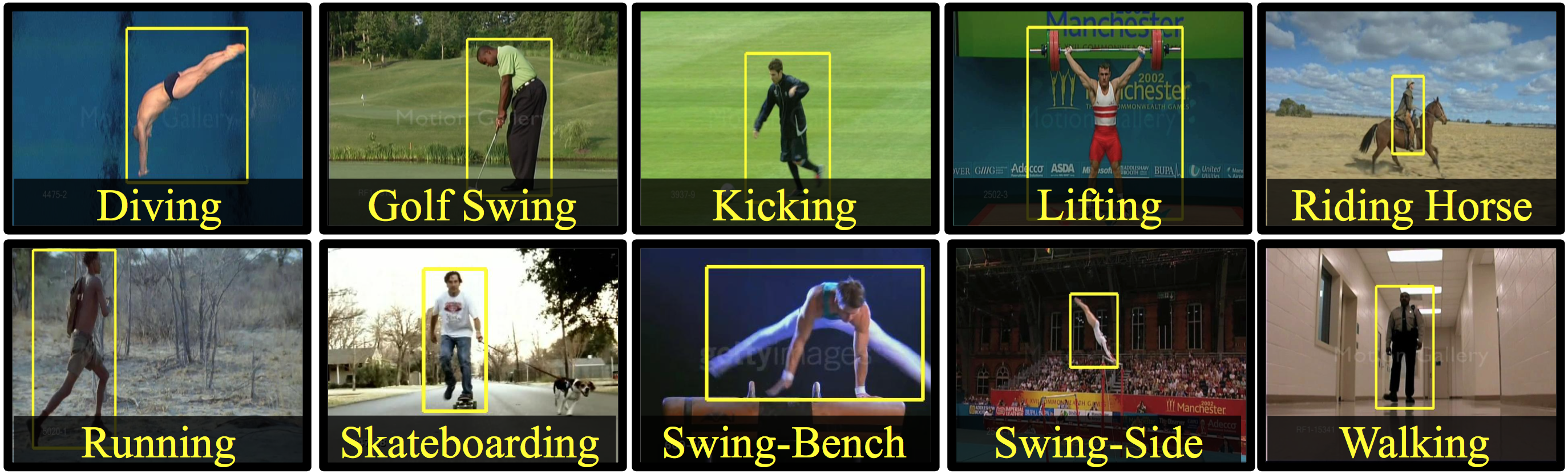}
				\caption{UCF Sports II. Select frames of ten actions (left-to-right, then top-to-bottom): \textit{diving}, \textit{golf swing}, \textit{kicking}, \textit{lifting}, \textit{horseback riding}, \textit{running}, \textit{skateboarding}, \textit{swinging on a pommel horse}, \textit{swinging on parallel bars}, and \textit{walking} from the latest version of the UCF Sports Dataset are illustrated. Redrawn from \cite{Soo12}.}
				\label{fig:ucf_sports2}
			\end{subfigure}
			\caption{UCF Sports Datasets. Two versions of the UCF Sports Dataset are illustrated.}
			\label{fig:ucf_sports}
			\end{center}
		\end{figure}
	\end{subsection}
	
	\begin{subsection}{The Olympic Dataset}
		The \textit{Olympic Dataset} \cite{Nie10} is a collection of Olympic sports videos extracted from YouTube. It contains 16 events that can be found in the Olympics: high jump, long jump, triple jump, pole vault, discus throw, hammer throw, javelin throw, shot put, basketball layup, bowling, tennis serve, platform (diving), springboard (diving), snatch (weightlifting), clean and jerk (weightlifting) and vault (gymnastics) (see Figure \ref{fig:olympics}), where each event contains approximately 50 sequences on average. It is suggested that the videos are split into 40:10 training:testing sequences for each action class as an experimental setup. The specific splits for training and testing can be found on their website: \url{http://vision.stanford.edu/Datasets/OlympicSports/}. All sequences in this dataset are stored in \textit{.seq} format, which requires special toolboxes to read. A summary of the file formats for these videos is omitted as the toolbox is difficult to use. Using the information obtained to split the data, this dataset is used to evaluate how accurately an algorithm can classify an action.
				
		\begin{figure}[h]
			\begin{center}
				\includegraphics[width=0.95\textwidth]{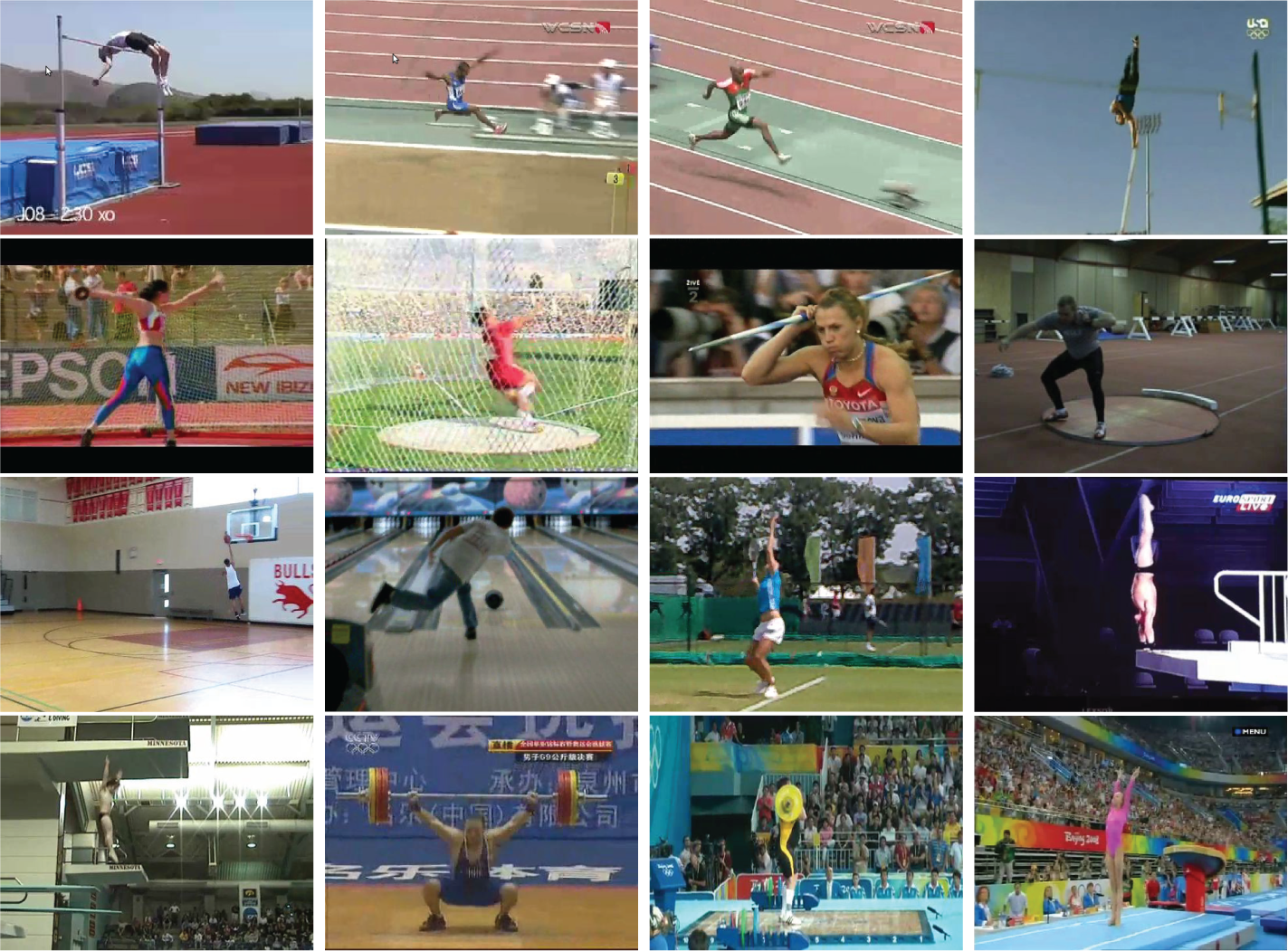}
				\caption{The Olympic Dataset. The Olympics Dataset contains 16 actions: \textit{high jump}, \textit{long jump}, \textit{triple jump}, \textit{pole vault}, \textit{discus throw}, \textit{hammer throw}, \textit{javelin throw}, \textit{shot put}, \textit{basketball layup}, \textit{bowling}, \textit{tennis serve}, \textit{platform (diving)}, \textit{springboard (diving)}, \textit{snatch (weightlifting)}, \textit{clean and jerk (weightlifting)}, and \textit{vault (gymnastics)} \cite{Nie10}.}
				\label{fig:olympics}
			\end{center}
		\end{figure}
	\end{subsection}

	\begin{subsection}{Sports-1M}
	The \textit{Sports-1M} \cite{AKar14} consists of over a million videos from YouTube. The videos in the dataset can be obtained through the YouTube URL specified by the authors. Unfortunately, approximately 7\% of the videos have been removed by the YouTube uploaders since the dataset was compiled \cite{Ng15}. This could change the training, validation, and/or testing set used in different experiments. 
However, there are still over a million videos in the dataset with 487 sports-related categories with $1,000$ to $3,000$ videos per category. 
The videos are automatically labelled with 487 sports classes using the YouTube Topics API \cite{YouTubeAPI} by analyzing the text metadata associated with the videos (e.g. tags, descriptions). 
While such large-scale dataset may be deemed useful to train CNN-based algorithms that are prone to overfitting on smaller datasets like UCF101 and HMDB51, the Sports-1M dataset must be used with caution. First, videos are gathered automatically and therefore labels are weak \cite{Feich16,Roh15}. Second, approximately 5\% of the videos are annotated with more than one class \cite{AKar14, Ng15}. Thus, the training video may not portray discriminative features of specific actions. Third, since users can post duplicate videos on YouTube, the same video could appear in both the training and testing sets \cite{AKar14}. \\

The spatial resolution of the videos range between $400 \times 240$ and $1280 \times 720$ pixels with a duration of $0$ to $37, 427$ frames. The Sports-1M dataset is split into 70\% training, 10\% validation, and 20\% testing sets. It is suggested that the videos are tested using a 10-fold cross-validation. The specific splits for each set can be found on the author's website: \url{http://cs.stanford.edu/people/karpathy/deepvideo/}.
	\end{subsection}
	
	\begin{subsection}{Discussion}
	Although these activity datasets have shown to be more difficult due to the presence of camera motion, the actions presented in these sets have shown to be relatively easy to identify. That is, by either analyzing the scene independent of the action or a pose of the actor in a single frame, an algorithm is likely to identify the action correctly \cite{Vu14}. 
This holds true because sports are location-specific (i.e. swimming-related events always occur in water and skiing on snow) and particular poses are only valid in specific sports (e.g. clean and jerk is specific to weightlifting) \cite{Der13, Kue11, Lan11,Soo14}.
	\end{subsection}
\end{section}

\begin{section}[Movies]{Action Recognition in Movies}
In efforts to create a dataset that meets the demands of applications in the real-world for action recognition, videos unrestricted of camera motions, scene context, spatial segmentation, and viewpoints had to be collected. The advent of unrestricted video dataset began with the collection of individuals ``drinking'' in movies ``Coffee and Cigarettes'' as well as ``Sea of Love'' \cite{Lap07}. Similarly, videos from eight different movies were gathered to collect 92 samples of ``kissing" and 112 samples of ``hitting/slapping'' \cite{Rod08}. The datasets extracted from movies gained popularity in the action recognition community when more actions were added to the datasets. The two most widely used datasets from movies are \textit{Hollywood1} \cite{Lap08} and \textit{Hollywood2} \cite{Mar09}.

	\begin{subsection}{Hollywood1}
		The \textit{Hollywood1} dataset \cite{Lap08} contains eight actions: answer the phone (AnswerPhone), get out of car (GetOutCar), handshake (HandShake), hug person (HugPerson), kiss, sit down (SitDown), sit up (SitUp), and stand up (StandUp) (see Figure \ref{fig:hollywood1}), extracted from 32 movies. The Hollywood1 dataset is randomly split into two sets: training and testing with 12 and 20 non-overlapping movies per set, respectively. The training set is further partitioned into \textit{automatic} and \textit{clean} datasets. The \textit{automatic training} set contains 233 action samples with 239 labels collected via unsupervised learning of automated script classification. The \textit{clean training} set, in contrast, contains 219 clips with 231 action labels and demonstrates supervised learning. That is, the clean training set has been manually selected to contain correct samples of the action classes retrieved from the text classification step. The \textit{test} set contains 211 clips with 217 action classes, which have been manually selected to discard false identifiers that arose from the script annotation step. Most clips in this dataset contain one action, and at most two actions per clip. The specific splits for training and test can be found on their website: \url{http://www.irisa.fr/vista/actions}. The videos in this dataset have a frame rate from 23 to 25 fps, spatial resolution from $180 \times 320$ to $240 \times 592$, and are 1 (41 frames) to 4 minutes and 48 seconds (7216 frames) long. The AP \eqref{eq:AP} and mAP \eqref{eq:mAP} scores are used to evaluate the performance of the system.
	\end{subsection}
	
	\begin{subsection}{Hollywood2}
		In addition to the actions in the Hollywood1 dataset, four new actions (drive a car (DriveCar), eat, fight a person (FightPerson), and run) were added from 69 movies to the \textit{Hollywood2} dataset \cite{Mar09} (see Figure \ref{fig:hollywood2}). Furthermore, to determine if algorithms benefit from drawing correlations between scene context and actions, ten scene settings: house, road, bedroom, car, hotel, kitchen, living room, office, restaurant, and shop were also provided in the dataset. The scenes were further categorized into either exterior (EXT) or interior (INT) scenes. Similar to the Hollywood1 dataset, the Hollywood2 dataset is split into automatic training, clean training, and testing sets. Again, the pre-defined splits can be found on the author's website: \url{http://www.di.ens.fr/~laptev/actions/hollywood2/}. The videos in this dataset have a frame rate of 23 to 29 fps, a spatial resolution of $224 \times 528$ to $576 \times 720$, and a duration ranging from 2 seconds (59 frames) to 8 minutes and 5 seconds (12131 frames). All clips within the dataset are trimmed such that it contains one of twelve actions. Furthermore, the ground truth data only provide the action label for each clip. Thus, this dataset is useful for the task of action recognition and cannot be used for action detection.
		
		\begin{figure}[h]
			\begin{center}
				\begin{subfigure}{0.95\textwidth}
					\includegraphics[width=\textwidth]{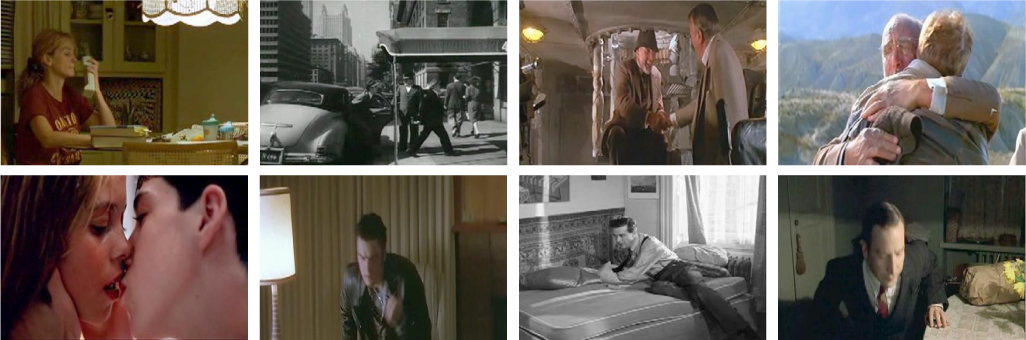}
					\caption{Hollywood1 Dataset. The Hollywood1 dataset contains eight actions (left-to-right): \textit{answer the phone} (AnswerPhone), \textit{get out of car} (GetOutCar), \textit{handshake} (HandShake), \textit{hug person} (HugPerson), \textit{kiss}, \textit{sit down} (SitDown), \textit{sit up} (SitUp), and \textit{stand up} (StandUp). Redrawn from \cite{Lap08}.}
					\label{fig:hollywood1}
				\end{subfigure}
				
				\begin{subfigure}{0.95\textwidth}
					\includegraphics[width=\textwidth]{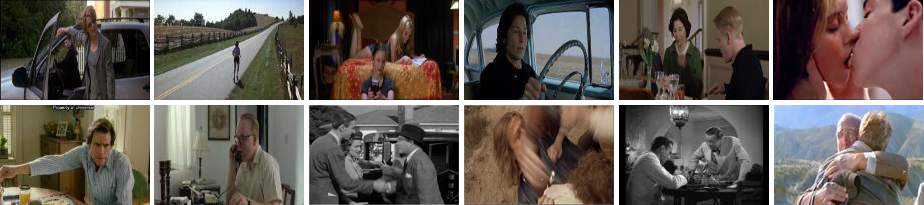}
					\caption{Hollwood2 Dataset. The Hollywood2 dataset contains twelve actions (left-to-right): \textit{get out of car} (GetOutCar), \textit{run} (Run), \textit{sit up} (SitUp), \textit{drive a car} (DriveCar), \textit{eat} (Eat), \textit{kiss} (Kiss), \textit{stand up} (StandUp), \textit{answer the phone} (AnswerPhone), \textit{shake hands} (HandShake), \textit{fight} (FightPerson), \textit{sit down} (SitDown), and \textit{hug} (HugPerson). Redrawn from \cite{Mar09a}.}
					\label{fig:hollywood2}
				\end{subfigure}
			\end{center}
			\caption{Hollywood1 and Hollywood2 Datasets. Select frames of actions in (a) Hollywood1 and (b) Hollywood2 datasets are illustrated.}
			\label{fig:hollywood}
		\end{figure}
	\end{subsection}
	\begin{subsection}{Discussion}
	Both datasets, Hollywood1 and Hollywood2, pose great challenges in the computer vision community as both databases contain diverse camera views, dynamic background, foreground clutter, frequent occlusions, and large intra-class variations. Although a plenitude parameter variations are considered, such as camera motion and clutter, all clips in these datasets are filmed by professional camera crew under controlled lighting conditions. These conditions are not very representative of the videos that we would encounter in the real-world. Furthermore, the parameter variations are not arranged in a systematic way, which brings difficulties in identifying the exact strengths and weaknesses of any action recognition approach. 
	\end{subsection}
\end{section}

\begin{section}[Home Videos]{Action Recognition in Home Videos}
With over 600 hours of home videos that are uploaded per minute on video-sharing websites like YouTube \cite{You}, categorization of videos is in great demand. Automated action recognition could be of great assistance in resolving this issue. Home videos are typically recorded in unconstrained environments, therefore contain diverse variations, such as random camera motion, poor lighting conditions, foreground clutter, movement in background, changes in scale, appearance, view points, and limited focus on the action of interest \cite{Red12}. Thus, to apply action recognition/detection algorithms in the real-world, scientists at the Centre for Research in Computer Vision at the University of Central Florida (UCF) collected videos from YouTube and other stock footage websites to construct a dataset that is more representative of real-world situations. Many datasets have been made publicly available by UCF to the computer vision community for non-commercial research purposes. 

	\begin{subsection}{UCF11 (YouTube Action), UCF50, and UCF101}\label{sec:UCF_dataset}
		Each of the \textit{UCF11} (also known as \textit{UCF YouTube Action}) \cite{Liu09}, \textit{UCF50} \cite{Red12}, and \textit{UCF101} \cite{Soo12} is an extension of the previous dataset. The videos for each action are assorted into 25 groups, where each group contains of 4-7 action clips. The clips are grouped according to common features videos share, such as the person in the video, background setting, and/or viewpoint. \\
		
		The original release of the UCF11 dataset contains videos with various spatial resolution, frame rate, and duration. In the latest release, the frame rate has been fixed to a constant rate of 29 fps, the spatial resolution ranges between $176 \times 144$ to $320 \times 240$, and the videos are less than a second (22 frames) to 29 seconds (900 frames) in length. The UCF50 and UCF101 datasets contain a total of $6,681$\footnotemark and $13,320$ videos, respectively, with at least 100 videos for each action class. All videos in both the UCF50 and UCF101 dataset have a spatial resolution of $240 \times 320$, and its frame rates are either 25 or 29 fps. The \textit{leave-one-out cross-validation} scheme is employed for all UCF11, UCF50, and UCF101 datasets and an additional experimental setup of \textit{train/test split} is recommended for the UCF101 dataset. Three specific train/test splits are suggested for the UCF101 dataset, in which each group is kept separate such that the clips from the same group are not shared in training and testing. Each test split has 7 different groups and their respective remaining 18 groups are used for training. 
		\footnotetext{The official report of the UCF50 dataset \cite{Red12} documents a total of 6676 videos in the UCF50 dataset. However, the downloadable UCF50 dataset contains 6681 videos.}\\

		The \textit{UCF101} dataset is a compilation of videos with the following actions: Apply Eye Makeup, Apply Lipstick, Archery, Baby Crawling, Balance Beam, Band Marching, Baseball Pitch, Basketball Shooting, Basketball Dunk, Bench Press, Biking, Billiards Shot, Blow Dry Hair, Blowing Candles, Body Weight Squats, Bowling, Boxing Punching Bag, Boxing Speed Bag, Breaststroke, Brushing Teeth, Clean and Jerk, Cliff Diving, Cricket Bowling, Cricket Shot, Cutting In Kitchen, Diving, Drumming, Fencing, Field Hockey Penalty, Floor Gymnastics, Frisbee Catch, Front Crawl, Golf Swing, Haircut, Hammer Throw, Hammering, Handstand Push-ups, Handstand Walking, Head Massage, High Jump, Horse Race, Horse Riding, Hula Hoop, Ice Dancing, Javelin Throw, Juggling Balls, Jump Rope, Jumping Jack, Kayaking, Knitting, Long Jump, Lunges, Military Parade, Mixing Batter, Mopping Floor, Nunchucks, Parallel Bars, Pizza Tossing, Playing Guitar, Playing Piano, Playing Tabla, Playing Violin, Playing Cello, Playing Daf, Playing Dhol, Playing Flute, Playing Sitar, Pole Vault, Pommel Horse, Pull Ups, Punch, Push Ups, Rafting, Rock Climbing Indoor, Rope Climbing, Rowing, Salsa Spins, Shaving Beard, Shot put, Skate Boarding, Skiing, Skijet, Sky Diving, Soccer Juggling, Soccer Penalty, Still Rings, Sumo Wrestling, Surfing, Swing, Table Tennis Shot, Tai Chi, Tennis Swing, Throw Discus, Trampoline Jumping, Typing, Uneven Bars, Volleyball Spiking, Walking with a dog, Wall Push-ups, Writing On Board, Yo-Yo (see Figure \ref{fig:ucf101}). These actions are divided into five groups: human-object interaction, body-motion only, human-human interaction, playing musical instruments, and sports. The categorization of each action into the groups are summarized in Table \ref{tab:ucf101}. The actions comprised in the UCF11 and UCF50 are summarized in Figures \ref{fig:ucf_youtube} and \ref{fig:ucf50}. 
		
		\begin{figure}[hp]
		\begin{center}
			\resizebox{.63\paperwidth}{!}{	
			\begin{subfigure}{0.225\textwidth}
				\includegraphics[width=\textwidth]{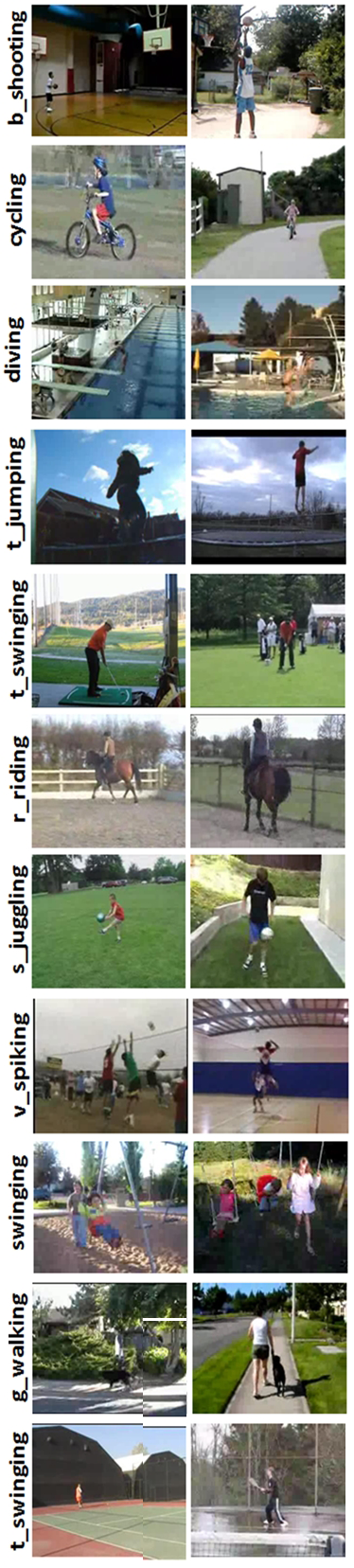}
			\caption{UCF11 Dataset}
			\label{fig:ucf_youtube}
			\end{subfigure}
			~~~
			\begin{subfigure}{0.67\textwidth}
				\includegraphics[width=\textwidth]{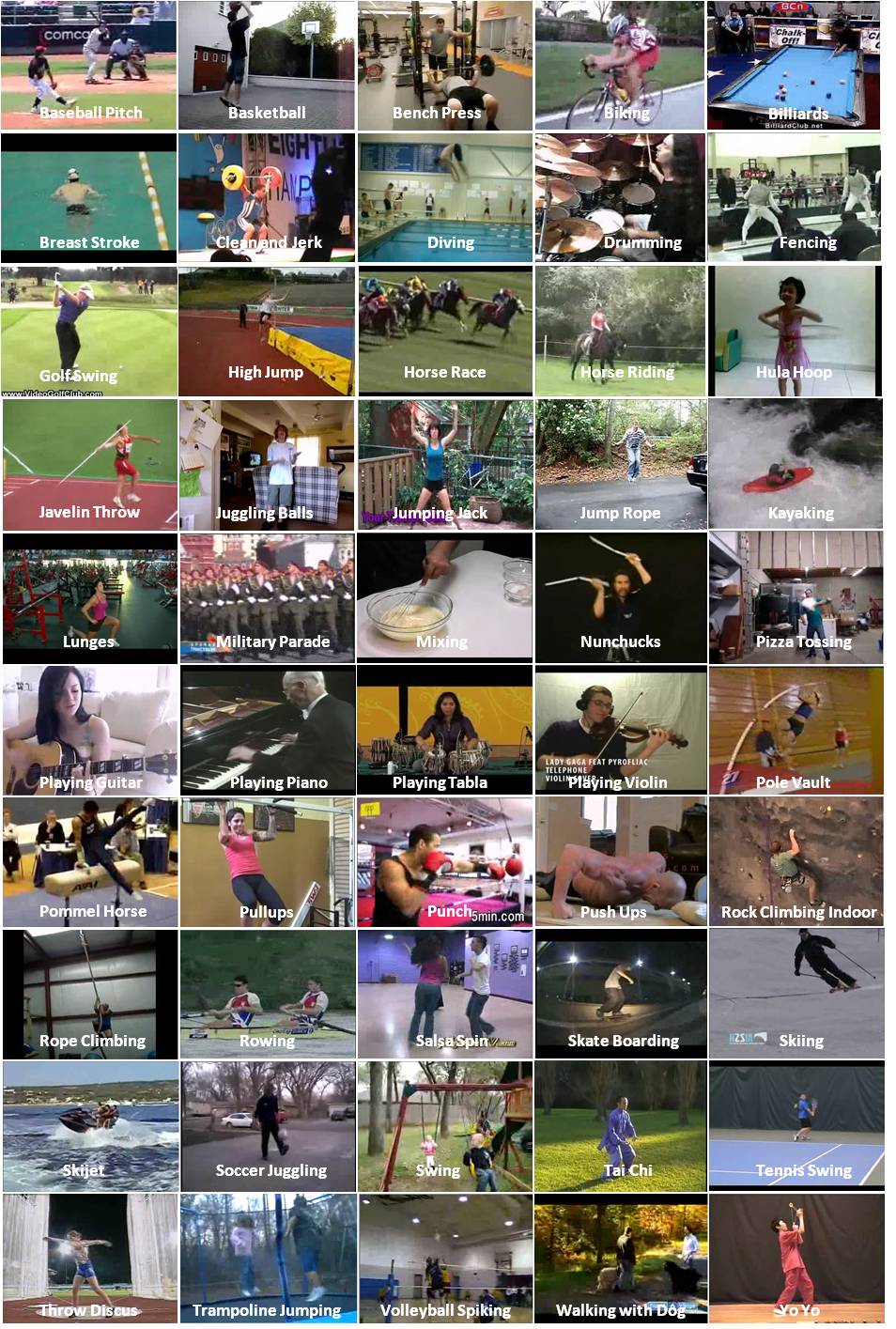}
				\caption{UCF50 Dataset}
				\label{fig:ucf50}
			\end{subfigure}
			}
		\end{center}
		\caption{UCF11 \cite{Liu09} and UCF50 \cite{Red12}. (a) Actions in the UCF11 dataset include (top-to-bottom): \textit{basketball shooting} (b\_shooting), \textit{cycling}, \textit{diving}, \textit{golf swinging} (t\_swinging), \textit{horse back riding} (r\_riding), \textit{soccer juggling} (s\_juggling), \textit{swinging}, \textit{tennis swinging} (t\_swinging), \textit{trampoline jumping} (t\_jumping), \textit{volleyball spiking} (v\_spiking), and \textit{walking with a dog} (g\_walking). Redrawn from \cite{Liu09}.
(b) Actions in the UCF50 dataset include (left-to-right, then top-to-bottom): \textit{Baseball Pitch}, \textit{Basketball Shooting}, \textit{Bench Press}, \textit{Biking}, \textit{Billiards Shot}, \textit{Breaststroke}, \textit{Clean and Jerk}, \textit{Diving}, \textit{Drumming}, \textit{Fencing}, \textit{Golf Swing}, \textit{High Jump}, \textit{Horse Race}, \textit{Horseback Riding}, \textit{Hula Hoop}, \textit{Javelin Throw}, \textit{Juggling Balls}, \textit{Jumping Jack}, \textit{Jump Rope}, \textit{Kayaking}, \textit{Lunges}, \textit{Military Parade}, \textit{Mixing Batter}, \textit{Nunchucks}, \textit{Pizza Tossing}, \textit{Playing Guitar}, \textit{Playing Piano}, \textit{Playing Tabla}, \textit{Playing Violin}, \textit{Pole Vault}, \textit{Pommel Horse}, \textit{Pull Ups}, \textit{Punch}, \textit{Push-Ups}, \textit{Rock Climbing Indoors}, \textit{Rope Climbing}, \textit{Rowing}, \textit{Salsa Spins}, \textit{Skate Boarding}, \textit{Skiing}, \textit{Ski-jet}, \textit{Soccer Juggling}, \textit{Swing}, \textit{TaiChi}, \textit{Tennis Swing}, \textit{Throwing a Discus}, \textit{Trampoline Jumping}, \textit{Volleyball Spiking}, \textit{Walking with a dog}, and \textit{Yo-Yo}. Redrawn from \cite{Red12a}.}
		\end{figure}

		\begin{figure}
		\begin{center}
				\includegraphics[width=0.85\textwidth]{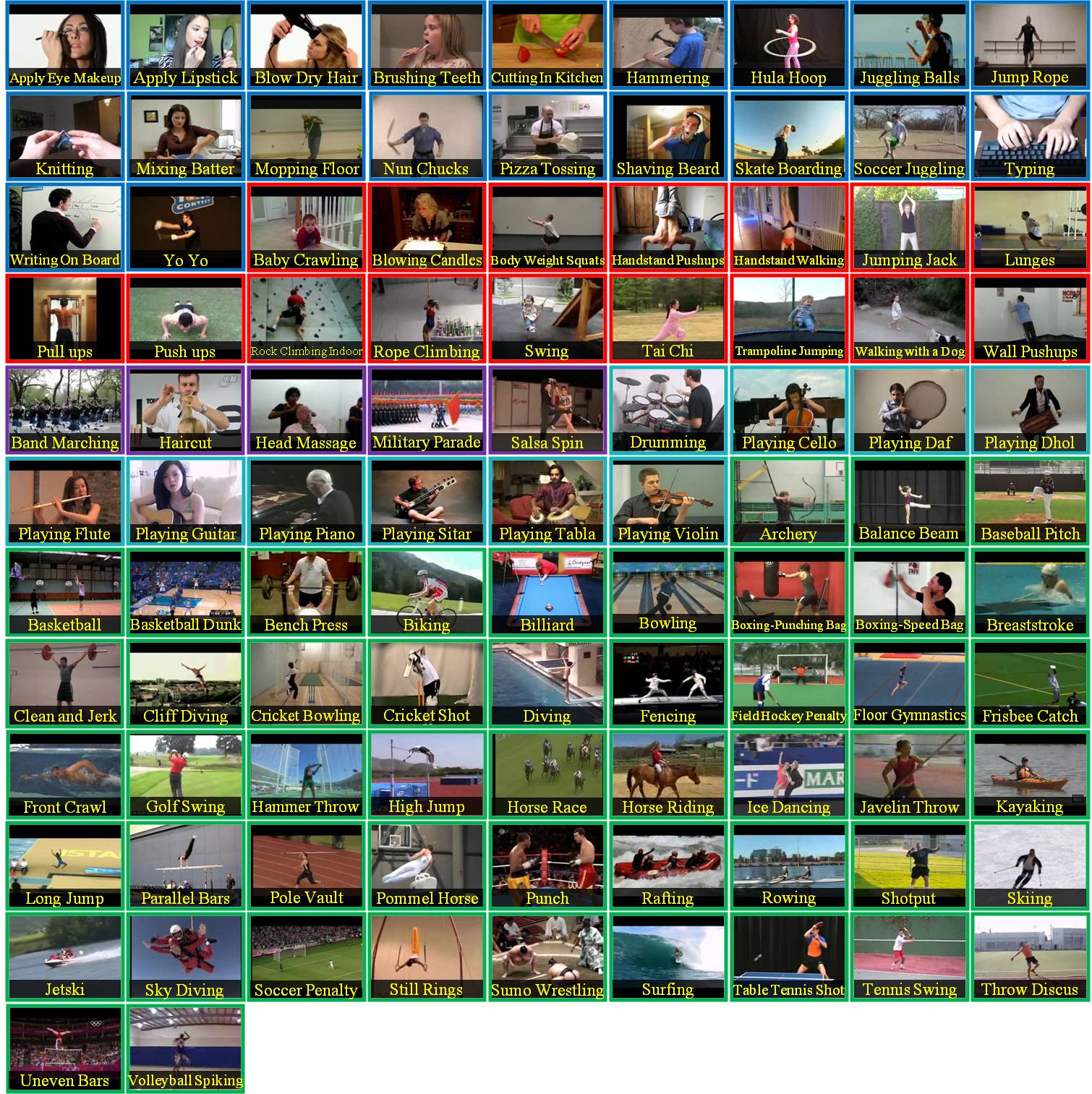}
				\caption{UCF101 Dataset \cite{Soo12}. Actions in the UCF101 dataset include (left-to-right then top-to-bottom): \textit{Apply Eye Makeup}, \textit{Apply Lipstick}, \textit{Blow Dry Hair}, \textit{Brushing Teeth}, \textit{Cutting In Kitchen}, \textit{Hammering}, \textit{Hula Hoop}, \textit{Juggling Balls}, \textit{Jump Rope}, \textit{Knitting}, \textit{Mixing Batter}, \textit{Mopping Floor}, \textit{Nun chucks}, \textit{Pizza Tossing}, \textit{Shaving Beard}, \textit{Skate Boarding}, \textit{Soccer Juggling}, \textit{Typing}, \textit{Writing On Board}, \textit{Yo-Yo}, \textit{Baby Crawling}, \textit{Blowing Candles}, \textit{Body Weight Squats}, \textit{Handstand Pushups}, \textit{Handstand Walking}, \textit{Jumping Jack}, \textit{Lunges}, \textit{Pull Ups}, \textit{Push-Ups}, \textit{Rock Climbing Indoor}, \textit{Rope Climbing}, \textit{Swing}, \textit{Tai Chi}, \textit{Trampoline Jumping}, \textit{Walking with a dog}, \textit{Wall Push-ups}, \textit{Band Marching}, \textit{Haircut}, \textit{Head Massage}, \textit{Military Parade}, \textit{Salsa Spins}, \textit{Drumming}, \textit{Playing Cello}, \textit{Playing Daf}, \textit{Playing Dhol}, \textit{Playing Flute}, \textit{Playing Guitar}, \textit{Playing Piano}, \textit{Playing Sitar}, \textit{Playing Tabla}, \textit{Playing Violin}, \textit{Archery}, \textit{Balance Beam}, \textit{Baseball Pitch}, \textit{Basketball Shooting}, \textit{Basketball Dunk}, \textit{Bench Press}, \textit{Biking}, \textit{Billiards Shot}, \textit{Bowling}, \textit{Boxing Punching Bag}, \textit{Boxing Speed Bag}, \textit{Breaststroke}, \textit{Clean and Jerk}, \textit{Cliff Diving}, \textit{Cricket Bowling}, \textit{Cricket Shot}, \textit{Diving}, \textit{Fencing}, \textit{Field Hockey Penalty}, \textit{Floor Gymnastics}, \textit{Frisbee Catch}, \textit{Front Crawl}, \textit{Golf Swing}, \textit{Hammer Throw}, \textit{High Jump}, \textit{Horse Race}, \textit{Horse Riding}, \textit{Ice Dancing}, \textit{Javelin Throw}, \textit{Kayaking},\textit{Long Jump}, \textit{Parallel Bars}, \textit{Pole Vault}, \textit{Pommel Horse}, \textit{Punch}, \textit{Rafting}, \textit{Rowing}, \textit{Shot put}, \textit{Skiing}, \textit{Skijet}, \textit{Sky Diving}, \textit{Soccer Penalty}, \textit{Still Rings}, \textit{Sumo Wrestling}, \textit{Surfing}, \textit{Table Tennis Shot}, \textit{Tennis Swing}, \textit{Throw Discus}, \textit{Uneven Bars}, and \textit{Volleyball Spiking}. Redrawn from \cite{Soo12}.}
				\label{fig:ucf101}
		\end{center}
		\end{figure}
		
		\begin{table}[hp]
			\begin{center}
				\begin{tabular}{| l | p{0.35\textwidth} | p{0.5\textwidth} |}
					\hline
					 & \textbf{Category} & \textbf{Actions} \\
					\hline
					1 & Human-Object Interaction & Apply eye makeup, apply lipstick, blow dry hair, brushing teeth, cutting in kitchen, hammering, hula hoop, juggling balls, jump rope, knitting, mixing batter, mopping floor, nun chucks, pizza tossing, shaving beard, skate boarding, soccer juggling, typing, writing on board, and yo-yo \\
					2 & Body-Motion Only & baby crawling, blowing candles, body weight squats, handstand push-ups, handstand walking, jumping jack, lunges, pull ups, push ups, rock climbing indoor, rope climbing, swing, tight, trampoline jumping, walking with a dog, and wall push-ups \\
					3 & Human-Human Interaction & band marching, haircut, head massage, military parade, and salsa spin\\
					4 & Playing musical instruments & drumming, playing cello, playing dad, playing dhol, playing flute, playing guitar, playing piano, playing sitar, playing tabla, and playing violin \\
					5 & Sports & Archery, balance beam, baseball pitch, basketball, basketball dunk, bench press, biking, billiard, bowling, boxing-punching bag, boxing-speed bag, breaststroke, clean and jerk, cliff diving, cricket bowling, cricket shot, diving, fencing, field hockey penalty, floor gymnastics, frisbee catch, front crawl, golf swing, hammer throw, high jump, horse race, horse riding, ice dancing, javelin throw, kayaking, long jump, parallel bars, pole vault, pommel horse, punch, rafting, rowing, shot-put, skiing, jets, sky diving, soccer penalty, still rings, sumo wrestling, surfing, table tennis shot, tennis swing, throw discus, uneven bars, and volleyball spiking\\
					\hline
				\end{tabular}
			\end{center}
			\caption{UCF101 Dataset categorization \cite{Soo12}.}
			\label{tab:ucf101}
		\end{table}	
	\end{subsection}
	
	\begin{subsection}{ActivityNet}
ActivityNet \cite{Hei15} is a large-scale video benchmark dataset for human activity understanding. Note, some instances of `activities' in the ActivityNet dataset are `events' by the definitions of this document as opposed to actions (see Chapter \ref{ch:intro}). Nevertheless, it covers a wide-range of complex human actions, with ample samples per class, that occur in our daily living. The classes are organized semantically according to social interactions and where the actions would generally take place (see Table \ref{tab:ActivityNet} for the ActivityNet semantic taxonomy). The actions are categorized in multiple levels. This hierarchical organization can be useful for (i) algorithms that are able to exploit hierarchy during model training, and (ii) precise analysis of actions that are more suited for certain algorithms over others. 
Two versions of the ActivityNet dataset have been released: ActivityNet 100 (release 1.2) and ActivityNet 200 (release 1.3). ActivityNet 100 contains 100 action classes, $4,819$ training videos with $7,151$ instances, $2,383$ validation videos with $3,582$ instances, and $2,480$ testing videos with the labels withheld for use in future challenges. ActivityNet 200 contains 203 action classes, $10,024$ training videos with $15,410$ instances, $4,926$ validation videos with $7,654$ instances, and $5,044$ testing videos with its labels withheld as well. The list of actions and the splits can be found on the author's website: \url{http://activity-net.org/index.html}.\\

All videos in ActivityNet are obtained from video sharing sites, such as YouTube. The videos are downloaded at the best quality available, approximately half of which have HD resolution of $1280 \times 720$. The majority of the videos in the dataset have a duration between 5 to 10 minutes with a frame rate of 30 fps. 
The dataset contains both temporally trimmed and untrimmed videos with an average of 1.41 trimmed video for each untrimmed video. This allows for classification of (i) trimmed action recognition, (ii) untrimmed action recognition, and (iii) temporal action detection. 
The trimmed action recognition set contains 203 classes of actions with an average of 193 samples per class, where each video contains a single instance of the action. Instances from a single video are forced to stay in the same training, validation, or test sets to avoid data contamination.
The untrimmed action recognition set contains $27,801$ videos belonging to 203 action classes, where each video can contain more than one activity. The set is randomly divided into 50\% training, 25\% validation, and 25\% test sets. 
The temporal action detection set contains 849 hours of video, where the detection algorithm should identify the start and end frames of all actions present in the untrimmed test video sequence. Like trimmed and untrimmed recognition sets, the set is randomly divided into 50\% training, 25\% validation, and 25\% test sets. 
mAP \eqref{eq:mAP} is used to measure the performance of all three tasks. A detection is considered a true positive if the IoU score \eqref{eq:eval_overlap} between a predicted temporal segment and the ground truth segment is greater than some constant $\kappa$. Authors report results on varying values of $\kappa$ from $0.1$ to $0.5$ in increments of 0.1.

\afterpage{
	\clearpage
	\begin{landscape}
		\begin{table}
			\begin{center} 
				\resizebox{.83\paperheight}{!}{
		\begin{tabular}{| l | p{0.25\textwidth} | p{0.35\textwidth} | p{\textwidth} |}
			\hline
			 & \textbf{Category} & \textbf{Sub-categories} & \textbf{Actions} \\
			\toprule\bottomrule
			1 & Eating and Drinking & Eating and Drinking & drinking coffee, drinking beer \\ 
				&	& Food and Drink Preparation & preparing pasta, preparing salad, making a sandwich, mixing drinks \\
				&	& Kitchen and Food Clean-up & washing dishes \\
			\midrule
			2 & Sports, Exercise, and Recreation & doing aerobics & zumba, step-aerobics\\
				&	& Martial arts & kickboxing, karate, tai chi \\
				&	& Playing sports & high jump, cricket, discus throw, javelin throw, paintball, long jump, bungee jumping, triple jump, shot put, dodgeball, hammer throw, skateboarding, motocross, campfire, archery, volleyball, kickball, pole vault,field hockey, basketball layup \\
				&	& Weightlifting & clean and jerk, snatch \\
				&	& Gymnastics & pommel horse, balance beam, tumbling, parallel bars, uneven bars \\
				&	& Cardiovascular equipment & spinning \\
				&	& Racket sports & table tennis, tennis serve, squash, lacrosse, racquetball, badminton \\
				&	& Equestrian sports & polo, horseback riding \\
				&	& Climbing, spelunking, caving & rock climbing \\
				&	& Water sports & springboard diving, sailing, platform diving, windsurfing, water polo, kayaking \\
			\midrule
			3 & Socializing, Relaxing, and Leisure & Dancing & tango, cheerleading, cumbia, breakdancing, belly dancing \\
				&	& Musical Instrument & playing bagpipes, harmonica, saxophone, guitar, flute, piano, violin, accordion \\
				&	& Arts and Entertainment & ballet \\
		\end{tabular}
				}
			\end{center}
		\end{table}	
	\end{landscape}

	\clearpage
	\begin{landscape}
		\begin{table}
			\begin{center} 
				\resizebox{.83\paperheight}{!}{
		\begin{tabular}{| l | p{0.25\textwidth} | p{0.35\textwidth} | p{\textwidth} |}
				&	& Tobacco and Drug Use & smoking hookah, smoking a cigarette \\
				&	& Playing Games & hopscotch \\
			\midrule
			4 & Personal Care & Washing, Dressing, and Grooming Oneself & putting on makeup, washing face, brushing hair, brushing teeth, doing nails, washing hands, shaving, shaving legs, removing curlers \\
				&	& Washing, Dressing, and Grooming & getting a tattoo, piercing, and a haircut \\
			\midrule
			5 & Household Activities & Household Management & wrapping presents \\
				&	& Animals and Pets & bathing dogs, grooming horse, walking the dog \\
				&	& Interior Maintenance, Repair, and Decoration & chopping wood, painting \\
				&	& Housework & cleaning windows, vacuuming floor, polishing furniture, cleaning shoes, polishing shoes, ironing clothes, handwashing clothes \\
				&	& Vehicles & fixing bicycle \\
				&	& Exterior Maintenance, Repair, and Decoration & shovelling snow \\
				&	& Lawn, Garden, and Houseplants & lawn mowing \\
			\toprule
		\end{tabular}
				}
			\end{center}
			\caption{ActivityNet Categorization \cite{Hei15}.}
			\label{tab:ActivityNet}
		\end{table}	
	\end{landscape}
}
	\end{subsection}

	\begin{subsection}{Discussion}
		The UCF101 dataset was one of the most challenging and largest datasets in action recognition and detection. Recently, the ActivityNet Dataset has taken the role and has become one of the most difficult for its large-scale and unconstrained characteristic of the videos. Both UCF101 and ActivityNet datasets contain videos that closely resemble videos that can be found in the real-world. Thus, algorithms that perform well in these datasets have great potential for use in real-life scenarios.
	\end{subsection}

\end{section}

\begin{section}{The Human Motion Databases}
	In efforts to collect videos that would capture the complexity of videos found in movies and videos online, the large Human Motion Database (\textit{HMDB51}) \cite{Kue11} was created by collecting videos from various sources, such as movies, YouTube, and Google videos. 

	\begin{subsection}{HMDB51}
		A total of 51 actions were selected for the HMDB51 database, where the actions were broadly categorized into five groups: 1) general facial actions, 2) facial actions with object manipulation, 3) general body movements, 4) body movements with object interaction, and 5) body movements for human interaction (see Table \ref{tab:hmdb51} and Figure \ref{fig:hmdb51}).  There are a total of $6,766$ clips in the HMDB51 dataset with each action containing at least 102 clips. To test the strengths and weaknesses in context of various nuisance factors, each video is annotated with a meta tag, which provides information like camera viewpoint, presence/absence of camera motion, video quality, number of actors involved in the action, and visible body part (see Table \ref{tab:hmdb51_metatag}). Three distinct training and testing splits are suggested for experimentation, where each split was generated to ensure that the clips from the same video did not appear in both the training and testing sets while there was an even distribution of meta tags across the sets. Each split contains 70 training and 30 testing videos with the excess videos excluded from the split. All the videos in the dataset have been normalized for a consistent height of 240 pixels and the widths have been scaled accordingly, ranging between 176 and 592 pixels, to maintain the original aspect ratio. All videos are trimmed to contain one of 51 actions, and the location of each action is not provided as a ground truth. Thus, this dataset is useful for testing classification.
		
		\begin{table}[!h]
			\begin{center}
				\begin{tabular}{| l | p{0.35\textwidth} | p{0.5\textwidth} |}
					\hline
					 & \textbf{Category} & \textbf{Actions} \\
					\hline
					1 & General facial actions & smile, laugh, chew, talk \\
					2 & Facial actions with object manipulation & smoke, eat, drink \\
					3 & General body movements & cartwheel, clap hands, climb, climb stairs, dive, fall on the floor, backhand flip, hand-stand, jump, pull up, push up, run, sit down, sit up, somersault, stand up, turn, walk, wave\\
					4 & Body movements with object interaction & brush hair, catch, draw sword, dribble, golf, hit something, kick ball, pick, pour, push something, ride bike, ride horse, shoot ball, shoot bow, shoot gun, swing baseball bat, sword exercise, throw \\
					5 & Body movements for human interaction & fencing, hug, kick someone, kiss, punch, shake hands, sword fight \\
					\hline
				\end{tabular}
			\end{center}
			\caption{HMDB51 Dataset categorization \cite{Kue11}.}
			\label{tab:hmdb51}
		\end{table}

		\begin{figure}[hp]
				\begin{subfigure}{\textwidth}
					\begin{center}
					\includegraphics[width=0.9\textwidth]{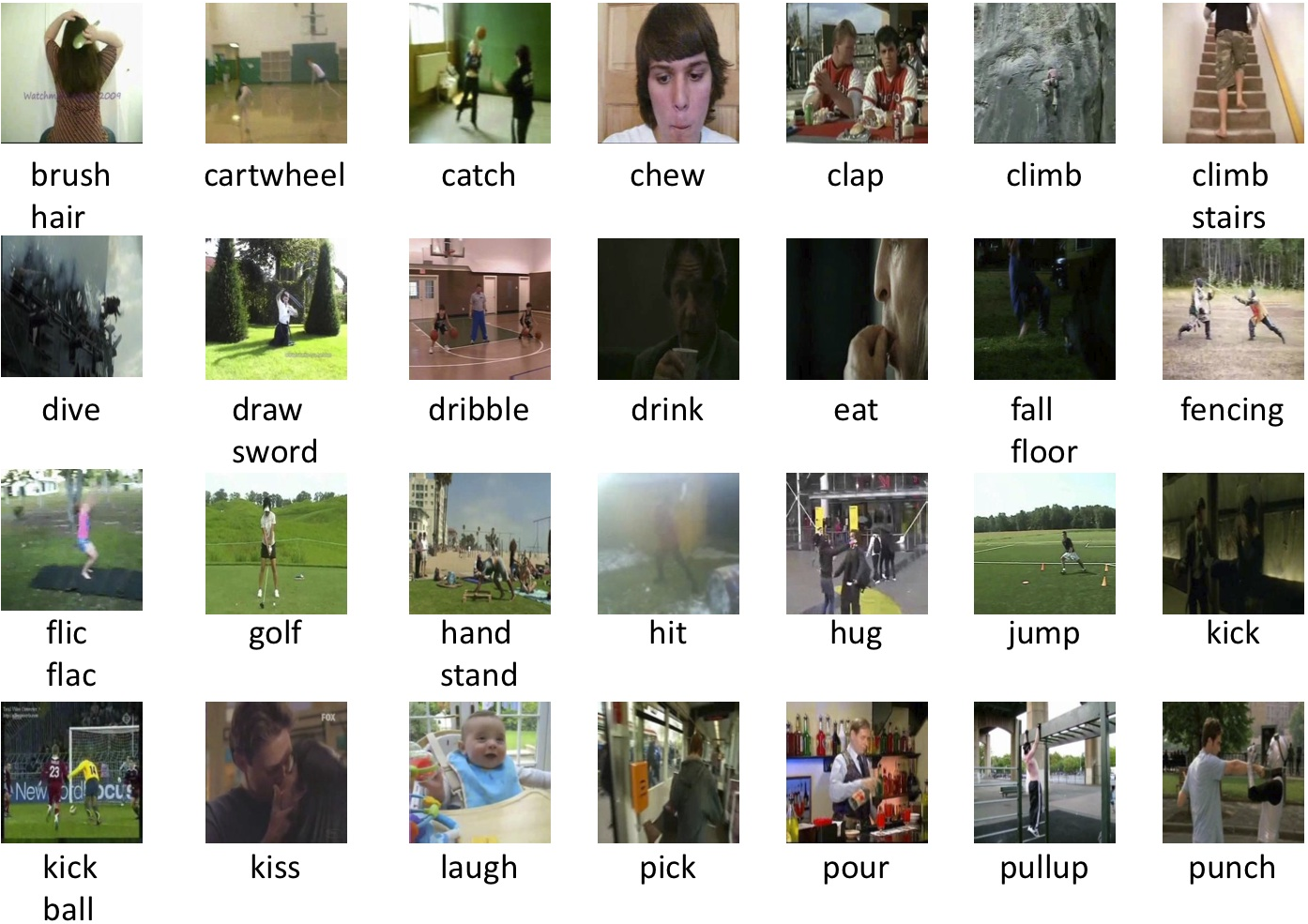}
					\end{center}
				\end{subfigure}

				\begin{subfigure}{\textwidth}
					\begin{center}
						\includegraphics[width=0.9\textwidth]{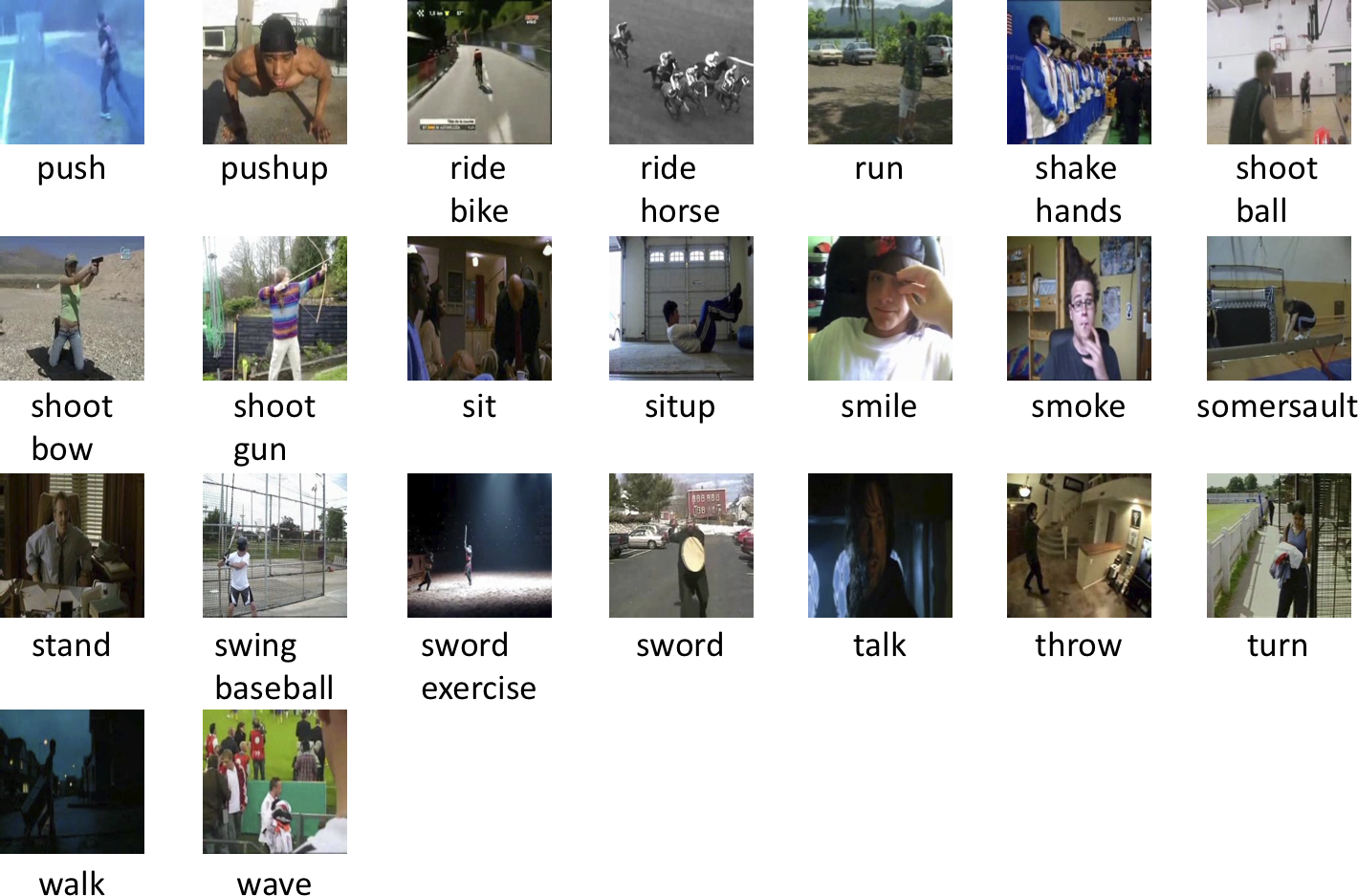}
					\end{center}
				\end{subfigure}
			\caption{HMDB51. 
	Actions in the HMDB51 dataset include (left-to-right): \textit{brush hair}, \textit{cartwheel}, \textit{catch}, \textit{chew}, \textit{clap}, \textit{climb}, \textit{climb stairs}, \textit{dive}, \textit{draw sword}, \textit{dribble}, \textit{drink}, \textit{eat}, \textit{fall floor}, \textit{fencing}, \textit{flic flac}, \textit{golf}, \textit{hand stand}, \textit{hit}, \textit{hug}, \textit{jump}, \textit{kick}, \textit{kick ball}, \textit{kiss}, \textit{laugh}, \textit{pick}, \textit{pour}, \textit{pull-up}, \textit{punch}, \textit{push}, \textit{push up}, \textit{ride bike}, \textit{ride horse}, \textit{run}, \textit{shake hands}, \textit{shoot ball}, \textit{shoot bow}, \textit{shoot gun}, \textit{sit}, \textit{sit-up}, \textit{smile}, \textit{smoke}, \textit{somersault}, \textit{stand}, \textit{swing baseball}, \textit{sword exercise}, \textit{sword}, \textit{talk}, \textit{throw}, \textit{turn}, \textit{walk}, and \textit{wave}. Redrawn from \cite{Kue11}.}
			\label{fig:hmdb51}	
	\end{figure}	

		\begin{table}[!h]
			\begin{center}
				\begin{tabular}{| l | p{0.35\textwidth} | p{0.5\textwidth} |}
					\hline
					 & \textbf{Property} & \textbf{Labels} \\
					\hline
					1 & Visible Body Parts & head, upper body, full body, lower body \\
					2 & Camera Motion & motion, static \\
					3 & Camera Viewpoint & front, back, left, right\\
					4 & Number of People involved in the Action & single, two, three \\
					5 & Video Quality & good, medium, ok \\
					\hline
				\end{tabular}
			\end{center}
			\caption{HMDB51 Dataset Meta Tag Labels \cite{Kue11}.}
			\label{tab:hmdb51_metatag}
		\end{table}
	\end{subsection}
	
	\begin{subsection}{J-HMDB}
To better understand and analyze the limitations and identify components of algorithms for improvement on overall accuracy on the HMDB51 dataset, a joint-annotated HMDB (\textit{J-HMDB}) dataset has been made available \cite{Jhu13}. Among the 51 different human action categories that were collected for the HMDB51 dataset, categories that mainly contain facial expressions (e.g. smiling), interaction with others (e.g. shaking hands), and very specific actions (e.g. cartwheels) were excluded. As a result, 21 classes that involve a single individual performing the action has been chosen, which includes: brush hair, catch, clap, climb stairs, golf, jump, kick ball, pick, pour, pull-up, push, run, shoot ball, shoot bow, shoot gun, sit, stand, swing baseball, throw, walk, and wave. \\
		
		There are 36 to 55 clips per action class with each clip containing about 15-40 frames, summing to a total of 928 clips in the dataset. Each clip is trimmed such that the first and last frames correspond to the beginning and end of an action. All clips have a spatial resolution of $320 \times 240$ with a frame rate of 30 fps. The dataset is randomly split into three distinct sets for evaluation with the condition that the clips from the same video file are not used for both training and testing. For each action category, 70\% of the videos are used for training, and 30\% for testing with a relatively even distribution of the meta tags (e.g. camera position, video quality, motion, etc.). A 2D puppet model for annotation, which represents the human body with a set of 10 body parts connected by 13 joints (shoulders, elbows, wrists, hips, knees, ankles, and neck) and 2 landmarks (the face and the core) are provided to allow researchers to test their algorithms on both the spatiotemporal localization and recognition of the specified actions.		
	\end{subsection}
\end{section}
\begin{section}[Challenges]{Action Recognition and Detection Challenges}
	In efforts to encourage researchers in the vision community to develop action recognition and detection algorithms that can be effectively and efficiently applied in natural settings, an international workshop called the \textit{THUMOS Challenge} took place annually from 2013 to 2015 and ActivityNet Challenge in 2016 in conjunction with various major conferences in computer vision \cite{thu13,thu14,thu15,ActivityNet16}.  Three THUMOS challenges: THUMOS' 13, THUMOS' 14, THUMOS' 15, along with the ActivityNet challenge will be surveyed in this section.
	\begin{subsection}{THUMOS' 13}
		The very first THUMOS challenge, \textit{THUMOS' 13}, which took place in conjunction with the International Conference on Computer Vision (ICCV) in 2013, consisted of two tasks: the \textit{recognition task} and the \textit{detection task}. Both the recognition and the detection tasks were based on videos from the UCF101 dataset (see section \ref{sec:UCF_dataset}). Three training and testing splits were randomly generated such that for each split, 18 of the 25 groups were used as training, and the rest as test data for each action. Each participating team had to submit results to all three training and testing splits that were provided to qualify for the competition. For evaluation, various low-level features (e.g. STIP \cite{Lap05}, SIFT \cite{Low04}, and DT \cite{Wan11} features (see section \ref{sec:feature_descriptors})) with location information, action attributes for the action classes (see Table \ref{tab:thumos13_classlevelattr}), and bounding box annotations (for the detection task) were provided. \\

The objective of the \textit{recognition task} was to predict which action amongst the 101 action classes were present in each test clip. Each team was allowed to submit multiple runs. 17 teams took part in the challenge, and a total of 30 runs were submitted. In this competition, 12 teams made use of low-level features (e.g. (improved) DT feature \cite{Wan11,Wan13}, triangulation of SURF \cite{Nga13}, 3D HOG \cite{Kla08} and HOF \cite{Lap08}, and LPM \cite{Shi11}) (see section \ref{sec:feature_descriptors}), and the rest used newly developed mid-level features (e.g. acton \cite{Zhu13}, online matrix factorization \cite{Cai12}). The most commonly used methods of encoding and pooling were bag-of-words \cite{Siv09} and/or FVs \cite{Jaa98} with a few using spatial/region pooling (see section \ref{sec:encode_features}). The top 10 performing algorithms used VLAD \cite{Jeg10} and/or FV encoding method along with (improved) DT features and an SVM classifier. All teams used either the non-linear or linear SVM for classification with one using neural networks (see section \ref{sec:eager_learner}). Even though action attribute information were provided for all videos, there were no submissions that made use of the class-level attributes to recognize the test data. The baseline recognition result reported on the UCF101 data by November of 2012 was 43.9\% \cite{Soo12}, and the winner of the THUMOS 2013 challenge achieved an overall accuracy of 87.46\% using VLAD+FV-encoded iDT features with a linear SVM \cite{Wan13_thumos13}, which is a significant improvement within a year. \\

The goal of the \textit{detection task} was to localize the bounding boxes provided in the test videos and to identify the 24 pre-defined action classes. 10 of the 24 classes were selected from the UCF11 dataset, which include: basketball shooting, cycling, diving, golf swing, tennis swing, trampoline jumping, volleyball spiking, and walking the dog; and 14 additional classes: basketball dunk, cliff diving, cricket bowling, fencing, floor gymnastics, horseback riding, ice dancing, long jump, pole vault, rope climbing, salsa spin, skateboarding, skiing, ski-jet, soccer juggling, and surfing; were added to the challenge. A detected result was considered correct if the action class was classified correctly and the intersection-over-union \eqref{eq:eval_overlap} was greater than or equal to $0.2$. 
Unfortunately, no team took part in the localization task of the THUMOS' 13 challenge. 
It is worth noting here that although no team took part in the detection task of the THUMOS' 13 challenge, there were algorithms that reported detection results on other datasets, such as the UCF Sports dataset and the MSR Action Dataset II \cite{Tian13}. 

\afterpage{
	\clearpage
	\begin{landscape}
		\begin{table}
			\begin{center} 
				\resizebox{.83\paperheight}{!}{
				\begin{tabular}{| l | l | p{1.3\textwidth} |}
						\bottomrule
						\multicolumn{2}{| l |}{\textbf{Class}} & \textbf{Attributes} \\
						\toprule \bottomrule
						\multicolumn{2}{| l |}{Body Motion} & flipping, walking, running, riding, up down, pulling, lifting, pushing, diving, jumping up, jumping forward, jumping over obstacle, spinning, climbing up, horizontal, vertical up, vertical down, bending \\
						\midrule
						\multicolumn{2}{| l |}{Body Parts Visible} & head close-up, face close-up, upper body, lower body, full body, one hand, two hands \\
						\midrule
						\multicolumn{2}{| l |}{Number of People} & one, two, many\\
						\midrule
						\multicolumn{2}{| l |}{Object} & ball-like, big ball-like, stick-like, rope-like, sharp, circular, cylindrical, musical instrument, portal musical instrument, animal, boat-like \\
						\midrule
						\multicolumn{2}{| l |}{Outdoor} & grass, water, ocean/lake, court, sky, street/road, track, general \\
						\midrule
						\multicolumn{2}{| l |}{Indoor} & pool, office, court, gym, home, track, general\\
						\midrule
						\multicolumn{2}{| l |}{Posture} & sitting, sitting in front of a table-like object, standing, lying, handstand \\
						\midrule
						\multicolumn{2}{| l |}{Body Parts Used} & head, hands, arms, legs, foot \\
						\midrule
						\multirow{6}{*}{Body Part Articulation} & Arm & one arm motion, two arms motion, synchronized arm motion, alternate arm motion, one arm raised over head, two arms raised over head, one arm raised chest level, two arms raised chest level, one arm open to the side, two arms open to the side, one arm down, two arms down, one arm bent, two arms bent, one arm stretched, two arms stretched, one arm swinging, two arms swinging \\
						\cline{2-3}
						 & Leg & synchronized leg motion, alternate leg motion, fold-unfold motion, up-down motion, up-forward motion, side-stretch motion, one leg raise, two legs raise, legs open to the side, one leg bent, two legs bent, one leg stretched, two legs stretched \\
						\cline{2-3}
						 & Hand & throw-release motion, synchronized hand motion, one hand closed, two hands closed, one hand grab, two hands grab, one hand open, two hands open\\
						\cline{2-3}
						 & Head & facing down, facing up, facing front, facing sideways, straight position, tilted position \\
						\cline{2-3}
						 & Torso & down-forward motion, twist motion, bent position, straight up position \\
						\cline{2-3}
						 & Feet & touching ground, in air \\
						\toprule
				\end{tabular}
				}
			\end{center}
		\caption{The 115 class-level attributes assigned to the 101 actions for the THUMOS' 13 Challenge \cite{thu13}.}
		\label{tab:thumos13_classlevelattr}
		\end{table}
	\end{landscape}
}
	\end{subsection}

	\begin{subsection}{THUMOS' 14}
		The second THUMOS challenge, \textit{THUMOS' 14}, took place the following year in conjunction with the 2014 European Conference on Computer Vision (ECCV). Similar to the previous THUMOS challenge, there were two main tasks in the THUMOS' 14 challenge: the \textit{recognition task} and the \textit{temporal action detection task}. The goal of the \textit{recognition task} remained the same as the previous year, which was to predict the presence/absence of an action class in a given sequence. The objective of the \textit{temporal action detection task}, however, was to identify \textit{when} \textit{which} of the pre-defined 20 actions had occurred in the test clip without providing the spatial location. For both tasks, four types of data were provided: training, validation, background, and test. The training data were videos extracted from the UCF101 dataset, which were temporally trimmed such that each sequence contained one instance of the action and all irrelevant frames were removed. The other three parts (validation, background, and test data), on the other hand, were collections of untrimmed videos. As in the THUMOS' 13 challenge, pre-computed low-level feature of the iDT features along with the spatiotemporal information were provided for all (training, validation, background, and test) datasets. Each team was granted at most five submissions of the results for each task, where the run with the best performance was used to rank across other results. \\
	
For the action recognition task, the entire UCF101 dataset of temporally trimmed videos was provided for training. The validation set contained 10 untrimmed videos for each class tallying $1,000$ videos in total to allow participants to fine-tune their algorithms and to use as further training data, if necessary. Each validation video contained a primary action with some containing one or more instances of other action classes. The background data, which contained $2,500$ clips, were videos relevant to each action, but did not contain an instance of any of the 101 action classes. For example, a clip of a basketball court without a basketball game taking place was provided as background data for ``basketball dunk''. Background data provided verification of the absence of action classes. The test data consisted of $1,574$ temporally untrimmed test videos, which contained one or multiple instances of one, multiple, or none of the action classes were provided as test data. 11 teams took part in the challenge and 35 runs were submitted. 10 participants used DT features while 4 used CNNs. In addition, 9 teams used FVs in conjunction with iDT features (see section \ref{sec:feature_descriptors} and section \ref{sec:assignment}). Beyond low-level features, participants used various mid-level features such as face, body and eye features, audio, saliency features, and shot boundary detection. 10 teams used SVM for classification and one team used extreme learning \cite{Hua12,Var14}. 
Using \eqref{eq:AP} and \eqref{eq:mAP}, the winner of the THUMOS' 14 action recognition challenge achieved  an mAP score of $0.71$ by using iDT features with CNN and SVM as a classifier. The THUMOS' 14 recognition task was deemed more challenging than the previous year's as the test videos were temporally untrimmed, which meant that significant portion of some videos did not contain any of the 101 actions.  Furthermore, variations of instances, where multiple or no instance of any actions were possibilities in test videos, was another factor that made the classification task more challenging than the previous year's. These added features in the test videos were embedded to the competition to guide the next generation of action recognition algorithms to be more useful in practical settings.
\\ 

From the task of spatial and temporal detection, the THUMOS' 14 detection challenge had been mitigated to temporal detection. The task mitigation led to computational complexity and annotation alleviation. Instead of 24 action classes as in the previous year's challenge, the detection task called for localization of 20 action classes (baseball pitch, basketball dunk, billiards, clean and jerk, cliff diving, cricket bowling, cricket shot, diving, frisbee catch, golf swing, hammer throw, high jump, javelin throw, long jump, pole vault, shot put, soccer penalty, tennis swing, throw discus, and volleyball spike). Similar to the recognition task, four datasets (training, validation, background, and test) were provided. The training data contained temporally trimmed videos from the UCF101 dataset of the 20 action classes, $200$ validation videos with temporal annotations (start and end time) of all instances of the 20 actions were provided in the validation set, the same set of background data as in the recognition task were provided for the 20 actions, and $1,574$ temporally untrimmed videos were provided as test data. As in the recognition task, interpolated AP and mAP metrics were used to measure performance of each action class and each run, respectively. A detection was considered correct if the IoU score \eqref{eq:eval_overlap} was greater than 0.5 for the predicted time range and ground truth time range. 
3 teams took part in the challenge with 11 submissions in total. All three teams utilized the FV-encoded iDT with CNN features and used 1-vs-rest SVM over temporal windows. The variation amongst the three approaches depended on using either the early or late fusion of the features, system parameters (e.g. window size, step size, hard negatives), post-processing (re-scoring, thresholding), and/or combining with classification scores. The top performing approach, which attained a score of $0.14$ was distinguished in the following three ways \cite{One14}. First, combining the window's detection score with video's classification score for the same action class. Second, using additional features such as SIFT, colour moments, CNN, and MFCC. Third, using ASR in their classification process.
	\end{subsection}
	
	\begin{subsection}{THUMOS' 15}
		The third annual THUMOS challenge, \textit{THUMOS' 15}, took place in conjunction with the 2015 Conference on Computer Vision and Pattern Recognition (CVPR). Identical to previous years' THUMOS challenges, the THUMOS' 15 challenge also comprised of two tasks: the \textit{recognition task} and the \textit{detection task}. The objectives of the recognition and detection tasks remained the same as the THUMOS' 14 tasks, to detect the presence/absence of an action in a given clip and to temporally localize and identify actions in a test video, respectively. Four datasets (training, validating, background, and testing) were provided, as before. The same temporally trimmed $13,320$ videos from the UCF101 dataset were provided for the recognition task and select videos for the chosen 20 actions of the localization task were provided in the training set. $2,140$ and $200$ temporally untrimmed validation videos were provided for the classification and detection tasks, respectively, the same $2,980$ background videos were provided for both tasks, and $5,613$ temporally untrimmed videos were provided for test in both tasks. \\

The same evaluation metrics, AP \eqref{eq:AP} and mAP \eqref{eq:mAP}, as in the THUMOS' 14 challenge, were employed to evaluate the results on each action class and to evaluate the performance of a single run, respectively. The intersection-over-union defined overlap as in the previous challenges \eqref{eq:eval_overlap} was used, where the detection was considered correct if the overlap was greater than $0.5$. 
A total of 11 teams participated in the recognition challenge and 52 runs were submitted. 10 of 11 teams used iDT features and ranked in the top 10 of the competition. Various other methods were employed such as deep networks, MFCC, and multi-granularity analysis (VGG, C3D, iDT, and MFCC). Use of enhanced iDT, multi-granularity analysis (VGG), CNN (LCD), along with MFCC and ASR features and a combination of SVM and a logistic regression fusion classifier allowed the winner of the recognition challenge to attain an mAP score of $ 0.7384$ \cite{Xu15}. Only one team took part in the temporal action detection challenge for which they utilized FV-encoded iDTs, performed multi-granular analysis using VGG and FV, embedded the shot boundary detection method, and used an SVM classifier to attain an mAP score of $0.1830$. With such low participation in the localization task, it is plausible that the datasets for the task had been too computationally demanding and not enough time had been granted for submission.
	\end{subsection}
	
	\begin{subsection}{ActivityNet Challenge}
		In conjunction with CVPR 2016, the \textit{ActivityNet Large Scale Activity Recognition Challenge} took place. Similar to the THUMOS challenges, the ActivityNet Challenge also comprised of two tasks: the \textit{classification task} and the \textit{detection task}. The objective of the classification challenge was to identify the label of the activities that were present in a given long untrimmed video. The detection challenge required an additional challenge of identifying the temporal extents of the activities that were present in the given video. 
Similar to the THUMOS challenges, pre-computed features were provided (e.g. ImageNetShuffle and MBH global features, C3D frame-based features, and agnostic temporal activity proposals).\\

	To evaluate the performance of each algorithm, mAP (equation \eqref{eq:mAP}) and top-$k$ classification accuracy metrics were used. 
The top-$k$ metric, which measures the probability of the correct class attaining the top $k$ confidence score for $k \in \mathbb{Z}^+$, provides additional information about the algorithm, but was not used to determine the winner of the challenge.  
A detection was considered correct if the IoU score \eqref{eq:eval_overlap} was greater than $0.5$.
Only one submission was permitted per participant. A total of 24 participants took part in the classification challenge and 6 in the temporal detection challenge. Algorithms that achieved top 10 performance in the classification challenge either used handcrafted iDT features, deep-learned convolutional features, or its combination to achieve an mAP score greater than 82.5. The winner of the untrimmed video classification challenge achieved an mAP score of 93.2 by analyzing two complementary components of a video: visual and auditory information. The visual system takes an altered two-stream approach adopting the ResNet and Inception V3 architectures, which are aggregated via top-k pooling and attention weighted pooling. The audio system, on the other hand, combines the FV-encoded standard MFCC features trained on SVMs with audio-based CNNs. Many algorithms in the detection task temporally localized actions by either utilizing (i) the sliding temporal window approach or (ii) using LSTM-RNNs. The winner of the action detection challenge achieved an mAP score of 42.5 using VLAD-encoded IDT combined with C3D features on SVM classifiers.
	\end{subsection}

	\begin{subsection}{Final Remarks on the Challenges}
	In this section, four action recognition and detection challenges that took place in conjunction with major conferences were examined. A quantitative summary of the THUMOS' 13, 14, 15, as well as the ActivityNet challenges are provided in Table \ref{tab:THUMOS_summary}. In the upcoming challenge, it is projected that the task of action proposal, whose goal is to retrieve temporal (or spatiotemporal) regions that are likely to contain actions, will be added. Furthermore, the classification task will be based on a larger dataset containing approximately $1,000$ action classes with more than 500 samples per class and the detection task may be extended to the spatiotemporal domain.
\afterpage{
	\clearpage
	\begin{landscape}
		\begin{table}
			\begin{center} 
				\resizebox{.7\paperheight}{!}{
				\begin{tabular}{| c | l | c | c |  p{0.85\textwidth} |}
				\bottomrule
					\textbf{Competition Name} & \textbf{Task} & \textbf{Datasets} & \textbf{No. Videos} & \textbf{Additional Comments} \\
					\bottomrule
					\toprule
					\multirow{4}{*}{\textbf{THUMOS' 13}} & \multirow{2}{*}{Recognition} & Training & $13,320$ & Same as the UCF101 dataset. 3 splits constructed; each split contains 18 training videos of the 25 groups per action. \\
					 & & Testing & $13,320$ & Same as the UCF101 dataset. 3 splits constructed; each split contains 7 test videos of the 25 groups per action.\\
					 \cline{2-5}
					 & \multirow{2}{*}{Spatiotemporal Detection} & Training & - & Videos from the UCF101 dataset of select 24 action classes. \\
					 & & Testing & - & Videos from the UCF101 dataset of select 24 action classes. \\
					\midrule
					\multirow{8}{*}{\textbf{THUMOS' 14}} & \multirow{4}{*}{Recognition} & Training & $13,320$ & Temporally trimmed videos from the UCF101 dataset. \\
					 & & Validation & $1,000$ & Temporally untrimmed data. \\
					 & & Background & $2,500$ &  \\
					 & & Testing & $1,574$ & Temporally untrimmed data. Videos may contain none, one, or multiple instances of a single or multiple action(s). \\
					 \cline{2-5}
					 & \multirow{4}{*}{Temporal Detection}  	& Training & - & Temporally trimmed videos from the UCF101 dataset. \\
					 & 								& Validation & 200 & Temporally untrimmed data. \\
					 & 								& Background & $2,500$ &  \\
					 & 								& Testing & $1,574$ & Temporally untrimmed data. \\
					 \midrule
					\multirow{8}{*}{\textbf{THUMOS' 15}} & \multirow{4}{*}{Recognition} & Training & $13,320$ & Temporally trimmed videos from the UCF101 dataset. \\
					 & & Validation & $2,104$ & Temporally untrimmed data. \\
					 & & Background & $2,980$ & \\
					 & & Testing & $5,613$ & Temporally untrimmed data. Videos may contain none, one, or multiple instances of a single or multiple action(s). \\
					 \cline{2-5}
					 & \multirow{4}{*}{Temporal Detection}  	& Training & - & Temporally trimmed videos from the UCF101 dataset for select 20 actions. \\
					 & 								& Validation & 200 & Temporally untrimmed videos. \\
					 & 								& Background & $2,980$ &  \\
					 & 								& Testing & $5,613$ & Temporally untrimmed data.\\
				\end{tabular}
				}
			\end{center}
		\end{table}
	\end{landscape}
	\clearpage
	\begin{landscape}
		\begin{table}
			\begin{center} 
				\resizebox{.7\paperheight}{!}{
				\begin{tabular}{| c | l | c | c |  p{0.85\textwidth} |}
					\midrule
					\multirow{6}{*}{\textbf{ActivityNet}} & \multirow{3}{*}{Classification} & Training & $10,024$ & Temporally untrimmed data. \\
					&  						& Validation & $4,926$ & Temporally untrimmed data. \\
					&  						& Testing & $5,044$ & Temporally untrimmed data. \\
					\cline{2-5}
					& \multirow{3}{*}{Temporal Detection} & Training & $10,024$ & Temporally untrimmed data. \\
					& 						& Validation & $4,926$ & Temporally untrimmed data. \\
					& 						& Testing & $5,044$ & Temporally untrimmed data. \\
					\toprule
				\end{tabular}
				}
			\end{center}
			\caption{Summary of the THUMOS and ActivityNet Challenges \cite{thu13, thu14, thu15, ActivityNet16}.}
			\label{tab:THUMOS_summary}
		\end{table}
		\end{landscape}
		}
	\end{subsection}
\end{section}

\begin{section}{Summary}
	In this chapter, numerous benchmark datasets have been introduced. Table \ref{tab:summary} summarizes the key features of the commonly used datasets. \\
	
	Although significant progress has been made in collecting data to test various action recognition algorithms, current major datasets are deemed too unrealistic and/or disorderly. The availability of a systematic dataset that consists of naturalistic videos is crucial since the next plausible step in action recognition and detection would be to implement the next generation of algorithms into the real-world. Thus, in constructing the next benchmark dataset, a set of useful actions that make frequent appearance in security, robotics, entertainment, and health care should be considered. Furthermore, the parameters should vary in a systematic way to allow researchers to quickly examine the effect caused by changes in illumination, viewing direction, scale, clutter, recording setting, and performance nuance. 

\afterpage{
	\clearpage
	\begin{landscape}
		\begin{table}
			\begin{center} 
				\resizebox{.83\paperheight}{!}{
				\begin{tabular}{| l | c | c | c | c | c | c | c | c | c |}
					\bottomrule
					\textbf{Dataset} 				& \textbf{Year} & \textbf{No. Actions} & \textbf{No. Actors} & \textbf{No. Videos} & \textbf{Frame Rate} (fps) &\textbf{Cam. View} & \textbf{Cam. Motion} & \textbf{Bckg clutter} &\textbf{Task} \\
					\toprule \bottomrule
					KTH \cite{Sch04}				& 2004	& 6 		& 25 	& 600 & 25 & Frontal/Side 	& No 	&  No	& Recognition, Temporal Detection* \\
					\hline
					Weizmann \cite{Bla05}			& 2005	& 10 		& 9  & 600 & 25 & Frontal/Side 	& No 	& No	& Recognition, Spatiotemporal Detection* \\
					\hline
					MPII Cooking Activities \cite{Roh12} 	& 2012	& 65		& 12	& 44 & 29.4  & Frontal/Side	& No & No	& Recognition, Temporal Detection \\	
					\hline
					MPII Cooking 2 \cite{Roh15} 		& 2015	& 67		& 30 & 273 & 29.4 & Frontal/Side 	& No & No	& Recognition, Temporal Detection \\
					\midrule
					\multirow{2}{*}{CMU Crowded Videos \cite{Ke07}} & \multirow{2}{*}{2007}	& \multirow{2}{*}{5} & \multirow{2}{*}{6} & 5 training & \multirow{2}{*}{25-30} & \multirow{2}{*}{Frontal/Side} 	& \multirow{2}{*}{No} & \multirow{2}{*}{Yes} & \multirow{2}{*}{Recognition, Spatiotemporal Detection} \\
												&		&		&	& 48 test	&		&			&		&	&		\\
					\hline
					MSR Action I \cite{Yua09}			& 2009 	& 3		& 10 	& 16 & 15 & Frontal/Side 	& No 	& Yes & Spatiotemporal Detection \\
					MSR Action II \cite{Cao10}		& 2010 	& 3		& 10+ & 54 & 14-15 & Frontal/Side	& No		& Yes & Spatiotemporal Detection\\
					\midrule
					CMU Sports (Ballet) \cite{Efr03}	& 		& 16 		& 6 	&  &  & Frontal/Side 	& No 	& No & \\
					CMU Sports (Tennis) \cite{Efr03}	& 2003	& 6		& 2 	& N/A & N/A & Side 		& No 	& No & - \\
					CMU Sports (Soccer) \cite{Efr03}	&		& 8		& R	&  &  & Multiple		& Yes	& Yes &  \\
					\hline
					UCF Sports \cite{Rod08,Soo14}	& 2008	& 10 		& R	& 150 & 10 & Multiple 		& Yes	& Yes 	& Spatiotemporal Detection \\
					\hline
					Olympic Sports	\cite{Nie10}		& 2010	& 16		& R	& 783 & N/A & Multiple		& Yes 	& Yes & Recognition \\
					\hline
					Sports-1M \cite{AKar14} & 2014 & 487 		& R	& $1,133,158$ & -	& Multiple& Yes & Yes 	& Recognition	\\
					\midrule
					Hollywood1 \cite{Lap08}			& 2008	& 8		& R	& 475 & 23-25 & Multiple		& Yes	& Yes & Recognition \\
					Hollywood2 \cite{Mar09}			& 2009	& 10 + 6 scenes & R & $2,517$ & 23-29 & Multiple	& Yes	& Yes & Recognition \\
					\midrule
					UCF11 (YouTube) \cite{Liu09}		& 2009	& 11 		& R	& $1,600$ & 29 & Multiple		& Yes 	& Yes & Spatiotemporal Detection \\
					UCF50 \cite{Red12}				& 2012	& 50 		& R	& $6,681^\dag$ & 25 or 29 & Multiple		& Yes	& Yes & Spatiotemporal Detection \\
					UCF101 \cite{Soo12}			& 2012	& 101 	& R	& $13,320$ & 25 or 29 & Multiple		& Yes 	& Yes & Spatiotemporal Detection \\
					\midrule
					ActivityNet \cite{Hei15} 			& 2015 & 203 	& R	& $19,994$ & mostly 30 & Multiple	& Yes & Yes & Recognition, Temporal Detection\\
					\hline
					HMDB51 \cite{Kue11}			& 2011	& 51		& R	& $6,766$ & 30 & Multiple		& Yes	& Yes & Recognition \\
					J-HMDB \cite{Jhu13}	& 2013	& 21	& R	& 928 & 30 & Multiple & Yes & Yes & Recognition, Spatiotemporal Detection \\
					\toprule
				\end{tabular}
				}
			\end{center}
			\caption{Summary of Benchmark Datasets. \textit{R} indicates that the datasets were extracted from realistic videos. Thus, the number of actors cannot be determined. (*) Although the intended use of these datasets is to recognize actions, the authors provide ground truths (e.g. start and end frames, silhouettes) allowing the evaluation of temporal/spatiotemporal detection possible. (\dag) \ The official report of the UCF50 dataset \cite{Red12} documents a total of 6676 videos in the UCF50 dataset. However, the downloadable UCF50 dataset contains 6681 videos.}
			\label{tab:summary}
		\end{table}
	\end{landscape}
}
\end{section}

%% file: chapters/chapter3_representation.tex
\begin{figure}[htbp]
	\begin{center}
	\includegraphics[width=0.9\textwidth]{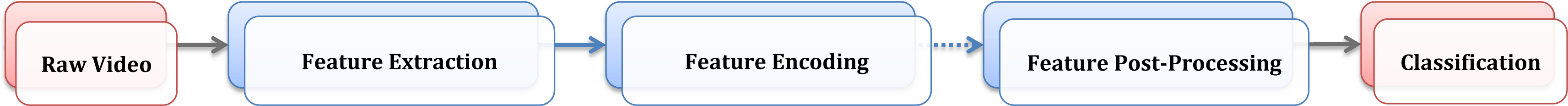}
	\end{center}
	\caption{General overview of the contents that this chapter will entail. Blue boxes indicate the steps that will be covered in this chapter and the dashed line indicates that \textit{feature post-processing} is an optional step prior to the classification stage. \label{fig:overview}}
\end{figure}

In order to categorize an action in an efficient and accurate manner, features that provide meaningful information must be gathered and encoded for classification. Ideally, the representation model should be robust to variation in appearance of the actor(s), background, viewpoint, and performance nuance while preserving sufficient information to accurately classify the action. To overcome this barrier, a plenitude of representation models have been introduced. 
In this review, representation models will be organized according to the general sequence of steps that are taken to extract features from raw input videos. This procedure involves transforming the raw data in videos into features then encoding these features before they enter the classification stage (see Figure \ref{fig:overview}). In this chapter, various methods to obtain useful features (section \ref{sec:feature_representation}) and encoding methods (section \ref{sec:encode_features}) that have appeared in the field of action recognition and detection will be explored. In some algorithms, the resulting feature representation or encoding model has led to excessive and redundant data, thus features have been post-processed to overcome this issue and will be examined in section \ref{sec:dim_reduce}. 

\begin{section}{Feature Extraction \label{sec:feature_representation}}
	A raw input video is made of voxels, where each voxel contains photometric information, such as intensity or RGB values. This lattice of raw information must be transformed into some representational model such that it can be processed in its subsequent classification stage. To transform this raw data into informative features, useful information must first be extracted then represented in some form. In this section, various approaches to sampling input video data and subsequently extracting primitive feature descriptors will be examined. 
	\begin{subsection}{Sampling Methods}\label{sec:sampling}
		\input{chapters/chapter3a_sampling}
	\end{subsection}

	\begin{subsection}{Feature Descriptors}\label{sec:feature_descriptors}
		\input{chapters/chapter3b_descriptors}
	\end{subsection}

\end{section}

\begin{section}{Encoding Methods}\label{sec:encode_features}
Primitive features extracted from videos are often selected in a generic way, which are not specific enough to directly serve the given task. Consequently, it can be beneficial to encode primitive features with a representation that is specifically designed to serve the assigned task through an encoding procedure. There are a variety of different encoding procedures to convert primitive features, $\mathbf{f}(\mathbf{x}) \in \mathbb{R}^d$, to a more effective encoded representations, $\mathbf{c}(\mathbf{x}) \in \mathbb{R}^k$, where $\mathbf{f}(\mathbf{x})$ is a $d$-dimensional local descriptor extracted from a video at $\mathbf{x}=[x \ y \ t]^\top$, and $\mathbf{c}(\mathbf{x})$ is a $k$-dimensional encoding vector of $\mathbf{f}(\mathbf{x})$\footnotemark \cite{Cha11}. In general, the descriptor space must initially be converted into a codespace via \textit{codebook generation}. Second, the features must be encoded to correspond to the newly defined space through \textit{feature assignment}. In some cases, the amount of encoded data needs to be reduced (\textit{pooled}) and/or \textit{normalized} such that the data type is consistent with other data. In this section, three key steps involved in encoding feature descriptors, codebook generation, feature assignment, and pooling and/or normalization, as illustrated in Figure \ref{fig:encoding_methods}, will be examined.\\
\footnotetext{From here on, $\mathbf{f}$ and $\mathbf{c}$ will be used in replacement of $\mathbf{f(\mathbf{x})}$ and $\mathbf{c}(\mathbf{x})$, respectively, for brevity.}
	
	\begin{figure}[htbp]
		\begin{center}
			\includegraphics[width=0.9\textwidth]{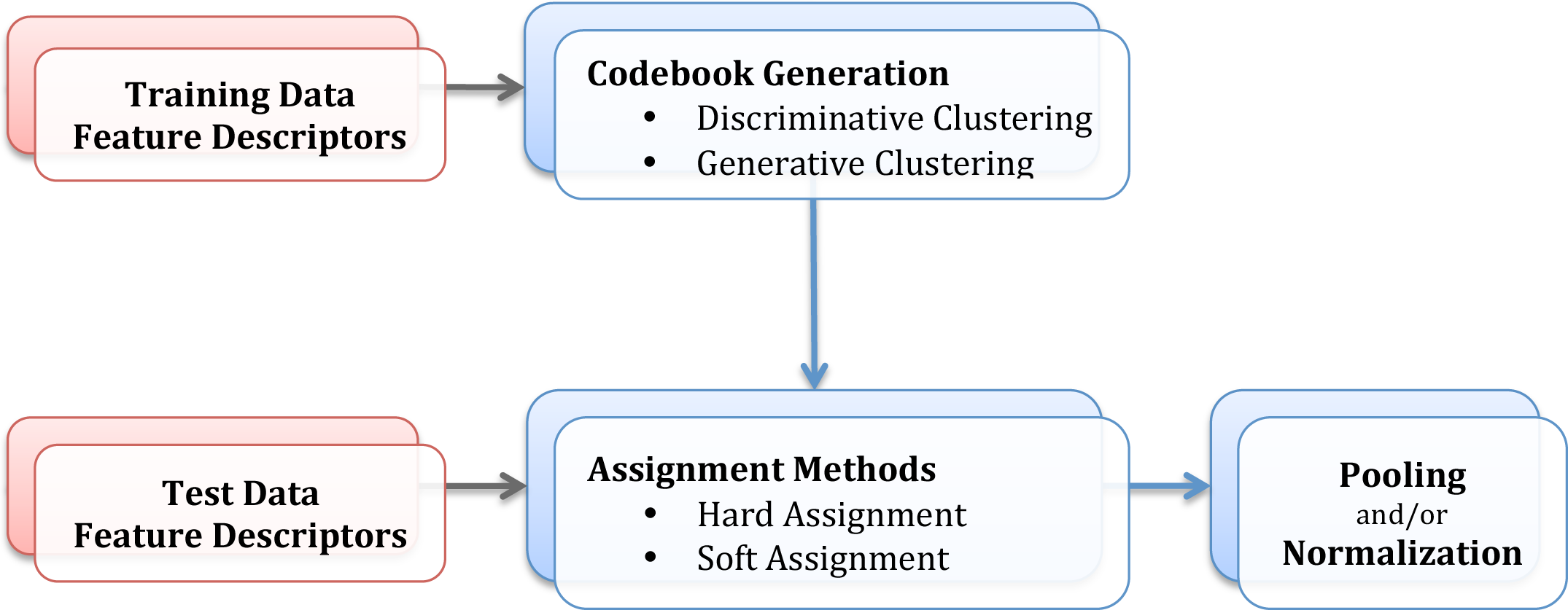}
		\end{center}
		\caption{General framework of encoding feature descriptors. The stages that are involved in feature encoding are marked in blue and its prior steps are marked in red. \label{fig:encoding_methods}}
	\end{figure}

	\begin{subsection}{Codebook Generation}\label{sec:grouping}
		\input{chapters/chapter3d_grouping}
	\end{subsection}
	
	\begin{subsection}{Assignment Methods}\label{sec:assignment}
		\input{chapters/chapter3e_encoding}
	\end{subsection}
	
	\begin{subsection}{Pooling and Normalization}\label{sec:poolNnorm}
		\input{chapters/chapter3f_pooling}
	\end{subsection}

	\begin{subsection}{Discussion on Encoding Methods}
		The order, choice, and combination of codebook generation, assignment, pooling, and normalization can all affect the final outcome of the classification problem. Even though the major stages of encoding were presented as: codebook generation, assignment, pooling and normalization, it does not suggest that the optimal performance will be attained by following this exact sequence of steps. In fact, pooling and/or normalization can appear at any stage of encoding, if either or both stages are deemed helpful at all. \\

		It was mentioned in the codebook generation section that increasing the size of the codebook (i.e. the number of codewords) to a certain point improves the accuracy of the recognition. Furthermore, it was pointed out that soft contrasting assignment methods retain richer information between codewords and feature vectors (e.g. dissimilarities between features and codewords). Together, they allow soft contrasting assignment methods to allow for a smaller codebook than other assignment models to achieve a similar level of performance \cite{Pen14}. \\

		In the normalization section, it was briefly discussed that power-normalization has a smoothing effect on histograms. When power-normalization is combined with sum- or average-pooling, a very good result can be obtained since sum-pooling produces sharp and unbalanced histogram. Thus, the smoothening effect of power-normalization and the sharpening effect of sum-pooling pair well together to balance the smooth-sharp effects. 
		Noting that FVs are based on a codebook constructed using GMMs, FVs implicitly perform an average pooling as it computes the first-order statistics to obtain the FV. As a result, FVs have been shown to perform well with average-pooling \cite{Fei16}. Consequently, power-normalization is the most well-suited normalization method with FVs \cite{Per10}. \\
		
		A synergistic relationship can be observed between a well-chosen pair of assignment and pooling methods. For assignment models that pursue a sparse representation, the optimal pooling method is via max-pooling \cite{Ser07}. Max-pooling couples well with sparse data since the distance between the nearest codeword and the feature vector is significantly closer than with other codewords inducing a strong response. The strong response is preserved and weaker responses are discarded through max-pooling \cite{Hua11}. In fact, it was empirically confirmed that SpC, an assignment model that pursues a sparse representation, and LLC, an assignment method that eventually leads to sparsity through its locality constraint, is best pooled via max-pooling \cite{Wan10,Yan09}.\\

		Another factor that cannot be overlooked when choosing the type of encoding is the type of classifier in the subsequent step. That is, to use a linear SVM over non-linear SVMs for its efficiency and smaller memory requirement, $l_2$-normalization would be the preferred normalization method since the inner product of any vector with itself is an identity in $l_2$-normalization, which ensures that the vector compared to itself is the most similar. This trait warrants stability during training \cite{Pen14, Wan10}. Thus, although the sharp characteristic induced by $l_2$-normalization on FVs can be resolved via $l_1$-normalization, power-normalization is preferred since $l_1$-normalization suggests the use of non-linear SVMs as opposed to linear SVMs in the succeeding classification step \cite{Per10}. \\

		There is a plethora of choices for each step in the encoding framework. The selection of encoding can greatly impact the final classification performance \cite{Cha11}. Since each choice within the pipeline are highly inter-related, they should be chosen with care. Although many gaps have been filled to determine which combination would yield the most ideal encoding framework (e.g. FVs with sum-pooling, power- and $l_2$-normalization for linear SVMs), extensive research is still in need to bridge the theoretical gap between all existing choices within each step. 
	\end{subsection}
\end{section}

\begin{section}[Feature Post-processing]{Feature Post-processing}\label{sec:dim_reduce}
	\input{chapters/chapter3c_dimreduce}
\end{section}

\begin{section}{Final Remarks}
	In this chapter, three major steps that are involved in representing images were examined: feature extraction, feature encoding, and feature post-processing. The feature extraction stage and the encoding stage can occur once or multiple times as needed before it enters the final classification stage \cite{Jhu07, Nin09}. Furthermore, although dimensionality reduction may improve the accuracy and efficiency of an algorithm, it is not a necessary procedure and can occur before or after the encoding stage. 
\end{section}

%% file: chapters/chapter3a_sampling.tex
\begin{figure}[htbp]
	\begin{center}
		\includegraphics[width=0.95\textwidth]{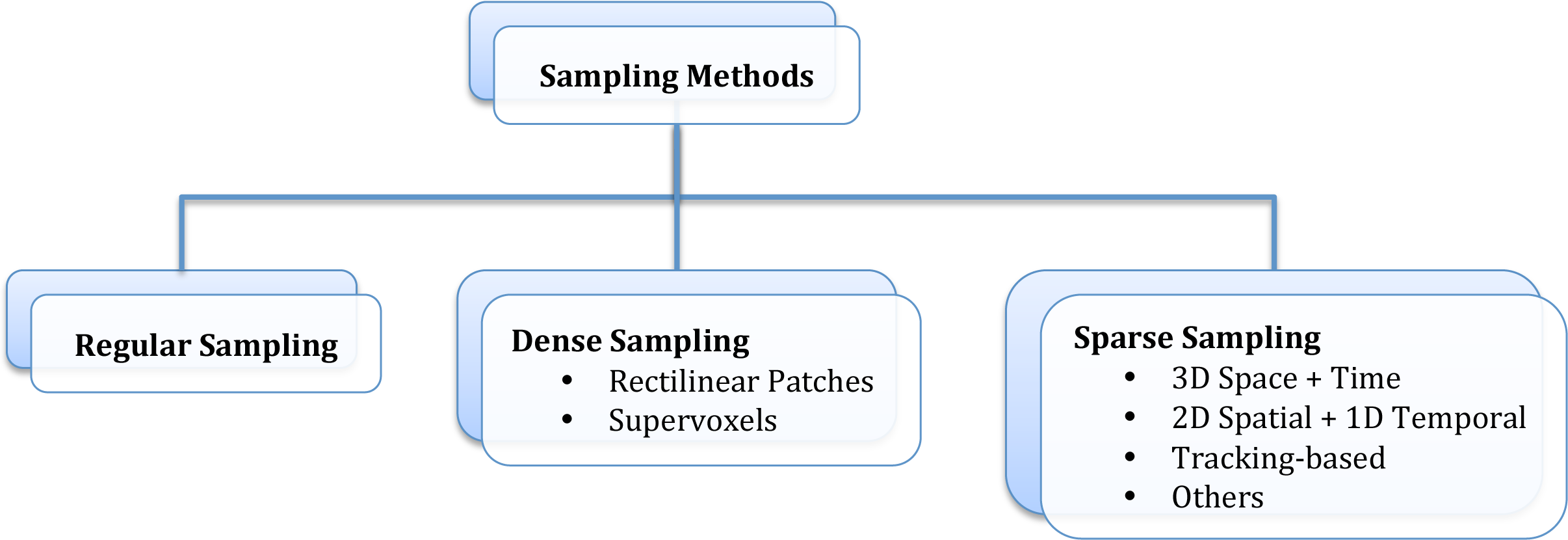}
	\end{center}
	\caption{General breakdown of the sampling methods. Data can be sampled from videos through regular, dense, or sparse sampling methods. Although these sample methods are described as independent entities, regular sampling at every interval is equivalent to dense sampling the entire video as would setting the threshold to zero for any response function in sparse sampling. \label{fig:sample_breakdown}}
\end{figure}

Information from a video can be sampled in three ways: through (i) regular sampling, (ii) dense sampling, or (iii) sparse sampling (see Figure \ref{fig:sample_breakdown}). 
In \textit{regular sampling}, data is obtained at every $n$ voxels, where $n \in \mathbb{Z}^+$, and if $n=1$ then the entire data of the video is used. In \textit{dense sampling}, a video is divided into either rectilinear patches or as more irregular supervoxels. In \textit{sparse sampling}, salient regions within a video are localized by optimizing some saliency function. In the following, various types of dense and sparse sampling techniques that have appeared in the field of action recognition and detection will be studied\footnotemark.
\footnotetext{Further details on regular sampling are omitted for its simplicity and lack of variability in the field of action recognition.}

\begin{subsubsection}{Dense Sampling Methods}
	Videos can be partitioned into simple rectilinear patches or supervoxel segments according to proximity, similarity, and continuation \cite{Xu13}. Numerous supervoxel algorithms have appeared in computer vision and various methods have been used as a pre-processing step to solve action recognition problems, such as \textit{mean shift} \cite{Ke07}, \textit{streaming hierarchical supervoxel method} \cite{Xu13}, and \textit{SLIC} \cite{Fei15}. 
Common to all, supervoxel region extractors is a critical parameter (or \textit{kernel bandwidth size}) that determines the size of the objects to be segmented. A small bandwidth correctly segments small objects but tends to over-segment large objects into multiple parts. Conversely, a large bandwidth correctly segments large objects but incorrectly groups small objects together. Therefore, even though a rich set of supervoxel methods have appeared in the field of computer vision, its utilization in action recognition remains under-explored partly because it is expected that an entire object will not be segmented as a single region in a typical realistic video. Thus, use of supervoxels is perceived as groupings of video-based features for object and region labelling \cite{Xu13}.
However, the borders created by the supervoxels can provide crude information on the boundaries between objects (see Figure \ref{fig:supervoxel}) without relying on the unsolved background-subtraction problem \cite{Ke07}. Furthermore, supervoxels can be used as weighing functions to distinguish motion created by the actor, camera, and the background \cite{Che14,Fei15}. 

	\begin{figure}[htbp]
		\begin{center}
			\includegraphics[width=0.95\textwidth]{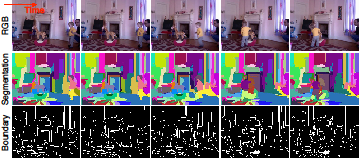}
			\caption{Example of an input video (top row), its corresponding supervoxel segmentation (middle row), and the boundaries of the supervoxel segmentation. Redrawn from \cite{Xu13}.}
			\label{fig:supervoxel}		
		\end{center}
	\end{figure}
\end{subsubsection}

\begin{subsubsection}{Sparse Sampling Methods}
	Representing every voxel of a video can be computationally taxing especially for benchmark datasets that contain thousands of videos, like UCF101, HMDB51, and ActivityNet. Correspondingly, there has been extensive research to avoid the computational burden of processing entire videos in large datasets \cite{Pen14,Sha12,Wan09,XWang13}. A video can be sampled sparsely at regular grid points or by extracting interest points or regions. In images, interest points often refer to regions with corners, blobs, and junctions. Likewise, \textit{spatiotemporal interest points} (STIPs) in videos can be considered as three-dimensional corners, blobs, and/or junctions, which can be detected by maximizing some response function. The construction of a three-dimensional response function for videos can be done by either generalizing a two-dimensional interest point detector in images to three-dimensions or by combining a two-dimensional interest point detector with a one-dimensional detector to compensate for the extra temporal domain in videos. In the following, various sparse sampling methods that extract STIP by (i) generalizing the two-dimensions in images to three-dimensions in videos, (ii) a combination of two-dimensional spatial domain with one-dimensional temporal domain, (iii) tracking two-dimensional interest points, and (iv) others, will be explored.
		
\begin{paragraph}{Direct Extensions of 2D Detectors}
	Sampling methods that have been successful at extracting interest points in images can be directly extended to the third-dimension by assuming that the temporal domain in videos is analogous to a third dimension of space. 
	In order to detect multi-scale interest points in videos, a spatiotemporal scale-space representation of a video sequence must initially be defined. Then a saliency map can be constructed to extract spatiotemporal interest points \cite{Sha12}. An image sequence, $I$, at point $\mathbf{x} = [x \ y \ t]^\top$ can be modelled in linear scale-space by taking the convolution of $I$ with a Gaussian kernel $g$: 
	\begin{equation}\label{eq:image_representation}
		L(\mathbf{x}| \sigma_0^2, \tau_0^2) = g(\mathbf{x}| \sigma_0^2, \tau_0^2) \ast I(\mathbf{x}) \text{,}
	\end{equation}
	where $\sigma_0$ and $\tau_0$ denote distinct spatial and temporal scales, respectively. \\

	One of the most common 2D corner detector for images is the Harris detector, which can be generalized to \textit{Harris 3D detectors} \cite{Lap05,Lap08} to detect 3D corners in videos by averaging the spacetime gradients $\mathbf{\nabla} L$ with a Gaussian weighting function:
	\begin{equation}
		H_1(\mathbf{x}| \sigma_1, \tau_1) = g(\mathbf{x}| \sigma_1^2, \tau_1^2) \ast 
						\begin{bmatrix}
							L_x^2 	& L_x L_y 	& L_x L_t \\
							L_x L_y	& L_y^2	& L_y L_t \\
							L_x L_t 	& L_y L_t 	& L_t^2
						\end{bmatrix}
		\text{,}
	\end{equation}
		where $L_x$, $L_y$, and $L_t$ denote first-order partial derivatives of $L$ with respect to $x$, $y$, and $t$, respectively.
	Spatiotemporal interest points are obtained by detecting the local positive maxima of the following function:
	\begin{equation}
		S_1 = \det{(H_1)} - k {[\Tr{(H_1)}]^3} \text{,}
	\end{equation}
	for some constant $k$. The Harris 3D detector is suited to detect spatial corners that change motion direction, like start or stop of some local motion in a video \cite{Sha12}. \\
		
	Another common interest point detector that appears often in images is the Hessian detector. The \textit{Hessian detector} \cite{Wil08} in images can be directly extended to videos by defining the Hessian matrix in 3D as:
		\begin{equation}\label{eq:3d_hessian}
			H_2(\mathbf{x}| \sigma_2^2, \tau_2^2) = 
				\begin{bmatrix}
					L_{xx} & L_{xy} & L_{xt} \\
					L_{yx} & L_{yy} & L_{yt} \\
					L_{tx} & L_{ty} & L_{tt}
				\end{bmatrix}
			\text{.}
		\end{equation}
	Regions with a local maxima of the determinant of the 3D Hessian (i.e. $S_2 = \left| \det{H_2} \right|$) for some particular position and scale correspond to a centre of a blob in a video \cite{Sha12}.
\end{paragraph}

\begin{paragraph}{2D (Spatial) Detector with a 1D (Temporal) Detector}
	Beyond varying the scale-space support in space and time separately via constants $\sigma$ and $\tau$, the temporal dimension can be managed by generating an even more distinct filter in the temporal domain. The temporal domain can be treated differently from the spatial domain 
	by applying distinct filters for each domain \cite{Dol05, Nie08}. The \textit{cuboid detector} \cite{Dol05} couples a Gaussian filter in the spatial domain and a Gabor filter in the temporal domain to create a response function that is applicable in the spatiotemporal domain. For a given video $I(\mathbf{x})$, the response function is defined as:
	\begin{equation}\label{eq:cuboid_detector}
		R(\mathbf{x}) = {\left[ I(\mathbf{x}) \ast g(x,y|\sigma) \ast h_{even}(t|\tau,\omega) \right]}^2 + {\left[ I(\mathbf{x}) \ast g(x,y|\sigma) \ast h_{odd}(t,\tau,\omega) \right]}^2 \text{,}
	\end{equation}
	where $g(x,y| \sigma)$ is the 2D Gassian smoothing kernel applied along the spatial dimensions $(x,y)$, and $h_{even}(t|\tau,\omega) = -\cos{(2\pi t \omega) e^{-\sfrac{t^2}{\tau^2}}}$ and $h_{odd}(t|\tau,\omega)=-\sin{(2\pi t \omega)e^{-\sfrac{t^2}{\tau^2}}}$ are quadrature pair of 1D Gabor filters applied along the temporal domain $t$. $\sigma$ and $\tau$ correspond to spatial and temporal scales of the detector, respectively, and $\omega$ the centre frequency\footnotemark. 
 It can be observed that the cuboid detector is best matched to an intensity pattern that oscillates sinusoidally along the temporal dimension and smoothed in the spatial dimension with a low-pass (Gaussian) filter.
Conversely, the smallest response would be generated in regions that lack temporally distinguishing features. 
Hence, it is well suited to detect temporally varying patterns even while providing little response to those that remain static.
In comparison to the aforementioned detectors, 3D Harris and Hessian, the cuboid detector extracts a denser set of features and is consequently computationally more expensive to follow-on processing \cite{Wan09}.
	\footnotetext{The centre frequency for the Gabor function refers to the frequency in which the filter yields the greatest response. $\omega$ can be set to $\sfrac{4}{\tau}$ to reduce the number of parameters involved in equation \eqref{eq:cuboid_detector} \cite{Dol05,Nie08}.}
\end{paragraph}

\begin{paragraph}{Tracking-based Detectors}
	Determining good features to track is an alternative approach to obtaining a useful set of sample points. Since points found in structureless regions are impossible to track, it would be helpful to remove them from the sampling set. The decision to retain or remove a point can be made using the \textit{good-features to track} criterion \cite{Shi94}, which is determined by the eigenvalues of the auto-correlation matrix, a matrix intimately related to 2D Harris. This sampling technique is incorporated in the (improved) dense trajectory features \cite{Wan11, Wan13}, which has shown to be very effective as it is one of the strongest contemporary features in application to action recognition.
\end{paragraph}

	\begin{paragraph}{Other Sparse Sampling Methods}
		There are many other sparse sampling methods that were not mentioned in detail, such as the Harris-Laplace \cite{Mik01}, Hessian-Laplace \cite{Mik04}, Difference of Gaussian (DoG) \cite{Low04} and maximally stable extremal region (MSER) \cite{Mat04} detectors. The \textit{Harris-Laplace}, which uses the Harris and Laplacian functions to find and select points, respectively, is capable of detecting corners and other junctions, pairs and triplets of edge segments to represent contours invariant of scale and rotation changes \cite{Mik08}. The \textit{Hessian-Laplace} localizes points in space and scale by taking the local maxima of the determinant of a Hessian and the Laplacian-of-Gaussian, respectively \cite{Mik05}. Since the shape of the Hessian kernel fits better to blob-like structures than corners, the Hessian-Laplace detector is used to extract various types of blobs \cite{Mik08}. The DoG detector, which is often used in accordance with a 3D histogram of gradient location and orientation and together referred to as SIFT, uses the difference of images of different scales convolved with a Gaussian function to identify the locations of edges and blob-like structures. \textit{MSER} extracts blobs by expanding regions according to their intensity levels by gradually increasing some threshold value. The value that enforces the smallest rate of change is selected as the threshold to extract MSER and has shown to provide useful detection results \cite{Lin09,Mik08,Uem08}.\\ 

The extracted features can be pruned using spatial, temporal, or motion statistical measures \cite{Liu09}. Excessive amount of features can be judged by comparing the number of features extracted in a single frame to the average amount of features present per frame. Spatial outliers can be spotted using neighbourhood information. Lastly, PageRank \cite{Mol02,Pag99} can be used to identify consistency of the extracted feature to others to classify them as inliers.
	\end{paragraph}
\end{subsubsection}

\begin{subsubsection}{Discussion on Sampling Methods}
Regular, dense, and sparse sampling methods have been described as independent entities in this section, but we must bear in mind that these methods are not disjoint. That is, regularly sampling at every interval would be equivalent to dense sampling the entire video, which is equivalent to setting the threshold to 0 for any response function in sparse sampling. \\

	Videos that largely consist of static backgrounds that pose no useful information to recognize actions (e.g. videos in the KTH dataset) benefit from sparse sampling as features obtained through dense sampling provide no useful data \cite{Wan09}. Furthermore, extracting features sparsely across videos provide data compactness leading to computational efficiency. When coupled with appropriate descriptors and classifiers (to be described in more detail in the following section and chapter, respectively), these detectors extract sufficient data to acceptably differentiate between human actions. However, it was observed that sparse sampling methods fall behind the accuracy in recognition that dense (or regular) sampling methods are able to provide, especially in videos with contextual information (e.g. UCF Sports, Hollywood2) \cite{Wan09}. This result may be due to the fact that (i) the data extracted using these detectors tend to be too sparse and (ii) the contextual information, such as equipment or scene, can provide additional information to improve classification results. Furthermore, many saliency functions that are used to extract features assume that videos contain several instances of motion or appearance that are significantly different in either direction of motion or the boundary between the background and the actor. This assumption leads to failure in capturing smooth motions (as in Figure \ref{fig:stip_fail1}) and generates spurious detects along object boundaries (see Figure \ref{fig:stip_fail2}) \cite{Ke05}. \\

The sparse motion detectors mentioned in this paper (e.g. cuboid detector, KLT tracker, DT) can be used in motion compensated or non-compensated videos. These detectors are expected to fire at the presence of motion whether it be camera motion or motion created by different body parts of an actor. Often in action recognition, it is understood that motion created by the object's body provides useful information. Thus, the output results of these detectors must be used with caution as they may respond to some dominant motion due to camera movement or an actor occupying a large portion of the field of view, which may or may not be the desired information that one wishes to obtain for their recognition algorithm. \\

The choice of data extraction can affect the computational efficiency but can also influence the accuracy of the recognition step as sampling is the first step in the recognition procedure. Thus, the data extraction technique must be chosen with caution as it can heavily influence or deter the outcome of the results in following processing steps.

\begin{figure}[htbp]
	\begin{subfigure}{0.95\textwidth}
		\begin{center}
			\includegraphics[width=0.5\textwidth]{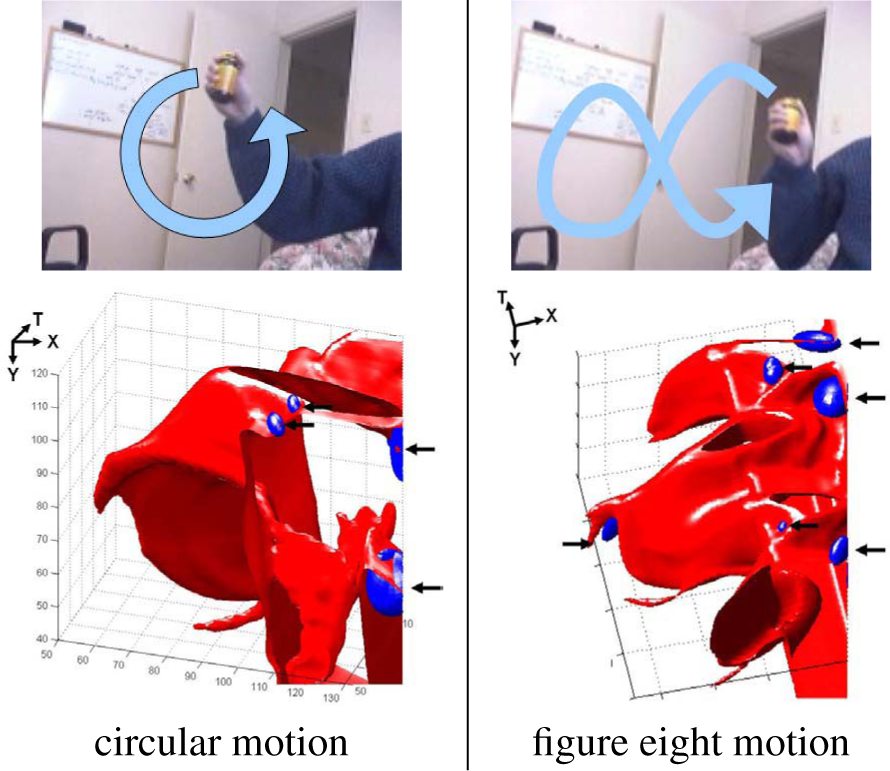}
			\caption{Blue arrow indicates the direction of motion. Two motions are illustrated in this example: circular motion (left) and the figure `8' motion (right). The 3D plots of motion through time are illustrated (bottom) with blue ellipsoids showing detected interest points. All detected interest points were non-informative, and were only detected due to the boundaries that formed as the arm moved with the edge of the frame. \label{fig:stip_fail1}}
		\end{center}
	\end{subfigure}
	
	\begin{subfigure}{0.9\textwidth}
		\begin{center}
			\includegraphics[width=0.8\textwidth]{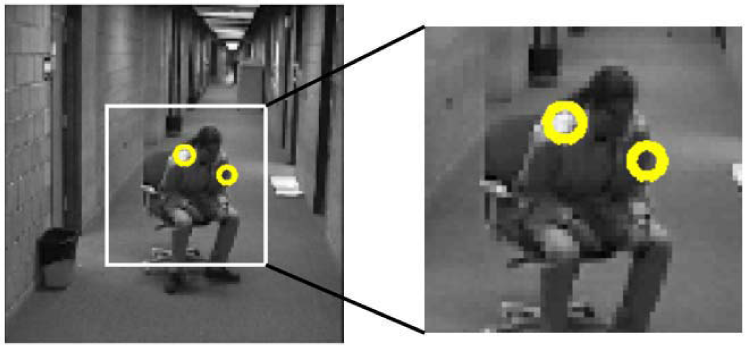}
			\caption{Spacetime interest points detected on regions affected by varying lighting conditions. STIP detectors are sensitive to lighting conditions, therefore are detected in regions with bright light or shadows. \label{fig:stip_fail2}}
		\end{center}
	\end{subfigure}
	\caption{Examples of commonly occurring motions that fail to produce useful interest points. Redrawn from \cite{Ke05}.}
\end{figure}
\end{subsubsection}

%% file: chapters/chapter3b_descriptors.tex
\begin{figure}[htbp]
	\begin{center}
		\includegraphics[width=0.9\textwidth]{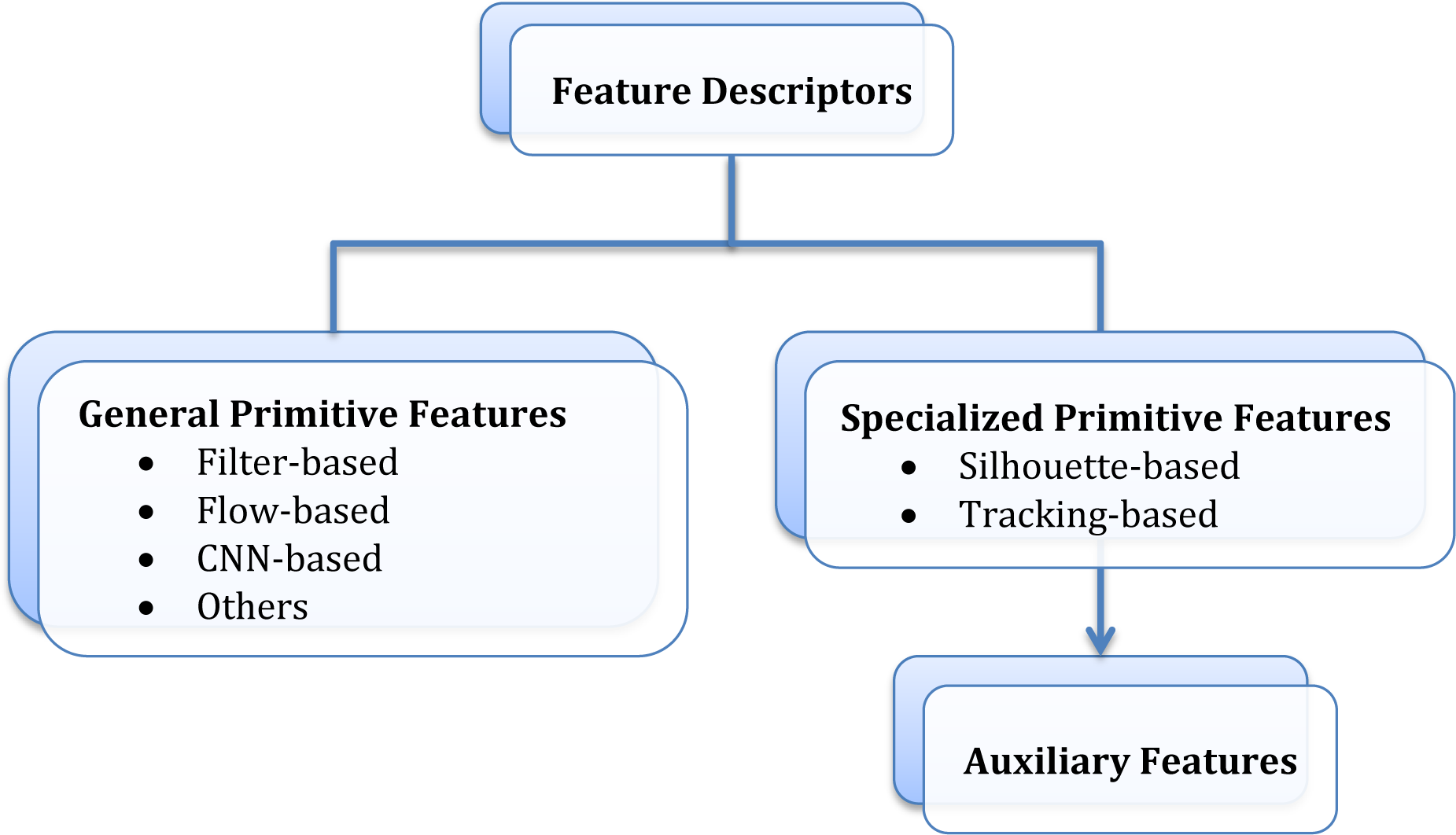}
	\end{center}
	\caption{General breakdown of feature descriptors. Features can be obtained from raw videos by describing them using \textit{general primitive features} or \textit{specialized primitive features}. While general primitive features can be used to train and test data immediately, specialized primitive features must be further processed into \textit{auxiliary features} before the features enter the classification stage.} \label{fig:descriptor_breakdown}
\end{figure}

Once a sampling method has been selected, information that would characterize the structure of the region must be represented in some useful way as a \textit{descriptor} before it enters the classification stage. 
In the following, the feature descriptors have been split into general primitive and specialized primitive features as illustrated in Figure \ref{fig:descriptor_breakdown}. \textit{General primitive features} refer to features that can be obtained directly from raw input videos, which then can be used directly in the classification module. \textit{Specialized primitive features} refer to features that are extracted from raw input videos and require additional processing into auxiliary features before they enter the classification stage. In this section, some common primitive feature descriptors as well as its associated auxiliary feature descriptors that have appeared in the action recognition and detection literature will be studied. 

\begin{subsubsection}{General Primitive Features}\label{sec:feature_descriptors_midlevel2}
	\input{chapters/chapter3b1_midfeat_representation_general}
\end{subsubsection}

\begin{subsubsection}{Specialized Primitive Features and Auxiliary Features}\label{sec:descriptor_special}
Some algorithms require extraction of primitive features and further refinement into auxiliary features before they can be useful to a classifier, especially the methods that were proposed in the earlier years of action recognition. Some examples of specialized primitive features include silhouettes/contours and object tracks. Silhouette-/contour- and tracking-based features and the corresponding auxiliary features are described in the following.
	 
	\begin{paragraph}{Silhouette-/Contour-based Models}\label{sec:silhouette_models}
		\input{chapters/chapter3b2a_specific_silhouette}
	\end{paragraph}

	\begin{paragraph}{Tracking-based Models}
		\input{chapters/chapter3b2b_specific_flow}
	\end{paragraph}
\end{subsubsection}

\begin{subsubsection}{Final Remarks on Feature Descriptors}
	In this section, a select number of popularly used feature descriptors for human actions were examined. Once the type of sampling method has been determined, primitive features can be obtained from raw videos. These primitive features can either be encoded directly or must be converted into auxiliary features before it is encoded to enter the classification stage. Historically, the field of action recognition approached the task of action recognition using specialized primitive features as it contained useful information. However, features that rely on these specialized primitive features were deemed unfavourable as background-subtraction and tracking remain unsolved problems in computer vision. A mixture of filter- and flow-based algorithms merged. Now, the state-of-the-art performance is achieved by CNN-based algorithms. 
\end{subsubsection}

%% file: chapters/chapter3b1_midfeat_representation_general.tex
\textit{General primitive features} refer to features that can be directly extracted from raw videos after some sampling method has been chosen (regular, dense, or sparse) and are transformed in a way such that it can be processed directly by some chosen classification method. General primitive features can be divided into four broad categories: filter-, flow-, convolutional neural network (CNN)-based, and others. Here, each of these categories will be examined. 

\begin{paragraph}{Filter-based Descriptors}\label{sec:gradient_descriptors}
	\input{chapters/chapter3b1a_general_filter}
\end{paragraph}

\begin{paragraph}{Optical Flow-based Descriptors}\label{sec:optflow_descriptors}
	\input{chapters/chapter3b1b_general_flow}
\end{paragraph}

\begin{paragraph}{Convolutional Neural Network-based Descriptors}
In recent years, there has been a surge of algorithms relying on \textit{Convolutional Neural Networks} (CNNs or ConvNets) in a wide variety of artificial intelligence-based problems, including action recognition. As its name suggests, CNNs are based on neural networks, which is a system that consists of a sequence of layers with a set of artificial ``neurons'' in each layer. 
	The first layer of the network, the \textit{input layer}, usually consists of raw pixels of an image/videos \cite{Bac11, Bil16, Don15, Fei16, Fer16, Ji13, Sim14, Wan16}, but pre-processed data, such as optical flow displacement fields \cite{Don15, Fei16, Ji13, Ng15, Sim14, Wan16}, can also be used. The last layer of the network, the \textit{output layer}, is typically interpreted as a softmax/logistic regression. Alternatively, the outcome of the output layer can be fed into a classifier (e.g. an SVM) to produce a class score or class rankings. 
The architecture of a CNN can be characterized by the local connections in the intermediate, \textit{hidden}, layers. The hidden layers often alternate between convolution, rectification, and pooling operations, with an optional normalization layer. On occasion, pooling is neglected altogether \cite{Spr15}. In conjunction with deep-learning, the network weights are learned via back-propagation with shared weights within a layer. Prototypically, the learned weights only pertain to the numerical values of the taps in the convolution's point-spread functions \cite{Fuk80,LeC98}. While the theoretical understanding of these architectures are limited, it appears to successfully extract descriptors that are well-suited to the domains on which they are trained (e.g. object parts and assemblies thereof) \cite{Zei14}. Currently, CNNs dominate the empirical evaluations in many image-based recognition tasks, including action recognition \cite{Bil16,Wan16}.\\ 

	Motivated by state-of-the-art performance on various image classification tasks, CNNs have been utilized in various ways on video classification tasks as well. A method to incorporate the temporal domain or motion information onto the well-established 2D CNN architecture has been the main branching point of many algorithms in video classification. The most intuitive approach would be to replace 2D convolution and/or pooling operations with 3D ones to account for the additional (temporal) domain in videos \cite{Bac11, Ji13, AKar14, Sho16,Tra15}. Alternatively, the temporal information in videos can be summarized into a single RGB image such that standard 2D CNNs can be applied to recognize actions \cite{Bil16}. 
Recurrent neural networks (RNN), which are capable of learning temporal dynamics by explicitly considering the sequences of CNN activations in a recurring manner, is another approach taken to account temporal dimension in videos \cite{Bac11,Don15,SMa16,Ng15,Sin16}. To account for RNN's inability to learn long-range temporal relationships, 
numerous algorithms suggest embedding long short-term memory (LSTM) units into the architecture to allow the network to learn to recognize and synthesize temporal dynamics \cite{Bac11,Don15, SMa16, Ng15}. Recent methods resort to CNNs to obtain feature vectors of images \cite{Ibr16,SMa16,Mah16,Yeu16} or iFV-encoded iDT features with HOG, HOF, and MBH feature descriptors \cite{Yua16} as inputs to LSTM-RNN. 
Processing images in a per frame basis keeps track of which features are occurring when, allowing temporal detection of actions to be possible \cite{Yeu16}.\\

Another route that has been explored is the two-stream model \cite{Sim14}, inspired by biology \cite{Goo92}, which decouples the appearance and motion components of a video \cite{Feich16,Sin16,Zhu16}. The appearance stream takes framewise spatial input (e.g. RGB values) while the motion stream takes motion input (e.g. optical flow values \cite{Don15, Feich16, Ji13, Ng15, Sim14, Wan16,Zhu16}, motion vectors \cite{Zha16}). The two streams can be fused at the final stage of their respective architectures \cite{Don15, Feich16, Ji13, Ng15, Sim14, Wan16}, or sooner via convolutional fusion to put the channel responses in two streams that occur at the same pixel location into correspondence \cite{Feich16}. Alternatively, the two streams can be fused via introduction of residual connections between the paths \cite{Fei16c}. In the standard two-stream approach, computing the optical flow is expensive and the most timely step. Thus, 
rather than employ the most sophisticated dense optical flow techniques, some have relied on cruder block-based matching approaches, as employed for compression, which the authors refer to as ``motion vectors'' \cite{Zha16}. These approaches, however, exhibit coarser structure than optical flow and may contain noise and inaccurate movements.\\

	One CNN-based algorithm takes a completely different approach by redefining ``action'' as a change that it brings to the environment (see Figure \ref{fig:actions_transformations}). Thus, features before the action (at the pre-conditioned state) and after the action (at the effect state) are aggregated using a Siamese network to represent an action \cite{Wan16}.\\
	\begin{figure}[htbp]
		\begin{center}
			\includegraphics[width=0.9\textwidth]{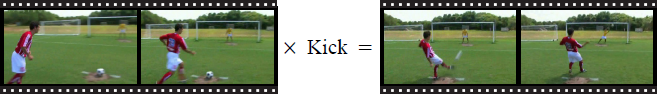}
			\includegraphics[width=0.9\textwidth]{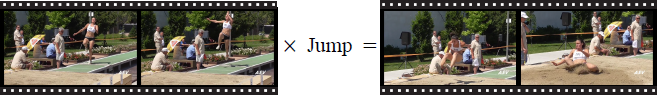}
		\end{center}
		\caption{One algorithm defines \textit{actions} as \textit{transformations brought to the environment} (i.e. pre-conditioned state $\times$ action $=$ effect). Two \textit{transformations}, kick (top row) and jump (bottom row), are illustrated with their respective pre-conditioned (left columns) and effect (right columns) states. Redrawn from \cite{Wan16}. \label{fig:actions_transformations}}
	\end{figure}

	The specificity of the features increases at higher layers of the network \cite{Spr15,Zei14}. Thus, reducing the number of layers and neurons in each layer could depreciate the overall performance of the system \cite{AKar14,Zei14}. 
Although the state-of-the-art performance in complex datasets are achieved using CNNs with many layers, it is done at a high computational cost \cite{AKar14}. 
To compensate for computational complexity, one approach applies PCA-whitening between layers on a stacked ISA network \cite{Le11}. 
Alternatively, a network can be separated into two streams that processes each frame of a video with two different spatial resolutions: (i) downsampled frames at half the original spatial resolution, and (ii) a smaller spatial window at the original resolution (e.g. centre region if videos are obtained from video sharing services to take advantage of the camera bias shot by amateur recorders) \cite{AKar14}.
Another obstacle that hinders the use of CNN-based methods is the amount of training data that is required to construct a reliable system \cite{AKar14,Sim14}. Two of the largest benchmark datasets available, UCF101 and HMDB51, are considered too small to train a CNN-based video classification program from scratch \cite{Sim14,Wan16}. Thus, \textit{Sports-1M} \cite{AKar14}, a dataset containing more than a million videos, is often used to train the system. Since datasets as large as Sports-1M are typically constructed with some degree of automaticity, it leads to corruption of data, accumulating even more challenges at the training and testing stages. Alternatively, the networks can be pre-trained on large static image recognition datasets (e.g. ImageNet \cite{Den09}). However, such pre-training may cause the final network to bias towards appearance information over motion, an undesirable trait for action recognition and detection in videos.
\end{paragraph}

\begin{paragraph}{Other General Primitive Feature Descriptors}
Not all descriptors that have appeared in the action recognition literature can be categorized as either filter-, \mbox{flow-,} or CNN-based representation models. Here, a select few other general primitive feature descriptors that do not fall under these categories that possess noteworthy characteristics are mentioned. They are: eSURF \cite{Bay06}, MACH filter \cite{Rod08}, and TCCA features \cite{Kim09}.\\

The \textit{extended Speeded Up Robust Features} (eSURF) is a descriptor based on Haar-wavelet responses $(d_x,d_y,d_t)$ along the three axes \cite{Wil08} based on SURF \cite{Bay06}. The feature vector is constructed by summing the weighed responses of the Haar-wavelets as sampled uniformly across each interest point $(\sum{d_x},\sum{d_y},\sum{d_t})$. The Haar-wavelet responses are weighed with a Gaussian to account for geometric deformations and localization errors \cite{Bay06}. \\

The \textit{maximum average correlation height} (MACH) filter \cite{Rod08} is one of few algorithms that considers condensing a collection of data into a single template. Intra-class variations of an action is generalized into a single template by optimizing four performance metrics: average correlation height (ACH), average correlation energy (ACE), average similarity measure (ASM), and output noise variance (ONV). It uses spatiotemporal regularity flow (SPREF) to obtain the direction that best represents the overall regularity of the volume (i.e. the direction in which the pixel intensities change the least) instead of other motion estimators to avoid challenges that occur due to motion discontinuities, aperture problems, and large illumination variations. The SPREF flow field volume of each example is converted using a Clifford Fourier Transform (CFT) for its efficiency, which is used to synthesize the MACH filter. The composite template video is obtained by combining the mean of the CFTs, the noise covariance matrix, the average power spectral density, and the average similarity matrix to minimize ACE, ASM, and ONV while maximizing the ACH. \\

\textit{Tensor canonical correlation analysis} (TCCA) features \cite{Kim09} consider videos as third-order \textit{tensors} with three modes (or axes). Third-order tensors can share any single or multiple modes. Thus, if a canonical transformation, a transformation that maximizes the correlation of two multi-dimensional arrays, is applied to the modes that are not shared, then two types of TCCA can be produced: the joint-shared mode and the single shared-mode. 
The \textit{joint-shared} mode allows any two modes (or axes) (i.e. a plane or section in the video) to be shared and applies the canonical transformation to the remaining single mode. It is found that a single pair of canonical directions would maximize the inner product of the output tensors (or canonical objects) for the joint-shared modes.
The \textit{single-shared} mode, on the other hand, allows any single mode (i.e. a scan line of a video) to be shared and applies the transformation to the remaining two modes. Here, two pairs of free transformations maximize the inner product of the canonical objects for the single-shared modes.
A single pairing of joint-shared mode TCCA preserves discriminative information, whereas the double pairing of single-shared mode TCCA preserves less original data resulting in more flexibility in its information. Thus, the joint-shared mode TCCA is used to filter inter-class differences (e.g. difference between actions) while the single-shared mode TCCA features are permissive to intra-class variations (e.g. difference in appearance). 
\end{paragraph}

%% file: chapters/chapter3b1a_general_filter.tex
Filter-based approaches can be categorized into two types: (i) gradient-based and (ii) oriented bandpass filter-based descriptors.
Gradient-based methods rely on the assumption that the local appearance and shapes of an object can be portrayed by their local intensity gradient or edge directions. Oriented bandpass filter-based approaches use oriented filters to decompose videos into basic components using local orientation and scale. Notably, gradient-based approaches are an example of (high-pass) oriented filters, which have received a particularly large research focus. Hence, they are dealt separately from the more general oriented bandpass filters in the following. \\

A rich set of gradient-based descriptors have appeared in the field of action recognition. Some descriptors that have made frequent appearance in the field include: histogram of oriented gradients (HOG) \cite{Dal05,JWang13}, HOG3D \cite{Kla08}, cuboid descriptor \cite{Dol05}, scale-invariant feature transform (SIFT) \cite{Low04}, gradient location-orientation histogram (GLOH) \cite{Mik05}, local trinary patterns (LTP) \cite{Yef09}, and spatiotemporal (ST) patches \cite{She07}. \textit{HOG}s store spatially oriented gradient to capture appearance information of the action. \textit{HOG3D} extends HOG descriptors by storing spatiotemporal oriented gradients to store shape and motion information together. The \textit{cuboid descriptor} \cite{Dol05} concatenates three gradient channels $(G_x,G_y,G_t)$ into a single vector to form a single feature vector for each neighbourhood. \textit{SIFT} \cite{Low04}, which is coupled with a scale-invariant region detector, DoG, uses 3D histograms to represent the gradient locations and orientations. The 2D SIFT descriptor uses polar coordinates to obtain the gradient magnitudes and orientations, and the 3D SIFT descriptor \cite{Sco07} uses an additional angle to represent the direction of the gradient to incorporate temporal information. The location and orientation bins in 2D/3D SIFT are weighed by the gradient magnitudes. Instead of quantizing the location information on a Cartesian grid as in 2D/3D SIFT, \textit{GLOH} quantizes them on a log-polar grid to increase robustness and distinctiveness \cite{Mik08}. \textit{LTP}s compare intensities of the neighbouring pixels between preceding and succeeding frames to the current frame to determine the direction of motion \cite{Yef09,Kli12}. \textit{ST patches} uses spatiotemporal gradients to estimate the motion of the sampled regions to obtain a rank of the ST patch. The constraint based on the rank provides information on motion without explicitly computing the optical flow and spatial information (e.g. uniform intensity, edge-, and corner-like features) \cite{She07}. \\

Although many of these oriented gradient-based descriptors provide computational efficiency to gather crucial information, such as appearance and/or motion, they are very sensitive to illumination changes. Often, these descriptors do not provide sufficient information and must be used in parallel with other descriptors that possess distinguishing characteristics (e.g. HOG is often found with HOF) to overcome its limitation. \\

Spatiotemporal oriented bandpass filters can decompose an image sequence into basic components using the dimension of local orientation and scale (i.e. angular and radial frequencies). Consequently, various types of oriented filters have been applied to a range of dynamic image understanding tasks, such as action recognition and detection \cite{Nin09}.
These representation models tend to be capable of characterizing image dynamics without explicitly requiring flow recovery nor segmentation of videos \cite{Cho99}.
Two particular approaches of spatiotemporal oriented filtering have been commonly applied to actions: 3D Gabor filters \cite{Cho99, Nin09} and Gaussian derivative filters \cite{Jhu07}. Both 3D Gabor and Gaussian derivative filters are typically applied in quadrature pairs and combined to produce some local energy measurement. Often subsequent processing is involved, such as normalization and/or combination of filter outputs. The normalization process provides robustness to photometric variations \cite{Cho99}, while combining filter outputs (e.g. appearance marginalization \cite{Der13}) attempt to gather information on image dynamics that is invariant to spatial appearance. The filter outputs can also be combined to yield explicit motion estimates or other measurements of image motion \cite{Gry09}.\\

Representations based on spatiotemporal oriented bandpass filters tend to be robust to illumination changes, in-class variations, and occlusion. Many researchers choose to use Gaussian derivative filters for its separability and recursive components to keep the representation computationally efficient \cite{Cho99,Der13}. However, some filter responses (e.g. bandpass filters) pose sensitivity to irrelevant appearance attributes. Furthermore, these filters tend to be sensitive to scale changes, which is problematic since the actor/action size is inconsistent between and within each video.

%% file: chapters/chapter3b1b_general_flow.tex
Optical flow-based algorithms have appeared frequently in various action recognition algorithms. Optical flow provides data that can be used in two ways: (i) to extract information on motion and (ii) for tracking purposes. Here, some common optical flow-based representation models that have appeared in the action recognition literature for each method will be explored. \\

\begin{figure}
	\begin{subfigure}{0.45\textwidth}
		\includegraphics[width=0.95\textwidth]{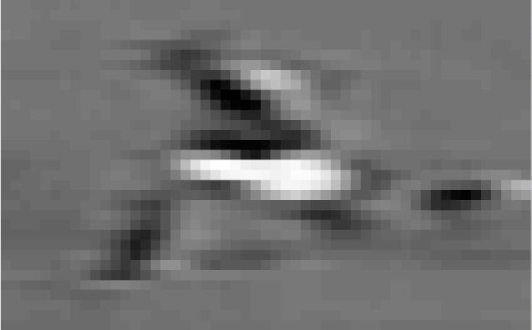}
		\caption{Original Frame. \label{fig:track_ori}}
	\end{subfigure}
	~
	\begin{subfigure}{0.45\textwidth}
		\includegraphics[width=0.95\textwidth]{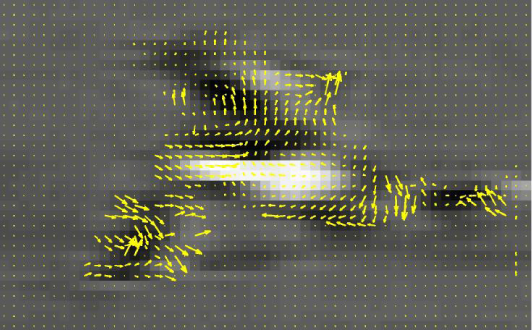}
		\caption{Optical Flow $\vec{F}={\begin{bmatrix} F_x & F_y\end{bmatrix}}^\top$. \label{fig:track_OptFlow}}
	\end{subfigure}
	
	\begin{subfigure}{0.2\textwidth}
		\includegraphics[width=0.95\textwidth]{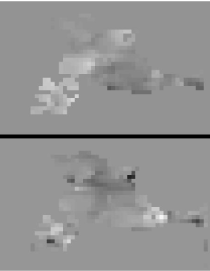}
		\caption{The optical flow vector $\vec{F}$ is split into horizontal ($F_x$ (top)) and vertical ($F_y$ (bottom)) components. \label{fig:track_HorVerOptFlow}}
	\end{subfigure}
	~
	\begin{subfigure}{0.35\textwidth}
		\includegraphics[width=0.95\textwidth]{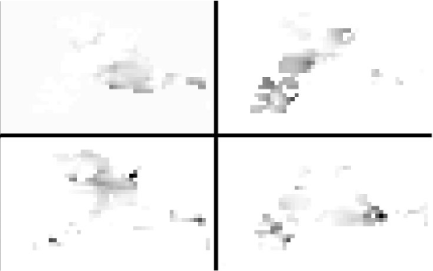}
		\caption{Horizontal and vertical optical flows are half-wave rectified to produce $F_x^+$ (top left), $F_x^-$ (top right), $F_y^+$ (bottom left), and $F_y^-$ (bottom right). \label{fig:track_halfwave_rect}}
	\end{subfigure}
	~
	\begin{subfigure}{0.35\textwidth}
		\includegraphics[width=0.95\textwidth]{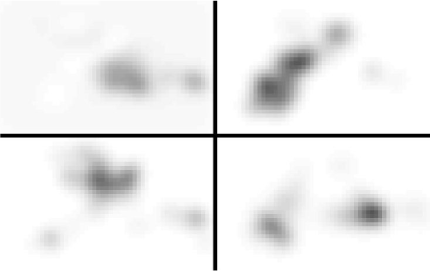}
		\caption{Half-wave rectified motions are blurred into $Fb_x^+$ (top left), $Fb_x^-$ (top right), $Fb_y^+$ (bottom left), and $Fb_y^-$ (bottom right). \label{fig:track_blurred}}
	\end{subfigure}
	\caption{The actor can be tracked to obtain a stabilized figure-centric volume. A standard optical flow algorithm applied on a stabilized volume captures the motion created by the local regions in the volume. Redrawn from \cite{Efr03}. \label{fig:tracking}}
\end{figure}

Optical flow can be used to recognize actions by describing the motion of the actor. A standard optical flow algorithm can be applied to stabilized figure-centric volumes to capture motion created by different parts of the body (see Figure \ref{fig:track_OptFlow}) \cite{Efr03,Lin09}. By separating the optical flow into horizontal and vertical components (as in Figure \ref{fig:track_halfwave_rect}) then blurring them (via Gaussian as in Figure \ref{fig:track_blurred}), an artificial set of motion channels are created \cite{Efr03,Fat08}. 
Often, the \textit{Kanade-Lucas-Tomasi} (KLT) tracker is used to estimate local motion in a hierarchical manner to obtain the initial flow for the next level \cite{Uem08}.  \\

\textit{Histograms of Optical Flow} (HOF) captures local motion of the pattern by quantizing the orientation of the optical flow vectors. 
While such characterization of motion is sufficient in distinguishing highly distinct actions (e.g. ``walk'' vs. ``wave'' in the KTH dataset), it fails to distinguish fine differences in actions (e.g. ``box'' vs. ``clap'' in the KTH dataset).
Thus, simple description of motion combined with information on appearance (e.g. HOG) can yield more accurate recognition results as has been observed in more complicated datasets, such as the Hollywood1 dataset \cite{Lap08}. \\

\begin{figure}
	\begin{center}
		\includegraphics[width=0.9\textwidth]{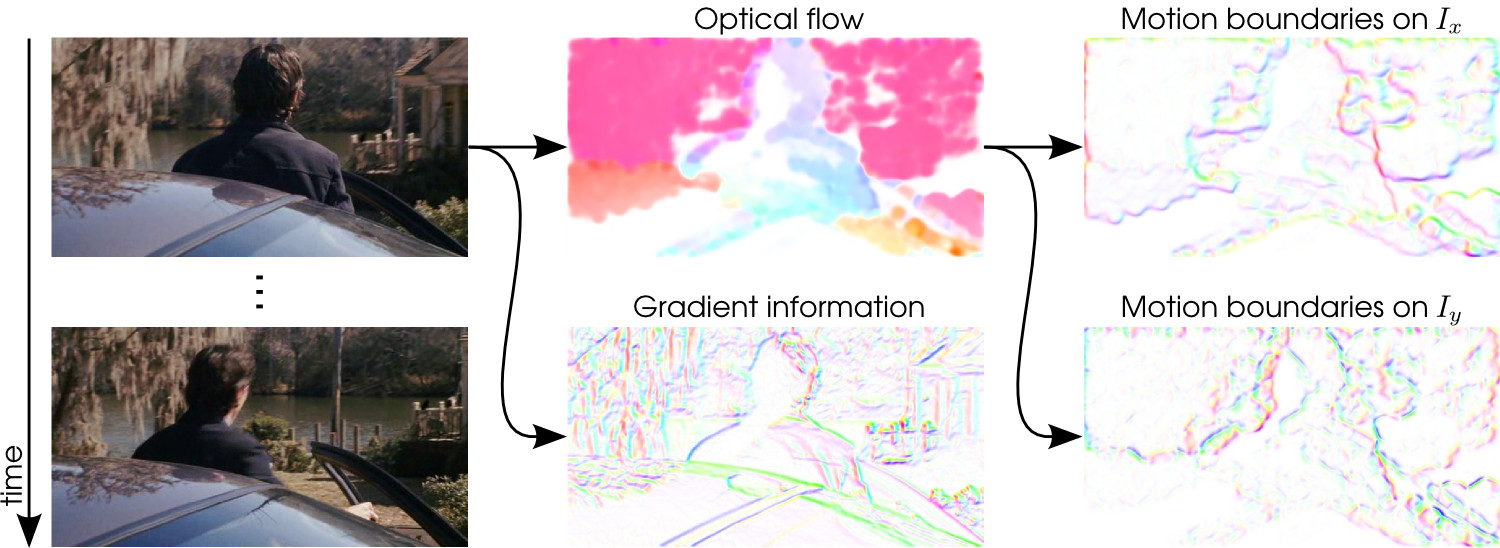}
	\end{center}
	\caption{Illustration of HOF, HOG, and MBH interest point descriptors. The gradient information (HOG) (bottom-centre) and flow orientation (HOF) (top-centre) is calculated for each frame in a video (left). Using the $x$ and $y$ components of optical flow, the spatial derivatives are calculated for each direction to obtain the motion boundaries on $I_x$ and $I_y$ (right). The gradient directions are indicated by the hue and magnitude by the saturation. Redrawn from \cite{Wan11}. \label{fig:stip_descriptors}}
\end{figure}

The \textit{Motion Boundary Histogram} (MBH) is a descriptor that uses derivatives of optical flow for each horizontal and vertical directions, $I_x$ and $I_y$, respectively \cite{Dal06,Wan11}. By computing the spatial derivatives for each flow field, the local gradient orientations and magnitudes can be found to construct a local orientation histogram. Since MBH computes the gradient of optical flow, constant motion is suppressed and only the information regarding changes in the flow field are kept. Thus, MBH provides a simple way to suppress constant motion (e.g. camera motion) while preserving local relative motion of pixels (e.g. motion boundaries/foreground motion) (see Figure \ref{fig:stip_descriptors} right). This is an appealing feature, especially for recognizing actions in realistic videos, since they tend to contain severe camera motion \cite{Wan11}. Furthermore, the majority of the texture information from the static background is eliminated as the derivatives of the trajectories are considered. \\

\begin{figure}[htbp]
	\begin{center}
		\begin{subfigure}{0.9\textwidth}
			\includegraphics[width=0.95\textwidth]{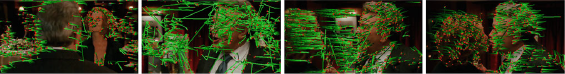}
			\caption{KLT Trajectories \label{fig:track_klt}}
		\end{subfigure}
		~
		\begin{subfigure}{0.9\textwidth}
			\includegraphics[width=0.95\textwidth]{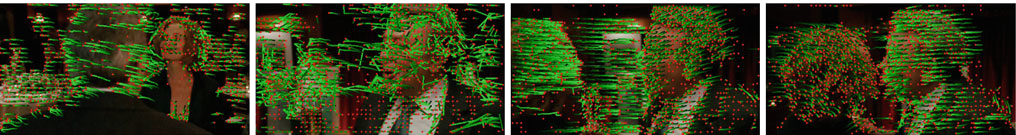}
			\caption{Dense Trajectories \label{fig:track_dt}}
		\end{subfigure}
	\end{center}
	\caption{Examples of KLT and dense trajectories of the ``kiss'' action from the Hollywood2 dataset. Redrawn from \cite{Wan13}. \label{fig:trajectories}}
\end{figure}

With optical flow, physical properties of the flow pattern can be extracted via \textit{kinematic features}, such as divergence, vorticity (or curl), symmetric and antisymmetric optical flow field, second and third principal invariants of flow gradient and rate of strain tensor \cite{Ali10, Jai13}. Kinematic features are perceived as independent forces that act on the object and capture information regarding motion only. For example, \textit{divergence} captures information on the amount of axial motion, expansion, and scaling effects. 
\textit{Vorticity} (or \textit{curl}), on the other hand, highlights the circular motion created by the human body or part of the human body.
Thus, motions of the hand toward the camera would be well captured by divergence; in contrast, rotary motions of the hand parallel to the image plane would be well characterized by a curl. 
The kinematic features collectively provide a unique spatiotemporal pattern description of the human action.\\
\textit{Dense trajectory} (DT) features \cite{Wan11} were introduced as another form of descriptors that track the path of sampled motion (see Figure \ref{fig:track_dt}), which have made frequent appearance in the field of action recognition and detection \cite{Wan11, Zho15}. DT features first require \textit{dense} sampling of feature points at each frame, which are pruned using good-features to track. Then each of the sampled points are tracked using optical flow to obtain its trajectory. The trajectory descriptor is obtained by concatenating the normalized displacement vectors. These features are often combined with other features (e.g. HOG, HOF, MBH) aggregated along the trajectories.
	Various dense trajectory models that would enhance the original DT model \cite{Wan11} have been proposed \cite{Jia12,Wan13}. 
	One approach was to cluster the dense trajectories to detect the dominant direction of motion and consider relative motion between the trajectories to gather object-background and object-object information \cite{Jia12}. 
	Another approach was to explicitly estimate camera motion \cite{Wan13} by matching feature points between frames using SURF descriptors \cite{Bay06} and dense optical flow \cite{Shi94}. This particular camera motion compensated trajectory feature is referred to as the \textit{improved dense trajectory} (iDT) feature and has appeared frequently in action recognition and detection literature \cite{Bil16}. 


Optical flow has been successful in various applications (e.g. tracking). In fact, some approaches have benefited from using optical flow-based algorithms to track humans, body parts, and interest points yielding good action recognition results (see under \textit{Specialized Primitive Features - Tracking-based Models}). However, the ability to estimate motion accurately and consistently has numerous challenges associated, such as motion discontinuities (e.g. occlusion), aperture problems, and large illumination variations (e.g. appearance changes).

%% file: chapters/chapter3b2a_specific_silhouette.tex
Numerous cognitive studies have shown that humans are capable of extracting various useful information from silhouettes, such as recognizing objects, labelling parts, and comparing similarities to other shapes \cite{Bat08,Bie85}. Thus, a video of silhouettes may provide sufficient information for recognition even while being robust to lighting conditions and invariant to the appearance of the person. Once the silhouettes of the actors are extracted, information can be described in various forms. Silhouettes can either be directly converted into 1D signals, converted into binary or scalar images then described using moments, or they can be stacked to form space-time volumes. A sample of each type of auxiliary silhouette features, which include $\mathcal{R}$ Transforms, motion energy images, motion history images, motion history volumes, and spacetime volumes, will be described below as a sample of such approaches. \\

\textit{$\mathcal{R}$ transforms} are shape descriptors that convert silhouette images to 1D signals. By taking the squared sum of the Radon transform, commonly used to detect lines in images, over varying radii, a translation invariant Radon transform is defined allowing video alignment to match the position of the actor unnecessary. Furthermore, to resolve the scale sensitivity problem of Radon transforms, $\mathcal{R}$ is normalized. This improved extension of the Radon transform, the $\mathcal{R}$ transform, attracted attention to earlier action recognition algorithms that were silhouette-based (see \cite{Sou08, Wan07}). \\

\begin{figure}[htbp]
	\begin{center}
	\begin{subfigure}{0.9\textwidth}
		\begin{center}
			\includegraphics[width=0.4\textwidth]{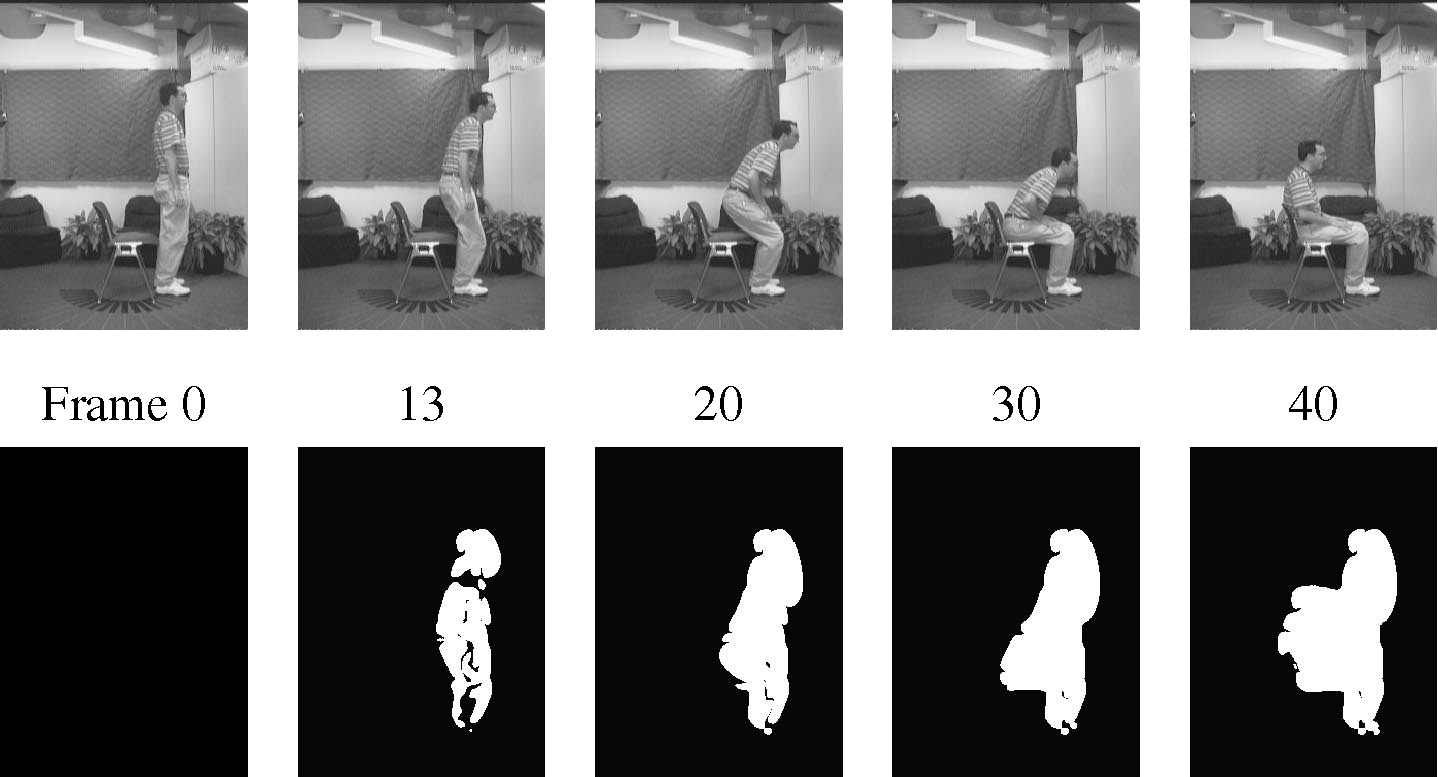}
			~
			\includegraphics[width=0.55\textwidth]{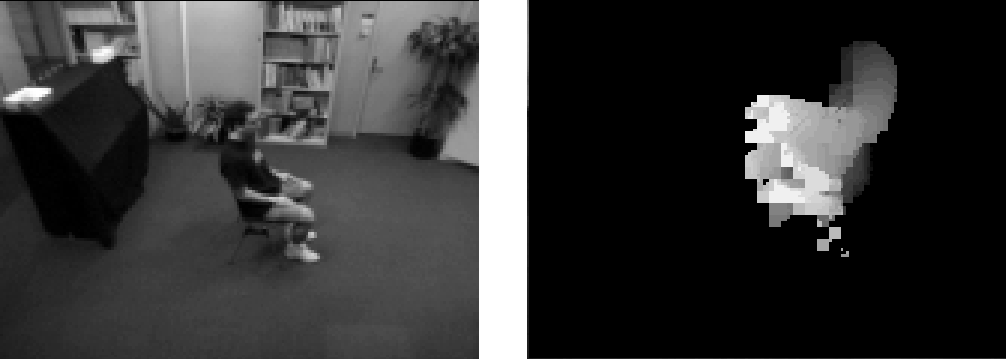}
		\caption{Examples of MEI (left) and MHI (right) of the sitting motion. Redrawn from \cite{Bob01}. \label{fig:MEI_MHI}}
		\end{center}
	\end{subfigure}

	\begin{subfigure}{0.9\textwidth}
		\begin{center}
			\includegraphics[width=0.9\textwidth]{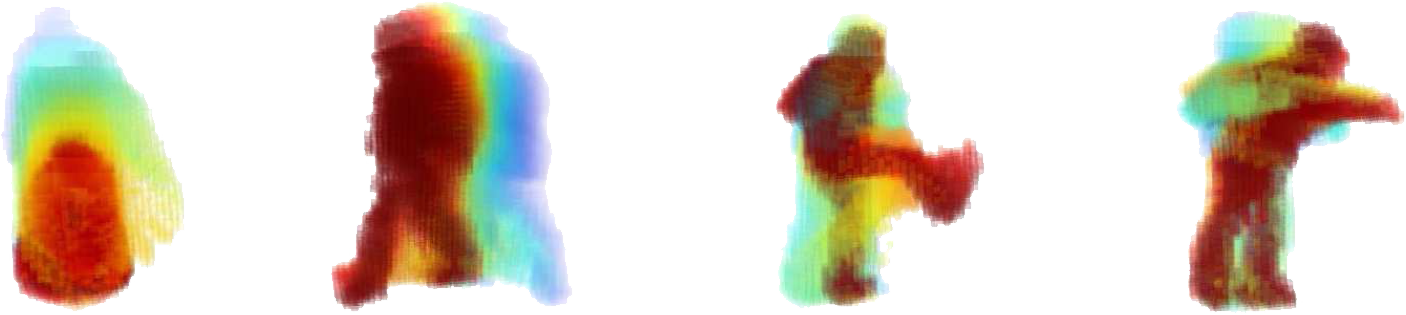}
		\caption{Examples of Motion History Volumes. Motion history volumes of actions (left-to-right): sit-down, walk, kick, and punch are illustrated using the colour spectrum, where blue indicates oldest motion and red indicates the most recent motion. Redrawn from \cite{Wei06}. \label{fig:MHV}}
		\end{center}
	\end{subfigure}

	\begin{subfigure}{0.9\textwidth}
		\begin{center}
			\includegraphics[width=0.8\textwidth]{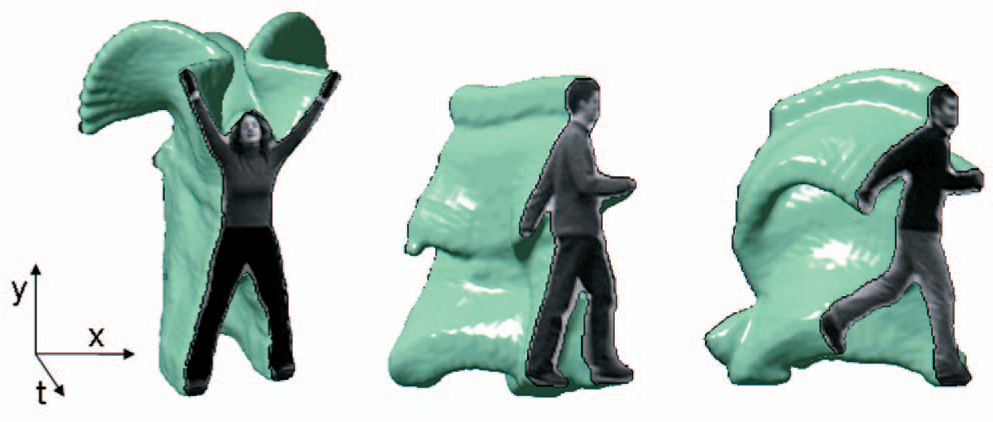}
		\caption{Examples of a spacetime volume (STV) for actors performing a \textit{jumping jack}, \textit{walk}, and \textit{run} actions. Redrawn from \cite{Bla05}.}
		\label{fig:stv}
		\end{center}
	\end{subfigure}
		
	\begin{subfigure}{0.9\textwidth}
		\begin{center}
			\includegraphics[width=0.8\textwidth]{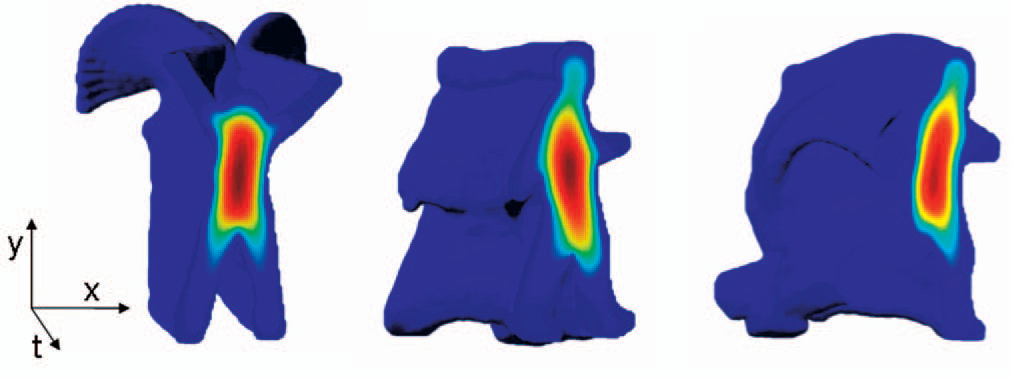}
		\caption{The solution to the Poisson equation reveal the shape of an actor. The values are encoded using the colour spectrum, where low values are encoded by blue and high values are encoded by red. Regions far from the core (the extremities and the head) have low values, therefore are encoded in blue.}
		\label{fig:stv_stfeature1}
		\end{center}
	\end{subfigure}
	\end{center}
	\caption{Examples of various silhouette-based models in action recognition.}
\end{figure}

Binary images of silhouettes called \textit{motion energy images} (MEI) can be constructed by accumulating the difference between silhouettes in subsequent frames and a scale-valued image, referred to as \textit{motion history images} (MHI) can be constructed to store the recency of motion that occurred at every pixel (see Figure \ref{fig:MEI_MHI}). MEIs and MHIs together provide information on the \textit{location} and the \textit{temporal history} of the motion, respectively. These images have been further described using Hu moments \cite{Hu62} to draw further comparisons with other actions \cite{Luo03}. 
Many silhouette-based algorithms have shown sensitivity to object's displacement and orientation to the camera. This problem can be resolved by replacing the silhouette motion indicating function with a silhouette occupancy function to create \textit{motion history volumes} (MHV) instead of MHIs (see Figure \ref{fig:MHV}) \cite{Wei06}. Although MHVs have this appealing feature of viewpoint invariance with the use of an occupancy function, it is a great challenge to obtain an accurate function that would precisely model $x$-, $y$-, $z$-coordinates of where the object of interest is especially in videos gathered in uncontrolled settings, such as the web. \\

A sequence of silhouettes or its contours/boundaries can be concatenated along the temporal axis to create an image feature that captures the relationship between space and time of a person's action called \textit{spacetime volumes} (STV) (see Figure \ref{fig:stv}). Information on the location of the general body parts (e.g. head, torso, and extremities) can be obtained by calculating the average time it takes for every point inside the STV to reach the contour via a random-walk process \cite{Bla05} or differential geometry \cite{Yil08}. The Poisson equation can be used to identify the motion saliency of moving parts and their orientations \cite{Bla05}. While MHVs and STVs appear similar, MHVs illustrate the recency function through its 3D reconstruction, while temporal information cannot be observed in STVs.\\

Although silhouettes/contours provide useful information, obtaining accurate segmentation of an actor is not guaranteed, especially in situations where the background is not static as background subtraction remains an unsolved problem in computer vision. Furthermore, the view angle can alter a person's silhouette drastically and the features inside the boundary cannot be delineated since a person is represented as a single region.

%% file: chapters/chapter3b2b_specific_flow.tex
\begin{figure}[htbp]
			\begin{center}
			\begin{subfigure}{0.5\textwidth}
				\begin{center}
					\includegraphics[width=\textwidth]{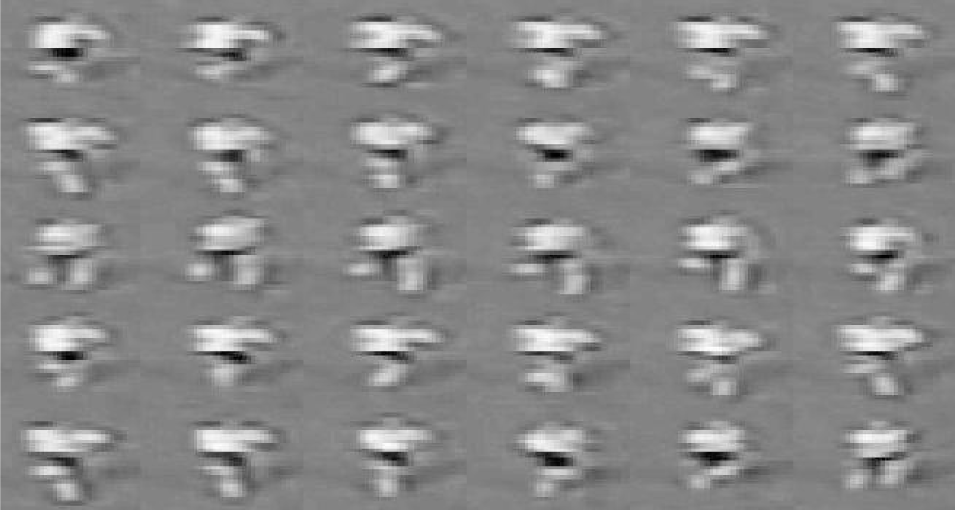}
				\end{center}
				\caption{A sequence of figure-centric frames that constitute a figure-centric volume of the actor. Redrawn from \cite{Efr03}. \label{fig:fig_centric}}
			\end{subfigure}
			~
			\begin{subfigure}{0.45\textwidth}
				\begin{center}
					\includegraphics[width=0.38\textwidth]{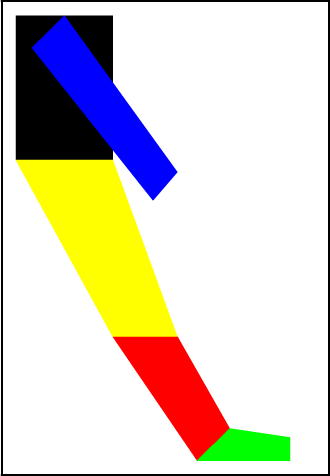}
				\end{center}
				\caption{Cardboard Person Model representing the major components of the human body: arm (blue), torso (black), thigh (yellow), calf (red), and foot (green). Redrawn from \cite{Ju96}. \label{fig:cardboard_figure}}
			\end{subfigure}
			\end{center}
			\caption{Utilizing tracking algorithms to extract the entire actor as a whole (left) and to track the movement of each body part (right).}
		\end{figure}

As briefly mentioned in the optical flow section, tracking can be perceived as an extreme example of optical flow. Tracking algorithms can be utilized in action recognition by (i) tracing the trajectory of the entire actor in a video to segment the actor from the background (see Figure \ref{fig:fig_centric}) \cite{Bod03,Efr03,Ibr16,Sin16} or (ii) by tracking body parts (see Figure \ref{fig:cardboard_figure}) \cite{Fan05,Hah08,Iki08,Ram03,Rao02,Yac99} or local interest regions \cite{Fan05,Iki09}. \\

Tracking-based methods are potentially robust to variations in appearance of each actor or local region and have been shown to yield impressive results on low-resolution videos \cite{Efr03}. Despite significant progress, however, tracking remains an unsolved problem in computer vision as initializing tracking can be difficult as can maintaining tracks over an extended period of time, especially in scenes with cluttered or dynamic backgrounds. Moreover, since feature trackers often assume constant appearance of image patches over time, this assumption can pose problems when the appearance of the object changes, especially when two objects merge (occlude) or split (deocclude) \cite{Lap05}. Furthermore, the output of a tracker tends to be noisy, susceptible to drifting and illumination changes, causing problems in its subsequent steps when representing the action. 

%% file: chapters/chapter3d_grouping.tex
Feature space encoding begins with the generation of a \textit{codebook} (also referred to as a \text{dictionary}) based on a set of training data.
A codebook can be generated in two ways: (i) by partitioning the features into regions (or \textit{clusters}) using a \textit{discriminative model} or (ii) by representing the space using a set of probability distributions using a \textit{generative model} \cite{XWang13}. 
In either case, the codebooks are constructed with respect to a set of training data.
In the following, one or more common approaches to each codebook generation model will be examined.

\begin{subsubsection}{Discriminative Clustering}
	A feature space can be divided into distinct regions (or \textit{clusters}) to form \textit{codewords}. Each cluster is comprised of objects that share similar characteristics to one another but different from objects in other clusters. Among many discriminative clustering algorithms that are available, $k$-means clustering is one of the most widely used techniques in action recognition \cite{Fen02,Iki08,Lap08,Lin09,Ram03,Sco07,JWang13}. \textit{$k$-means clustering} divides a given set of features into $k$ clusters for $k \in \mathbb{Z}^+$, such that the total distance between each categorized feature and the centre of its cluster (\textit{centroid}), which is referred to as a \textit{codeword}, is minimized. $k$-means clustering partitions the space into non-overlapping regions. As a result, each feature in the feature space is assigned to one specific cluster.
$k$-means clustering is implemented frequently in practice for its simplicity and performance.\\ 

	Another discriminative clustering method that appears in the action recognition literature is agglomerative clustering \cite{Lin09,Mik08}. In \textit{agglomerative clustering}, data points are clustered to their nearest cluster in a hierarchical manner to form a larger cluster. The results are usually presented as a dendrogram to record the sequences of merges \cite{Dou13}. 
A dendrogram exempts the need to select a specific number of clusters at the outset \cite{Dou13}. In fact, the optimal number of clusters can be determined using a scree plot of the dendrogram, where the optimal number of clusters is indicated by the high curvature in a scree plot. Despite this benefit, not too many recognition algorithms rely on agglomerative clustering due to its computational burden and its requirement on storage space \cite{Dou13}.
\end{subsubsection}

\begin{subsubsection}{Generative Clustering}
	A feature space can be represented using probability distributions such as the \textit{Gaussian Mixture Model} (GMM). Given a set of feature descriptors (from a training set), a weighted sum of Gaussian functions can be used to model the (training set) feature space. 
Typically, the parameters (i.e. the weight, mean vector and covariance matrix of individual Gaussian distribution) that would optimally represent the feature space are trained through maximum-likelihood (ML) estimation using the expectation-maximization (EM) algorithm. 
The learned parameters of the GMM (e.g. mean vectors and covariance matrices) provide information on the mean information of the codewords as well as the shape of their distributions \cite{XWang13}. While first- and second-order statistical information provides information that would assist in improving the accuracy of the classification procedure, it is computationally expensive to obtain and store first- and second-order statistical information compared to discriminative models and not as compact.
\end{subsubsection}

\begin{subsubsection}{Discussion on Codebook Generation}
	The size of the codebook (i.e. number of clusters or GMMs) is a crucial parameter in codebook generation as it affects the computational cost and classification accuracy. Up to a certain point, recognition performance has been empirically shown to improve with the growth of the codebook size (i.e. number of clusters or GMMs). Exceptions to this general point can be observed as the performance plateaus when the size of the codebook exceeds some threshold \cite{Pen14}. Moreover, an excessively large codebook size can harm the accuracy level due to over-fitting of the data or over-partition of the feature space. The thresholds to yield an optimal codebook is dependent on the dimension and sampling strategy of the feature descriptor \cite {Pen14}. \\

	 Features with higher dimensions require more codewords to divide the feature space. Thus, a larger codebook size would be necessary for optimal performance.  Sparsely sampled feature points tend to be more scattered in the feature space than densely sampled feature points. Thus, to avoid over-partitioning of the codebook (i.e. to ensure that every cluster is affiliated with a feature), the codebook size should be smaller in data obtained via sparse sampling as opposed to data obtained through dense sampling.
	Moreover, the distribution of densely sampled descriptors in the feature space would not provide useful high-order statistics (e.g. variance), which would affect the type of information that should be obtained in the subsequent assignment step. Thus, although generative models provide more information, discriminative clustering would be the preferred choice with densely sampled features as they provide a more compact clustering leading to computational efficiency. \\

	While the codebook size is a key parameter, the optimal codebook size is dependent on many factors. Unfortunately, there is no theoretical solution that would find the optimal codebook size. Thus, readers should bear in mind that many algorithms that use $k$-means clustering or GMMs often report best results based on $k$ that was obtained through trial-and-error.
\end{subsubsection}

%% file: chapters/chapter3e_encoding.tex
With a codebook generated using a set of features from the training set, a new set of features can be quantized according to the clusters (or \textit{codewords}) in the pre-defined codebook. Features can either be assigned to a single word through \textit{hard assignment} or into multiple words through \textit{soft assignment}. Here, some examples of these two types of quantization assignment methods are examined.

\begin{subsubsection}{Hard Assignment}
	\textit{Hard assignment} methods assign feature descriptors from videos to a single codeword in the codebook. The most common hard assignment quantization method that appears in action recognition algorithms is \textit{vector quantization} (VQ), which assigns a feature descriptor to the nearest codeword in the codebook. 
Instead of assigning a binary value to the closest codeword as in VQ, a weight can be assigned to the nearest codeword 
to quantitatively indicate the similarity between the feature and a small subset of close codewords as in \textit{salient coding} (SC) \cite{Hua11}.
For its simplicity and efficiency, VQ is widely used in many action recognition algorithms \cite{Fen02,Iki08,Kov10,Mar09,Ram03,Sch04,Yam92}. 
Since hard assignments represent each feature by the nearest codeword, features that are nearly equidistant to multiple codewords are prone to change even when small adjustments are made at the codebook generation stage. This ambiguity causes hard assignment-based methods to be unstable, which can aggravate recognition accuracy rates \cite{XWang13}.
\end{subsubsection}

\begin{subsubsection}{Soft Assignment}
	To overcome the ambiguity that hard assignment quantization techniques pose, features can be assigned to multiple codewords instead of one through \textit{soft assignment}. Soft assignment methods can be further broken down into two categories: combinatorial and contrasting. \textit{Combinatorial methods} express features as a combination of the codewords while \textit{constrasting methods} describe features by alluding to the differences between features and codewords. Here, some common approaches of each soft assignment methods that have appeared in the action recognition literature are considered.

	\begin{paragraph}{Combinatorial}
		Features can be expressed as a combination of all or just a few codewords in the codebook. 
 		To naively encode a feature vector based on all codewords would yield an unreliable feature assignment to the codespace, especially the linkages that are made with distant codewords \cite{LLiu11}. Thus, a select number of codewords in the codebook should be considered. The weight to assign the degree of membership of feature, $\mathbf{f}$, to codeword, $\mathbf{c}_i$, can be determined by solving the following optimization problem \cite{Pen14}:
		\begin{equation}\label{eq:soft_combinatorial}
			\arg{\min_{\mathbf{c}}{ 
				{\| \mathbf{f} - C \mathbf{c} \|}_2^2 + \lambda \psi{ \left(\mathbf{c}\right) } }} \text{,} 
		\end{equation}
		where $C=[\mathbf{c}_1 \dots \mathbf{c}_k]$ is a codebook with $k$ codewords $\mathbf{c}_i$ for $i=1, \dots, k$, and $\lambda$ is a constant that controls the strength of the regularization term $\psi(\mathbf{c})$. 
		Some examples of assignment methods that assign features, $\mathbf{f}$, to codewords, $\mathbf{c}$, using \eqref{eq:soft_combinatorial} include: orthogonal matching pursuit (OMP) \cite{Tro07}, sparse coding (SpC) \cite{Yan09}, local coordinate coding (LCC) \cite{Yu09}, and locality-constrained linear coding (LLC) \cite{Wan10}, which differ by their regularization term, $\psi$. The regularization term enforces varying properties of $\mathbf{c}$. \\ 

		The \textit{orthogonal matching pursuit} (OMP) approximates $\mathbf{c}$ by considering the number of nonzero elements of $\mathbf{c}$, the $l_0$-norm of $\mathbf{c}$. Unfortunately, $l_0$-norms are non-convex and to obtain a solution to \eqref{eq:soft_combinatorial} with $\psi(\mathbf{c})=\| \mathbf{c}\|_0$ requires some heuristic strategy. 
Thus, to counter the non-convexity of the $l_0$-norm, the regularization term in \eqref{eq:soft_combinatorial} can be replaced with an $l_1$-norm (i.e. $\psi(\mathbf{c}) = \| \mathbf{c} \|_1$), which is referred to as \textit{sparse coding} (SpC).\\ 

		It was empirically observed that SpC is helpful when the codewords are local (i.e. when non-zero coefficients are assigned to codewords (or bases) near the feature vector (the data to be encoded)) \cite{Wan10,Yu09}. Since this locality is not guaranteed the way \eqref{eq:soft_combinatorial} is set up in SpC, the locality constraint of SpC can be explicitly enforced by modifying the regularization term as $\psi(\mathbf{c}) = \| \hat{\mathbf{e}} \odot |\mathbf{c}| \|_1$, such that $\mathbf{1}^\top \mathbf{c} = 1$ as in \textit{local coordinate coding} (LCC)\footnotemark. 
Unfortunately, SpC and LCC require solving an $l_1$-optimization problem, which is computationally expensive and problematic for large-scale problems. As a result, a practical assignment scheme called the \textit{locality-constrained linear coding} (LLC) \cite{Wan10} was designed as a fast-implementation of LCC by defining the regularization term as
$\psi(\mathbf{c}) = \| \mathbf{e} \odot \mathbf{c} \|_2^2$, such that $\mathbf{1}^\top \mathbf{c}= 1$ for 
$\mathbf{e} = \exp{ \left( \frac{\hat{\mathbf{e}}}{\sigma} \right)}$, where $\mathbf{e}$ ensures that similar patches have similar codes by assigning weights proportional to how similar each codeword is to the feature vector. \\
\footnotetext{$\hat{\mathbf{e}}$ in LCC and LLC denote $dist(\mathbf{f},C)=[dist(\mathbf{f},\mathbf{c}_1) \ \cdots \ dist(\mathbf{f},\mathbf{c}_k)]^\top$, where $dist(\mathbf{f},\mathbf{c}_i)$ is the Euclidean distance between $\mathbf{f}$ and $\mathbf{c}_i$ for $\mathbf{c}_i \in C$.}

		Among various soft combinatorial assignments that were introduced in this section (see Table \ref{tab:soft_assmnt_cmb} for a summary), LLC is the most popularly used for its fast implementation. Put simply, LLC assigns each feature as a linear combination of $m$-nearest codewords in the codebook of size $k$ for $m \ll k \in \mathbb{Z}^+$. As a point of comparison, note that VQ and LLC base their assignments on the 1-nearest and $m$-nearest codewords, respectively. However, the weighted sum of multiple codes allow LLCs to better capture the relationship between similar descriptors that share the same codewords than the hard assignment quantization methods \cite{Wan10}.
Although LLC is faster than other combinatorial methods, the least square problem \eqref{eq:soft_combinatorial} that needs to be solved to find the $m$ nearest words remains a computational burden of the LLC combinatorial assignment method. \\

		Unlike the combinatorial assignment methods that were introduced earlier, the localized soft-assignment \cite{LLiu11} does not involve solving the least-squares problem \eqref{eq:soft_combinatorial}, rather a normalized weight is assigned with respect to $m$-nearest codewords for $m < k$ in a codebook of size $k$. Although it has a computational advantage over LLC, and is the most computationally efficient combinatorial assignment approach, with a comparable accuracy rate, a constant value that determines the softness of the assignment is present as a free parameter.

\begin{table}[htbp]
	\begin{center}
	\begin{tabular}{| l | c |}
		\bottomrule
		\textbf{Assignment Type} 		& \textbf{Regularization Term} $\psi(\mathbf{c})$ \\
		\toprule \bottomrule
		Orthogonal Matching Pursuit (OMP) 	& $\| \mathbf{c} \|_0$ \\
		\hline
		Sparse Coding (SpC)			& $\| \mathbf{c} \|_1$ \\
		\hline
		Local Coordinate Coding (LCC) 		& $\| \hat{\mathbf{e}} \odot |\mathbf{c}| \|_1$ 
									 such that $\mathbf{1}^\top \mathbf{c} = 1$\\
		\hline
		Locality-Constrained Linear Coding (LLC) & $\| \mathbf{e} \odot |\mathbf{c}| \|_2^2$, where 
									$\mathbf{e} = \exp{\left( \frac{\hat{\mathbf{e}}}{\sigma} \right)}$ \\
		\toprule
	\end{tabular}
	\end{center}
	\caption{List of regularization terms for combinatorial soft assignment methods. The coefficients that determines the degree of membership between feature $\mathbf{f}$ and codeword $\mathbf{c}_i$ is determined by solving the least-squares problem: $\arg \min_\mathbf{c}{ \| \mathbf{f} - C\mathbf{c} \|_2^2 } + \lambda \psi(\mathbf{c})$ given a codebook $C=[\mathbf{c}_1 \ \cdots \ \mathbf{c}_k]$. Assignment type varies with regularization $\psi$. $\hat{\mathbf{e}}$ in LCC and LLC denotes $dist(\mathbf{f},C)=[dist(\mathbf{f},\mathbf{c}_1) \ \cdots \ dist(\mathbf{f},\mathbf{c}_k)]^\top$, where $dist(\mathbf{f},\mathbf{c}_i)$ is the Euclidean distance between $\mathbf{f}$ and $\mathbf{c}_i$, and $\sigma$ in LLC is a constant that controls the weight of $\mathbf{c}_i$ for $1 \leq i \leq k$. \label{tab:soft_assmnt_cmb}}
\end{table}
	\end{paragraph}

	\begin{paragraph}{Contrasting}
		Alternate to analyzing direct affiliations between features and codewords, dissimilarities between descriptor mean and codewords can provide useful information. Some examples of this type of soft assignment encoding methods are Fisher vectors (FV) and vector of linearly aggregated descriptors (VLAD). Here, FV and VLAD will be examined in detail as well as their relationships. \\
		
		\textit{Fisher vectors} (FVs) \cite{Jaa98} are soft assignment methods that are derived from Fisher kernels (FKs) \cite{XWang13}. FVs rely on a codebook defined using a generative model (e.g. GMMs) such that the set of training features can be described by the gradient of the log-likelihood. A Fisher kernel, which measures the similarity between two sets of data, training and test, is defined as the product of the gradient of the log-likelihood functions of the sets and the Fisher information matrix. Finally, the Fisher vectors are obtained by concatenating the derivatives of the Fisher vectors with respect to the mean and the covariance. 
The use of Fisher kernels 
allows use with any kernel-based classifiers, such as SVMs.
Since Fisher vectors include information on deviation and covariance using GMMs, first- and second-order statistics of the feature descriptors are encoded providing generative information \cite{Wan13}. Like generative models, FKs are also capable of processing data of varying lengths (i.e. FK support addition or removal of data) and like discriminative methods, FKs have flexible criteria and yield better results. 
The number of Gaussians selected at the codebook generation step can affect the smoothness/sharpness of the histogram. As the number of Gaussians increase, there would be less descriptors assigned to a Gaussian with a significant probability. Noting that no descriptor assigned to some Gaussian yields a zero gradient vector, there would be more Gaussians that are not assigned to any descriptors. As a result, the histogram would be sharp around zero (cf. Figure \ref{fig:cf_ngaussians} (a)-(c)).
To reduce the sensitivity of FVs to the number of Gaussians, FVs can be improved into an \textit{improved FV} (iFV) \cite{Per10} by applying power-normalization to each element in FV. To ensure that the quantization is not affected by a free parameter, $l_2$ normalization is applied to iFV (to be discussed in greater detail in the \textit{normalization} section). That is, the dependency on a parameter that represents the object-to-background ratio, where small objects with a small parameter are not represented well, can be removed. 
The posterior probability calculation that is involved in FV and iFV slows down the computation, but is compensated through its use of small codebook. \\

		\begin{figure}[htbp]
			\begin{center}
				\includegraphics[width=0.9\textwidth]{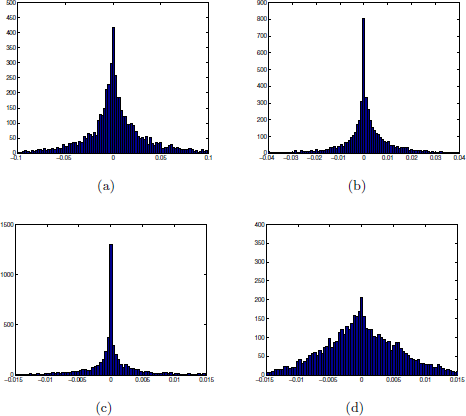}
			\end{center}
			\caption{(a)-(c) Comparing L2-normalized Fisher vectors (FVs) with a different number of Gaussians: (a) 16, (b) 64, and (c) 256 Gaussians.
				    (c)-(d) Comparing L2-normalization with power normalization: (c) L2-normalized FV, and (d) power-normalized FV. Redrawn from \cite{Per10}. \label{fig:cf_ngaussians}}
		\end{figure}
		
		\textit{Vector of Linearly Aggregated Descriptors} (VLAD) \cite{Jeg10} is another quantization method based on dissimilarities between new features and codewords that appear in action recognition and detection algorithms \cite{Jai13}. VLAD encoding methods typically rely on a codebook generated using $k$-means clustering, but GMMs can be used as well. The VLAD representation is obtained by summing, for each codeword, the differences between the feature vectors and the codeword, where each feature vector is associated with the nearest codeword in the codebook. 
That is, $\mathbf{c} = \sum_{j=1}^{k}{\mathbf{f}_i - \mathbf{c}_j}$, where $\mathbf{c}_j$ is the closest codeword to local feature $\mathbf{f}_i$. 
VLAD can be perceived as a simplified version of FV in that VLAD only keeps the first-order statistics (i.e. the mean) as opposed to first- and second-order statistics in FV. 
The additional second-order information in FVs typically lead to better performance than VLAD. However, VLAD can overcome the difference in the case that features appear more densely in the space of interest and thereby yield a more stable codebook \cite{Pen14}. Consequently, with a set of densely sampled features, it would be more beneficial to encode via VLAD rather than FV since the second-order statistics do not assist in obtaining higher accuracy, but adds computational cost.
	\end{paragraph}
\end{subsubsection}

\begin{subsubsection}{Discussion on Assignment Methods}
	The high-order statistical information that the encoding methods retain (e.g. difference of means and variances in FV vs. difference of means in VLAD) allows soft assignment methods to better capture the distribution shape of the descriptors in the feature space than hard assignment methods \cite{Pen14}. However, storing more information comes at a cost of higher dimension. Notice that the final dimensions of VQ, LLC, FV, and VLAD, are $k$, $k$, $2dk$, and $kd$, respectively, where $d$ is the dimension of the descriptor and $k$ is the codebook size (i.e. number of clusters if based on $k$-means clustering and number of mixture if based on the GMM). 
Thus, the computational cost of training FVs tend to be much larger than any other encoding method mentioned in this paper and often requires feature reduction in its subsequent steps.
\end{subsubsection}

%% file: chapters/chapter3f_pooling.tex
Some algorithms face too much repeated data or inconsistent representations of the data. Thus, further processing is needed to reduce and stabilize the data through pooling and normalization. Here, some common pooling and normalization operations that appear at the encoding stage are examined. Their role and effects in various quantization methods will be discussed as well.

\begin{subsubsection}{Pooling}
Processing responses of all features can be expensive. Thus, the statistics of the features can be aggregated (or \textit{pooled}) at various regions to yield a summary statistic (e.g. histogram). These summary statistics tend to be much lower in dimension and prevents over-fitting of the data. Furthermore, data with large variations can be condensed into a more compact representation by either removing or weighing the outliers less. Thus, an ideal pooling method must preserve important information and discard irrelevant materials while allowing invariance to small transformations of the input \cite{Bou10}. 
Typical pooling methods include: max-, sum-, and average-pooling. The feature with the largest response is chosen in \textit{max-pooling}, and the responses are combined additively or averaged in \textit {sum-pooling} and \textit{average-pooling}, respectively.
 The appropriate pooling operation depends on the sampling method, features type, and codebook size \cite{Bou10}. Max-pooling is the preferred method for sparsely sampled features \cite{Bou10,Ser07}. \\

Although max-, sum-, and average-pooling are simple ways to aggregate data, they have some obvious drawbacks. Responses that are slightly weaker than the strongest are discarded in max-pooling even though their weaker responses could provide additional useful information. Every response within a region is considered in sum- and average-pooling with equal importance, which would be undesirable since the responses with low magnitudes can down weight the responses with high magnitudes. Consequently, instead of considering one or all responses in a region, a probabilistic form of average-pooling and a weighted response can be considered during training and testing phases, respectively, as in \textit{stochastic pooling} \cite{Zei13}. The probabilities and the weights in stochastic pooling are determined by the magnitude with respect to other responses within the region (see Figure \ref{fig:pool_norm}). Alternatively, other mixture of pooling methods (e.g. taking the max over the fraction of all available feature points) can sometimes yield more accurate results \cite{Bou10}. \\

\begin{figure}[htbp]
	\begin{center}
		\begin{subfigure}{0.28\textwidth}
			\begin{center}
			\includegraphics[scale=0.7]{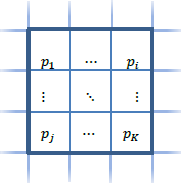}
			\end{center}
			\caption{General region $R_i=\{p_1, \dots, p_K\}$}
		\end{subfigure}
		~
		\begin{subfigure}{0.35\textwidth}
			\begin{center}
			\includegraphics[scale=0.7]{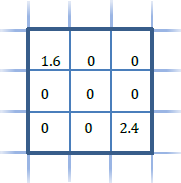}
			\end{center}
			\caption{A concrete example of $R_i = \{1.6, 0, 0, 0, 0, 0, 0, 0, 2.4 \}$ \label{fig:pool_eg}}
		\end{subfigure}
		~
		\begin{subfigure}{0.28\textwidth}
			\begin{center}
			\includegraphics[scale=0.7]{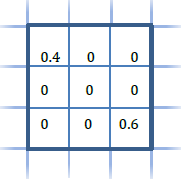}
			\end{center}
			\caption{Normalized $R_i$ \label{fig:pool_norm}}
		\end{subfigure}
	\end{center}
	\caption{Illustration of pooling regions with (a) general responses, (b) an example of responses in region $R_i$, and (c) a normalization of (b). \label{fig:pool}}
\end{figure}

\begin{table}[htbp]

	\resizebox{\textwidth}{!}
		{\begin{minipage}{1.3\textwidth}
			\begin{center}
			\begin{tabular}{| c | c | c |}
				\bottomrule
				\multirow{2}{*}{\textbf{Pooling Method}} 	& \textbf{Equation for Pooling Region} &  \textbf{Pooled value from} \\
											& $R_i=\{p_1, \dots, p_K\}$ & $R_i = \{ 1.6, 0, 0, 0, 0, 0, 0, 0, 2.4\}$ \\
				\toprule
				\bottomrule
				\textbf{Max} & $\max_k{\left\{p_k \right\}}$ & $\max{\left\{1.6,0, \dots, 0, 2.4\right\}}=2.4$ \\
				\hline
				\textbf{Sum} & $\sum_{k=1}^{K}{p_k}$ & $1.6 + 0 + \dots + 0 + 2.4 = 4.0$  \\
				\hline
				\textbf{Average} & $\frac{1}{K}\sum_{k=1}^{K}{p_k}$ & $\frac{1}{9}\left(1.6 + 0 + \dots + 0 + 2.4\right)=0.44$  \\
				\hline
				\textbf{Stochastic} at training \cite{Zei13} & $p_k$ with $P(p_k) = \frac{p_k}{\sum_k{p_k}}$ & $1.6$ or $2.4$ with prob. of $0.4$ and $0.6$, resp. \\
				\textbf{Stochastic} at testing \cite{Zei13} & $\sum_{k=1}^{K}w_k p_k$, where $w_k = \frac{p_k}{\sum_k{p_k}}$ & $0.4 \cdot 1.6 + 0 \cdot 0 + \dots + 0 \cdot 0 + 0.6 \cdot 2.4 = 2.08$ \\
				\toprule
			\end{tabular}
			\end{center}
		\end{minipage}}
	\caption{Summary of pooling methods. Refer to Figure \ref{fig:pool_eg} for an illustration of  the example in the rightmost column.}
\end{table}

The aforementioned pooling techniques aggregate data over some pre-defined region disregarding spatial layout and temporal order.
At a global scale, spatial invariance can be beneficial since the location of an action within a video should not change the class of an action. However, the spatial layout at a local scale, such as shape and location of body parts with respect to each part, 
can provide crucial information \cite{One13}. 
Motivated by the fact that varying spatial scale retains the order of the features in locally orderless images (or histograms) \cite{Koe99}, \textit{spatial pyramid pooling} \cite{Laz06} employs a hierarchy of rectangular windows to preserve spatial orders. It partitions each frame of a video into increasingly finer spatial subregions and computes the histograms of local features from each sub-region to concatenate into a single final vector \cite{Wan10}. 
Reconsideration of spatial order have shown to strengthen the descriptive power of the features. Pyramid pooling can be extended to the spatiotemporal domain from the spatial domain by partitioning videos into increasingly finer spatiotemporal subregions instead of spatial subregions \cite{Lap08,Ull10,HWang13}. This variation would preserve both the spatial as well as temporal orders of the features for finer discrimination between actions with similar structure that vary in temporal sequence (e.g. fall down vs. get up). \\

Pooling regions can also be more meaningfully defined by identifying regions that are more likely to contain actions (or \textit{actionness} \cite{Che14}) (see Figure \ref{fig:pool_dyn_region}). In fact, it was confirmed that pooling from a ground truth pose mask improves the accuracy of action recognition algorithms \cite{Jhu13}. There are many ways of explicitly decomposing videos. One intuitive way would be to split the video into foreground/background \cite{Ull10}. In a similar manner, action-, actor-, or object-specific detectors can be applied per frame of the video to  detect actions, actors, or specific objects \cite{Ull10}. One canny approach restricts pooling regions to areas that the human observers look at by collecting the human eye movement using an eye tracker as they view a video \cite{Vig12}. 
Alternatively, features can be pooled from saliency regions\footnotemark. Here, the premise is that saliency regions are likely to contain an actor. 
Various combinations and variants have appeared in literature to create a binary or real-valued saliency map (e.g. interest point detectors \cite{Bal13,Sul14}, structure tensors \cite{Vig12}, SOEs \cite{Fei15}). 
Features pooled from different salient regions but the same fixed grid segmentation (as in Figure \ref{fig:pool_dyn_region}) would have low similarities, especially if these features correspond to actions with spatial change over time. Thus, pooling from saliency regions allow features to undergo a more fair comparison as they are aggregated from similar regions. Furthermore, real-valued saliency maps \cite{Che14, Fei15, Sul14} can be used as weights since the features pooled from these regions are that much likely to contain an action.
\footnotetext{Here, saliency information is used to \textit{pool} features rather than to \textit{sample} them. That is, saliency information is used to select a few features that will be used to train or test the classifier after they have been extracted and represented as some feature vector.}
\begin{figure}[htbp]
	\begin{center}
		\begin{subfigure}{0.4\textwidth}
			\includegraphics[width=0.8\textwidth]{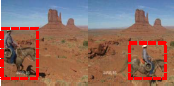}
			\caption{Action changing in spatial location in a single video sequence highlighted in red.}
		\end{subfigure}
		~
		\begin{subfigure}{0.55\textwidth}
			\includegraphics[width=0.8\textwidth]{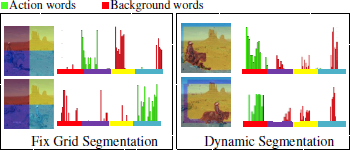}
			\caption{Fixed Grid Segmentation vs. Dynamic Segmentation}
		\end{subfigure}
	\caption{Comparing fixed grid segmentation and dynamic segmentation on a video that contains an action that has spatial variation. The action words (green histograms) fall in different cells (purple region followed by cyan region) of the fixed grid (left) as the action changes spatial location throughout the sequence. On the other hand, the action words remain in the same (red) region in a video that is segmented dynamically (right). Redrawn from \cite{Bal13}. \label{fig:pool_dyn_region}}
	\end{center}
\end{figure}
\end{subsubsection}

\begin{subsubsection}{Normalization}
To ensure consistency amongst the collected data, a normalization procedure can be applied to a database of features. 
Some common normalization techniques include \cite{Pen14}: $l_1$-, $l_2$-, \textit{power-}, and \textit{intra-}normalization. As its name suggests, in $l_1$- and $l_2$-normalizations, the features are divided by the $l_1$- and $l_2$-norms, respectively, of the vectors. The power-normalization \cite{Per10} computes the sign root of each element. That is, the power-norm of an encoded vector $\mathbf{c}(\mathbf{x})$ is defined as: $\|\mathbf{c}(x_i)\|=sign(x_i)|x_i|^\alpha$, where $0 \leq \alpha \leq 1$, for $x_i \in \mathbf{x}$. 
The operation of power has the tendency to reduce the difference between a large value and a small value in a histogram (cf. Figure \ref{fig:cf_ngaussians} (c)-(d)), which results in a smoothing of a histogram \cite{Pen14, Per10}. This smoothing effect can allow more frequently occurring codewords to have less impact, while a less frequently occurring codeword has more impact, which would be useful in data obtained through dense sampling especially if majority of the features correspond to the background. 
The power-normalization technique can be combined with $l_1$- or $l_2$-normalization techniques as in iFV. \\

\textit{Intra-normalization} \cite{Ara13} is different from other normalization techniques in that it is specific to codebook-based methods. Each codeword (or the $k$th Gaussian) is perceived as a block and $l_1$- or $l_2$-normalization is applied to each block. Intra-normalization is an effective way of balancing the weight of different codewords instead of being bias towards bursty features \cite{Ara13}. Burst of features can occur in features that contain repeated structures, which are prevalent in the background, as would be in the case of data obtained through dense sampling. Thus, intra-normalization has shown to be helpful in suppressing irrelevant information (e.g. background information) and putting greater emphasis on useful information especially in features obtained through dense sampling \cite{Pen14}. On the contrary, under the assumption that the data obtained through sparse sampling correspond to information that is a crucial component of an action, intra-normalization has shown to be decrease the discriminative power of action-related codewords degrading the final performance of the recognition algorithm \cite{Pen14}.
\end{subsubsection}

%% file: chapters/chapter3c_dimreduce.tex
Extracted features tend to have high dimensionality, correlated, and/or vary in duration. High dimensionality makes training difficult and computationally expensive at the classification stage. Redundant information could add bias in the training data affecting the accuracy of the algorithm. Difference in temporal duration or action execution rate can cause incorrect comparison of the data (e.g. extending vs. contracting arm in boxing have opposing motions). Thus, although it is not necessary, many recognition algorithms can benefit from dimensionality reduction, removal of redundant information, and/or temporal alignment of the videos.\\

There has been extensive research in the area of dimensionality reduction \cite{van09}. One of the oldest and most widely used post-processing procedure in action recognition and detection is \textit{Principal Component Analysis} (PCA) \cite{Dol05,Kol08,Wei06}. PCAs use orthogonal transforms (via computing the eigenvalues and eigenvectors of the covariance matrix of the feature vectors) to capture the variation amongst the features using principal components. Original features can be represented by a linear combination of \textit{principle components}, which are a set of linearly uncorrelated variables. These principal components are computed in decreasing order of importance, where the first principal component accounts for majority of the variation in the original data. Thus, the number of used principle components is typically less than the number of original variables resulting in dimension reduction. The ability of PCA to uncorrelate the data saves computation cost by removing redundancy \cite{XWang13}.\\ 

Features can be further processed such that they are more distinct while differing by the same amount. Variance between the data can be unified by rescaling the data. Using the eigenvalues obtained at the PCA stage, each feature, $\mathbf{f}_i\in \mathbb{R}^d$, can be rescaled by its respective eigenvalue, $\lambda_i$ for $i=1,\dots,d$, to ensure that each feature has a unit variance. This process of rescaling the feature is referred to as \textit{whitening} (i.e. $\mathbf{f}^{whiten}_i=\frac{1}{\sqrt{\lambda_i}}\mathbf{f}_i)$. It is important to keep in mind that some eigenvalues tend to be numerically close to zero, especially the latter few in a set of eigenvalues arranged in descending order. Thus, it is common practice to add a small constant, $\varepsilon$, to the eigenvalues before the features are rescaled (i.e. $\mathbf{f}_i^{whiten} = \frac{1}{\sqrt{\lambda_i + \varepsilon}} \mathbf{f}_i$) to prevent data inflation or numerical instability. \\

	Within the same action, the temporal duration of the snippet containing the single action can vary due to variations in action execution rate or different frame rate of videos. \textit{Dynamic time warping} (DTW) can be used to align sequences with variable durations \cite{Iki09,Lin09, Lv07,Vee06}. DTW aligns the two time series by warping the time axes to align the samples to the corresponding points. It simultaneously takes into account a pairwise distance between corresponding frames and the sequence alignment cost using dynamic programming. A low alignment cost results when the two sequences are segmented similarly in time and performed at similar rates. \\

Post-processing is not necessary for all methods and is seldom done on many encoding methods other than FV-based methods \cite{Pen14}. However, empirical evaluations show that applying PCA-whitening greatly improves algorithms that do not usually apply PCA-whitening, such as VQ and LLC-encoded methods \cite{Pen14}.

%% file: chapters/chapter4_classification.tex
\begin{figure}[htbp]
	\begin{center}
		\includegraphics[width=0.8\textwidth]{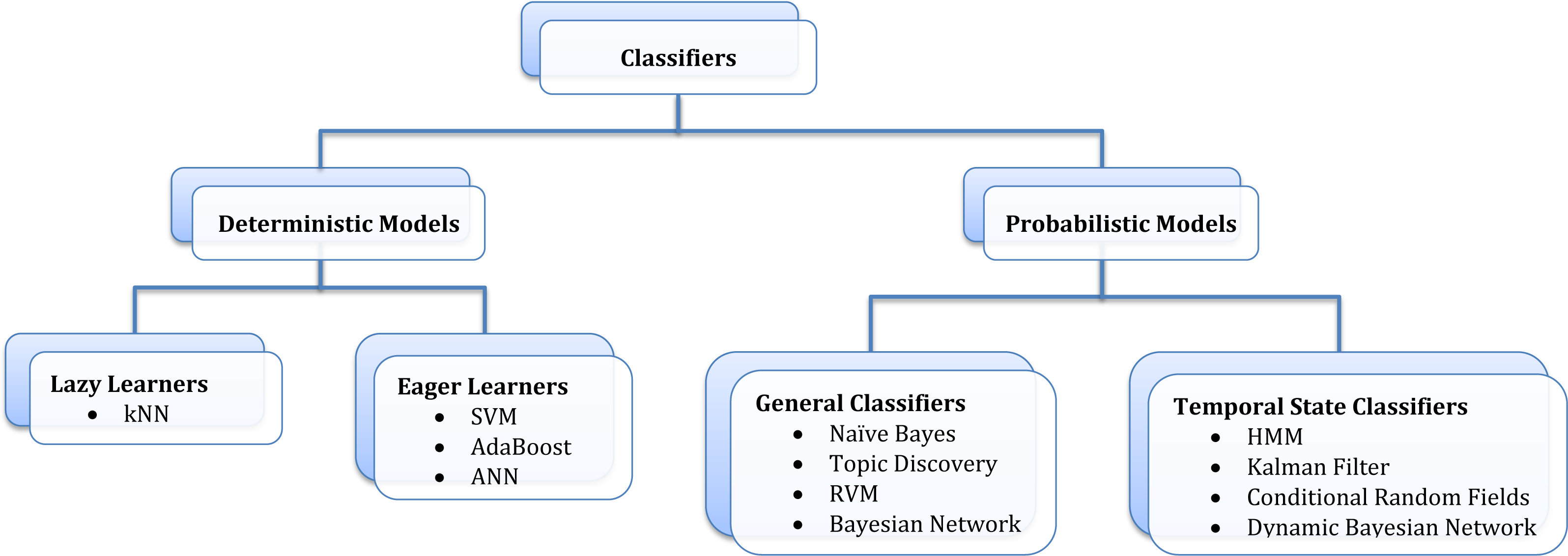}
	\end{center}
	\caption{General breakdown of the types of classifiers that appear in various action recognition algorithms. Features can be classified using a \textit{deterministic} or \textit{probabilistic} model. While deterministic models assigns query features to one class or another, probabilistic models learn the probability distribution over the set of classes to use them to make predictions on the query features. \label{fig:classification_overview}}
\end{figure}

Once a raw video has been transformed into a set of features representative of an action, the query features must be classified. 
A set of training data (labelled or unlabelled) can be used to categorize the test data into some pre-defined class. 
The action class of a query data can be assigned to a single class using \textit{deterministic models} or to a set of classes by modelling probability distributions between classes using \textit{probabilistic models} (see Figure \ref{fig:classification_overview}). We will begin by examining how one could measure the similarity/dissimilarity between features in section \ref{sec:classification_measure}. Then some common deterministic and probabilistic models that appear in the literature of action recognition and detection will be covered in sections \ref{sec:classification_deterministic} and \ref{sec:classification_probabilistic}, respectively.

\begin{section}{Comparison Metrics}\label{sec:classification_measure}
	Given a pair of samples, one must measure how similar (or dissimilar) two patterns are in order to cluster similar (or dissimilar) training samples together (or apart) or to associate (or dissociate) the query data with the same class as the training data. One way to compare sets of data would be to measure the distance between the two. \\

	The \textit{$L_p$-norm} (or Minkowski metric), $d_{L_p}$, is one of the most general classes of metric that measure dissimilarity between two $n$-dimensional features $\mathbf{f}, \mathbf{g} \in \mathbb{R}^n$, which is defined as: 
		\begin{equation}
			d_{L_p}(\mathbf{f},\mathbf{g}) = \left[ \sum_{i=1}^{n}{ {| f_i - g_i |}^p } \right]^{\sfrac{1}{p}} \text{,}
		\end{equation}
	where the value of $p \in \mathbb{Z}^+$ determines the type of distance that is measured between $\mathbf{f}$ and $\mathbf{g}$. $p=1$ measures the shortest distance between $\mathbf{f}$ and $\mathbf{g}$, while $p=\infty$ measures the largest distance between the projected distances of $\mathbf{f}$ and $\mathbf{g}$ (see Figure \ref{fig:Lp_norm}). When $p$ is set to 2, $L_2$-norm is the familiar \textit{Euclidean distance}, which is used in various algorithms \cite{Dol05,Lap08,Lin09,Lv07,JWang13,Wei07,Yil08}.\\ 

	\begin{figure}[htbp]
		\begin{center}
		\includegraphics[width=0.45\textwidth]{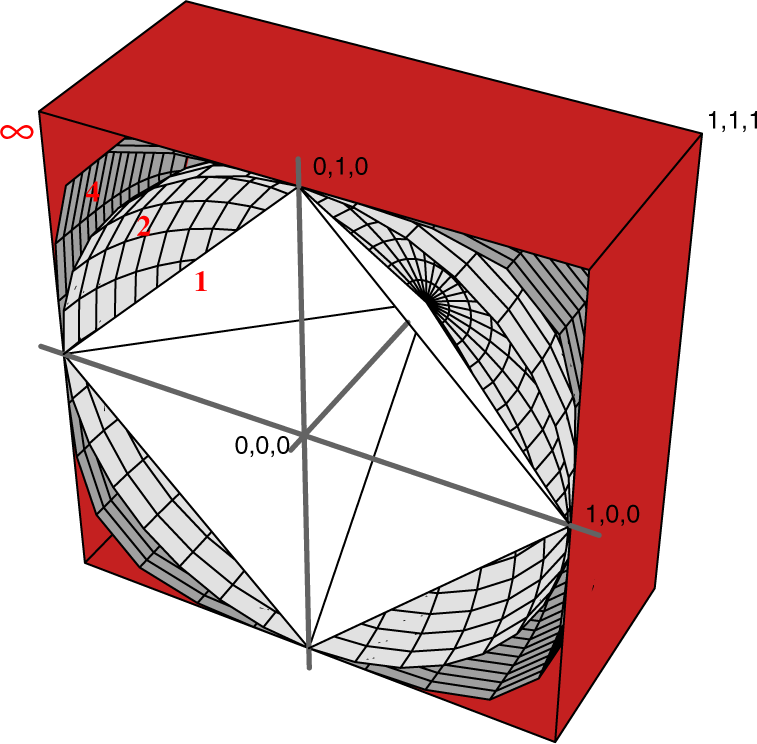}
		\end{center}
		\caption{Illustration of the $L_p$-norm with varying values of $p$ measuring the distance from the origin to point $\mathbf{g}$, a unit away on the coordinate axes. The $L_1$-norm, illustrated in white, is the shortest distance from the origin to point $\mathbf{g}$ while the $L_\infty$-norm is the maximum distance between the projected distances of the origin and $\mathbf{g}$ onto each of the $n$-coordinate axes. Redrawn from \cite{Dud01}. \label{fig:Lp_norm}}
	\end{figure}

While the Euclidean distance is a widely used comparison metric, it is only useful if the data are isotropic and distributed evenly along all directions in the feature space. A common way of standardizing data with different measurements is to apply some weight. A weighted Euclidean distance that uses the mean of the variables as its weight is referred to as the \textit{chi-square distance}, $d_{\chi^2}$, which is defined as:
\begin{equation*}
	d_{\chi^2}(\mathbf{f},\mathbf{g}) = \left( \frac{1}{2} \sum_{i=1}^{n}{ \frac{(f_i - g_i)^2}{f_i + g_i} } \right)^{\sfrac{1}{2}} \text{.}
\end{equation*} \\

	Alternatively, correlated data with varying scales can be accommodated by considering the covariance as in the \textit{Mahalanobis distance}, $d_M$, which is defined as:
		\begin{equation*}	
			d_M(\mathbf{f},\mathbf{g}) = {\left[ (\mathbf{f}-\mathbf{g}) \boldsymbol{\Sigma}^{-1} (\mathbf{f}-\mathbf{g})^\top \right]}^{\sfrac{1}{2}} \text{,}
		\end{equation*}
		where $\boldsymbol{\Sigma}$ is the covariance matrix corresponding to the typical distribution of interest points in the training data \cite{Lap05}. Thus, when the data is scattered in all directions around the centre of the cluster, the convariance matrix is a diagonal matrix, which is the normalized Euclidean distance and an identity covariance matrix would be the standard normalized Euclidean distance. 
The Mahalanobis distance provides a useful measure to calculate the amount of separation between two classes of features (e.g. Hu moments \cite{Bob01} or Fourier projections of MHVs \cite{Wei06}) by measuring the distance between their respective centres \cite{Dou13}. \\

There are various comparison metrics that measure the difference (or similarity) of two probability distributions. 
The \textit{Kullback-Leibler (KL) distance}, $d_{KL}$, which measures the difference between two probability distributions, is defined as: 
\begin{equation*}
	d_{KL}(\mathbf{f},\mathbf{g}) = \sum_{i=1}^{n}{f_i \cdot \ln{\left( \frac{f_i}{g_i} \right)}} \text{.}
\end{equation*}
The KL distance is nonzero and is equal to zero if and only if $\mathbf{f}=\mathbf{g}$ \cite{Dud01}. KL distance is used in various action recognition algorithms \cite{Liu09,Nin09}. KL distance lacks symmetry (i.e. $d_{KL}(\mathbf{f},\mathbf{g}) \neq d_{KL}(\mathbf{g},\mathbf{f})$), which is undesirable in action recognition algorithms because two features should be equally similar or dissimilar to be (part of) an action regardless of the order of comparison (i.e. action $a$ is similar to action $b$ as much as action $b$ is similar to action $a$). Asymmetry can be overcome by redefining the KL distance as 
$d'_{KL}(\mathbf{f},\mathbf{g}) = d_{KL}(\mathbf{f},\mathbf{g}) + d_{KL}(\mathbf{g},\mathbf{f})$ \cite{Nin09}. Alternatively, the KL distance can be modified into: 
\begin{equation*}
	d_J(\mathbf{f},\mathbf{g}) = \sum_{i=1}^{n}{ (f_i - g_i)(\ln{f_i} - \ln{g_i}) } \text{,}
\end{equation*}
referred to as the Jeffreys divergence, which is numerically stable, symmetric, and robust to noise \cite{Puz97,Rub00}. \\

The \textit{Bhattacharyya coefficient}, $d_B$, which measures the overlap between two probability distributions is defined as: 
\begin{equation*}
	d_B(\mathbf{f},\mathbf{g}) = \sum_{i=1}^{n}{ { \left[ f_i \cdot g_i \right] }^{\sfrac{1}{2}} } \text{.}
\end{equation*}
The Bhattacharyya coefficient, which is not to be confused with the Bhattacharyya distance, is bounded below by zero and above by one. Zero indicates no overlap and one indicates a perfect match between two normalized distributions $\mathbf{f}$ and $\mathbf{g}$. The bounded nature of the Bhattacharyya coefficient makes the measure robust to small outliers, which is favourable in action recognition application due to occlusion that could affect the overall distribution \cite{Der13,Yef09}. \\

The partial matches between two histograms in their corresponding bins can be modelled using a histogram intersection (HI) \cite{Swa91}. Histogram Intersection (HI) \cite{Swa91}, $d_{HI}$, is defined as:
\begin{equation*}
	d_{HI}(\mathbf{f},\mathbf{g})= 1 - \frac{ \sum_{i=1}^{n}{ \min{(f_i,g_i)} } }{ \sum_{i=1}^{n}{ f_i } } \text{.}
\end{equation*}
Interestingly, when the two histograms have the same size (i.e. $\sum_i f_i = \sum_i g_i$), then the histogram intersection of $\mathbf{f}$ and $\mathbf{g}$ is equivalent to the normalized $L_1$-distance \cite{Swa91}. \\

So far, all the measures that were mentioned in this section measured the similarity (or dissimilarity) between histograms bin-to-bin (i.e. compare $f_i$ and $g_i$ $\forall \ i$ but never $f_i$ and $g_j$ for $i \neq j$). This forces the two histograms to have the same bin sizes, which could cause the histogram to lack the discriminating power due to coarse binning or grouping of similar features due to fine binning. Thus, the flexibility for histograms to have different sizes and the ability to compare them across bins could be more robust and more useful \cite{Rub00}. \\

The \textit{Earth Mover's distance} (EMD) \cite{Rub98} is a cross-bin comparison metric that computes the minimal amount of work needed to transform one distribution to another. 
EMD can be broken down into a two-step process: (i) given two distributions, $\mathbf{f}\in\mathbb{R}^m$ and $\mathbf{g}\in\mathbb{R}^n$, find the flow with the smallest overall cost of transferring the distributional masses from $\mathbf{f}$ to $\mathbf{g}$ (or from $\mathbf{g}$ to $\mathbf{f}$), then 
(ii) use the flow to determine the amount of work required to transfer the distribution masses.
To find the optimal flow, $\mathbf{\phi}^*$, is to solve the following transportation problem:
\begin{equation}\label{eq:emd_transport}
	\mathbf{\phi}^* = \arg\min_{\phi_{ij}}{ \sum_{i=1}^{m}{ \sum_{j=1}^{n}{ \phi_{ij}\delta_{ij} } } } \text{,}
\end{equation}
where $\phi_{ij}$ is the flow between $f_i$ and $g_j$ for $1 \leq i \leq m, 1 \leq j \leq n$, and $\delta_{ij}$ is the ``ground distance'' between $f_i$ and $g_j$ $\forall \ i,j$, which can be any distance measure between single elements (e.g. $L_1$-norm \cite{Rub00}, $L_2$-norm \cite{Rub00,CYuan09}) depending on the features. 
Since \eqref{eq:emd_transport} is a transportation problem (see Figure \ref{fig:emd}), the optimal flow, $\mathbf{\phi}^*$, can be found using linear programming \cite{Rub00}.
Then the EMD between two histograms, $\mathbf{f}$ and $\mathbf{g}$, is defined as the work normalized by the total flow: 
\begin{equation*}\label{eq:emd}
	d_{EMD}(\mathbf{f},\mathbf{g}) = \frac{ \sum_{i}^{m}{ \sum_{j=1}^{n}{\phi^*_{ij} \delta_{ij}}} }{ \sum_{i=1}^{m}{ \sum_{j=1}^{n}{ \phi^*_{ij} } } } \text{,}
\end{equation*}
where the normalization factor (total flow) is equivalent to the total weight of the smaller distribution, which prevents the measure from favouring the smaller distribution \cite{Rub00}.\\

\begin{figure}
	\begin{center}
		\begin{subfigure}{0.8\textwidth}
			\begin{center}
				\includegraphics[width=0.7\textwidth]{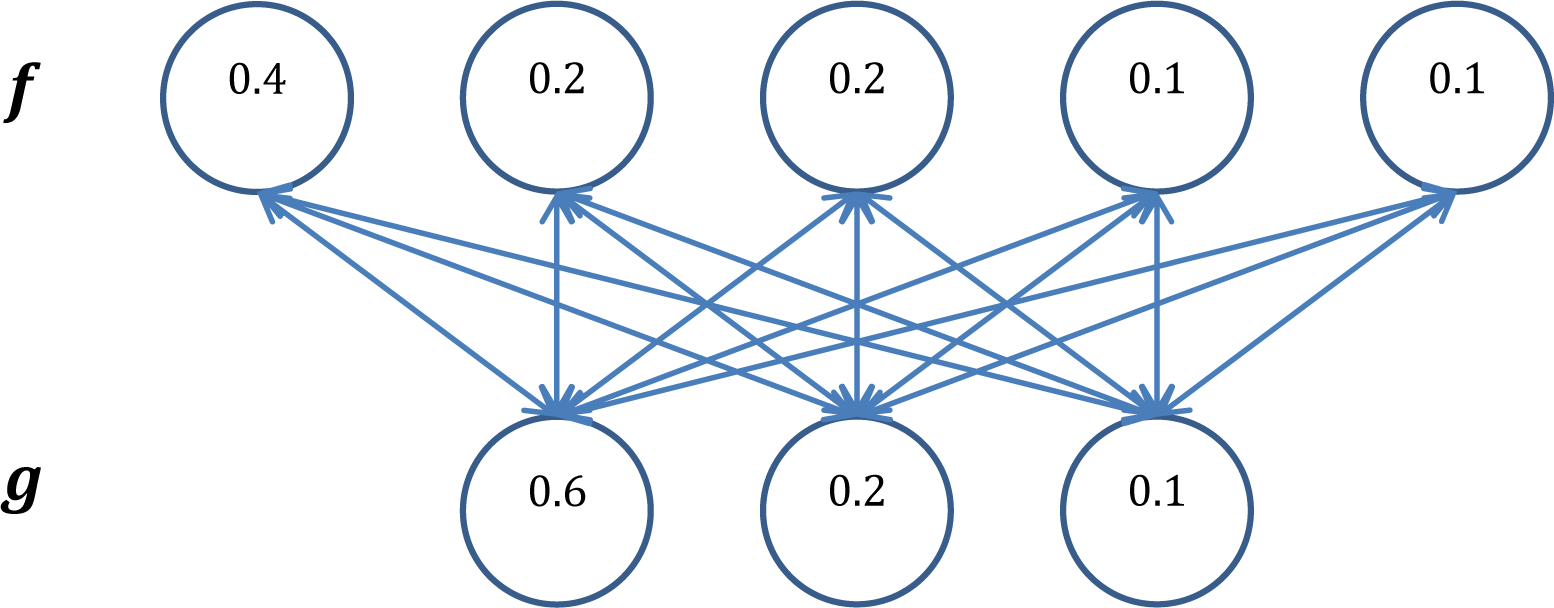}
				\caption{Transportation Problem \label{fig:emd_trnsprt}}
			\end{center}
		\end{subfigure}
		~
		\begin{subfigure}{0.15\textwidth}
			\begin{center}
				\includegraphics[width=\textwidth]{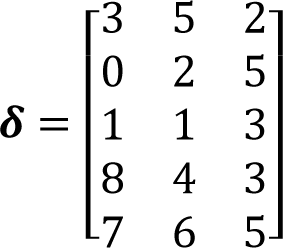}				
			\end{center}
		\end{subfigure}

		\begin{subfigure}{0.8\textwidth}
			\begin{center}
				\includegraphics[width=0.7\textwidth]{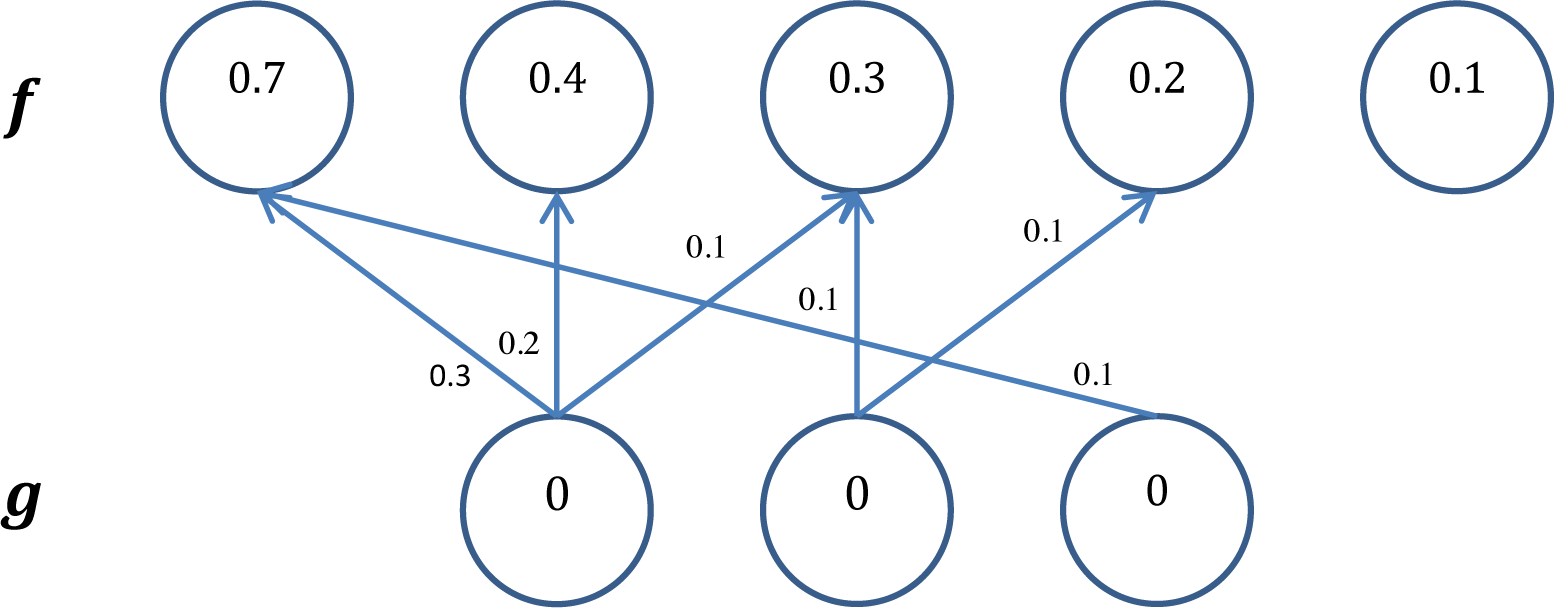}
				\caption{Solution to the Transportation Problem \label{fig:emd_cost}}
			\end{center}
		\end{subfigure}
		~
		\begin{subfigure}{0.15\textwidth}
			\begin{center}
				\includegraphics[width=\textwidth]{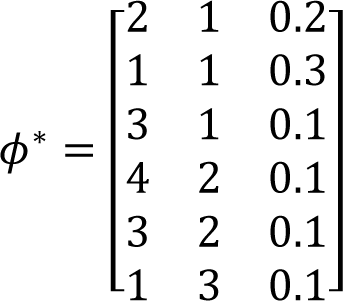}
			\end{center}
		\end{subfigure}
	\end{center}
	\caption{Example of the Earth Mover's Distance (EMD). 
	To calculate the EMD of $\mathbf{f}={[0.4 \ 0.2 \ 0.2 \ 0.1 \ 0.1]}^\top$ and $\mathbf{g}={[0.6 \ 0.2 \ 0.1]}^\top$, 
	(a) convert $\mathbf{f}$ and $\mathbf{g}$ into a transportation problem, where the cost (or ground distance), $\delta_{ij}$ between $f_i$ and $g_j$ for $1 \leq i \leq m, 1 \leq j \leq n$ is pre-defined. 
(b) The optimal flow $\phi^*$ of the transportation problem is found through linear programming. The columns of optimal flow, $\phi^*=[\phi_1 \ \phi_2 \ \phi_3]$, represents the amount of flow $\phi_3$ that is transferred from node $\phi_2$ to node $\phi_1$.
$d_{EMD}(\mathbf{f},\mathbf{g}) = \frac{0.2 \cdot 0 + 0.3 \cdot 3 + 0.1 \cdot 1 + 0.1 \cdot 4 + 0.1 \cdot 1 + 0 \cdot 5 + 0.1 \cdot 2}{0.2+0.3+0.1+0.1+0.1+0.1}=\frac{1.7}{0.9}=1.8889$. \label{fig:emd}}
\end{figure}

There are many cross-bin similarity measures \cite{Rub00}, but only the Earth Mover's distance is surveyed here. Other cross-bin measures are omitted since they are not as frequently used in the field of action recognition and detection. Comparison metrics of two histograms $\mathbf{f}$ and $\mathbf{g}$ that were described in this section are summarized in Table \ref{tab:metrics}.
\begin{table}[htbp]
	\begin{center}
	\begin{tabular}{| l | p{8cm} |}
		\bottomrule
		\textbf{Metric Type} 			& \textbf{Comparison Metric}, $d(\mathbf{f},\mathbf{g})$ \\
		\toprule \bottomrule
		$L_p$-norm  ($d_{L_p}$)			& ${\left[ \sum_{i=1}^{n}{ {|f_i - g_i|}^p } \right]}^{\sfrac{1}{p}}$ \\
		\hline
		$\chi^2$-distance ($d_{\chi^2}$)	& ${\left( \frac{1}{2}{\sum_{i=1}^{n}{ \frac{(f_i - g_i)^2}{f_i + g_i} } } \right)}^{\sfrac{1}{2}}$ \\
		\hline
		Mahalanobis distance ($d_M$) 		& $ {\left[ (\mathbf{f}-\mathbf{g}) \Sigma^{-1} (\mathbf{f} - \mathbf{g})^\top \right]}^{\sfrac{1}{2}} $\\
		\hline
		Kullback-Leibler distance ($d_{KL}$) 	& $ \sum_{i=1}^{n}{f_i \cdot \ln{\left( \frac{f_i}{g_i} \right)}} $ \\
		\hline
		Jeffreys divergence ($d_J$)		& $\sum_{i=1}^{n}{ (f_i-g_i)(\ln{f_i}-\ln{g_i}) }$ \\
		\hline
		Bhattacharyya coefficient ($d_B$)	& $\sum_{i=1}^{n}{ {\left[ f_i \cdot g_i \right]}^{\sfrac{1}{2}} }$ \\
		\hline
		Histogram Intersection ($d_{HI}$)	& $1-\frac{ \sum_{i=1}^{n}{ \min{(f_i,g_i)} } }{ \sum_{i}^{n}{f_i} }$ \\
		\hline
		Earth Mover's distance ($d_{EMD}$)	& $ \frac{ \sum_{i=1}^{m}{ \sum_{j=1}^{n}{ \phi_{ij}^*\cdot\sigma_{ij} } } }
									   { \sum_{i=1}^{m}{ \sum_{j=1}^{n}{ \phi_{ij} } } } $,
								where $\phi^*_{ij}$ is the optimal flow that minimizes the cost of $\sum_{i=1}^{m}{ \sum_{j=1}^{n}{ \phi_{ij} \delta_{ij} } }$, and $\delta_{ij}$ is the ground distance between each element in $\mathbf{f}$ and $\mathbf{g}$\\
		\toprule
	\end{tabular}
	\end{center}
	\caption{Histogram Comparison Metric Summary. All metrics, but the Earth Mover's distance, described in this section measure similarity (or dissimilarity) between two histograms $\mathbf{f}$ and $\mathbf{g}$ bin-to-bin. Thus, $\mathbf{f},\mathbf{g} \in \mathbb{R}^n$. The Earth Mover's distance compares the two histograms in a cross-bin manner. Thus, the sizes of the two histograms can vary (i.e. $\mathbf{f} \in \mathbb{R}^m$ and $\mathbf{g} \in \mathbb{R}^n$ for $m \neq n$). \label{tab:metrics}}
\end{table}

\end{section}

\begin{section}{Deterministic Models}\label{sec:classification_deterministic}
	Query data can be assigned to one action class or another without considering the probability distribution between classes of the training data in deterministic models. A set of training data can be learned in either a (i) lazy, or (ii) eager manner. \textit{Lazy learning classifiers} makes generalizations only when query data appears. \textit{Eager learning classifiers}, on the other hand, makes generalizations using the training data before it sees the query data. Thus, it takes more time to train eager learning algorithms, but less time to predict the class of the test data than lazy learning algorithms \cite{Dou13}. Here, some common lazy and eager learners that are used in various action recognition and detection algorithms will be studied. \\

	\begin{subsection}{Lazy Learners}
		Lazy-based learning classifiers defer data processing until they receive a request to classify an unlabelled test example \cite{Dou13}. The classifier waits for query data before it makes any generalizations about the data. One common lazy learning classifier used in action recognition is the $k$-nearest neighbour (kNN) classifier \cite{Lin09,Oga06}.
It determines the class of the test sample by growing a spherical region centred at the sample until the region contains $k \in \mathbb{Z}^+$ training data. The test data is labelled by the class with the majority vote in the enclosed space (see Figure \ref{fig:knn}) \cite{Dou13}. Many earlier algorithms set $k=1$, to find the nearest neighbour (i.e. template) to the query (i.e. test) vector \cite{Efr03,Lin09}. The distance between the training set and the test data can be obtained via a comparison metric mentioned in the previous section. Thus, computing can be expensive with a large training set. When there are two classes in the training set, an odd $k$ value is used to avoid ties between the classes. With more classes, larger $k$ values are used since they are more likely to break the ties \cite{Dou13}. Although the kNN classifier is simple to implement, it is prone to local noise. Furthermore, with an increase in the number of features, more training data is required leading to the case of \textit{curse of dimensionality}. To avoid bias when there are an unbalanced amount of training data from different classes or to assign more weight on false negatives over false positives, the standard kNN algorithm can be modified to assign a particular class to the test data if at least $l$ of the $k$ nearest neighbours are in that class for $l < k$ \cite{Dou13}.

		\begin{figure}[htbp]
			\begin{center}
				\includegraphics[width=0.6\textwidth]{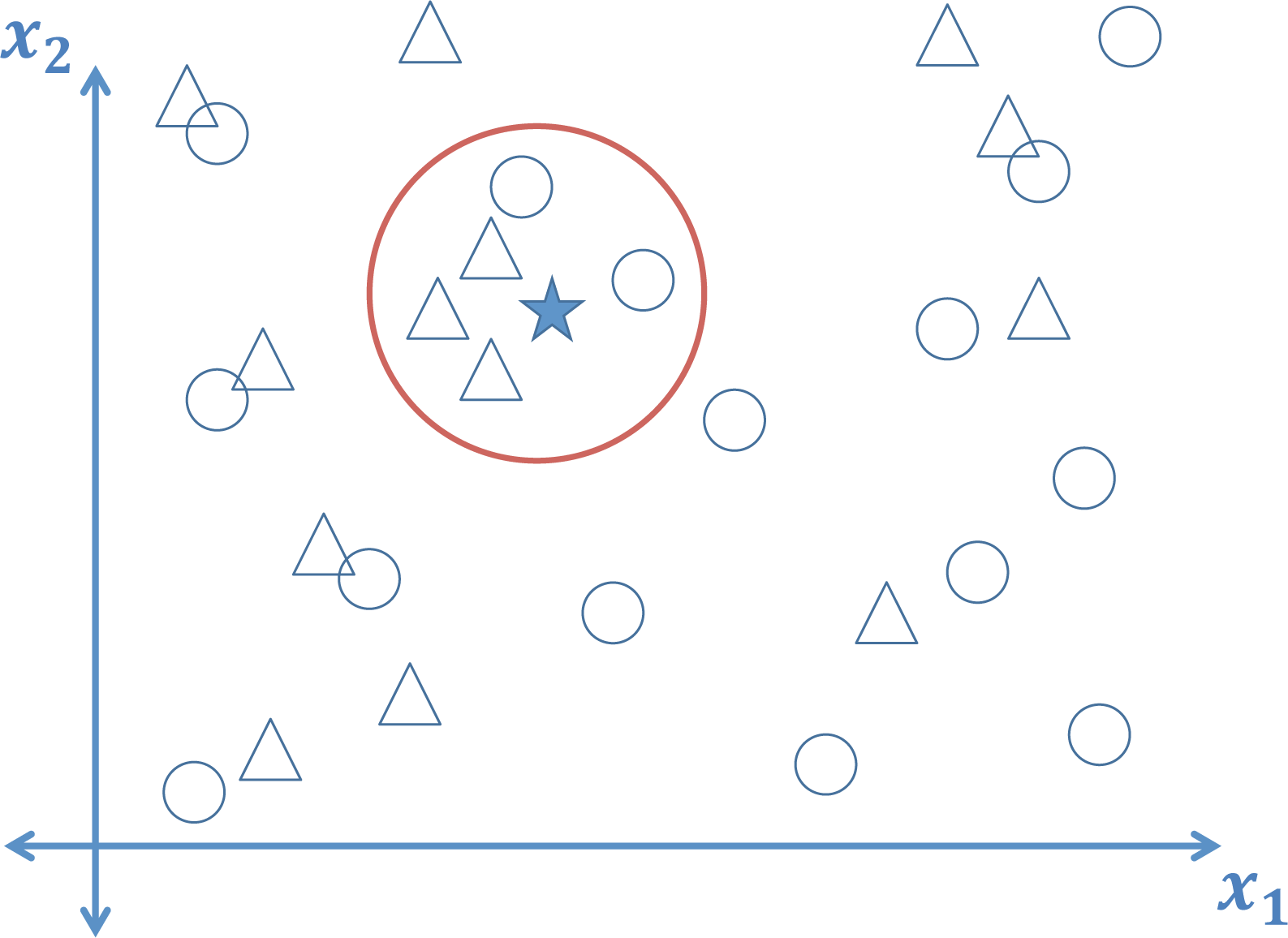}
			\end{center}
			\caption{$k$-nearest neighbours with $k=5$. A circular region (red) centred around the test sample (star) is expanded until $k=5$ samples (circles and triangles) are contained within the circular region. The test sample is labelled as the same class as triangle since there are more triangles (3) than circles (2) inside the bound region. \label{fig:knn}}
		\end{figure}
	\end{subsection}

	\begin{subsection}{Eager Learners}\label{sec:eager_learner}
		Given a collection of training data, \textit{eager learning classifiers} learn a model that would generalize the data as soon as it becomes available before the test data must be categorized. 
A model can be generated by partitioning the feature space of the data into a set of decision regions (see Figure \ref{fig:classify_decision_bdry}) \cite{Dou13}. These regions provide a guideline to classify the query feature into one of the classes. The decision regions are separated by decision boundaries, which can be described by a set of discriminant functions.
Some eager learning algorithms that are commonly used in action recognition and detection algorithms include: support vector machines (SVMs), AdaBoost, and artificial neural networks (ANNs). \\

\begin{figure}[htbp]
	\begin{center}
		\begin{subfigure}{0.45\textwidth}
			\begin{center}
				\includegraphics[width=0.8\textwidth]{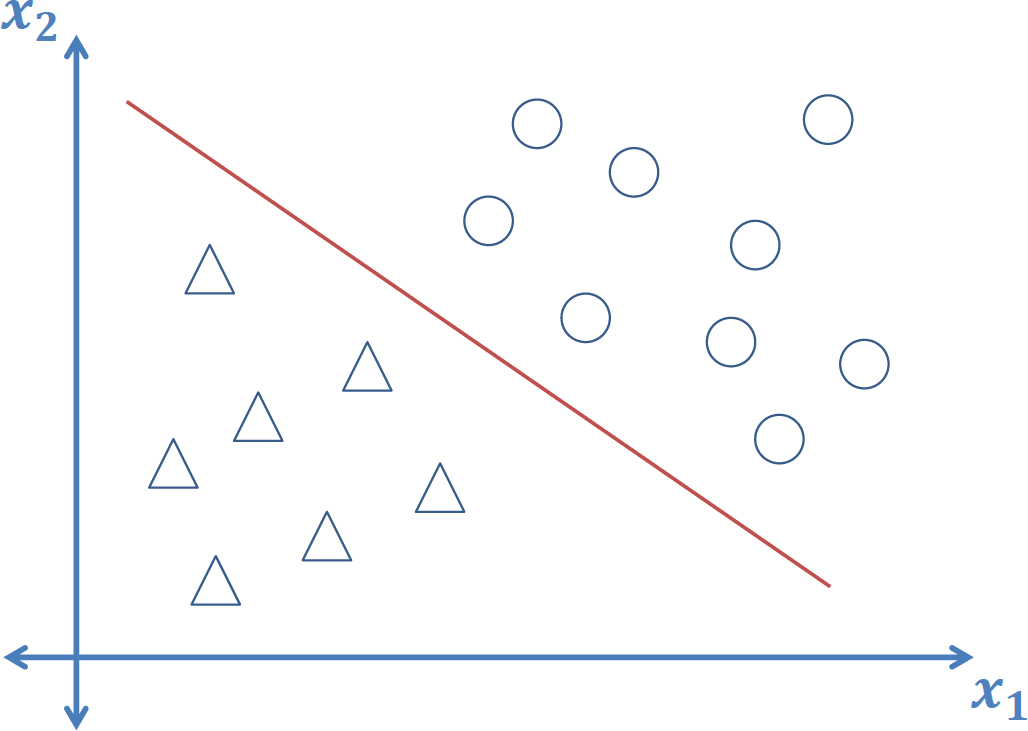}
				\caption{Linearly separable data}
			\end{center}
		\end{subfigure}
		~
		\begin{subfigure}{0.45\textwidth}
			\begin{center}
				\includegraphics[width=0.8\textwidth]{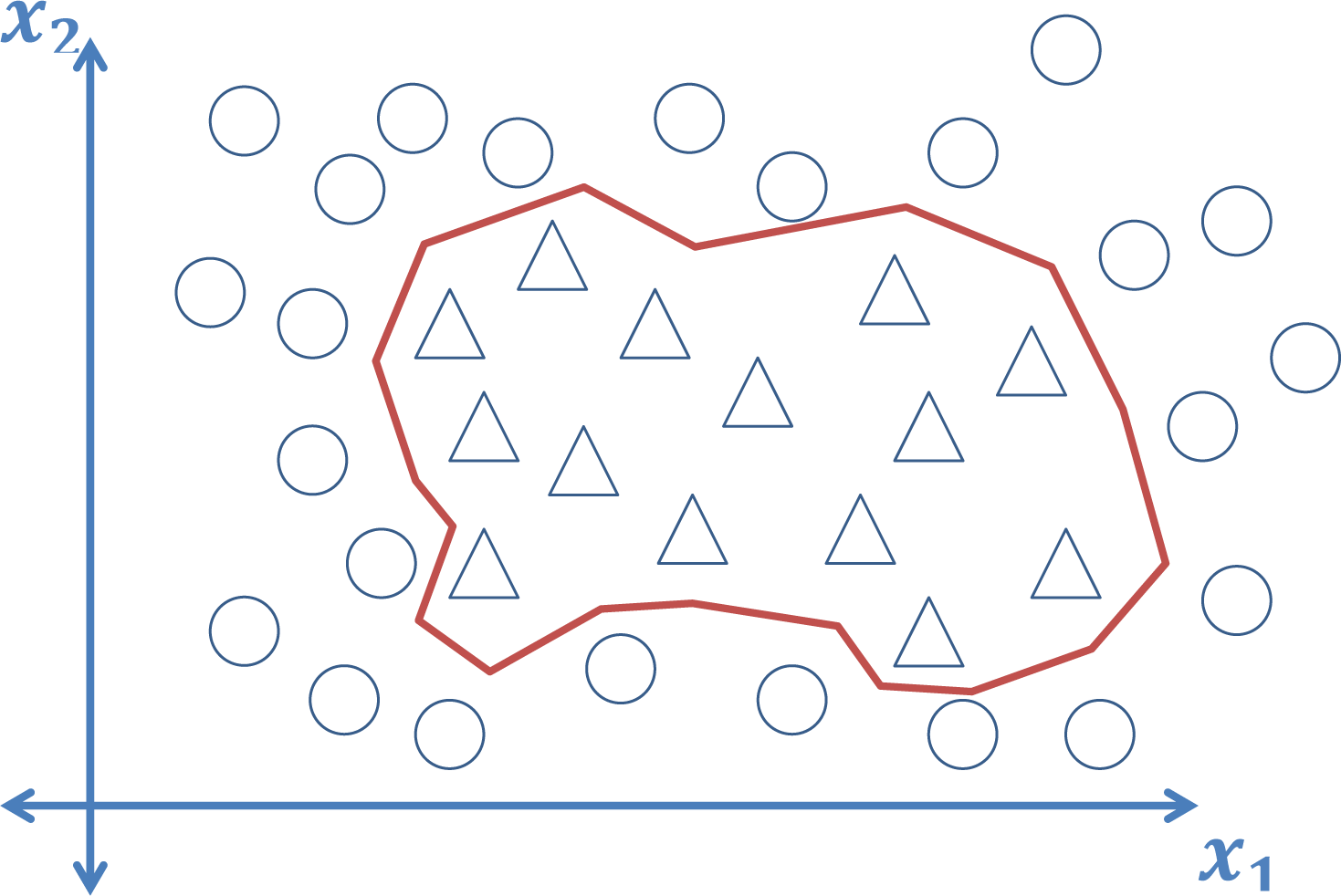}
				\caption{Non-linearly separable data}
			\end{center}
		\end{subfigure}
	\end{center}
	\caption{Decision Boundaries. Red lines indicate the decision boundary, which separates the samples of different classes (triangles and circles) into decision regions. (a) A linear decision boundary is the simplest decision boundary, which can be described by a linear (discriminant) function. (b) A non-linear decision boundary can be obtained with a set of complex polynomials. \label{fig:classify_decision_bdry}}
\end{figure}

		A \textit{support vector machine} (SVM) is one of the most common supervised classification tools used in action recognition and detection, e.g. \cite{Jai13, Jhu07, Jia12, Kov10, Lap08, Ma13, Mar09, Rap09, Sch08, JSun09, Tian13, Wan09, Wan11, Wan13, Yef09, Zha13}. An SVM is trained to find a hyperplane (or a decision boundary) that separates labelled data from two classes into its respective groups. The best hyperplane is the one that separates the two classes with the largest distance between the nearest point from each class to the hyperplane (see Figure \ref{fig:svm}). Since action recognition involves classifying videos into multiple actions (classes), a multi-class SVM must be employed, which can be done by applying the \textit{one-versus-all} approach \cite{Lap08,Zha07}. The \textit{one-versus-all} approach takes the training data from class $k$ labelled as positive and the rest as negative examples to train the $k$th model. Kernels enable implicit operation in a higher dimensional feature space, where hyperplane separability may be possible. There are two types of kernels: (i) linear, and (ii) non-linear. 
To determine what would be an appropriate kernel for the algorithm, one should examine the ratio between the number of features and the training data. A linear kernel is preferred when the number of features is large (i.e. high dimensional feature space) (e.g. DT/iDT features) relative to the number of training samples to prevent over-fitting in the feature space. When there are a few features with a lot of samples, a non-linear kernel would be a better choice. 
Although non-linear kernels typically achieve a lower error rate, linear SVMs are less computationally expensive and require less storage than non-linear SVMs allowing real-time detections possible \cite{Dal06,Yan09}. By adding more features, a linear SVM can be used. \\

	\begin{figure}[htbp]
		\begin{center}
					\includegraphics[width=0.6\textwidth]{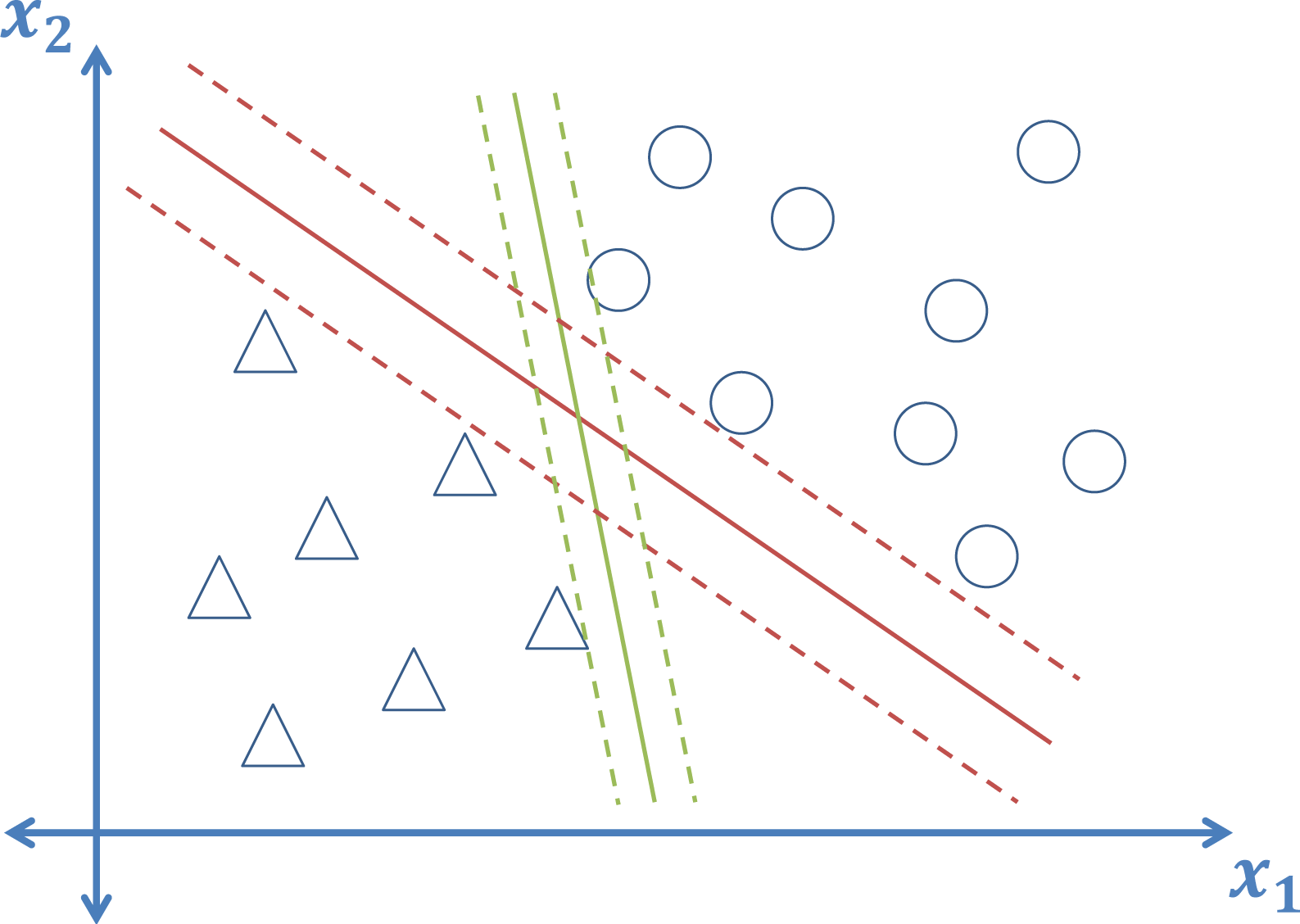}
		\end{center}
		\caption{Support Vector Machine (SVM). Solid lines indicate the decision boundary separating the samples of different classes (triangles and circles). Dashed lines are lines parallel to the decision boundary closest to the data of one class. SVM seeks a line that would maximize the margin, the distance between the dashed and solid line (i.e. red line). \label{fig:svm}}
	\end{figure}

	\textit{Adaptive Boosting} (AdaBoost) is a learning algorithm that takes several weak classifiers, classifiers that are slightly better than random guessing, and constructs a meta-classifier. By assigning different weights to training samples, different classifiers would pay more attention to different samples. The weights of an individual classifier is assigned depending on its accuracy \cite{Che14}. This approach has been applied with some success in various action recognition algorithms \cite{Fat08, Lap07, Liu09, Oga06}.\\

	Artificial neural networks (ANNs) are another widely used classification algorithm. The artificial neuron (\textit{perceptron}, or more generally referred to as \textit{units}) in each layer computes the weighted sum of its inputs. If the sum exceeds some specified threshold, the unit outputs a value \cite{Dou13}. 
A unit models a linear discriminant function partitioning the feature space using a decision boundary. 
Using a multilayer network, nonlinearly separable functions can be learned (see Figure \ref{fig:ann_decision_bdry}). 
The network is trained via backpropagation, which involves repeatedly presenting the training data to the network and adjusting the weights in the network to obtain a desired output \cite{Dou13,Dud01}. 
The number of units in the hidden layers govern the expressive power of the network \cite{Dud01}. A small number of hidden units is sufficient for well-separated or linearly separable patterns, but highly interspersed patterns with complicated densities require more hidden units. While a large number of hidden units produces a discriminative network lessening the training error, training becomes extremely time-consuming. Furthermore, it can lead to overfitting of the data, causing random noise in the test data to be modelled and poor generalization to the test data \cite{Dou13}. An ANN with too few hidden units would not have enough parameters to fit the training data, yielding poor classification results on the test data. Thus, finding an intermediate number of hidden units is key to obtaining good classification results with such powerful classification tool. \\

ANNs and CNNs (mentioned in Section \ref{sec:feature_descriptors}) have very similar architectures. Both networks output class scores of a feature vector by processing the components of a feature vector into a sequence of input, hidden, and output layers \cite{Dud01}. Each layer consists of a set of units, where each unit in the hidden layer receives some input, performs a dot product, and optionally follows it with a non-linearity. 
Based on an assumption that input signals from the domain of interest (e.g. images) are locally correlated (e.g. spatially neighbouring pixels), CNNs allow their receptive fields of the hidden units to have a relatively local support \cite{LeC98}, while more general ANNs do not. This allows units in the hidden layers of a CNN to be connected to a local neighbourhood of the previous layer, while all units in every layer of a general ANN is allowed to be fully-connected. Fewer connections between units significantly reduces the number of parameters (weights) that must be learned \cite{LeC98}. Consequently, fewer weights reduces the number of training that is required to cover the space of possible variations. Furthermore, it reduces the amount of memory required to store the weights in the hardware \cite{LeC98}. Remark, the last layers of a typical CNN architecture can be fully-connected. This allows for an output of a class, a class probability, or features that can be fed into another classifier (e.g. SVM).

\begin{figure}[htbp]
	\begin{center}
		\begin{subfigure}{0.7\textwidth}
			\begin{center}
				\includegraphics[width=0.8\textwidth]{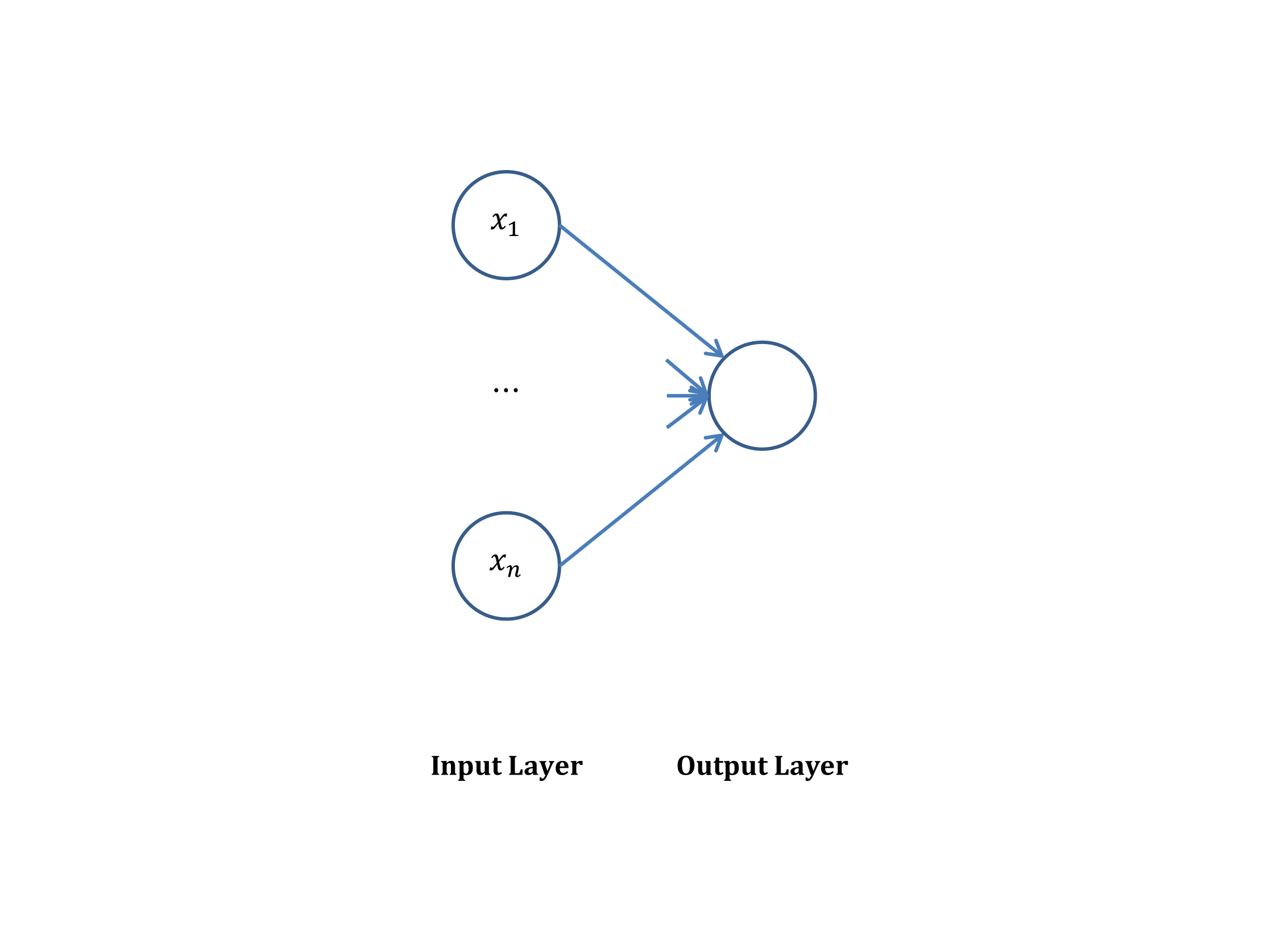}
				\caption{Two-Layer Neural Network \label{fig:ANN_2layer}}
			\end{center}
		\end{subfigure}
		~
		\begin{subfigure}{0.25\textwidth}
			\begin{center}
				\includegraphics[width=0.8\textwidth]{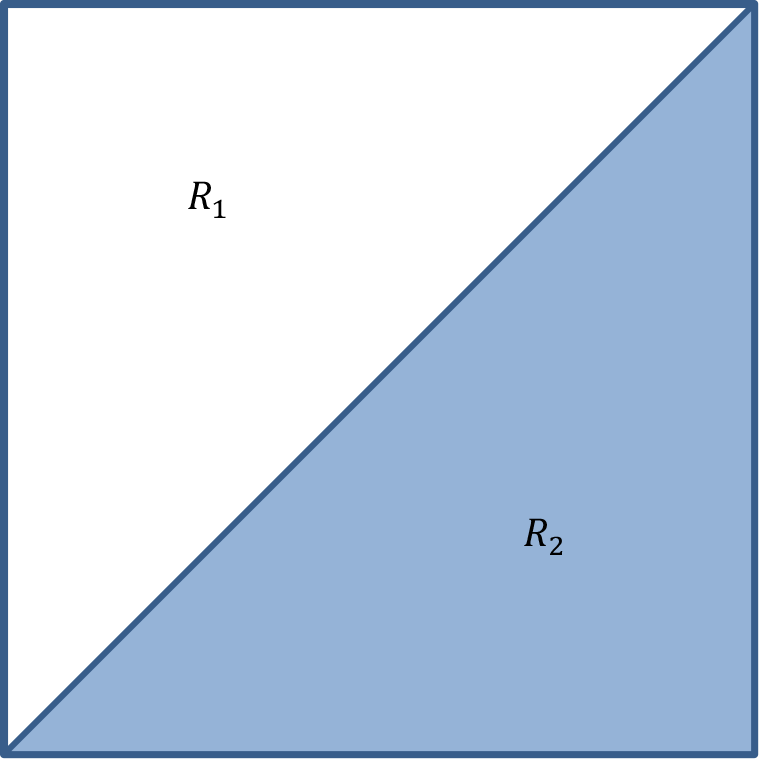}
				\caption{Linear Decision Boundary \label{fig:ANN_2layer_seg}}
			\end{center}
		\end{subfigure}

		\begin{subfigure}{0.7\textwidth}
			\begin{center}
				\includegraphics[width=0.8\textwidth]{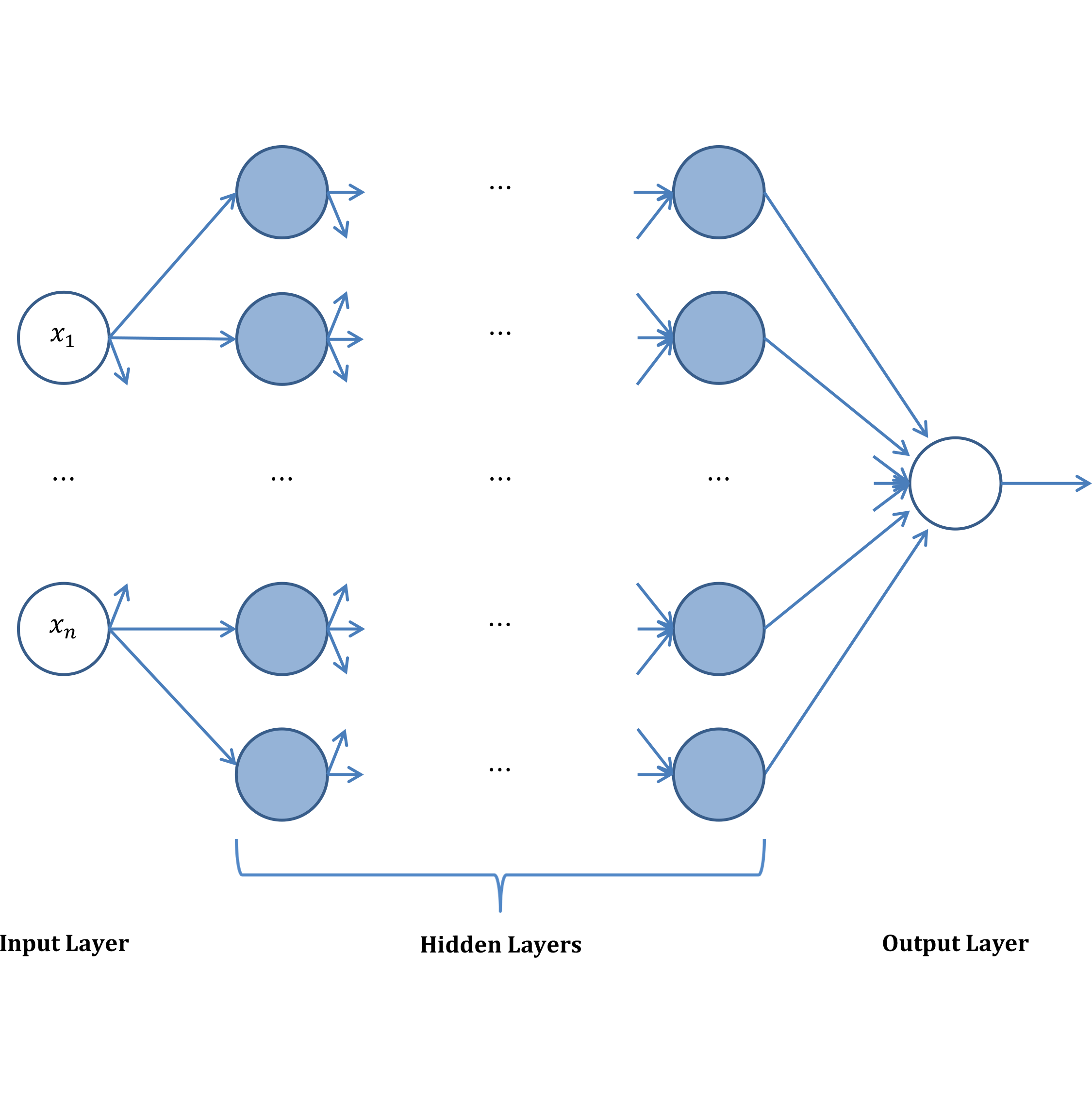}
				\caption{Multi-layer Neural Network \label{fig:ANN_multilayer}}
			\end{center}
		\end{subfigure}
		~
		\begin{subfigure}{0.25\textwidth}
			\begin{center}
				\includegraphics[width=0.8\textwidth]{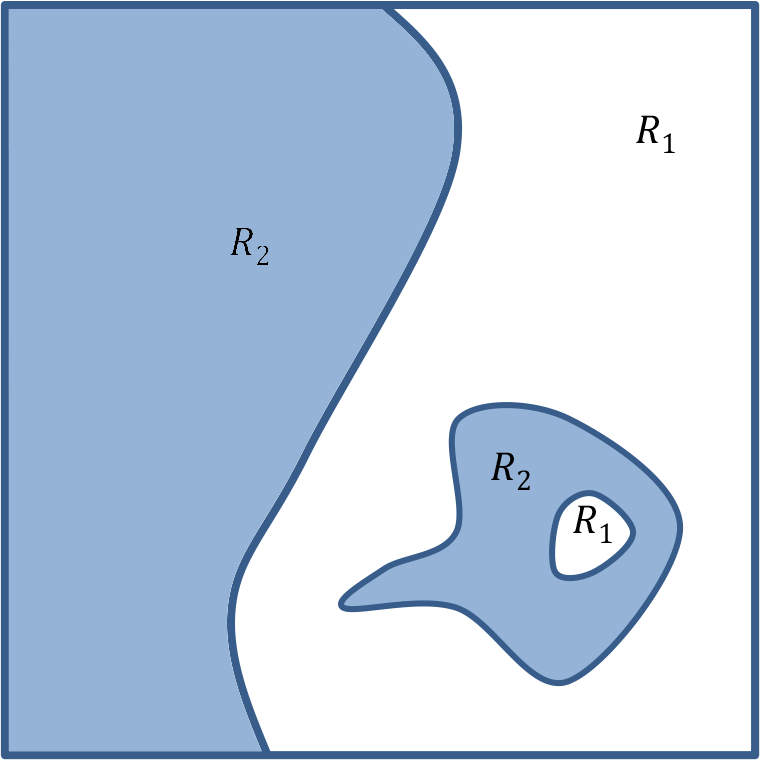}
				\caption{Arbitrary Decision Boundaries \label{fig:ANN_multilayer_seg}}
			\end{center}
		\end{subfigure}		
	\end{center}
	\caption{Artificial Neural Networks (ANNs) with different number of layers. While a two-layer neural network classifier (\ref{fig:ANN_2layer}) is only capable of implementing linear decision boundaries (\ref{fig:ANN_2layer_seg}), a multi-layer neural network (\ref{fig:ANN_multilayer}) with an appropriate number of hidden units can implement arbitrary decision boundaries (\ref{fig:ANN_multilayer_seg}), which do not necessarily have to be convex nor simply connected. Adapted from \cite{Dud01}. \label{fig:ann_decision_bdry}}
\end{figure}

	\end{subsection}

%
\end{section}
	
\begin{section}{Probabilistic Models}\label{sec:classification_probabilistic}
Probabilistic models learn the probability distribution over the set of classes to determine the probability of the query data belonging to each action class. These probabilistic models can be broadly categorized into two types: general classifiers and temporal state-space classifiers. \textit{General classifiers} categorize features without explicitly modelling variations in time while \textit{temporal state-space models} use temporal order information of features. Here, we look at probabilistic models that fall under general or temporal state-space models.

	\begin{subsection}{General Classifiers}
		The relationship between features and their respective action class can be modelled using probabilities. Here, we examine some common general probabilistic models that have been implemented in the field of action recognition, such as the naive Bayes classifier, latent topic discovery models, relevance vector machines, and the Bayesian network. \\

		The \textit{naive Bayes classifier} is one of the simplest probabilistic models that assigns a feature, $\mathbf{x}$, to some action class $c$ by comparing the posterior probability $P(c_k|\mathbf{x})$ $\forall \ c_k \in C$ \cite{Dou13}. Applying the Bayes' rule, the conditional posterior probability can be written as:
		\begin{equation}\label{eq:bayes}
			P(c_k|\mathbf{x}) = \frac{P(\mathbf{x}|c_k) P(c_k)}{P(\mathbf{x})} \text{,}
		\end{equation}	
		where $P(\mathbf{x}|c_k)$ represents the probability of feature $\mathbf{x}$ (e.g. filter bank \cite{Cho99}) belonging to class $c_k$, $P(c_k)$ and $P(\mathbf{x})$ represent probabilities of observing class $c_k$ and feature $\mathbf{x}$, respectively. $P(\mathbf{x}|c_k)$, $P(c)$, and $P(\mathbf{x})$ can all be trained from observing the distributions within the training set. The naive Bayes classifier makes a naive assumption that the features are conditionally independent to one another given its class (i.e. $P(x_1, \dots, x_n|c_k) = P(x_1|c_k) \dots P(x_n|c_k)$). Then the test feature $\mathbf{x}$ can be assigned to the class with the maximum a posterior probability $P(c|\mathbf{x})$ \cite{Dud01}, which is formulated as 
		\begin{equation}
			c_k 	= \arg{ \max_{c_k \in C}{ P(c_k|\mathbf{x}) } }  
				= \frac{1}{ P(\mathbf{x}) }\arg{ \max_{c_k \in C}{ P(x_1|c_k) \dots P(x_n|c_k) P(c_k) } } \text{.}
		\end{equation}
	Through the naive Bayes independence assumption, which may not necessarily be true, naive Bayes classifier is a simple classifier that is a good candidate for implementation for its simplicity and efficiency. \\ 
	
		Latent topic discovery models are statistical models that were originally popularized for the discovery of topics in a text. This approach can be extended to discover any latent classes in a collection of data, such as actions in videos. Two latent topic discovery models, probabilistic Latent Semantic Analysis (pLSA) \cite{Hof99} and Latent Dirichlet Allocation (LDA) \cite{Ble03}, have commonly appeared in various action recognition algorithms \cite{Nie08,JWang13,Zha08} \cite{Nie08,YWang09}. 
pLSA and LDA model the distribution of classes in sets of videos, such that the model can be used to classify the latent topics (i.e. action classes) in the new videos. 
		pLSA assumes that a video sequence, $v_i$, and a feature, $f_j$, are conditionally independent given an action class, $c_k$, (see Figure \ref{fig:pLSA}), 
		then the action class of the test data $v'$ can be best described by solving 
		\begin{equation*}
			c^* = \arg{\max_{k}{ P(c_k|v') }} \text{,}
		\end{equation*}
		which can be computed using the EM algorithm \cite{Nie08}.
		To determine a model that best represents a mixture of actions that could occur in a single video, the most optimal action proportion in videos, $p(c_k | v_i)$ for $i = 1, \dots, N$, must be learned. pLSA learns the class mixture probabilities by going through each video in the training set to describe the process of generating videos with action class distributions that was in some video in the set. \\

		The pLSA approach enforces the model to be stringent, placing new (unseen) videos at points within the pre-defined action distribution, leading to an overfit model (see Figure \ref{fig:pLSAvsLDA}) \cite{Ble03}.  To overcome this stringency, LDA sets arbitrary topic proportions (a prior probability distribution) $\theta_i$ for each video $v_i$. To ensure that the number of parameters to be learned does not grow linearly with the number of videos in the training set, the mixing proportions of actions per video are controlled by a parameter $\alpha$, which is specified per set of videos (see Figure \ref{fig:LDA}). 
Then the joint distribution of the action class mixture, $\theta$, a set of classes $\mathbf{c}$, and a set of features, $f$, is given by:
		\begin{equation*}
			p(\theta,\mathbf{c},f|\alpha,\beta) = p(\theta,\alpha) \prod_{k=1}^{K}{p(c_k|\theta) p(f_n|c_k,\beta) } \text{,}
		\end{equation*}
where $\mathbf{\beta}$ parameterizes the distribution of the features within a particular action label (i.e. $\beta$ corresponds to $p(f_i | c_k)$). 
		Parameters $\alpha$ and $\beta$ are found using the EM algorithm for a given collection of video sequences \cite{Ble03}. 
		The feature $f_j$ is classified to be action $c^*$ if 
		\begin{equation}
			c^* = \arg{ \max_{k}{ p(c_k|f_j,\alpha,\beta) } } \text{.}
		\end{equation}
		Since the number of topics is fixed to a particular value in LDA, it prevents overfitting from occurring, especially if a video contains a small amount of features to train, since it can rely on the prior to give a more reasonable guess about the actions for that video \cite{Nie08}. Conversely, if videos are known to produce a large amount of features, then the data would dominate the priors. Finally, it is worth noting that pLSA is more computationally efficient than LDA \cite{Mas08}.\\

		\begin{figure}[htbp]
			\begin{center}
				\begin{subfigure}{0.45\textwidth}
					\includegraphics[width=0.8\textwidth]{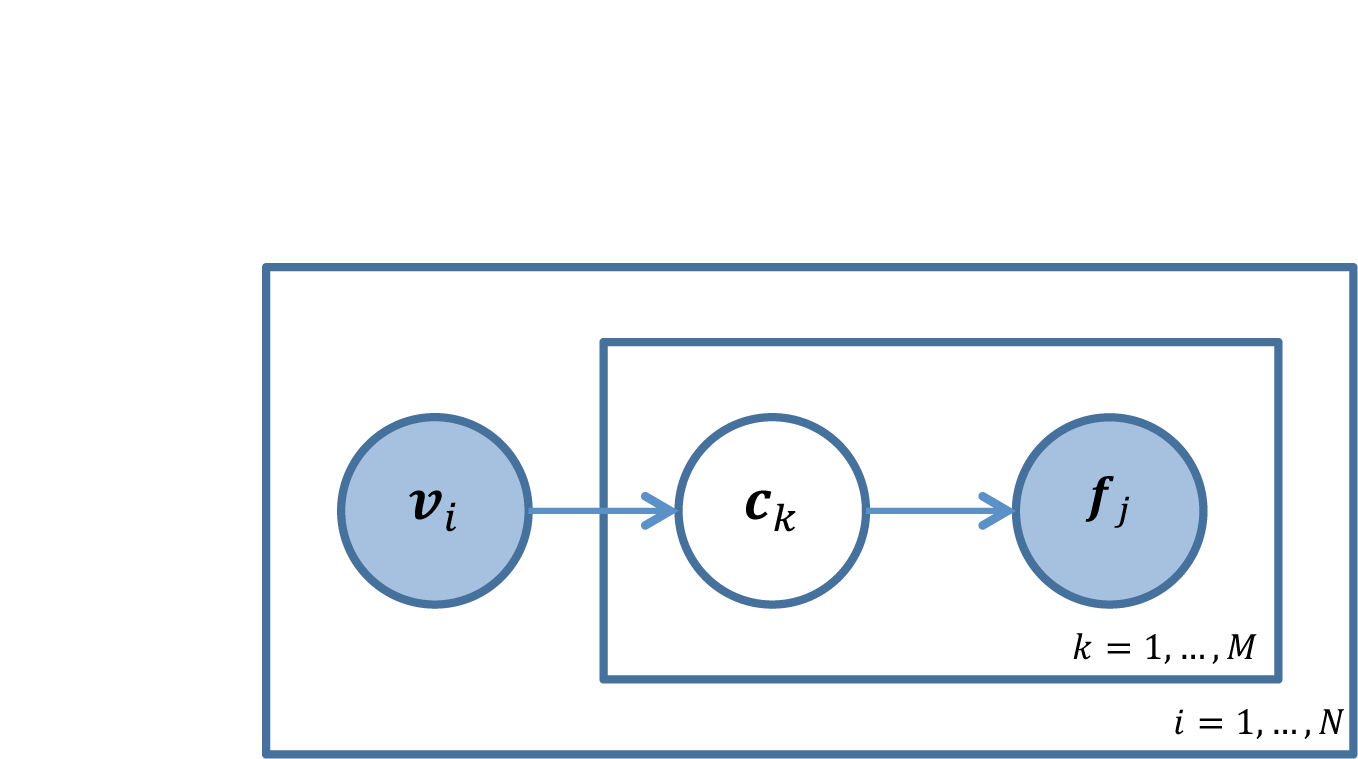}
					\caption{pLSA \label{fig:pLSA}}
				\end{subfigure}
				~
				\begin{subfigure}{0.45\textwidth}
					\includegraphics[width=0.8\textwidth]{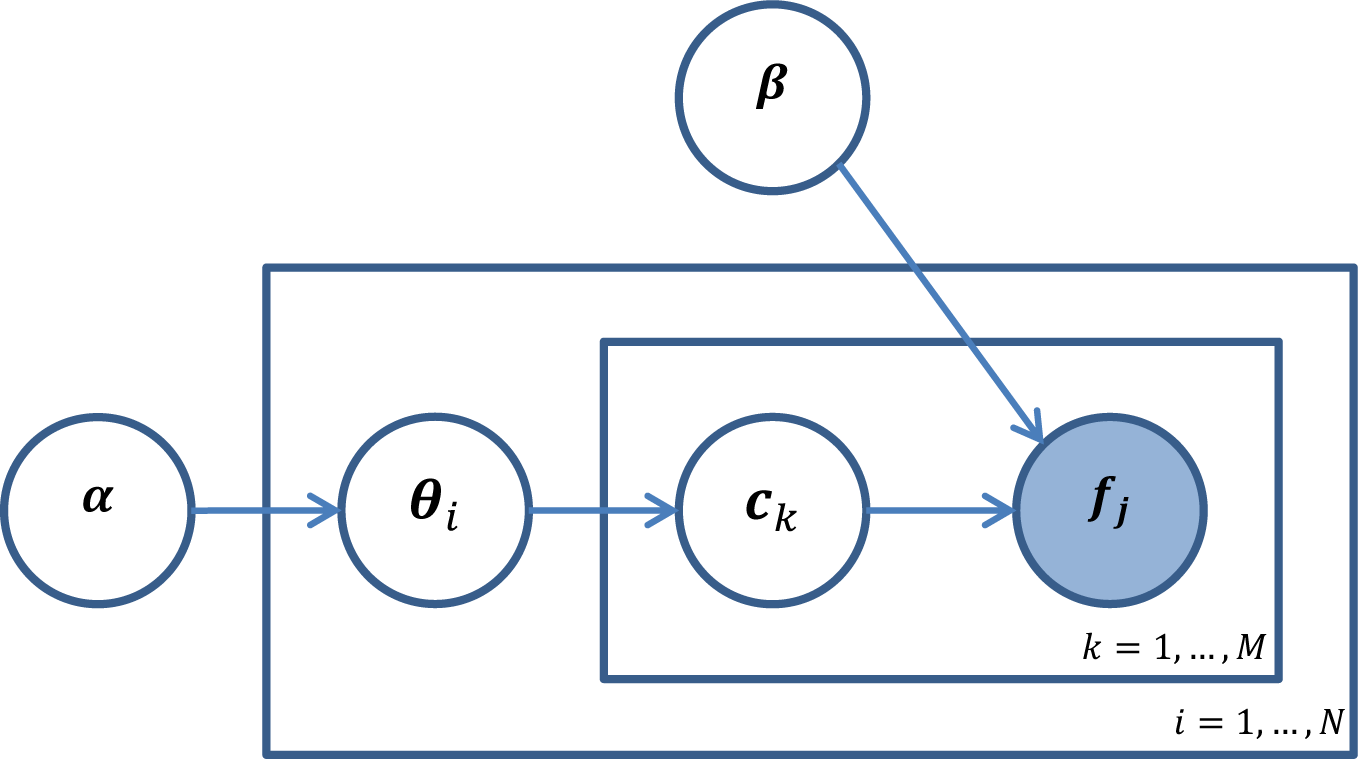}
					\caption{LDA \label{fig:LDA}}
				\end{subfigure}
			\end{center}
			\caption{A graphical model of latent topic discovery models: pLSA and LDA. 
			A graphical model provides a layout of the causal and independent relationships between each variables in a system.
			Shaded regions are observed and unshaded regions are unobserved (hidden/latent) variables. 
(a) pLSA. pLSA assumes that feature $f_j$ is conditionally independent of video $v_i$ given the action class $c_k$. For each video $v_i$, a latent class $c_k$ is chosen from the video's class multinomial distribution, $P(c_k|v_i)$ where $k=1, \dots, M$, to draw a feature $f_j$ from the class' multinomial distribution of the words, $P(f_j|c_k)$. 
(b) LDA. For each video, the vector of topic proportions, $\theta_i$, is sampled according to a Dirichlet distribution with parameter $\alpha$. For each feature, $f_k$, in a video, class $c_k$ is selected from the multinomial distribution over the classes with parameter $\theta$, $p(c_k|\theta)$, to choose a feature $f_k$ from a multinomial distribution conditioned on class $c_k$, $p(f_k|c_k,\beta)$. $\alpha$ and $\beta$ are sampled once in the process of generating a set of $N$ videos, while $f_j$ and $c_k$ are sampled for every feature in each video. Adapted from \cite{Ble03}. \label{fig:topic_discovery}}
		\end{figure}

		\begin{figure}[htbp]
			\begin{center}
				\includegraphics[width=0.5\textwidth]{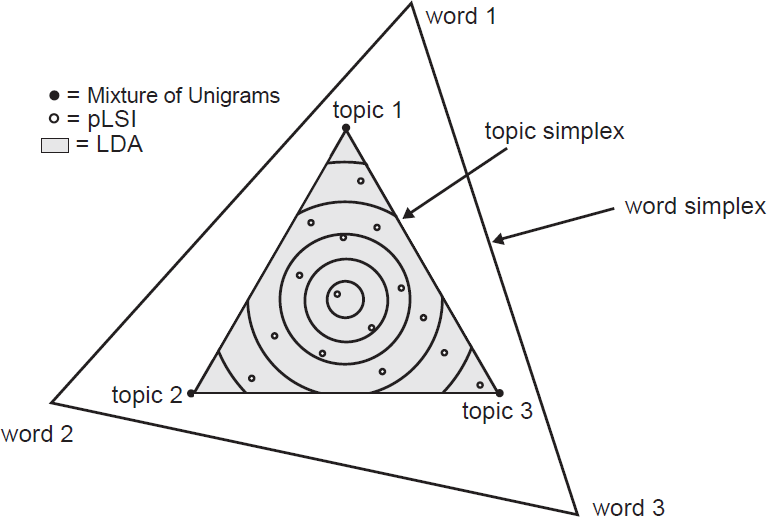}
			\end{center}
			\caption{Latent topic discovery model comparison. The \textit{mixture of unigrams} places documents at the corners of the topic simplex as the model permits one topic assignment to each document. \textit{pLSA} and \textit{LDA}, on the other hand, allow multiple topics to be assigned to a document. Therefore, the empty circles (pLSA) and the shaded area (LDA) lies within the topic simplex (triangle). In contrast to LDA, which can place the document anywhere in the shaded region, the topics for pLSA must be placed at one of the specified points. The smooth Dirichlet distribution determined by parameter $\alpha$ determines the contour of the topic simplex. Redrawn from \cite{Gim08}. \label{fig:pLSAvsLDA}}
		\end{figure}
	Relevance vector machines (RVMs) have an identical functional form as the deterministic model, SVM. It finds a hyperplane that separates the relevance vectors into two classes. Different from SVMs, RVMs provide a probabilistic classification instead of a deterministic decision. Furthermore, the hyperplane separates \textit{relevance vectors}, prototypical representations of classes (e.g. action class), instead of \textit{support vectors} (examples close to the decision boundary) \cite{Oik06}. 
RVMs tend to have a longer training time than SVMs. However, since RVMs result in a sparser set of support vectors, the computation time for test points is much less than on an SVM \cite{Bis06}.\\

		Some algorithms design probabilistic models using a \textit{Bayesian network} suited to incorporate the necessary variables to recognize the action \cite{Fan05}. Using a graphical representation, a complex system can be decomposed into simpler parts to provide a causal relationship between the variables. In addition, graphical models factorize variables into several conditional probability distributions that are simpler to compute \cite{Luo03,Sut12,Tom08}. For example, Figure \ref{fig:graphical_model} suggests the following factorized joint probability: 
	\begin{equation*}
		p(\theta,x,y,a,b,s,\delta) 
		= p(\sigma) p(s|\delta) p(y|x,s,\delta) p(b|a,s,\delta) p(a) p(\theta) p(x|\theta) \text{,}
	\end{equation*}
	where $\theta,x,y,a,b,s,\delta$ are parameters that indicate the centroid, position and velocity of body parts, position and velocity of detections, appearance of body parts, appearance of a detection, map of body parts to detects, and detection of body parts, respectively. Then each factor can be modelled through training, or by making appropriate assumptions between variables (e.g. $N$ labels $s_i$ of $M$ body parts are equally likely to be detected and mutual independence implies that $p(s|\delta)=\prod_{i=1}^{M}{ p(s_i|\delta_i) }=(\frac{1}{N})^M$). By combining probability and graph theories into the system, uncertainty and complexity can be dealt with simultaneously \cite{Tom08}.

		\begin{figure}[htbp]
			\begin{center}
				\includegraphics[width=0.4\textwidth]{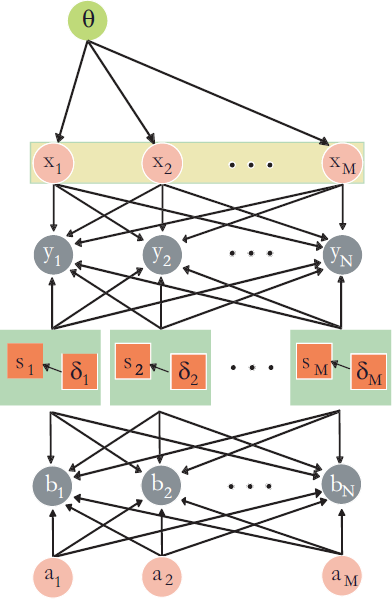}
			\end{center}
			\caption{A graphical model of the Bayesian Network (BN). A BN represents the joint probability of the variables in a complex systems as a factor of simpler parts to simplify the computation. In this graphical model of the BN, each node represents a random variable (e.g. $\theta, x, y, a, b, s, \delta$) and each directed edge indicates causal relationship between variables (e.g. $x$ is dependent on $\theta$). The joint probability of the variables can be decomposed as a product of simpler parts: $p(\theta,x,y,a,b,s,\delta) = p(\sigma) p(s|\delta) p(y|x,s,\delta) p(b|a,s,\delta) p(a) p(\theta) p(x|\theta)$. Redrawn from \cite{Fan05}. \label{fig:graphical_model}}
		\end{figure}
	\end{subsection}

	\begin{subsection}{Temporal State-Space Classifiers}
		Features obtained from videos can be perceived as temporal sequential data. \textit{Temporal state-space classifiers} model temporal sequential data by assuming that observations are generated through some underlying hidden (or latent) state and they utilize sequential information by acknowledging that states evolve over time. An observation corresponds to some feature vector and a hidden state represents an action performance at a specific moment in time. Temporal state-space classifiers model the relationships between state-to-state and state-to-observation using probabilities. In this section, some common temporal state-space models that have appeared in the action recognition and detection literature will be reviewed. \\
	
		An action can be perceived as a sequence of states that is directly influenced by its previous state(s), 
		and each state of the action can be observed by some feature representation. Correspondingly, a video of an action can be modelled using the \textit{Hidden Markov Model} (HMM), where each state of an action corresponds to the hidden/latent state, $z_t$, with observation, $f_t$ (see Figure \ref{fig:hmm}). Then the task of action recognition/detection can be formulated by finding the most probable set of sequences, $Z=\{z_t | t \in \mathbb{Z}^+\}$, that corresponds to a set of observations, $F=\{f_t | t \in \mathbb{Z}^+\}$. That is, maximize the joint probability of the paired observation and label sequences: 
		\begin{equation*}
			P(\mathbf{c},\mathbf{f}) = \prod_{t=1}^{n}{ P(z_t|z_{t-1}) P(f_t|z_t) } \text{,}
		\end{equation*}
		where $P(z_t|z_{t-1})$ and $P(f_t|z_t)$ denote transition probabilities and emission probabilities, respectively, for $t=1, \dots, n$. 
		The \textit{transition probability} models the probability of a state transitioning from $z_{t-1}$ to $z_t$, and the \textit{emission probability} models the probability of observation $f_t$ being emitted from state $z_t$.
		Transition probabilities can be trained using $k$-means clustering \cite{Fen02,Ram03} for supervised data and the Baum-Welch algorithm for unsupervised data \cite{Ahm08,Iki08,Lu06,Wei07,Yam92}. Each HMM represents an action category \cite{Iki08,Wei07,Yam92}. The observation can be of the entire body \cite{Ahm08,Wei07}, a body part \cite{Iki08,Ram03}, or an interest point (e.g. mesh \cite{Fen02}, HOG \cite{Lu06}). The probability that a sequence of hidden states would yield a set of observations is referred to as the \textit{decoding problem} \cite{Dud01}. The most likely action class that the test data would belong to among the $c$ HMMs, where $c$ denotes the number of action classes, can be evaluated using the Viterbi algorithm \cite{Ahm08}, or maximum likelihood estimation (MLE) \cite{Lu06}.
		Since HMMs are designed to deal with time-sequence data, they are robust to time scale shift and variance \cite{Yam92}. \\
		\begin{figure}[htbp]
			\begin{center}
				\includegraphics[width=0.7\textwidth]{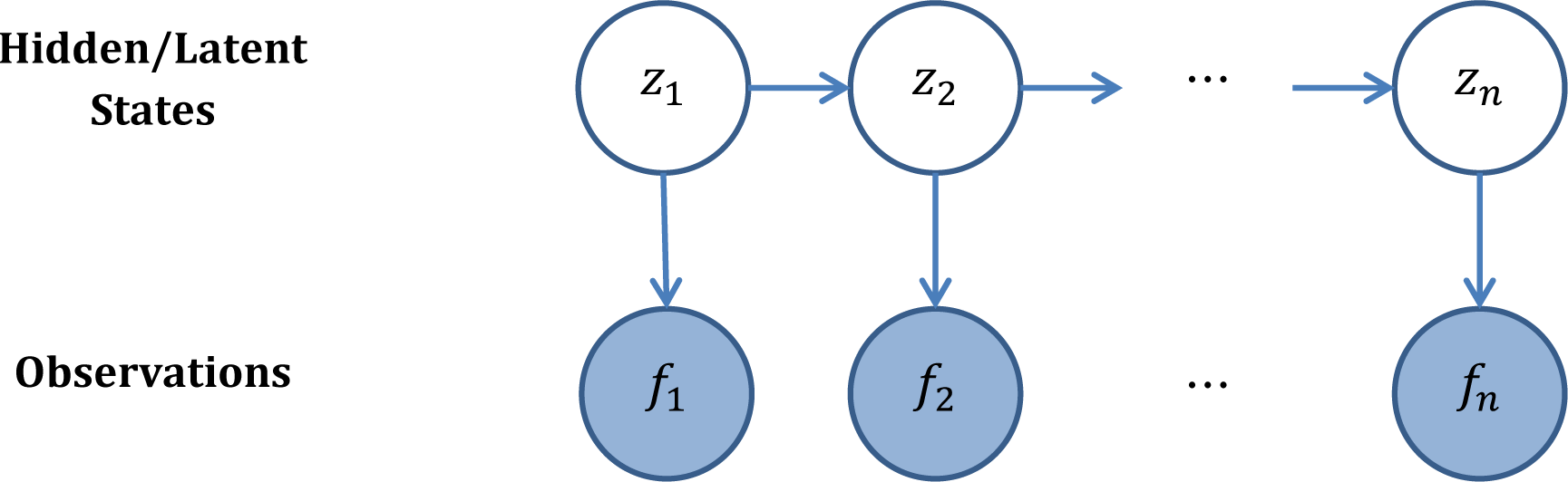}
			\end{center}
			\caption{Hidden Markov Model (HMM). The transition between states, $z_t$ to $z_{t+1}$, are represented by horizontal lines illustrating temporal causality between states whose likelihood is represented by the \textit{transition probability}, $P(z_{t+1}|z_t)$. The emission of a particular observation, $f_t$, from a specific state, $z_t$, is illustrated by the vertical lines whose likelihood is represented by the \textit{emission probability}, $P(f_t|z_t)$. A set of parameters - the initial state probability, transition probabilities, and emission probabilities - constitute an HMM, which represents one action class. A test data that matches an HMM with the highest probability is assigned to the class that the HMM represents. \label{fig:hmm}}
		\end{figure}
		
		The features obtained from videos can be noisy and extracted at random intervals. The hidden state corresponding to noisy data does not need to be constrained to discrete variables, but can be estimated as continuous variables using \textit{Kalman filters} \cite{Wel06} \cite{Rus10}. A Kalman filter models each state variable as continuously distributed using a Gaussian distribution \cite{Rus10}. Since the convolution and the product of a Gaussian also yields a Gaussian, all probabilities (transition and emission) are also Gaussians. The Kalman filter works in a two-step process: the prediction step and the correction step. At the prediction step, the Kalman filter estimates the current (hidden) state, $z_t$, along with its uncertainties and the future state, $z_{t+1}$. At the correction step, the weighted average of the new observation, $f_{t+1}$, and the predicted value is used to update the new hidden state $z_{t+1}$. The new hidden state is assigned with more emphasis placed on the predicted value if the new observation is deemed unreliable (noisy), and the observation is more favoured if the process deem unpredictable and unreliable \cite{Rus10}. The prediction and correction phases run recursively to update the current estimate based on all of the past measurements. Kalman filters are often used in conjunction with tracking-based algorithms \cite{Bod03,Du07,Hah08}. Kalman filters are not very capable of handling occlusion, therefore, require good foreground segmentation \cite{Ke13}. \\
		
		A \textit{conditional random field} (CRF) \cite{Laf01} is an undirected graphical model that is used to calculate the probability of a label sequence 
		given an observation sequence (see Figure \ref{fig:crf}). The conditional probability is factorized into a product of real-valued functions, where each function is described by log-linear combinations of feature functions. That is, the conditional probability distribution of observation, $\mathbf{f}$, and state sequence, $\mathbf{z}$, is described as:
		\begin{equation*}
			P(\mathbf{z}|\mathbf{f}) = \frac{1}{Z_0}\exp{ \left( 
						\sum_{t=1}^{n}{ \sum_{k=1}^{m}{ \lambda_k g_k(z_{t-1},z_t,\mathbf{f})} }
						+ \sum_{t=1}^{n}{ \sum_{k=1}^{m}{ \mu_k h_k(z_t,\mathbf{f})} 
					} \right) } 
			\text{,}
		\end{equation*}
		where $Z_0$ denotes a normalization factor of all possible state sequences, and $\lambda_k$ and $\mu_k$ are associated weights of the feature functions, $g_k$ and $h_k$, respectively. 
		There is a strong connection between HMMs and CRFs. That is, the feature function $g_k$ coupled with $\lambda_k$ is analogous to transition probabilities in HMMs, while $\mu_k$ and $h_k$ is analogous to emission probabilities. Since CRFs directly model the conditional distribution over hidden states given the observations, the conditional independence assumption between observations given the class labels to ensure tractability in HMMs can be relaxed. 
		This difference allows observations at different time instances to be jointly considered, allowing CRFs to handle large contextual dependencies among observations, multiple overlapping observations, and long-range interactions between observations \cite{Men08,Smi05}. Considering the context and long-term dependencies helps remove ambiguities between similar actions (e.g. walk vs. jog) \cite{Che14,Smi05}. 
		CRFs generally require many training sequences to robustly determine all parameters \cite{Pop10}. \\

		\begin{figure}[htbp]
			\begin{center}
				\includegraphics[width=0.7\textwidth]{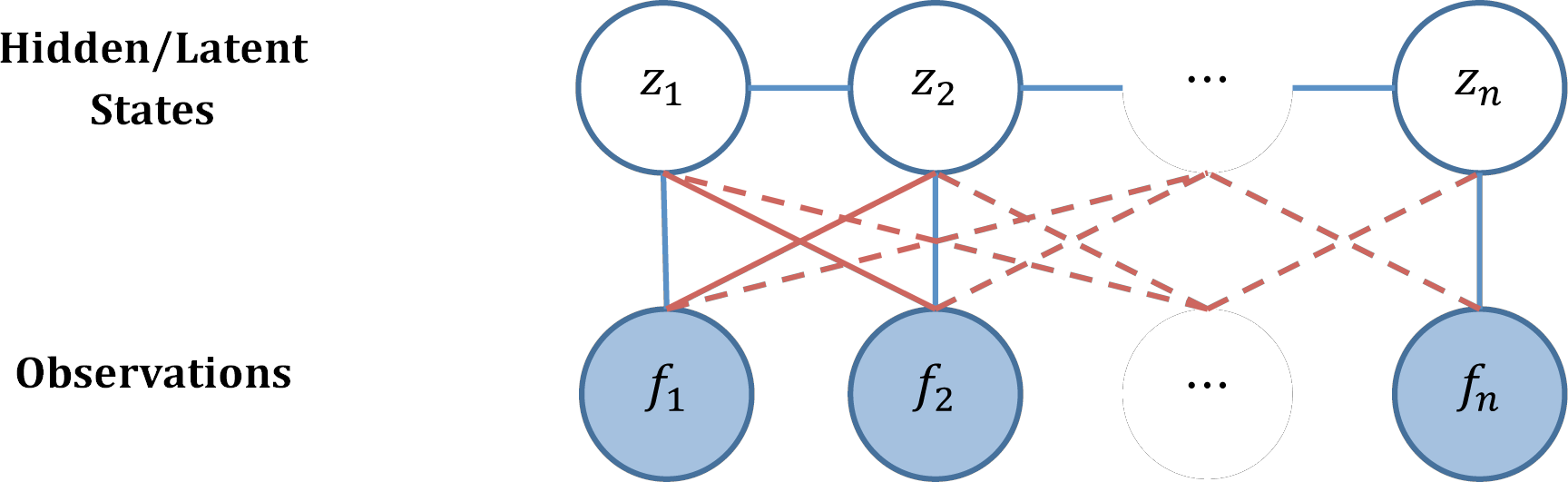}
			\end{center}
			\caption{Conditional Random Field (CRF). In an HMM, observation $f_t$ only depends on the current hidden state $z_t$ $\forall \ t$ (illustrated by blue links between hidden states and observations). Therefore, successive observations are independent. CRFs, on the other hand, directly model conditional distributions over hidden states given observations at different time instants (illustrated by blue and red links). This capability allows CRFs to relax the independence assumptions between observations and consider observations at different time instants. The undirected graphical model in a CRF allows the family of probability of distributions to factorize into a given collection of factors. HMMs, on the other hand, use a directed graphical model, which factorizes the probability of possible assignments  into local conditional probability distributions. \label{fig:crf}}
		\end{figure}

		A \textit{dynamic Bayesian network} (DBN) \cite{Mur02} is a system that models the relationship between and amongst hidden and observation variables using a Bayesian network with dynamic temporal states. That is, a Bayesian network in DBN models the causal relationship between hidden variables, $z_t^i$ for $i \in \mathbb{Z}^+$, and observation variables, $f_t^j$ for $j \in \mathbb{Z}^+$ at each state, $t$ for $t=1,\dots,n$ (see Figure \ref{fig:dbn}). The Bayesian network structure at each state is repeated to keep calculations simple through its periodic structure \cite{Luo03}. A system with a Bayesian network structure that changes per state is referred to as a dynamic Bayesian multinet (DBM) \cite{Bil00}, which is not the focus here. HMMs (or Kalman filters) can be considered as special types of DBNs \cite{Rus10}. While a DBN allows any number of hidden and observation variables per state, HMMs (or Kalman filters) only allow one discrete (or continuous) hidden and observation variables \cite{Luo03}. Although allowing more variables per state might lead to a larger computational complexity, the Bayesian network structure allows flexibility between variables, which simplifies the computation of the joint probability (i.e. some variables are independent from one another because some pairs do not have a causal relationship in the physical world). The hidden variable, $z_t^i$, can represent body parts (e.g. head, hands, feet) \cite{Luo03}, global and local activity state of the actor \cite{Du07}, objects present \cite{Lax07} at state $t$ and the observation variable can represent extracted features (e.g. Hu moments \cite{Luo03}, an MHI extension \cite{Xia06}, human poses \cite{Du07}). One major drawback of DBNs is the inevitable need of a very long training time \cite{Ke13}.

		\begin{figure}[htbp]
			\begin{center}
				\includegraphics[width=0.8\textwidth]{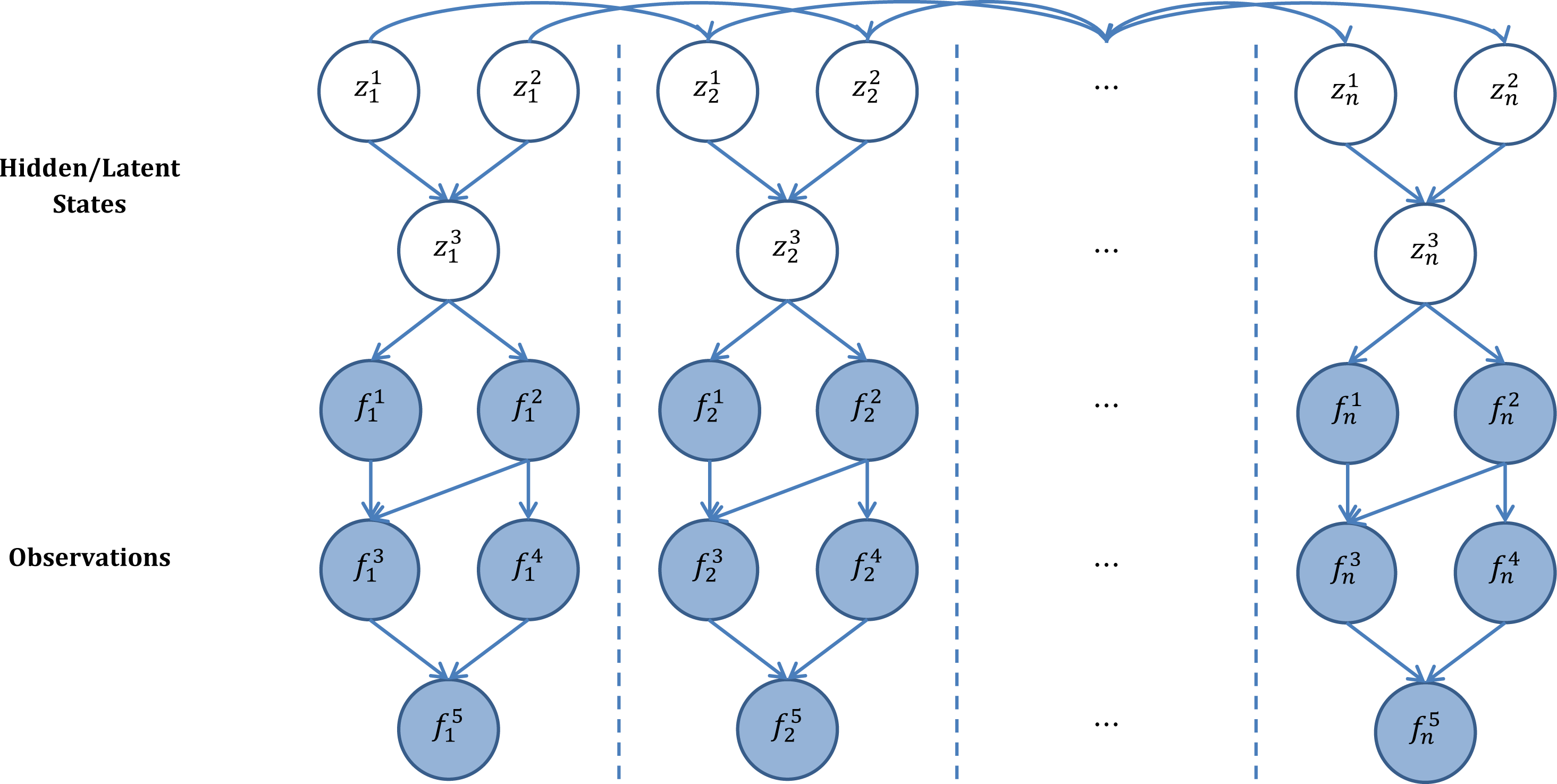}
			\end{center}
			\caption{Dynamic Bayesian Network (DBN). The Bayesian network (BN) is modelled in a sequential manner in a DBN, where the BN is the same for all states $1 \leq t \leq n$. Like other temporal state-space models, each BN at state $t$ has observed (shaded) and unobserved (unshaded) variables. Different from other temporal state-space models, however, a DBN can have $k_1$ observed variables (i.e. $f_t^i$ are unobserved variables for $1 \leq i \leq k_1$), and $k_2$ unobserved variables (i.e. $z_t^j$ are observed variables for $1 \leq j \leq k_2$) per state $t$ for $k_1,k_2 \in \mathbb{Z}^+$. \label{fig:dbn}}
		\end{figure}
	\end{subsection}
\end{section}
	
\begin{section}{Final Remarks}
	In this chapter, various classification algorithms that have appeared in the field of action recognition and detection were surveyed. 
A classifier determines the final accuracy of the overall action recognition algorithm. Thus, it is important to choose the right one. However, as stated by the \textit{No Free Lunch Theorem} \cite{Wol97}, it is a difficult task to find a classifier that is guaranteed to perform well since there are many factors to consider. Some factors to consider include: type of features, amount of training data, cost of the function, prior distributions (for probabilistic models), hyperparameters (e.g. type of norm in kNN classifier, number of hidden units, length of training, and the training rate in ANNs). For example, while a linear SVM is most suited for data with large number of features with a small amount of training examples, a Gaussian kernel is better suited on data with a small amount of features with an intermediate amount of training examples. While ANNs, on the other hand, are able to model complicated class distinctions, they require overly large datasets for training. Furthermore, while SVMs solve a convex problem, hence find a global solution; ANNs generally do not share that feature. Overall, while it is important to choose a classifier that would most accurately classify the action class given the features, it is more important to obtain and feed in useful features into the classifier and thereby simplify the classification problem itself. 
\end{section}

%% file: chapters/chapter5_currentstatus.tex
Throughout this report, various action recognition and detection algorithms have been surveyed. 
To conclude the report, the current trends as well as future direction of action recognition and detection research will be explored. 
Some commonly explored methods will be recognized for their top performance on benchmark datasets in section \ref{sec:current_trends} followed by some outstanding challenges that remain in the field, which will be addressed in section \ref{sec:open_problems}.

\begin{section}{Current Trends}\label{sec:current_trends}
Action recognition and detection continues to be a popular research topic in computer vision. In this section, we review some top performing action recognition and detection algorithms on benchmark datasets. A quantitative summary of the state-of-the-art action recognition and detection results on benchmark datasets can be found in Table \ref{tab:soa_ar_results} and Table \ref{tab:soa_ad_results}, respectively.\\

Before CNN-based algorithms took the field of action recognition and detection by storm, iFV-encoded iDT features with HOG, HOF, and MBH descriptors using a linear SVM classifier (see \textit{iFV-encoded iDT + linear SVM} in Table \ref{tab:soa_ar_results}) were top performing hand-crafted features achieving an accuracy of 57.2\% and 85.9\% on HMDB51 and UCF101 datasets, respectively \cite{Wan13,Wan13_thumos13}. Higher dimensional FV-encodings were implemented to further improve iDT features (see \textit{high-dim. FV-encoded iDT + linear SVM} in Table \ref{tab:soa_ar_results}), outperforming iFV-encoded features by 3.9\% and 2.0\% on the benchmark datasets \cite{Pen14}. These results suggest the power of combining appearance and motion features, as well as the importance of tuning encoding methods suited to serve the task of action recognition. Indeed, with such features and encodings, even simple classifiers, such as linear SVMs, are able to achieve outstanding results. \\ 

Due to its success in various classification tasks, there has been constant strive for success using deep-learned convolutional features in the field of action recognition and detection. The two-stream approach \cite{Sim14}, which decouples the appearance and motion components of a video by taking image and motion inputs and fusing them at the end via linear SVM (see \textit{two-stream CNN + linear SVM} in Table \ref{tab:soa_ar_results}), is able to achieve a comparable result to the hand-crafted features (cf. \textit{high-dim. FV-encoded iDT + linear SVM} and \textit{two-stream CNN + linear SVM} in Table \ref{tab:soa_ar_results}). Consequently, the two-stream approach became one of the most persistently pursued paths amongst other CNN-based approaches \cite{Don15,Feich16,Ng15,Sim14,Sin16,Wan16,Zhu16}. The hand-crafted and deep-learned features have demonstrated complementarity, achieving more accurate results than when either one is implemented independently (cf. \textit{traj. pooled two-stream CNN} and \textit{traj. pooled two-stream CNN + iDT} in Table \ref{tab:soa_ar_results}) \cite{Feich16,Fer16,Wan15,Wang16,Zhou16}. In fact, the top performing algorithm on both benchmark datasets to date is achieved by combining the high-dimensional FV-encoded iDT hand-crafted features with the two-streams of CNNs interacting through residual connections (see \textit{two-stream CNN + ResNet + iDT} in Table \ref{tab:soa_ar_results}) \cite{Fei16c}. \\

As can be witnessed in Table \ref{tab:soa_ar_results}, the results on the Sports-1M dataset are not as widely reported as HMDB51 or UCF101. 
Aside from the fact that the Sports-1M dataset was released two years prior to this report, there are many factors that could be limiting algorithms from reporting results on the Sports-1M dataset: (i) the training data is extremely large (multiple terabytes) \cite{Sim14}, (ii) automatic collection of the data does not permit the data from being free of label noise \cite{Feich16,AKar14} (e.g. a supposed training video for the class \textit{women's lacrosse} with video ID \verb+EaOvsVdbhhE+ does not contain a single instance of a person playing lacrosse, rather it is an interview of a women's lacrosse game), (iii) portions of the dataset have been removed by YouTubers that uploaded the original videos, which can no longer be accessed since the authors of the Sports-1M dataset provide URL links to each video \cite{Ng15}, and (iv) there is a very low inter-class variation in some cases (e.g. lacrosse vs. women's lacrosse).\\

The aggregate results reported in Table \ref{tab:soa_ar_results} reveal how algorithms perform in general relative to others. Unfortunately, these quantitative data do not reveal any details of the algorithm or dataset. Comparing how each algorithm performs on individual actions, using a confusion matrix for example, can reveal how an algorithm responds to particular actions and perhaps motions and/or appearances. Comparing how algorithms perform with varying viewing conditions would reveal their robustness to viewpoint, background clutter, occlusion, performance nuance, slight variations in pose, and/or illumination. However, current benchmark datasets lack systematic variation of such parameters, limiting algorithms from revealing the relative impact of these parameters. As a result, unfortunately, many recent works only report overall results limiting our ability to make any insightful observations within each technique. \\

	\begin{table}[htbp]
		\begin{center}
			\begin{tabular}{| p{0.5\textwidth} || c | c | c |}
				\bottomrule
					\backslashbox[80mm]{\textbf{Method}}{\textbf{Dataset}}& HMDB51 & UCF101 & Sports-1M \\
				\toprule
				\bottomrule
					iFV-encoded iDT with linear SVM \cite{Wan13,Wan13_thumos13}		& 57.2\% (10) 	& 85.9\% (11) & - \\
					high-dim. FV-encoded iDT with linear SVM \cite{Pen14} 			& 61.1\% (8)	& 87.9\% (10) & - \\
				\midrule
					2D CNN + slow-fusion \cite{AKar14}						& - 			& 65.4\% (13) & 60.9\% (2) \\
				\midrule
					two-stream CNN + linear SVM \cite{Sim14} 					& 59.4\% (9) 	& 88.0\% (9) & - \\
					\hline
					CNN + hier. pooling \cite{Fer16}							& 47.5\% (11) 	& 78.8\% (12) & - \\
					CNN + hier. pooling	+ FV-encoded iDT with non-linear SVM \cite{Fer16}	& 66.9\% (3) 	& 91.4\% (6) & - \\
					\hline
					two-stream CNN + key-volume mining \cite{Zhu16} 				& 63.3\% (6) 	& 93.1\% (3) & - \\
					\hline
					traj. pooled two-stream CNN \cite{Wan15} 					& 63.2\% (7) 	& 90.3\% (7) & - \\
					traj. pooled two-stream CNN + iDT \cite{Wan15} 				& 65.9\% (4) 	& 91.5\% (5) & - \\
					\hline
					two-stream CNN + conv. fusion \cite{Feich16} 				& 65.4\% (5) 	& 92.5\% (4) & - \\
					two-stream CNN + conv. fusion + iDT \cite{Feich16} 				& 69.2\% (2) 	& 93.5\% (2) & - \\
					\hline
					two-stream CNN + ResNet + iDT \cite{Fei16c}	& 70.3\% (1) & 94.6\% (1) & - \\
					\hline
					two-stream CNN + LSTM + conv. pooling \cite{Ng15}			& - 	& 88.6\% (8) & 73.1\% (1) \\
				\toprule	
			\end{tabular}
		\end{center}
		\caption{State-of-the-Art Action Recognition Results. The HMDB51 and UCF101 datasets have three splits for training and testing. The average accuracy over the three splits are reported. Numbers inside the parentheses indicates the rank in decreasing order for each dataset (i.e. ($k$) indicates that the algorithm performs $k$th best on the dataset). 
\label{tab:soa_ar_results}}
	\end{table}

In the classical two-stream approach, the computation of optical flow for the motion stream is the most time costly component of the algorithm \cite{Zha16}. Alternative to optical flow, motion vectors, which describes macro block movements from one frame to the next, can be used to significantly lower the computational cost of the algorithm, from 14.3 fps to 390.7 fps, which is approximately 27 times faster than the standard optical flow (see Table \ref{tab:soa_ar_efficiency}) \cite{Zha16}. However, since motion vectors exhibit coarser structure, lacking fine and accurate motion information than optical flow, slight degradation in performance does occur (from 88.0\% to 86.4\%) \cite{Zha16}. Intuitively, combining the most accurate approach with the most efficient approach could achieve an ideal algorithm. Thus, it would be worth incorporating convolutional fusion into the two-stream approach that uses motion vectors in its motion stream to achieve an efficient yet accurate algorithm.\\

	\begin{table}[htbp]
		\begin{center}
			\begin{tabular}{| l || c | c |}	
				\bottomrule
					\textbf{Method} & \textbf{Accuracy} & \textbf{FPS} \\
				\toprule
				\bottomrule
					iFV-encoded iDT + lin. SVM \cite{Wan13_thumos13} 		& 85.9\% 	& 2.1\\
				\midrule
					two-stream CNN (RGB + opt. flow) + lin. SVM \cite{Sim14} 	& 88.0\%	& 14.3\\
				\hline
					two-stream CNN (RGB + motion vec.) + lin. SVM \cite{Zha16} 	& 86.4\%	& 390.7\\
				\toprule
			\end{tabular}
		\end{center}
		\caption{State-of-the-Art Action Recognition Accuracy and Efficiency Comparison. The performance and speed of the classical hand-crafted feature (iFV-encoded iDT + lin. SVM), classical deep-learned convolutional two-stream feature (two-stream CNN (RGB+opt. flow) + lin. SVM), and the two-stream approach with optical flow replaced by motion vector (two-stream CNN (RGB+motion vec.) + lin. SVM) is compared on the UCF101 dataset. The speed of the algorithms are measured as frames per second (fps) on a single-core CPU (E5-2640-v3) and a K40 GPU. Results extracted from \cite{Zha16}. \label{tab:soa_ar_efficiency}}
	\end{table}

In general, action detection is a more complex task than action recognition. Moreover, spatiotemporal localization is a more demanding task than temporal localization. 
As a result, there have been more papers on action recognition and comparably less on detection. In recent years, however, a handful of research was done on temporal localization reporting results on THUMOS '14, MPII Cooking Activities and MPII Cooking 2 Activities, as well as the ActivityNet datasets (see Table \ref{tab:soa_ad_results}). However, these algorithms reported results on a select few and not all of these benchmark datasets. Therefore, it is a difficult task to compare and analyze the strengths and weaknesses of the algorithm relative to each other. Thus, we remark on some common traits amongst these temporal action detection algorithms.\\

As in the case of action recognition tasks, CNN-based algorithms also remain a popular choice in temporal action detection \cite{SMa16,Ng15,Ni16,Sho16,Sin16}.
Recent research localizes actions temporally by either: (i) sliding a temporal window to determine the action proposal and class \cite{Hei15,Ni16,Sho16}, or (ii) using LSTM-RNNs \cite{SMa16,Ni16,Yeu16}. Many top performing temporal action detection algorithms rely on CNNs to represent features and LSTM-RNNs to model temporal transition of the actions, which allow for temporal detection \cite{SMa16,Sho16,Yeu16,Yua16}. 
However, LSTM-RNNs are not limited to localize actions or objects of interest temporally. They can be used to sequentially refine the detected result, which is particularly useful for detecting fine-grained actions as in the MPII Cooking Activities dataset \cite{Ni16}. 

\afterpage{
	\clearpage
	\begin{landscape}
		\begin{table}
			\begin{center} 
				\resizebox{.83\paperheight}{!}{
					\begin{tabular}{| l || c | c | c | c |}
						\bottomrule
							\backslashbox[150mm]{\textbf{Method}}{\textbf{Dataset}} & THUMOS' 14 & MPII Cooking \cite{Roh12} & MPII Cooking 2 \cite{Roh15} & ActivityNet \\
						\toprule
						\bottomrule
							CNN features + LSTM-RNN \cite{Yeu16} & 17.1\% & - & - & 36.7 \\
							3D CNN for proposal, classification, localization networks \cite{Sho16} & 47.7\% & - & - & - \\
							detection score computed by LSTM on CNN-based features of the frame \cite{SMa16} & - & - & - & 54.0 mAP \\
							two-stream CNN (still frame + pixel trajectories) \cite{Sin16} & - & - & 41.2 mAP & - \\
							LSTM-RNN + FV-encoded iDT \cite{Ni16} & - & 58.9 mAP & - & - \\
							iFV-encoded iDT + LSTM-RNN \cite{Yua16} & - & 36.3 precision, 59.7 recall & - & - \\
						\toprule	
					\end{tabular}
				}
			\end{center}
			\caption{State-of-the-Art Temporal Action Detection Results. \label{tab:soa_ad_results}}
		\end{table}
	\end{landscape}
}
\end{section}

\begin{section}{Open Problems}\label{sec:open_problems}
While there has been significant progress in the field of action recognition and detection, computer vision-based algorithms are still far from identifying actions as well as humans. Provided that the video contains enough information for humans to visualize the actions of interest, we have the ability to classify actions irrespective of variations in viewpoint, background clutter, occlusion, performance nuance, considerable variations in pose, and illumination. 
Computer vision-based algorithms, on the other hand, are not able to overcome all of these obstacles yet.
It is then appropriate to question \textit{where} and \textit{why} these systems fall short. In this section, we address some open problems that remain in the field to direct ongoing research that would allow computer vision-based algorithms to reach the capabilities of humans.\\

%
%
Algorithms that are able to achieve accuracies of over 85\% on benchmark datasets collected from the ``wild'', like UCF101, may suggest that they have solved the invariance problems (see Table \ref{tab:soa_ar_results}). However, these same algorithms achieve just over 65\% on HMDB51, performing not as impressively on a similarly wild dataset with more variation in viewpoints. 
This result suggests that the proposed algorithms are not robust to viewpoints and that viewpoint invariance remains a crucial problem to be addressed.\\ 

The current widely used deep-learned convolutional features have demonstrated state-of-the-art results on both action recognition and detection tasks. However, there lacks a theoretical understanding of \textit{how} and \textit{why} these algorithms are so successful. A scientific understanding is in great need of these algorithms such that it would help researchers develop algorithms that are even more accurate and efficient.\\

Empirical results suggest that with copious amounts of data, CNN-based algorithms are able to learn similar features between different actors performing the same action (i.e. performance nuance) \cite{Feich16}. However, many real-world problems (e.g. surveillance scenarios) are not able to provide such massive amounts of data nor time for training. There is a need for algorithms to work in real-time respectably with small amounts of data that would progressively improve its confidence as more data is learned, most desirably in an unsupervised fashion.\\

Currently, many algorithms report overall results on benchmark datasets. Beyond aggregate results, we should be able to distinguish the specific categories in which algorithms perform well. This may be achieved with a systematic dataset with a hierarchical categorization. Although the ActivityNet dataset is organized in a hierarchical fashion (see Table \ref{tab:ActivityNet}), there are still commonalities between inter-category classes (e.g. \textit{dancing} in the \textit{socializing} category vs. \textit{doing aerobics} in the \textit{exercise} category or \textit{smoking} in the \textit{leisure activities} category vs. \textit{drinking} in the \textit{eating and drinking activities} category). A dataset that is more distinct between the categories could provide a principled way of tuning parameters that would serve particularly well to its application domain (e.g. surveillance, home monitoring, video indexing, and sports analysis). Furthermore, the videos within each class should vary viewing parameters systematically to specifically indicate where the proposed algorithms are failing. Even with more systematically collected datasets, however, deeper understanding of how and why algorithms perform the way they do will only occur once research efforts are refrained from simply chasing after performance numbers and focused on analyzing internal algorithm operations and representations.\\
%

Another matter worth addressing is that there are comparably less publications in action detection in contrast to action recognition. Correspondingly, the state-of-the-art quantitative results are significantly worse with 58.9 mAP \cite{Ni16} on the MPII Cooking Dataset for temporal action detection, whereas its counterpart, action recognition, is able to routinely achieve an accuracy of over 65\% \cite{Feich16,Fer16,Wang15} and 85\% \cite{Feich16,Fer16,Ng15,Pen14,Sim14,Wang16,Zha16,Zhu16} on much more complicated datasets, HMDB51 and UCF101, respectively. 
Temporal and/or spatiotemporal localization of actions can be useful in alerting caregivers or security personnel of abnormal or suspicious activities in home or security settings. It is worth directing our attention to detecting actions or regions that are likely to contain actions than to remain focused on the task of action categorization in well trimmed videos. In this light, we point out that datasets suited for action detection tasks are very limited. Thus, many algorithms that do try to tackle the detection problem rely on THUMOS, MSR Action Dataset I/II, CMU Crowded Videos, or MPII Cooking Activities Datasets. A systematic large-scale benchmark action detection dataset that is not only catered towards approaches that require a large set of videos (e.g. CNN-based algorithms), but is more resemblant of real-world scenarios could help improve the current status of action detection.\\

Significant research has been done in the field of action recognition and detection. Perhaps its potential fields of application in the real-world has attracted much attention. However, current algorithms still lack robustness to variations in real-world conditions. A better theoretical understanding of well-performing algorithms, and algorithms that can reliably detect actions efficiently and accurately with less training data are some of the most urgent steps to take from hereon. This enhancement in technology could better assist and benefit many people in the real-world.
%
%
\end{section}

%% file: chapters/appendix_related_fields.tex
There are many fields that are closely related to action recognition and detection. Their algorithms can either (i) help improve current recognition and detection algorithms, (ii) benefit from the results of existing recognition or detection algorithms, or (iii) use an approach similar to those in action recognition and detection algorithms to serve its own task. Here, each type of these related fields will be briefed.\\

Localization of actions in long untrimmed videos can be done efficiently by obtaining (or removing) snippets that are likely (or unlikely) to contain actions of interest. These candidate regions can be temporal \cite{Heilbron16,SMa16,Sho16,Wang16} or spatiotemporal snippets \cite{van15,Gki15,Jai14,Sul16,Yu15,Zhu16} and are referred to as \textit{action proposals}. Action proposals do not identify the class of an action but localize regions that are likely to contain an action (i.e. regions with high actionness scores \cite{Che14}). Conversely, regions that are unlikely to contain actions of interest, referred to as non-action shots \cite{Wang16}, can provide similarly useful information. Various approaches have been taken to obtain action proposals, such as supervoxel-based approaches \cite{Jai14,Oneata14, Sul14, van15}, combination of static and kinematic cues \cite{Gki15}, combination of human detectors and dense trajectory features \cite{Yu15}, as well as lattice conditional random fields \cite{Che14}. Once an action proposal has been found (or non-action shots have been removed), typical action recognition algorithms (e.g. FV-encoded iDT features with SVM classifier \cite{Heilbron16,van15,Jai14,Yu15} or two-stream CNN \cite{Gki15}) can be used to identify an action in the proposals \cite{Heilbron16, Sul16}. Action proposals (or non-action shots) prevent going through an exhaustive search space and helps detect temporal or spatiotemporal locations of actions.\\

Detection of an unusual behaviour, or more specifically \textit{anomalous action}, is another interesting related task to action recognition and detection. Some examples of anomalous behaviours are: 
(i) motion in an area where no motion is expected to occur (e.g. secured storage facility), and
(ii) motion of an object moving in an unexpected direction (e.g. car travelling in the ``wrong way'' on a one-way road, a person falling down on a sidewalk).
Anomalous behaviours can be detected by matching the observation to a database of normal videos or learning expected patterns from such a database, then flagging regions that deviate from learned patterns \cite{Bas08,Mor00,Sta00,Zah10}.\\

Forecasting the future rather than interpreting the present actions, or \textit{action prediction}, is another closely related field to action recognition. Like recognition and detection algorithms, HOG, HOF, and MBH \cite{Lan14}, cuboids \cite{Ryoo11}, or CNNs \cite{Von16} can be used to represent features, and the action class to occur can be specified using the nearest neighbour approach \cite{Von16}, or SVM \cite{Hoai14, Lan14, Von16}. 
The action prediction model can progressively transition into an action recognition model. That is, as more of the action is observed, the entire action would be seen. Progressive transition from action prediction to recognition can be indicated by a gradual increase in confidence score of the action class \cite{SMa16,Ryoo11}. This tack could provide very useful information in the real-world. For example, in autonomous navigation, prediction of when an accident could occur would divert a vehicle from causing any (further) damage, while recognition of an accident would alert emergency medical technicians (e.g. paramedics). In geriatric care, a robot that detects an elderly patient trying to stand up can help them stand without falling, while recognition of a fallen senior would alert the caregiver.\\

With an emergence of first-person point-of-view cameras (e.g. GoPro, Google Glass), recognition of actions from an egocentric view has recently been an emerging field \cite{Ma16,Singh16,Zhou16}. Egocentric action recognition has many interesting applications (e.g. extreme sports, law enforcement). Some overlap in recognition techniques can be found between first- and third-person recognition algorithms, such as CNN-based model \cite{Ma16,Singh16,Zhou16}. Different from third-person action recognition models, first-person models place emphasis on hand and object motions. Furthermore, the benchmark datasets used in first- and third-person action recognition algorithms differ greatly. In this report, the main focus is placed on third-person action recognition.\\

\begin{figure}[htbp]
	\begin{center}
		\includegraphics[width=0.7\textwidth]{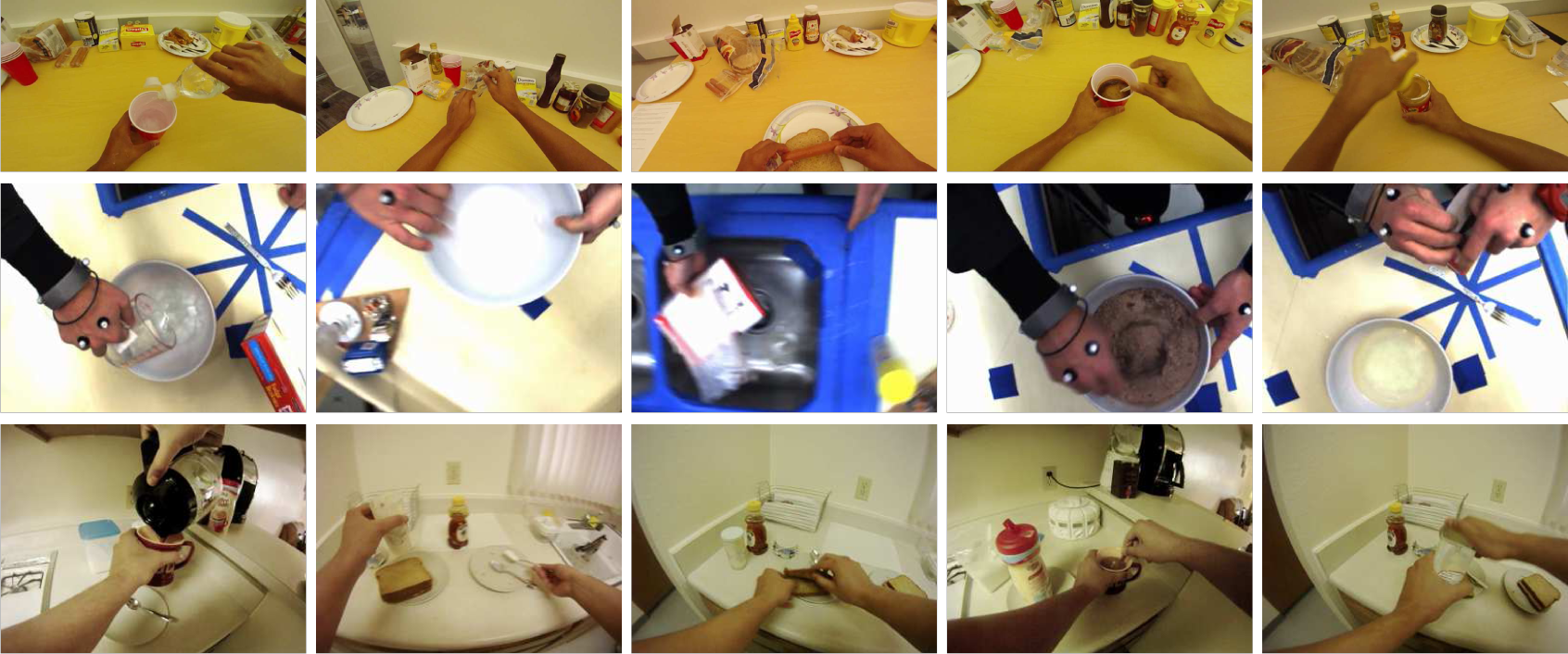}
	\end{center}
	\caption{Egocentric Action Recognition. Select frames of egocentric actions (left-to-right): pour, take, put, stir, and open from first-person action recognition datasets (top-to-bottom): GTEA \cite{Fat11}, Kitchen \cite{Spr09}, and ADL \cite{Pir12}. Redrawn from \cite{Singh16}. \label{fig:egocentric_action_recognition}}
\end{figure}

\textit{Activity recognition} \cite{Hu06} is another closely related active field of research. Activity recognition in itself is a vast field as it can be broken down into single person activity, group activity, and team activity as in sports \cite{Ibr16}. As actions are basic components that piece together to yield an activity, collectively recognizing primitive actions can provide strong indication of the activity that is occurring in a video. Thus, action recognition and detection remains a fundamental problem for its related recognition and classification tasks.
%

%% file: Literature_Review.bbl
\begin{thebibliography}{100}

\bibitem{Aha11}
M.A.R. Ahad, J.~Tan, H.~Kim, and S.~Ishikawa.
\newblock {Action Dataset - A Survey}.
\newblock In {\em SICE Annual Conference}, pages 1650--1655, 2011.

\bibitem{Ahm08}
M.~Ahmad and S.W. Lee.
\newblock {Human Action Recognition using Shape and CLG-motion flow from
  Multi-view Image Sequences}.
\newblock {\em The Journal of Pattern Recognition Society}, 41(7):2237--2252,
  2008.

\bibitem{Ali10}
S.~Ali and M.~Shah.
\newblock {Human Action Recognition in Videos Using Kinematic Features and
  Multiple Instance Learning}.
\newblock {\em IEEE Transactions on Pattern Analysis and Machine Intelligence
  (PAMI)}, 32(2):288--303, 2010.

\bibitem{Ara13}
R.~Arandjelovic and A.~Zisserman.
\newblock {All about VLAD}.
\newblock In {\em IEEE Conference on Computer Vision and Pattern Recognition
  (CVPR)}, pages 1578--1585, 2013.

\bibitem{Bac11}
M.~Baccouche, F.~Mamalet, C.~Wolf, C.~Garcia, and A.~Baskurt.
\newblock {Sequential Deep Learning for Human Action Recognition}.
\newblock In {\em Human Behavior Understanding}, pages 29--39, 2011.

\bibitem{Bal13}
N.~Ballas, Y.~Yang, Z.Z. Lan, B.~Delezoide, F.~Preteux, and A.~Hauptmann.
\newblock {Space-Time Robust Video Representation for Action Recognition}.
\newblock In {\em IEEE International Conference on Computer Vision (ICCV)},
  pages 2704--2711, 2013.

\bibitem{Bas08}
A.~Basharat, A.~Gritai, and M.~Shah.
\newblock {Learning Object Motion Patterns for Anomaly Detection and Improved
  Object Detection}.
\newblock In {\em IEEE Conference on Computer Vision and Pattern Recognition
  (CVPR)}, pages 1--8, 2008.

\bibitem{Bat08}
D.~Batra, T.~Chen, and R.~Sukthankar.
\newblock {Space-Time Shapelets for Action Recognition}.
\newblock In {\em IEEE Workshop on Motion and Video Computing (WMVC)}, pages
  1--6, 2008.

\bibitem{Bay06}
H.~Bay, T.~Tuytelaars, and L.~Van Gool.
\newblock {SURF: Speeded Up Robust Features}.
\newblock In {\em European Conference on Computer Vision (ECCV)}, pages
  404--417, 2006.

\bibitem{Bie85}
I.~Biederman.
\newblock {Human Image Understanding: Recent Research and a Theory}.
\newblock {\em Computer Vision, Graphics, and Image Processing}, 32(1):29--73,
  1985.

\bibitem{Bil16}
H.~Bilen, B.~Fernando, E.~Gavves, A.~Vedaldi, and S.~Gould.
\newblock {Dynamic Image Networks for Action Recognition}.
\newblock In {\em IEEE Conference on Computer Vision and Pattern Recognition
  (CVPR)}, 2016.

\bibitem{Bil00}
J.A. Bilmes.
\newblock {Dynamic Bayesian Multinets}.
\newblock In {\em Uncertainty in Artificial Intelligence}, 2000.

\bibitem{Bis06}
C.M. Bishop.
\newblock {\em {Pattern Recognition and Machine Learning}}.
\newblock Springer, 2006.

\bibitem{Bla05}
M.~Blank, L.~Gorelick, E.~Shechtman, M.~Irani, and R.~Basri.
\newblock {Actions as Space-Time Shapes}.
\newblock In {\em IEEE International Conference on Computer Vision (ICCV)},
  pages 1395--1402, 2005.

\bibitem{Ble03}
D.M. Blei, A.Y. Ng, and M.I. Jordan.
\newblock {Latent Dirichlet Allocation}.
\newblock {\em Journal of Machine Learning Research (JMLR)}, 3:993--1022, 2003.

\bibitem{Bob01}
A.~Bobick and J.~Davis.
\newblock {The Recognition of Human Movement Using Temporal Templates}.
\newblock {\em IEEE Transactions on Pattern Analysis and Machine Intelligence
  (PAMI)}, 23(3):257--267, 2001.

\bibitem{Bod03}
R.~Bodor, B.~Jackson, and N.~Papanikolopoulos.
\newblock {Vision-based Human Tracking and Activity Recognition}.
\newblock In {\em Mediterranean Conference on Control and Automation}, 2003.

\bibitem{Bou10}
Y.L. Boureau, J.~Ponce, and Y.~LeCun.
\newblock {A Theoretical Analysis of Feature Pooling in Visual Recognition}.
\newblock In {\em International Conference on Machine Learning (ICML)}, 2010.

\bibitem{Cai12}
J.C. Caicedo and F.A. Gonzalez.
\newblock {Online Matrix Factorization for Multimodal Image Retrieval}.
\newblock In {\em Progress in Pattern Recognition, Image Analysis, Computer
  Vision, and Applications}, pages 340--347, 2012.

\bibitem{Cao10}
L.~Cao, Z.~Liu, and T.S. Huang.
\newblock {Cross-dataset Action Detection}.
\newblock In {\em IEEE Conference on Computer Vision and Pattern Recognition
  (CVPR)}, pages 1998--2005, 2010.

\bibitem{Cha13}
J.M. Chalet, E.J. Carmon, and A.~Fernandez-Caballero.
\newblock {A Survey of Video Datasets for Human Action and Activity
  Recognition}.
\newblock {\em Computer Vision and Image Understanding (CVIU)},
  117(6):633--659, 2013.

\bibitem{Cha11}
K.~Chatfield, V.~Lemtexpitsky, A.~Vedaldi, and A.~Zisserman.
\newblock {The Devil is in the Details: An Evaluation of Recent Feature
  Encoding Methods}.
\newblock In {\em British Machine Vision Conference (BMVC)}, pages 76.1--76.12,
  2011.

\bibitem{Che14}
W.~Chen, C.~Xiong, R.~Xu, and J.J. Corso.
\newblock {Actionness Ranking with Lattice Conditional Ordinal Random Fields}.
\newblock In {\em IEEE Conference on Computer Vision and Pattern Recognition
  (CVPR)}, pages 748--755, 2014.

\bibitem{Cho99}
O.~Chomat and J.~Crowley.
\newblock {Probabilistic Recognition of Activity Using Local Appearance}.
\newblock In {\em IEEE Conference on Computer Vision and Pattern Recognition
  (CVPR)}, 1999.

\bibitem{Dal05}
N.~Dalal and B.~Triggs.
\newblock {Histograms of Oriented Gradients for Human Detection}.
\newblock In {\em IEEE Conference on Computer Vision and Pattern Recognition
  (CVPR)}, volume~1, pages 886--893, 2005.

\bibitem{Dal06}
N.~Dalal, B.~Triggs, and C.~Schmid.
\newblock {Human Detection Using Oriented Histograms of Flow and Appearance}.
\newblock In {\em European Conference on Computer Vision (ECCV)}, volume 3952,
  pages 428--441, 2006.

\bibitem{Den09}
J.~Deng, W.~Dong, R.~Socher, L.J. Li, K.~Li, and L.~Fei-Fei.
\newblock {ImageNet: A Large-scale Hierarchical Image Database}.
\newblock In {\em IEEE Conference on Computer Vision and Pattern Recognition
  (CVPR)}, 2009.

\bibitem{Der13}
K.~Derpanis, M.~Sizintsev, K.J. Cannons, and R.P. Wildes.
\newblock {Action Spotting and Recognition Based on a Spatiotemporal
  Orientation Analysis}.
\newblock {\em IEEE Transactions on Pattern Analysis and Machine Intelligence
  (PAMI)}, 35(3):1--8, 2013.

\bibitem{Dol05}
P.~Dollar, V.~Rabaud, G.~Cottrell, and S.~Belongie.
\newblock {Behavior Recognition via Sparse Spatio-Temporal Features}.
\newblock In {\em Joint IEEE International Workshop on Visual Surveillance and
  Performance Evaluation of Tracking and Surveillance}, pages 65--72, 2005.

\bibitem{Don15}
J.~Donahue, L.A. Hendricks, S.~Guadarrama, and M.~Rohrbach.
\newblock {Long-term Recurrent Convolutional Networks for Visual Recognition
  and Description}.
\newblock In {\em IEEE Conference on Computer Vision and Pattern Recognition
  (CVPR)}, pages 2625--2634, 2015.

\bibitem{Dou13}
G.~Dougherty.
\newblock {\em {Pattern Recognition and Classification}}.
\newblock Springer, 2013.

\bibitem{Du07}
Y.~Du, F.~Chen, and W.~Xu.
\newblock {Human Interaction Representation and Recognition Through Motion
  Decomposition}.
\newblock {\em IEEE Signal Processing Letters}, 14(12):952--955, 2007.

\bibitem{Dud01}
R.O. Duda, P.E. Hart, and D.G. Stork.
\newblock {\em {Pattern Classification}}.
\newblock John Wiley \& Sons, 2001.

\bibitem{Efr03}
A.A. Efros, A.C. Berg, G.~Mori, and J.~Malik.
\newblock {Recognizing Action at a Distance}.
\newblock In {\em IEEE International Conference on Computer Vision (ICCV)},
  pages 726 -- 733, October 2003.

\bibitem{Fan05}
C.~Fanti, L.~Zelnik Manor, and P.~Perona.
\newblock {Hybrid Models for Human Motion Recognition}.
\newblock In {\em IEEE Conference on Computer Vision and Pattern Recognition
  (CVPR)}, pages 1166--1173, 2005.

\bibitem{Fat08}
A.~Fathi and G.~Mori.
\newblock {Action Recognition by Learning Mid-Level Motion Features}.
\newblock In {\em IEEE Conference on Computer Vision and Pattern Recognition
  (CVPR)}, 2008.

\bibitem{Fat11}
A.~Fathi, X.~Ren, and J.M. Rehg.
\newblock {Learning to Recognize Objects in Egocentric Activities}.
\newblock In {\em IEEE Conference on Computer Vision and Pattern Recognition
  (CVPR)}, pages 3281--3288, 2011.

\bibitem{Fei15}
C.~Feichtenhofer, A.~Pinz, and R.P. Wildes.
\newblock {Dynamically Encoded Actions based on Spacetime Saliency}.
\newblock In {\em IEEE Conference on Computer Vision and Pattern Recognition
  (CVPR)}, pages 2755--2764, 2015.

\bibitem{Fei16}
C.~Feichtenhofer, A.~Pinz, and R.P. Wildes.
\newblock {Dynamic Scene Recognition with Complementary Spatiotemporal
  Features}.
\newblock {\em IEEE Transactions on Pattern Analysis and Machine Intelligence
  (PAMI)}, pages 1--14, 2016.

\bibitem{Fei16c}
C.~Feichtenhofer, A.~Pinz, and R.P. Wildes.
\newblock {Spatiotemporal Residual Networks for Video Action Recognition}.
\newblock In {\em Advances in Neural Information Processing Systems (NIPS)},
  2016.

\bibitem{Feich16}
C.~Feichtenhofer, A.~Pinz, and A.~Zisserman.
\newblock {Convolutional Two-Stream Network Fusion for Video Action
  Recognition}.
\newblock In {\em IEEE Conference on Computer Vision and Pattern Recognition
  (CVPR)}, pages 1933--1941, 2016.

\bibitem{Fen02}
X.~Feng and P.~Perona.
\newblock {Human Action Recognition by Sequence of Movelet Codewords}.
\newblock In {\em First International Symposium on 3D Data Processing
  Visualization and Transmission}, pages 717--721, 2002.

\bibitem{Fer16}
B.~Fernando, P.~Anderson, M.~Hutter, and S.~Gould.
\newblock {Discriminative Hierarchical Rank Pooling for Activity Recognition}.
\newblock In {\em IEEE Conference on Computer Vision and Pattern Recognition
  (CVPR)}, pages 1924--1932, 2016.

\bibitem{Fuk80}
K.~Fukushima.
\newblock {Neocognitron: A self-organizing neural network model for a mechanism
  of pattern recognition}.
\newblock {\em Biological Cybernetics}, 36(4):193--202, 1980.

\bibitem{Gim08}
K.~Gimpel.
\newblock {Modeling Topics}.
\newblock Technical report, Carnegie Mellon University, 2008.

\bibitem{Gki15}
G.~Gkioxari and J.~Malik.
\newblock {Finding Action Tubes}.
\newblock In {\em IEEE Conference on Computer Vision and Pattern Recognition
  (CVPR)}, pages 759--768, 2015.

\bibitem{Goo92}
M.A. Goodale and A.D. Milner.
\newblock {Separate Visual Pathways for Perception and Action}.
\newblock {\em Trends in Neurosciences}, 15(1):20--25, 1992.

\bibitem{thu15}
A.~Gorban, H.~Idrees, Y.G. Jiang, A.~Roshan Zamir, I.~Laptev, M.~Shah, and
  R.~Sukthankar.
\newblock {THUMOS Challenge: Action Recognition with a Large Number of
  Classes}.
\newblock \url{http://www.thumos.info/}, 2015.

\bibitem{Gry09}
J.M. Gryn, R.P. Wildes, and J.K. Tsotsos.
\newblock {Detecting Motion Patterns via Direction Maps with Application to
  Surveillance}.
\newblock {\em Computer Vision and Image Understanding (CVIU)},
  113(2):291--307, 2009.

\bibitem{Hah08}
M.~Hahn, L.~Kruger, and C.~Wohler.
\newblock {3D Action Recognition and Long-Term Prediction of Human Motion}.
\newblock In {\em Computer Vision Systems}, pages 23--32, 2008.

\bibitem{Hei15}
F.C. Heilbron, V.~Escorcia, B.~Ghanem, and J.C. Niebles.
\newblock {ActivityNet: A Large-Scale Video Benchmark for Human Activity
  Understanding}.
\newblock In {\em IEEE Conference on Computer Vision and Pattern Recognition
  (CVPR)}, pages 961--970, 2015.

\bibitem{Heilbron16}
F.C. Heilbron, J.C. Niebles, and B.~Ghanem.
\newblock {Fast Temporal Activity Proposals for Efficient Detection of Human
  Actions in Untrimmed Videos}.
\newblock In {\em IEEE Conference on Computer Vision and Pattern Recognition
  (CVPR)}, pages 1914--1923, 2016.

\bibitem{Hoai14}
M.~Hoai and F.~de~la Torre.
\newblock {Max-Margin Early Event Detectors}.
\newblock {\em International Journal of Computer Vision (IJCV)},
  107(2):191--202, 2014.

\bibitem{Hof99}
T.~Hofmann.
\newblock {Probabilistic Latent Semantic Indexing}.
\newblock In {\em Annual International ACM SIGIR Conference on Research and
  Development in Information Retrieval}, pages 50--57, 1999.

\bibitem{Hu62}
M.K. Hu.
\newblock {Visual Pattern Recognition by Moment Invariants}.
\newblock {\em IRE Transactions on Information Theory}, 8(2):179--187, 1962.

\bibitem{Hu06}
W.~Hu, T.~Tan, L.~Wang, and S.~Maybank.
\newblock {A Survey on Visual Surveillance of Object Motion and Behaviors}.
\newblock In {\em IEEE Transactions on Systems, Man, and Cybernetics}, pages
  334--352, 2006.

\bibitem{Hua12}
G.B. Huang, H.~Zhou, X.~Ding, and R.~Zhang.
\newblock {Extreme Learning Machine for Regression and Multiclass
  Classification}.
\newblock In {\em IEEE Transactions on Systems, Man, and Cybernetics},
  volume~42, pages 513--529, 2012.

\bibitem{Hua11}
Y.~Huang, K.~Huang, Y.~Yu, and T.~Tan.
\newblock {Salient Coding for Image Classification}.
\newblock In {\em IEEE Conference on Computer Vision and Pattern Recognition
  (CVPR)}, pages 1753--1760, 2011.

\bibitem{Ibr16}
M.S. Ibrahim, S.~Muralidharan, Z.~Deng, A.~Vahdat, and G.~Mori.
\newblock {A Hierarchical Deep Temporal Model for Group Activity Recognition}.
\newblock In {\em IEEE Conference on Computer Vision and Pattern Recognition
  (CVPR)}, pages 1971--1980, 2016.

\bibitem{Iki09}
N.~Ikizler and P.~Duygulu.
\newblock {Histogram of oriented rectangles: A new pose descriptor for human
  action recognition}.
\newblock {\em Image Vision Computing}, 27(10):1515--1526, 2009.

\bibitem{Iki08}
N.~Ikizler and D.A. Forsyth.
\newblock {Searching for Complex Human Activities with No Visual Examples}.
\newblock {\em International Journal of Computer Vision (IJCV)},
  80(3):337--357, 2008.

\bibitem{Jaa98}
T.S. Jaakkola and D.~Haussler.
\newblock {Exploiting Generative Models in Discriminative Classifiers}.
\newblock In {\em Advances in Neural Information Processing Systems (NIPS)},
  pages 487--493, 1998.

\bibitem{Jai13}
M.~Jain, H.~Jegou, and P.~Bouthemy.
\newblock {Better Exploiting Motion for Better Action Recognition}.
\newblock In {\em IEEE Conference on Computer Vision and Pattern Recognition
  (CVPR)}, pages 2555--2562, 2013.

\bibitem{Jai14}
M.~Jain, J.~van Gemert, H.~Jegou, P.~Bouthemy, and C.G.M. Snoek.
\newblock {Action Localization with Tubelets from Motion}.
\newblock In {\em IEEE Conference on Computer Vision and Pattern Recognition
  (CVPR)}, pages 740--747, 2014.

\bibitem{Jeg10}
H.~Jegou, M.~Douze, C.~Schmid, and P.~Perez.
\newblock {Aggregating Local Descriptors into a Compact Image Representation}.
\newblock In {\em IEEE Conference on Computer Vision and Pattern Recognition
  (CVPR)}, pages 3304--3311, 2010.

\bibitem{Jhu13}
H.~Jhuang, J.~Gall, S.~Zuffi, C.~Schmid, and M.J. Black.
\newblock {Towards Understanding Action Recognition}.
\newblock In {\em IEEE International Conference on Computer Vision (ICCV)},
  pages 3192--3199, 2013.

\bibitem{Jhu07}
H.~Jhuang, T.~Serre, L.~Wolf, and T.~Poggio.
\newblock {A Biologically Inspired System for Action Recognition}.
\newblock In {\em IEEE International Conference on Computer Vision (ICCV)},
  2007.

\bibitem{Ji13}
S.~Ji, W.~Xu, M.~Yang, and K.~Yu.
\newblock {3D Convolutional Neural Networks for Human Action Recognition}.
\newblock {\em IEEE Transactions on Pattern Analysis and Machine Intelligence
  (PAMI)}, 35(1):221--231, 2013.

\bibitem{Jia12}
Y.G. Jiang, Q.~Dai, X.~Xue, W.~Liu, and C.W. Ngo.
\newblock {Trajectory-Based Modeling of Human Actions with Motion Reference
  Points}.
\newblock In {\em European Conference on Computer Vision (ECCV)}, volume 7576,
  pages 425--438, 2012.

\bibitem{thu13}
Y.G. Jiang, J.~Liu, A.~Roshan Zamir, I.~Laptev, M.~Piccardi, M.~Shah, and
  R.~Sukthankar.
\newblock {THUMOS Challenge: Action Recognition with a Large Number of
  Classes}.
\newblock \url{http://crcv.ucf.edu/ICCV13-Action-Workshop/}, 2013.

\bibitem{thu14}
Y.G. Jiang, J.~Liu, A.~Roshan Zamir, G.~Toderici, I.~Laptev, M.~Shah, and
  R.~Sukthankar.
\newblock {THUMOS Challenge: Action Recognition with a Large Number of
  Classes}.
\newblock \url{http://crcv.ucf.edu/THUMOS14/}, 2014.

\bibitem{Ju96}
S.X. Ju, M.J. Black, and Y.~Yacoob.
\newblock {Cardboard People: A Parameterized Model of Articulated Image
  Motion}.
\newblock In {\em Proceedings of the Second International Conference on
  Automatic Face and Gesture Recognition}, pages 38--44, 1996.

\bibitem{AKar14}
A.~Karpathy, G.~Toderici, S.~Shetty, T.~Leung, R.~Sukthankar, and L.~Fei-Fei.
\newblock {Large-scale Video Classification with Convolutional Neural
  Networks}.
\newblock In {\em IEEE Conference on Computer Vision and Pattern Recognition
  (CVPR)}, pages 1725 -- 1732, 2014.

\bibitem{Ke13}
S.R. Ke, H.L.U. Thuc, Y.J. Lee, J.N Hwang, J.H. Yoo, and K.H. Choi.
\newblock {A Review on Video-Based Human Activity Recognition}.
\newblock {\em Activity Detection and Novel Sensing Technologies},
  2(2):88--131, 2013.

\bibitem{Ke05}
Y.~Ke, R.~Sukthankar, and M.~Hebert.
\newblock {Efficient Visual Event Detection using Volumetric Features}.
\newblock In {\em IEEE International Conference on Computer Vision (ICCV)},
  pages 166--173, 2005.

\bibitem{Ke07}
Y.~Ke, R.~Sukthankar, and M.~Hebert.
\newblock {Event Detection in Crowded Videos}.
\newblock In {\em IEEE International Conference on Computer Vision (ICCV)},
  pages 1--8, 2007.

\bibitem{Kim09}
T.~Kim and R.~Cupola.
\newblock {Canonical Correlation Analysis of Video Volume Tensors for Action
  Categorization and Detection}.
\newblock {\em IEEE Transactions on Pattern Analysis and Machine Intelligence
  (PAMI)}, 31:1415--1428, 2009.

\bibitem{Kla08}
A.~Klaser, M.~Marszalek, and C.~Schmid.
\newblock {A Spatio-Temporal Descriptor Based on 3D-Gradients}.
\newblock In {\em British Machine Vision Conference (BMVC)}, 2008.

\bibitem{Kli12}
O.~Kliper-Gross, Y.~Gurovich, T.~Hassner, and L.~Wolf.
\newblock {Motion Interchange Patterns for Action Recognition in Unconstrained
  Videos}.
\newblock In {\em European Conference on Computer Vision (ECCV)}, 2012.

\bibitem{Koe99}
J.J. Koenderink and A.J.~Van Doom.
\newblock {The Structure of Locally Orderless Images}.
\newblock {\em International Journal of Computer Vision (IJCV)},
  31(2):159--168, 1999.

\bibitem{Kol08}
B.~Kolman and D.R. Hill.
\newblock {\em {Elementary Linear Algebra with Applications}}.
\newblock Pearson Education Inc., ninth edition edition, 2008.

\bibitem{Kov10}
A.~Kovashka and K.~Grauman.
\newblock {Learning a Hierarchy of Discriminative Space-Time Neighborhood
  Features for Human Action Recognition}.
\newblock In {\em IEEE Conference on Computer Vision and Pattern Recognition
  (CVPR)}, pages 2046--2053, 2010.

\bibitem{Kue11}
H.~Kuehne, H.~Jhuang, E.~Garrote, T.~Poggio, and T.~Serre.
\newblock {HMDB: A Large Video Database for Human Motion Recognition}.
\newblock In {\em IEEE International Conference on Computer Vision (ICCV)},
  pages 2556--2563, 2011.

\bibitem{Laf01}
J.~Lafferty, A.~McCallum, and F.~Pereira.
\newblock {Conditional Random Fields: Probabilistic Models for Segmenting and
  Labeling Sequence Data}.
\newblock In {\em International Conference on Machine Learning (ICML)}, pages
  282--289, 2001.

\bibitem{Lan14}
T.~Lan, T.C. Chen, and S.~Savarese.
\newblock {A Hierarchical Representation for Future Action Prediction}.
\newblock In {\em European Conference on Computer Vision (ECCV)}, pages
  689--704, 2014.

\bibitem{Lan11}
T.~Lan, Y.~Wang, and G.~Mori.
\newblock {Discriminative Figure-centric Models for Joint Action Localization
  and Recognition}.
\newblock In {\em IEEE International Conference on Computer Vision (ICCV)},
  pages 2003--2010, 2011.

\bibitem{Lap05}
I.~Laptev.
\newblock {On Space-Time Interest Points}.
\newblock {\em International Journal of Computer Vision (IJCV)},
  64(2):107--123, 2005.

\bibitem{Lap08}
I.~Laptev, M.~Marszalek, C.~Schmid, and B.~Rozenfeld.
\newblock {Learning Realistic Human Actions from Movies}.
\newblock In {\em IEEE Conference on Computer Vision and Pattern Recognition
  (CVPR)}, 2008.

\bibitem{Lap07}
I.~Laptev and P.~Perez.
\newblock {Retrieving Actions in Movies}.
\newblock In {\em IEEE International Conference on Computer Vision (ICCV)},
  pages 1--8, 2007.

\bibitem{Lax07}
B.~Laxton, L.~Lim, and D.~Kriegman.
\newblock {Leveraging Temporal, Contextual and Ordering Constraints for
  Recognizing Complex Activities in Video}.
\newblock In {\em IEEE Conference on Computer Vision and Pattern Recognition
  (CVPR)}, pages 1--8, 2007.

\bibitem{Laz06}
S.~Lazebnik, C.~Schmid, and J.~Ponce.
\newblock {Beyond Bags of Features: Spatial Pyramid Matching for Recognizing
  Natural Scene Categories}.
\newblock In {\em IEEE Conference on Computer Vision and Pattern Recognition
  (CVPR)}, pages 2169--2178, 2006.

\bibitem{Le11}
Q.V. Le, W.Y. Zou, S.Y. Yeung, and A.Y. Ng.
\newblock {Learning Hierarchical Invariant Spatio-temporal Features for Action
  Recognition with Independent Subspace Analysis}.
\newblock In {\em IEEE Conference on Computer Vision and Pattern Recognition
  (CVPR)}, pages 3361--3368, 2011.

\bibitem{LeC98}
Y.~LeCun, L.~Bottou, Y.~Bengio, and P.~Haffner.
\newblock {Gradient-based Learning Applied to Document Recognition}.
\newblock In {\em Proceedings of the IEEE}, volume~86, 1998.

\bibitem{Lin09}
Z.~Lin, Z.~Jiang, and L.~Davis.
\newblock {Recognizing Actions by Shape-Motion Prototype Trees}.
\newblock In {\em IEEE International Conference on Computer Vision (ICCV)},
  pages 444--451, 2009.

\bibitem{Liu11}
H.~Liu, R.~Feris, and M.T. Sun.
\newblock {\em {Visual Analysis of Humans}}, chapter 20 - Benchmarking Datasets
  for Human Activity Recognition, pages 411--427.
\newblock Springer, 2011.

\bibitem{Liu09}
J.~Liu, J.~Luo, and M.~Shah.
\newblock {Recognizing Realistic Actions from Videos ``in the wild''}.
\newblock In {\em IEEE Conference on Computer Vision and Pattern Recognition
  (CVPR)}, 2009.

\bibitem{LLiu11}
L.~Liu, L.~Wang, and X.~Liu.
\newblock {In Defense of Soft-Assignment Coding}.
\newblock In {\em IEEE International Conference on Computer Vision (ICCV)},
  pages 2486--2493, 2011.

\bibitem{Low04}
D.~Lowe.
\newblock {Distinctive Image Features from Scale-Invariant Keypoints}.
\newblock {\em International Journal of Computer Vision (IJCV)}, 60(2):91--110,
  2004.

\bibitem{Lu06}
W.L. Lu and J.J. Little.
\newblock {Simultaneous Tracking and Action Recognition using the PCA-HOG
  Descriptor}.
\newblock In {\em Canadian Conference on Computer and Robot Vision (CRV)},
  pages 1--6, 2006.

\bibitem{Luo03}
Y.~Luo, T.D. Wu, and J.N. Hwang.
\newblock {Object-based Analysis and Interpretation of Human Motion in Sports
  Video Sequences by Dynamic Bayesian Networks}.
\newblock {\em Computer Vision and Image Understanding (CVIU)},
  92(2-3):196--216, 2003.

\bibitem{Lv07}
F.~Lv and R.~Nevatia.
\newblock {Single View Human Action Recognition using Key Pose Matching and
  Viterbi Path Searching}.
\newblock In {\em IEEE Conference on Computer Vision and Pattern Recognition
  (CVPR)}, pages 1--8, 2007.

\bibitem{Ma16}
M.~Ma, H.~Fan, and K.M. Kitani.
\newblock {Going Deeper into First-Person Activity Recognition}.
\newblock In {\em IEEE Conference on Computer Vision and Pattern Recognition
  (CVPR)}, pages 1894--1903, 2016.

\bibitem{Ma13}
S.~Ma, S.~Sclaroff, J.~Zhang, and N.~Ikizler-cinbis.
\newblock {Action Recognition and Localization by Hierarchical Space-Time
  Segments}.
\newblock In {\em IEEE International Conference on Computer Vision (ICCV)},
  2013.

\bibitem{SMa16}
S.~Ma, L.~Sigal, and S.~Sclaroff.
\newblock {Learning Activity Progression in LSTMs for Activity Detection and
  Early Detection}.
\newblock In {\em IEEE Conference on Computer Vision and Pattern Recognition
  (CVPR)}, pages 1942--1950, 2016.

\bibitem{Mah16}
B.~Mahasseni and S.~Todorovic.
\newblock {Regularizing Long Short Term Memory with 3D Human-Skeleton Sequences
  for Action Recognition}.
\newblock In {\em IEEE Conference on Computer Vision and Pattern Recognition
  (CVPR)}, pages 3054--3062, 2016.

\bibitem{Mar09a}
M.~Marszalek, I.~Laptev, and C.~Schmid.
\newblock {Hollywood2: Human Actions and Scenes Dataset}.
\newblock \url{http://www.di.ens.fr/~laptev/actions/hollywood2/}.

\bibitem{Mar09}
M.~Marszalek, I.~Laptev, and C.~Schmid.
\newblock {Actions in Context}.
\newblock In {\em IEEE Conference on Computer Vision and Pattern Recognition
  (CVPR)}, pages 2929--2936, 2009.

\bibitem{Mas08}
T.~Masada, S.~Kiyasu, and S.~Miyahara.
\newblock {Comparing LDA with pLSI as a Dimensionality Reduction Method in
  Document Clustering}.
\newblock In {\em Conference on Large-Scale Knowledge Resources}, pages 13--26,
  2008.

\bibitem{Mat04}
J.~Matas, O.~Chum, M.~Urban, and T.~Pajdla.
\newblock {Robust Wide-Baseline Stereo from Maximally Stable Extremal Regions}.
\newblock In {\em British Machine Vision Conference (BMVC)}, pages 384--396,
  2004.

\bibitem{Men08}
M.A. Mendoza and N.P. de~la Blanca.
\newblock {Applying Space State Models in Human Action Recognition: A
  Comparative Study}.
\newblock In {\em Articulated Motion and Deformable Objects (AMDO)}, pages
  53--62, 2008.

\bibitem{Mik01}
K.~Mikolajczyk and C.~Schmid.
\newblock {Indexing based on Scale Invariant Interest Points}.
\newblock In {\em IEEE International Conference on Computer Vision (ICCV)},
  2001.

\bibitem{Mik04}
K.~Mikolajczyk and C.~Schmid.
\newblock {Scale and Affine Invariant Interest Point Detectors}.
\newblock {\em International Journal of Computer Vision (IJCV)}, 60(1):63--86,
  2004.

\bibitem{Mik05}
K.~Mikolajczyk and C.~Schmid.
\newblock {A Performance Evaluation of Local Descriptors}.
\newblock {\em IEEE Transactions on Pattern Analysis and Machine Intelligence
  (PAMI)}, 27(10):1615--1630, 2005.

\bibitem{Mik08}
K.~Mikolajczyk and H.~Uemura.
\newblock {Action Recognition with Motion-Appearance Vocabulary Forest}.
\newblock In {\em IEEE Computer Vision and Pattern Recognition (CVPR)}, 2008.

\bibitem{Mol02}
C.~Moler.
\newblock {The World's Largest Matrix Computation}.
\newblock Technical articles and newsletters, MathWorks, 2002.

\bibitem{Mor00}
R.J. Morris and D.C. Hogg.
\newblock {Statistical Models of Object Interaction}.
\newblock {\em International Journal of Computer Vision (IJCV)},
  37(2):209--215, 2000.

\bibitem{Mur02}
K.~Murphy.
\newblock {\em {Dynamic Bayesian Networks: Representation, Inference and
  Learning}}.
\newblock PhD thesis, University of California, Berkeley, 2002.

\bibitem{Ng15}
J.Y.H. Ng, M.~Hausknecht, S.~Vijayanarasimhan, O.~Vinyals, R.~Monga, and
  G.~Toderici.
\newblock {Beyond Short Snippets: Deep Networks for Video Classification}.
\newblock In {\em IEEE Conference on Computer Vision and Pattern Recognition
  (CVPR)}, pages 4694--4702, 2015.

\bibitem{Nga13}
D.H. Nga and K.~Yanai.
\newblock {A Spatio-Temporal Feature based on Triangulation of Dense SURF}.
\newblock In {\em IEEE International Conference on Computer Vision Workshops
  (ICCVW)}, pages 420--427, 2013.

\bibitem{Ni16}
B.~Ni, X.~Yang, and S.~Gao.
\newblock {Progressively Parsing Interactional Objects for Fine Grained Action
  Detection}.
\newblock In {\em IEEE Conference on Computer Vision and Pattern Recognition
  (CVPR)}, pages 1020--1028, 2016.

\bibitem{Nie10}
J.C. Niebles, C.W. Chen, and L.~Fei-Fei.
\newblock {Modeling Temporal Structure of Decomposable Motion Segments for
  Activity Classification}.
\newblock In {\em European Conference on Computer Vision (ECCV)}, volume~2,
  pages 392--405, 2010.

\bibitem{Nie08}
J.C. Niebles, H.~Wang, and L.~Fei-Fei.
\newblock {Unsupervised Learning of Human Action Categories Using
  Spatial-Temporal Words}.
\newblock {\em International Journal of Computer Vision (IJCV)}, 79:299--318,
  2008.

\bibitem{Nin09}
H.~Ning, T.~Han, D.~Walther, M.~Liu, and T.~Huang.
\newblock {Hierarchical Space-Time Model Enabling Efficient Search for Human
  Actions}.
\newblock In {\em IEEE Transactions in Circuits and Systems for Video
  Technology}, volume~19, pages 808--820, 2006.

\bibitem{Oga06}
T.~Ogata, W.~Christmas, J.~Kittler, and S.~Ishikawa.
\newblock {Improving Human Activity Detection by Combining Multi-dimensional
  Motion Descriptors with Boosting}.
\newblock In {\em International Conference on Pattern Recognition (ICPR)},
  volume~1, pages 295--298, 2006.

\bibitem{Oik06}
A.~Oikonomopoulos, I.~Patras, and M.~Pantic.
\newblock {Spatiotemporal Salient Points for Visual Recognition of Human
  Actions}.
\newblock In {\em IEEE Transactions on Systems, Man, and Cybernetics - Part B:
  Cybernetics}, volume~36, pages 710--719, 2006.

\bibitem{Oneata14}
D.~Oneata, J.~Revaud, J.~Verbeek, and C.~Schmid.
\newblock {Spatio-temporal Object Detection Proposals}.
\newblock In {\em European Conference on Computer Vision (ECCV)}, pages
  737--752, 2014.

\bibitem{One13}
D.~Oneata, J.~Verbeek, and C.~Schmid.
\newblock {Action and Event Recognition with Fisher Vectors on a Compact
  Feature Set}.
\newblock In {\em IEEE International Conference on Computer Vision (ICCV)},
  pages 1817--1824, 2013.

\bibitem{One14}
D.~Oneata, J.~Verbeek, and C.~Schmid.
\newblock {The {LEAR} submission at {THUMOS} 2014}.
\newblock In {\em {THUMOS} Challenge: Action Recognition with a Large Number of
  Classes}, 2014.

\bibitem{Pag99}
L.~Page, S.~Brin, R.~Motwani, and T.~Winograd.
\newblock {The PageRank Citation Ranking: Bringing Order to the Web}.
\newblock Technical Report 422, Stanford University, 1999.

\bibitem{Pen14}
X.~Peng, L.~Wang, X.~Wang, and Y.~Qiao.
\newblock {Bag of Visual Words and Fusion Methods for Action Recognition:
  Comprehensive Study and Good Practice}.
\newblock {\em Computer Vision and Image Understanding (CVIU)}, 150:109--125,
  2016.

\bibitem{Per10}
F.~Perronnin, J.~Sanchez, and Thomas Mensink.
\newblock {Improving the Fisher Kernel for Large-Scale Image Classification}.
\newblock In {\em European Conference on Computer Vision (ECCV)}, pages
  143--156, 2010.

\bibitem{Pir12}
H.~Pirsiavash and D.~Ramanan.
\newblock {Detecting Activities of Daily Living in First-person Camera Views}.
\newblock In {\em IEEE Conference on Computer Vision and Pattern Recognition
  (CVPR)}, pages 2847--2854, 2012.

\bibitem{Pop10}
R.~Poppe.
\newblock {A Survey on Vision-based Human Action Recognition}.
\newblock {\em Image and Vision Computing}, 28(6):976--990, 2010.

\bibitem{Puz97}
J.~Puzicha, T.~Hofmann, and J.M. Buhmann.
\newblock {Non-parametric Similarity Measures for Unsupervised Texture
  Segmentation and Image Retrieval}.
\newblock In {\em IEEE Conference on Computer Vision and Pattern Recognition
  (CVPR)}, pages 267--272, 1997.

\bibitem{Ram03}
D.~Ramanan and D.A. Forsyth.
\newblock {Automatic Annotation of Everyday Movements}.
\newblock In {\em Advances in Neural Information Processing Systems (NIPS)},
  2003.

\bibitem{Rao02}
C.~Rao, A.~Yilmaz, and M.~Shah.
\newblock {View-Invariant Representation and Recognition of Actions}.
\newblock {\em International Journal of Computer Vision (IJCV)},
  50(2):203--226, 2002.

\bibitem{Rap09}
K.~Rapantzikos, Y.~Avrithis, and S.~Kollias.
\newblock {Dense Saliency-Based Spatiotemporal Feature Points for Action
  Recognition}.
\newblock In {\em IEEE Conference on Computer Vision and Pattern Recognition
  (CVPR)}, 2009.

\bibitem{Red12a}
K.K. Reddy and M.~Shah.
\newblock {UCF50 - Action Recognition Data Set}.
\newblock \url{http://crcv.ucf.edu/data/UCF50.php}.

\bibitem{Red12}
K.K. Reddy and M.~Shah.
\newblock {Recognizing 50 Human Action Categories of Web Videos}.
\newblock {\em Machine Vision and Applications Journal}, 24(5):971--981, 2012.

\bibitem{Rod08}
M.~Rodriguez, J.~Ahmed, and M.~Shah.
\newblock {Action MACH: A Spatio-temporal Maximum Average Correlation Height
  Filter for Action Recognition}.
\newblock In {\em IEEE Conference on Computer Vision and Pattern Recognition
  (CVPR)}, 2008.

\bibitem{Roh12}
M.~Rohrbach, S.~Amin, M.~Andriluka, and B.~Schiele.
\newblock {A Database for Fine Grained Activity Detection of Cooking
  Activities}.
\newblock In {\em IEEE Conference on Computer Vision and Pattern Recognition
  (CVPR)}, pages 1194--1201, 2012.

\bibitem{Roh15}
M.~Rohrbach, A.~Rohrbach, M.~Regneri, S.~Amin, M.~Andriluka, M.~Pinkal, and
  B.~Schiele.
\newblock {Recognizing Fine-Grained and Composite Activities Using Hand-Centric
  Features and Script Data}.
\newblock {\em International Journal of Computer Vision (IJCV)},
  119(3):346--373, 2015.

\bibitem{Rub98}
Y.~Rubner, C.~Tomasi, and L.J. Guibas.
\newblock {A Metric for Distributions with Applications to Image Databases}.
\newblock In {\em IEEE International Conference on Computer Vision (ICCV)},
  pages 59--66, 1998.

\bibitem{Rub00}
Y.~Rubner, C.~Tomasi, and L.J. Guibas.
\newblock {The Earth Mover's Distance as a Metric for Image Retrieval}.
\newblock {\em International Journal of Computer Vision (IJCV)}, 40(2):99--121,
  2000.

\bibitem{Rus10}
S.~Russell and P.~Norvig.
\newblock {\em {Artificial Intelligence: A Modern Approach}}.
\newblock Prentice Hall, third edition edition, 2010.

\bibitem{Ryoo11}
M.S. Ryoo.
\newblock {Human Activity Prediction: Early Recognition of Ongoing Activities
  from Streaming Videos}.
\newblock In {\em IEEE International Conference on Computer Vision (ICCV)},
  pages 1036--1043, 2011.

\bibitem{Sch08}
K.~Schindler and L.~van Gool.
\newblock {Action Snippets: How many frames does human action recognition
  require?}
\newblock In {\em IEEE Conference on Computer Vision and Pattern Recognition
  (CVPR)}, pages 1--8, 2008.

\bibitem{Sch04}
C.~Schuldt, I.~Laptev, and B.~Caputo.
\newblock {Recognizing Human Action: A Local {SVM} Approach}.
\newblock In {\em IEEE Conference on Computer Vision and Pattern Recognition
  (CVPR)}, pages 32--36, 2004.

\bibitem{Sco07}
P.~Scovanner, S.~Ali, and M.~Shah.
\newblock {A 3-Dimensional SIFT Descriptor and Its Applications to Action
  Recognition}.
\newblock In {\em Proceedings of the 15th ACM International Conference on
  Multimedia}, pages 357--360, 2007.

\bibitem{Ser07}
T.~Serre, L.~Wolf, S.~Bileschi, M.~Riesenhuber, and T.~Poggio.
\newblock {Robust Object Recognition with Cortex-like Mechanisms}.
\newblock {\em IEEE Transactions on Pattern Analysis and Machine Intelligence
  (PAMI)}, 29(3):411--426, 2007.

\bibitem{Sha12}
A.H. Shabani, D.A. Clausi, and J.S. Zelek.
\newblock {Salient Feature Detectors for Human Action Recognition}.
\newblock In {\em Ninth Conference on Computer and Robot Vision (CRV)}, pages
  468--475, 2012.

\bibitem{She07}
E.~Shechtman and M.~Irani.
\newblock {Space-Time Behavior-Based Correlation - OR - How to Tell If Two
  Underlying Motion Fields are Similar without Computing Them?}
\newblock {\em IEEE Transactions on Pattern Analysis and Machine Intelligence
  (PAMI)}, 29:2045--2056, 2007.

\bibitem{Shi11}
F.~Shi, E.M. Petriu, and A.~Cordeiro.
\newblock {Human Action Recognition from Local Part Model}.
\newblock In {\em IEEE International Workshop on Haptic Audio Visual
  Environments and Games (HAVE)}, 2011.

\bibitem{Shi94}
J.~Shi and C.~Tomasi.
\newblock {Good Features to Track}.
\newblock In {\em IEEE Conference on Computer Vision and Pattern Recognition
  (CVPR)}, pages 593--600, 1994.

\bibitem{Sho16}
Z.~Shou, D.~Wang, and S.F. Chang.
\newblock {Temporal Action Localization in Untrimmed Videos via Multi-stage
  CNNs}.
\newblock In {\em IEEE Conference on Computer Vision and Pattern Recognition
  (CVPR)}, pages 1049--1058, 2016.

\bibitem{Sim14}
K.~Simonyan and A.~Zisserman.
\newblock {Two-Stream Convolutional Networks for Action Recognition in Videos}.
\newblock In {\em Advances in Neural Information Processing Systems (NIPS)},
  2014.

\bibitem{Sin16}
B.~Singh, T.K. Marks, O.~Tuzel M.~Jones, and M.~Shao.
\newblock {A Multi-Stream Bi-Direction recurrent Neural Network for
  Fine-Grained Action Detection}.
\newblock In {\em IEEE Conference on Computer Vision and Pattern Recognition
  (CVPR)}, pages 1961--1970, 2016.

\bibitem{Singh16}
S.~Singh, C.~Arora, and C.V. Jawahar.
\newblock {First Person Action Recognition Using Deep Learned Descriptors}.
\newblock In {\em IEEE Conference on Computer Vision and Pattern Recognition
  (CVPR)}, pages 2620--2628, 2016.

\bibitem{Siv09}
J.~Sivic and A.~Zisserman.
\newblock {Efficient Visual Search of Videos Cast as Text Retrieval}.
\newblock {\em IEEE Transactions on Pattern Analysis and Machine Intelligence
  (PAMI)}, 31:591--606, 2009.

\bibitem{Smi05}
C.~Sminchisescu, A.~Kanaujia, L.~Li, and D.~Metaxas.
\newblock {Conditional Models for Contextual Human Motion Recognition}.
\newblock In {\em IEEE International Conference on Computer Vision (ICCV)},
  pages 1808--1815, 2005.

\bibitem{ActivityNet16}
C.~Snoek, B.~Ghanem, J.C. Niebles, F.C. Heilbron, W.~Barrios, V.~Escorcia, and
  P.~Mettes.
\newblock {ActivityNet: A Large-Scale Activity Recognition Challenge}.
\newblock \url{http://activity-net.org/challenges/2016/index.html}, 2016.

\bibitem{Soo14}
K.~Soomro and A.R. Zamir.
\newblock {\em {Computer Vision in Sports}}, chapter 9 - Action Recognition in
  Realistic Sports Videos, pages 181--208.
\newblock Springer, 2014.

\bibitem{Soo12}
K.~Soomro, A.R. Zamir, and M.~Shah.
\newblock {UCF101: A Dataset of 101 Human Actions Classes from Videos in the
  Wild}.
\newblock Technical Report CRCV-TR-12-01, University of Central Florida, 2012.

\bibitem{Sou08}
R.~Souvenir and J.~Babbs.
\newblock {Learning the Viewpoint Manifold for Action Recognition}.
\newblock In {\em IEEE Conference on Computer Vision and Pattern Recognition
  (CVPR)}, 2008.

\bibitem{Spr09}
E.H. Spriggs, F.~de~la Torre, and M.~Hebert.
\newblock {Temporal Segmentation and Activity Classification from First-Person
  Sensing}.
\newblock In {\em IEEE Conference on Computer Vision and Pattern Recognition
  Workshops (CVPRW)}, pages 17--24, 2009.

\bibitem{Spr15}
J.~T. Springenberg, A.~Dosovitskiy, T.~Brox, and M.~Riedmiller.
\newblock {Striving for Simplicity: The All Convolutional Net}.
\newblock In {\em International Conference on Learning Representation (ICLR)},
  2015.

\bibitem{Sta00}
C.~Stauffer and E.L. Grimson.
\newblock {Learning Patterns of Activity Using Real-Time Tracking}.
\newblock {\em IEEE Transactions on Pattern Analysis and Machine Intelligence
  (PAMI)}, 22:747--757, 2000.

\bibitem{Sul14}
W.~Sultani and I.~Saleemi.
\newblock {Human Action Recognition across Datasets by Foreground-weighted
  Histogram Decomposition}.
\newblock In {\em IEEE Conference on Computer Vision and Pattern Recognition
  (CVPR)}, pages 764--771, 2014.

\bibitem{Sul16}
W.~Sultani and M.~Shah.
\newblock {What if we do not have multiple videos of the same action? - Video
  Action Localization Using Web Images}.
\newblock In {\em IEEE Conference on Computer Vision and Pattern Recognition
  (CVPR)}, pages 1077--1085, 2016.

\bibitem{JSun09}
J.~Sun, X.~Wu, S.~Yan, L.~Cheong, T.~Chua, and J.~Li.
\newblock {Hierarchical Spatio-Temporal Context Modeling for Action
  Recognition}.
\newblock In {\em IEEE Conference on Computer Vision and Pattern Recognition
  (CVPR)}, 2009.

\bibitem{Sut12}
C.~Sutton and A.~McCallum.
\newblock {An Introduction to Conditional Random Fields}.
\newblock {\em Foundations and Trends in Machine Learning}, 4(4):267--373,
  2012.

\bibitem{Swa91}
M.J. Swain and D.H. Ballard.
\newblock {Color Indexing}.
\newblock {\em International Journal of Computer Vision (IJCV)}, 7(1):11--32,
  1991.

\bibitem{Tian13}
Y.~Tian, R.~Sukthankar, and M.~Shah.
\newblock {Spatiotemporal Deformable Part Models for Action Detection}.
\newblock In {\em IEEE Conference on Computer Vision and Pattern Recognition
  (CVPR)}, 2013.

\bibitem{Tom08}
E.~Di Tomaso and J.F. Baldwin.
\newblock {An Approach to Hybrid Probabilistic Models}.
\newblock {\em International Journal of Approximate Reasoning}, 47(2):202--218,
  2008.

\bibitem{Tra15}
D.~Tran, L.~Bourdev, R.~Fergus, L.~Torresani, and M.~Paluri.
\newblock {Learning Spatiotemporal Features with 3D Convolutional Networks}.
\newblock In {\em IEEE International Conference on Computer Vision (ICCV)},
  pages 4489--4497, 2015.

\bibitem{Tro07}
J.A. Tropp and A.C. Gilbert.
\newblock {Signal Recovery from Random Measurements via Orthogonal Matching
  Pursuit}.
\newblock In {\em IEEE Transactions on Information Theory}, volume~53, pages
  4655--4666, 2007.

\bibitem{Uem08}
H.~Uemura, S.~Ishikawa, and K.~Mikolajczyk.
\newblock {Feature Tracking and Motion Compensation for Action Recognition}.
\newblock In {\em British Machine Vision Conference (BMVC)}, 2008.

\bibitem{Ull10}
M.M. Ullah, S.N. Parizi, and I.~Laptev.
\newblock {Improving Bag of Features Action Recognition with Non-Local Cues}.
\newblock In {\em British Machine Vision Conference (BMVC)}, pages 95.1--95.11,
  2010.

\bibitem{van09}
L.~van~der Maaten, E.~Postma, and J.~van~den Herik.
\newblock {Dimensionality Reduction: A Comparative Review}.
\newblock Technical Report 005, Tilburg University, 2009.

\bibitem{van15}
J.C. van Gemert, M.~Jain, E.~Gati, and C.G.M. Snoek.
\newblock {APT: Action Localization Proposals from Dense Trajectories}.
\newblock In {\em British Machine Vision Conference (BMVC)}, pages 1--12, 2015.

\bibitem{Var14}
G.~Varol and A.A. Salah.
\newblock {Extreme Learning Machine for Large-Scale Action Recognition}.
\newblock In {\em {THUMOS} Challenge: Action Recognition with a Large Number of
  Classes}, 2014.

\bibitem{Vee06}
A.~Veeraraghavan, R.~Chellappa, and A.K. Roy-Chowdhury.
\newblock {The Function Space of an Activity}.
\newblock In {\em IEEE Conference on Computer Vision and Pattern Recognition
  (CVPR)}, pages 959--968, 2006.

\bibitem{Vig12}
E.~Vig, M.~Dorr, and D.~Cox.
\newblock {Space-Variant Descriptor Sampling for Action Recognition Based on
  Saliency and Eye Movements}.
\newblock In {\em European Conference on Computer Vision (ECCV)}, pages 84--97,
  2012.

\bibitem{Von16}
C.~Vondrick, H.~Pirsiavash, and A.~Torralba.
\newblock {Anticipating Visual Representations from Unlabeled Video}.
\newblock In {\em IEEE Conference on Computer Vision and Pattern Recognition
  (CVPR)}, pages 98--106, 2016.

\bibitem{Vu14}
T.H. Vu, C.~Olsson, I.~Laptev, A.~Oliva, and J.~Sivic.
\newblock {Predicting Actions from Static Scenes}.
\newblock In {\em European Conference on Computer Vision (ECCV)}, pages
  421--436, 2014.

\bibitem{Wan11}
H.~Wang, A.~Klaser, C.~Schmid, and C.L. Liu.
\newblock {Action Recognition by Dense Trajectories}.
\newblock In {\em IEEE Conference on Computer Vision and Pattern Recognition
  (CVPR)}, pages 3169--3176, 2011.

\bibitem{HWang13}
H.~Wang, A.~Klaser, C.~Schmid, and C.L. Liu.
\newblock {Dense Trajectories and Motion Boundary Descriptors for Action
  Recognition}.
\newblock {\em International Journal of Computer Vision (IJCV)}, 103:60--79,
  2013.

\bibitem{Wan13}
H.~Wang and C.~Schmid.
\newblock {Action Recognition with Improved Trajectories}.
\newblock In {\em IEEE International Conference on Computer Vision (ICCV)},
  2013.

\bibitem{Wan13_thumos13}
H.~Wang and C.~Schmid.
\newblock {LEAR-INRIA Submission for the THUMOS Workshop}.
\newblock In {\em {THUMOS} Challenge: Action Recognition with a Large Number of
  Classes}, 2013.

\bibitem{Wan09}
H.~Wang, M.~Ullah, A.~Klaser, I.~Laptev, and C.~Schmid.
\newblock {Evaluation of Local Spatio-Temporal Features for Action
  Recognition}.
\newblock In {\em British Machine Vision Conference (BMVC)}, 2009.

\bibitem{JWang13}
J.~Wang, P.~Liu, M.F.H. She, A.~Kouzani, and S.~Nahavandi.
\newblock {Supervised Learning Probabilistic Latent Semantic Analysis for Human
  Motion Analysis}.
\newblock {\em Neurocomputing}, 100:134--143, 2013.

\bibitem{Wan10}
J.~Wang, J.~Yang, K.~Yu, F.~Lv, T.~Huang, and Y.~Gong.
\newblock {Locality-constrained Linear Coding for Image Classification}.
\newblock In {\em IEEE Conference on Computer Vision and Pattern Recognition
  (CVPR)}, pages 3360--3367, 2010.

\bibitem{Wang15}
L.~Wang, Y.~Qiao, and X.~Tang.
\newblock {Action Recognition with Trajectory-Pooled Deep Convolutional
  Descriptors}.
\newblock In {\em IEEE Conference on Computer Vision and Pattern Recognition
  (CVPR)}, pages 4305--4314, 2015.

\bibitem{Wan15}
L.~Wang, Z.~Wang, Y.~Xiong, and Y.~Qiao.
\newblock {CUHK\&SIAT Submission for THUMOS15 Action Recognition Challenge}.
\newblock In {\em {THUMOS} Challenge: Action Recognition with a Large Number of
  Classes}, 2015.

\bibitem{Wan16}
X.~Wang, A.~Farhadi, and A.~Gupta.
\newblock {Actions $\sim$ Transformations}.
\newblock In {\em IEEE Conference on Computer Vision and Pattern Recognition
  (CVPR)}, pages 2658--2667, 2016.

\bibitem{XWang13}
X.~Wang, L.M. Wang, and Y.~Qiao.
\newblock {A Comparative Study of Encoding, Pooling and Normalization Methods
  for Action Recognition}.
\newblock In {\em 11th Asian Conference on Computer Vision (ACCV)}, pages
  572--585, 2013.

\bibitem{Wang16}
Y.~Wang and M.~Hoai.
\newblock {Improving Human Action Recognition by Non-action Classification}.
\newblock In {\em IEEE Conference on Computer Vision and Pattern Recognition
  (CVPR)}, pages 2698--2707, 2016.

\bibitem{Wan07}
Y.~Wang, K.~Huang, and T.~Tan.
\newblock {Human Activity Recognition Based on R Transform}.
\newblock In {\em IEEE Computer Vision and Pattern Recognition (CVPR)}, 2007.

\bibitem{YWang09}
Y.~Wang and G.~Mori.
\newblock {Human Action Recognition by Semilatent Topic Models}.
\newblock {\em IEEE Transactions on Pattern Analysis and Machine Intelligence
  (PAMI)}, 31(10):1762--1774, 2009.

\bibitem{Wei07}
D.~Weinland, E.~Boyer, and R.~Ronfard.
\newblock {Action Recognition from Arbitrary Views using 3D Exemplars}.
\newblock In {\em IEEE International Conference on Computer Vision (ICCV)},
  pages 1--7, 2007.

\bibitem{Wei06}
D.~Weinland, R.~Ronfard, and E.~Boyer.
\newblock {Free Viewpoint Action Recognition Using Motion History Volumes}.
\newblock In {\em Computer Vision and Image Understanding (CVIU)}, pages
  249--257, 2006.

\bibitem{Wel06}
G.~Welch and G.~Bishop.
\newblock {An Introduction to the Kalman Filter}.
\newblock Technical Report 95-041, University of North Carolina, 2006.

\bibitem{Wil08}
G.~Willems, T.~Tuytelaars, and L.~Van Gool.
\newblock {An Efficient Dense and Scale-Invariant Spatio-Temporal Interest
  Point Detector}.
\newblock In {\em European Conference on Computer Vision (ECCV)}, volume 5303,
  pages 650--663, 2008.

\bibitem{Wol97}
D.H. Wolpert and W.G. Macready.
\newblock {No Free Lunch Theorems for Optimization}.
\newblock In {\em IEEE Transactions on Evolutionary Computation}, pages 67--82,
  1997.

\bibitem{Xia06}
T.~Xiang and S.~Gong.
\newblock {Beyond Tracking: Modelling Activity and Understanding Behaviour}.
\newblock {\em International Journal of Computer Vision (IJCV)}, 67(1):21--51,
  2006.

\bibitem{Xu13}
C.~Xu, R.~F. Doell, S.J. Hanson, C.~Hanson, and J.J. Corso.
\newblock {A Study of Actor and Action Semantic Retention in Video Supervoxel
  Segmentation}.
\newblock {\em International Journal of Semantic Computing}, 2013.

\bibitem{Xu15}
Z.~Xu, L.~Zhu, Y.~Yang, and A.G. Hauptmann.
\newblock {UTS-CMU at THUMOS 2015}.
\newblock In {\em {THUMOS} Challenge: Action Recognition with a Large Number of
  Classes}, 2015.

\bibitem{Yac99}
Y.~Yacoob and M.~Black.
\newblock {Parameterized Modeling and Recognition of Activities}.
\newblock In {\em IEEE International Conference on Computer Vision (ICCV)},
  pages 120--127, 1998.

\bibitem{Yam92}
J.~Yamato, J.~Ohya, and K.~Ishii.
\newblock {Recognizing Human Action in Time-Sequential Images using Hidden
  Markov Model}.
\newblock In {\em IEEE Conference on Computer Vision and Pattern Recognition
  (CVPR)}, pages 379--385, 1992.

\bibitem{Yan09}
J.~Yang, K.~Yu, Y.~Gong, and T.~Huang.
\newblock {Linear Spatial Pyramid Matching Using Sparse Coding for Image
  Classification}.
\newblock In {\em IEEE Conference on Computer Vision and Pattern Recognition
  (CVPR)}, pages 1794--1801, 2009.

\bibitem{Yef09}
L.~Yeffet and L.~Wolf.
\newblock {Local Trinary Patterns for Human Action Recognition}.
\newblock In {\em 12th IEEE International Conference in Computer Vision
  (ICCV)}, pages 492--497, 2009.

\bibitem{Yeu16}
S.~Yeung, O.~Russakovsky, G.~Mori, and L.~Fei-Fei.
\newblock {End-to-end Learning of Action Detection from Frame Glimpses in
  Videos}.
\newblock In {\em IEEE Conference on Computer Vision and Pattern Recognition
  (CVPR)}, pages 2678--2687, 2016.

\bibitem{Yil08}
A.~Yilmaz and M.~Shah.
\newblock {A Differential Geometric Approach to Representing the Human
  Actions}.
\newblock In {\em Computer Vision and Image Understanding (CVIU)}, volume 109,
  pages 335--351, 2008.

\bibitem{You}
YouTube.
\newblock {Statistics}.
\newblock \url{https://www.youtube.com/yt/press/statistics.html}, May 2005.

\bibitem{YouTubeAPI}
YouTube.
\newblock {Search with Freebase Topics}.
\newblock
  \url{https://developers.google.com/youtube/v3/guides/searching_by_topic}, May
  2015.

\bibitem{Yu15}
G.~Yu and J.~Yuan.
\newblock {Fast Action Proposals for Human Action Detection and Search}.
\newblock In {\em IEEE Conference on Computer Vision and Pattern Recognition
  (CVPR)}, pages 1302--1311, 2015.

\bibitem{Yu09}
K.~Yu, T.~Zhang, and Y.~Gong.
\newblock {Nonlinear Learning using Local Coordinate Coding}.
\newblock In {\em Advances in Neural Information Processing Systems (NIPS)},
  2009.

\bibitem{CYuan09}
C.~Yuan, W.~Hu, X.~Li, S.~Maybank, and G.~Luo.
\newblock {Human Action Recognition under Log-Euclidean Riemannian Metric}.
\newblock In {\em Asian Conference on Computer Vision (ACCV)}, pages 343--353,
  2009.

\bibitem{Yua09}
J.~Yuan, Z.~Liu, and Y~Wu.
\newblock {Discriminative Subvolume Search for Efficient Action Detection}.
\newblock In {\em IEEE Conference on Computer Vision and Pattern Recognition
  (CVPR)}, pages 2442--2449, 2009.

\bibitem{Yua16}
J.~Yuan, B.~Ni, X.~Yang, and A.A. Kassim.
\newblock {Temporal Action Localization with Pyramid of Score Distribution
  Features}.
\newblock In {\em IEEE Conference on Computer Vision and Pattern Recognition
  (CVPR)}, pages 3093--3102, 2016.

\bibitem{Zah10}
A.~Zaharescu and R.P. Wildes.
\newblock {Anomalous Behaviour Detection Using Spatiotemporal Oriented
  Energies, Subset Inclusion Histogram Comparison and Event-Driven Processing}.
\newblock In {\em European Conference on Computer Vision (ECCV)}, pages
  563--576, 2010.

\bibitem{Zei13}
M.D. Zeiler and R.~Fergus.
\newblock {Stochastic Pooling for Regularization of Deep Convolutional Neural
  Networks}.
\newblock In {\em International Conference on Learning Representations (ICLR)},
  2013.

\bibitem{Zei14}
M.D. Zeiler and R.~Fergus.
\newblock {Visualizing and Understanding Convolutional Neural Networks}.
\newblock In {\em European Conference on Computer Vision (ECCV)}, pages
  818--833, 2014.

\bibitem{Zha16}
B.~Zhang, L.~Wang, Z.~Wang, Y.~Qiao, and H.~Wang.
\newblock {Real-time Action Recognition with Enhanced Motion Vector CNNs}.
\newblock In {\em IEEE Conference on Computer Vision and Pattern Recognition
  (CVPR)}, pages 2718--2726, 2016.

\bibitem{Zha07}
J.~Zhang, M.~Marszalek, S.~Lazebnik, and C.~Schmid.
\newblock {Local Features and Kernels for Classification of Texture and Object
  Categories: A Comprehensive Study}.
\newblock {\em International Journal of Computer Vision (IJCV)},
  73(2):213--238, 2007.

\bibitem{Zha13}
W.~Zhang, K.~Derpanis, and M.~Zhu.
\newblock {From Actemes to Action: A Strongly-supervised Representation for
  Detailed Action Understanding}.
\newblock In {\em IEEE International Conference on Computer Vision (ICCV)},
  2013.

\bibitem{Zha08}
Z.~Zhang, Y.~Hu, S.~Chan, and L.~Chia.
\newblock {Motion Context: A New Representation for Human Action Recognition}.
\newblock In {\em European Conference on Computer Vision (ECCV)}, pages
  817--829, 2008.

\bibitem{Zho15}
Y.~Zhou, B.~Ni, R.~Hong, M.~Wang, and Q.~Tian.
\newblock {Interaction Part Mining: A Mid-Level Approach for Fine-Grained
  Action Recognition}.
\newblock In {\em IEEE International Conference on Computer Vision (ICCV)},
  pages 3323--3331, 2015.

\bibitem{Zhou16}
Y.~Zhou, B.~Ni, R.~Hong, X.~Yang, and Q.~Tian.
\newblock {Cascaded Interactional Targeting Network for Egocentric Video
  Analysis}.
\newblock In {\em IEEE Conference on Computer Vision and Pattern Recognition
  (CVPR)}, pages 1904--1913, 2016.

\bibitem{Zhu13}
J.~Zhu, B.~Wang, X.~Yang, and W.~Zhang.
\newblock {Action Recognition with Actons}.
\newblock In {\em IEEE International Conference on Computer Vision (ICCV)},
  pages 3559--3566, 2013.

\bibitem{Zhu16}
W.~Zhu, J.~Hu, G.~Sun, X.~Cao, and Y.~Qiao.
\newblock {A Key Volume Mining Deep Framework for Action Recognition}.
\newblock In {\em IEEE Conference on Computer Vision and Pattern Recognition
  (CVPR)}, pages 1991--1999, 2016.

\end{thebibliography}
